%% file: main_small.tex
\documentclass[12pt]{book}

\usepackage[utf8]{inputenc}
\usepackage{amsmath,amssymb}
\usepackage[amsmath]{ntheorem}
\usepackage[bitstream-charter]{mathdesign}
\usepackage[dvipsnames]{xcolor}
\usepackage{listings}
\usepackage{courier}
\usepackage[hyphens]{url}
\usepackage{textcomp}
\usepackage{upquote}
\usepackage[backref=page]{hyperref}
\usepackage{graphicx}
\usepackage{tikz}
\usepackage{anyfontsize}
\usepackage{tcolorbox}
\usepackage{setspace}
\usepackage[skip=10pt plus1pt, indent=0pt]{parskip}
\usepackage{booktabs}
\usepackage{svg}
\usepackage{mathtools}
\usepackage{annotate-equations}
\usepackage{subcaption}
\usepackage[leftcaption,raggedright]{sidecap}
\usepackage{soul}
\usepackage{relsize}
\usepackage{wrapfig}
\usepackage{marginnote}
 \usepackage{microtype}
 
\emergencystretch=\maxdimen
\usepackage[paperheight=9.25in,paperwidth=6.125in,inner=1in,outer=1.125in,bmargin=1.125in,tmargin=1.125in,marginparsep=0.5cm]{geometry}

\sidecaptionvpos{figure}{c}

\definecolor{titlecolor}{RGB}{245, 242, 238} 
\definecolor{darkgray}{HTML}{404040}
\definecolor{drawgray}{HTML}{666666}
\definecolor{drawgray_l}{HTML}{F5F5F5}
\definecolor{drawblue}{HTML}{6C8EBF}
\definecolor{drawblue_l}{HTML}{DAE8FC}
\definecolor{drawgreen}{HTML}{82B366}
\definecolor{drawgreen_l}{HTML}{D5E8D4}
\definecolor{draworange}{HTML}{D79B00}
\definecolor{draworange_l}{HTML}{FFE6CC}
\definecolor{drawyellow}{HTML}{D6B656}
\definecolor{drawyellow_l}{HTML}{FFF2CC}
\definecolor{drawred}{HTML}{B85450}
\definecolor{drawred_l}{HTML}{EA6B66}
\definecolor{drawviolet}{HTML}{9673A6}
\definecolor{drawviolet_l}{HTML}{E1D5E7}

\tcbset {
  base/.style={
    arc=0mm, 
    bottomtitle=0.5mm,
    boxrule=0mm,
    colbacktitle=drawblue_l, 
    colframe=drawblue,
    coltitle=black,
    fonttitle=\bfseries, 
    fontlower=\scriptsize,
    left=0.5cm,
    leftrule=1mm,
    right=0.5cm,
    top=0.5cm,
    bottom=0.5cm,
    title={#1},
    toptitle=0.75mm, 
  }
}

\newtcolorbox{supportbox}[1]{
  colframe=drawblue, 
  base={#1},
  fontupper=\small
}

\usepackage{fancyhdr}

\newcommand*{\idx}[2]{{\color{gray!70}\left[\kern-\nulldelimiterspace\right.}#1{\color{gray!70}\left.\kern-\nulldelimiterspace\right]}_{\color{gray!70}#2}}
\DeclareMathOperator*{\argmax}{arg\,max}

\DeclareCaptionFormat{custom}
{%
    \textbf{#1#2}\textit{\small #3}
}
\usepackage[style=0,ntheorem]{mdframed}
\mdfsetup{%
topline=false,
rightline=false,
bottomline=false,
linewidth=3pt,
innerleftmargin=15pt,
innerrightmargin=0pt,
skipabove=\baselineskip,
skipabove=1.2\baselineskip,
}
\newmdtheoremenv[linecolor=drawred]{definition}{Definition}[chapter]

\usepackage[style=0,ntheorem]{mdframed}
\mdfsetup{%
topline=false,
rightline=false,
bottomline=false,
linewidth=3pt,
innerleftmargin=15pt,
innerrightmargin=0pt,
skipabove=\baselineskip,
skipabove=1.2\baselineskip,
}
\newmdtheoremenv[linecolor=drawviolet]{theorem}{Theorem}[chapter]

\usepackage[newfloat=true,frozencache,cachedir=mintedcache]{minted}
\usepackage[newfloat=true]{minted}
\SetupFloatingEnvironment{listing}{name=Box}
\surroundwithmdframed[linecolor=drawgreen]{minted}
\newenvironment{mypy}[2]{%
\VerbatimEnvironment
\def\myenvargumentI{#1}%
\def\myenvargumentII{#2}%
\begin{listing}[t]
\footnotesize%
\begin{minted}{python}%
}{%
\end{minted}%
\caption{\myenvargumentI}
\label{\myenvargumentII}
\end{listing}
}
\setmintedinline{fontsize=\footnotesize}

\usepackage{environ}
\usepackage{footnote}
\NewEnviron{TCBx}{ 
    \begin{savenotes}
        \begin{tcolorbox}
                \BODY   
        \end{tcolorbox}
    \end{savenotes}
}

\hypersetup{
    colorlinks=true,
    linkcolor=drawred_l,
    breaklinks=true,
    citecolor = drawred_l,
    urlcolor=darkgray,
}

\usepackage[raggedright]{titlesec}
\usepackage{blindtext, color}
\definecolor{gray75}{gray}{0.75}
\newcommand{\hsp}{\hspace{20pt}}
\titleformat{\chapter}[hang]{\Huge\bfseries}{\selectfont \thechapter\hsp\textcolor{gray75}{|}\hsp}{0pt}{\Huge\bfseries}

\newcommand{\addclock}{\marginnote{\includegraphics[width=1.27cm]{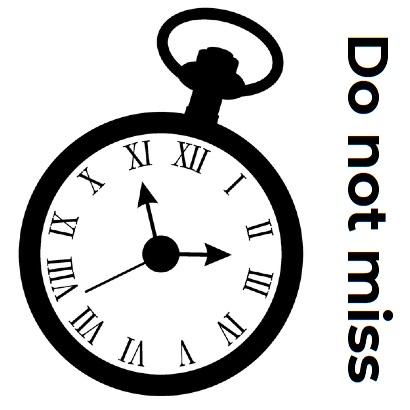}}}
\newcommand{\addteacup}{\marginnote{\includegraphics[width=1.25cm]{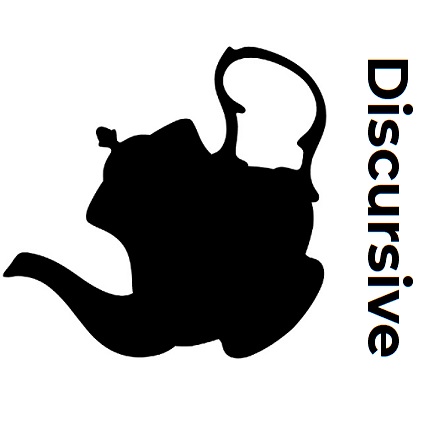}}}
\newcommand{\addbottle}{\marginnote{\includegraphics[width=1.25cm]{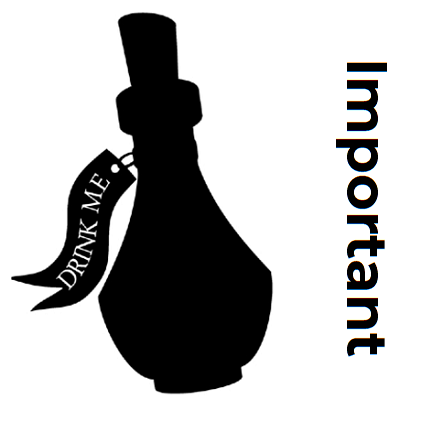}}}

\begin{document}

\begin{titlepage}
\pagecolor{titlecolor}

    \begin{flushright}
    {\Huge
    \vspace{20em}
        {\fontsize{25}{25}\selectfont Alice's Adventures in a {\color{Peach}\textbf{differentiable}} wonderland}\\
        \vskip0.5cm
        {\fontsize{15}{15}\selectfont \textit{A primer on designing neural networks}} \\ {\fontsize{12}{12}\selectfont \textbf{Vol. I - A tour of the land}}\\
        \vskip1cm
       \large Simone Scardapane
    }
    \end{flushright}
    \tikz[remember picture,overlay] \node[opacity=0.1,inner sep=0pt] at (2, -3){\includegraphics[width=9cm]{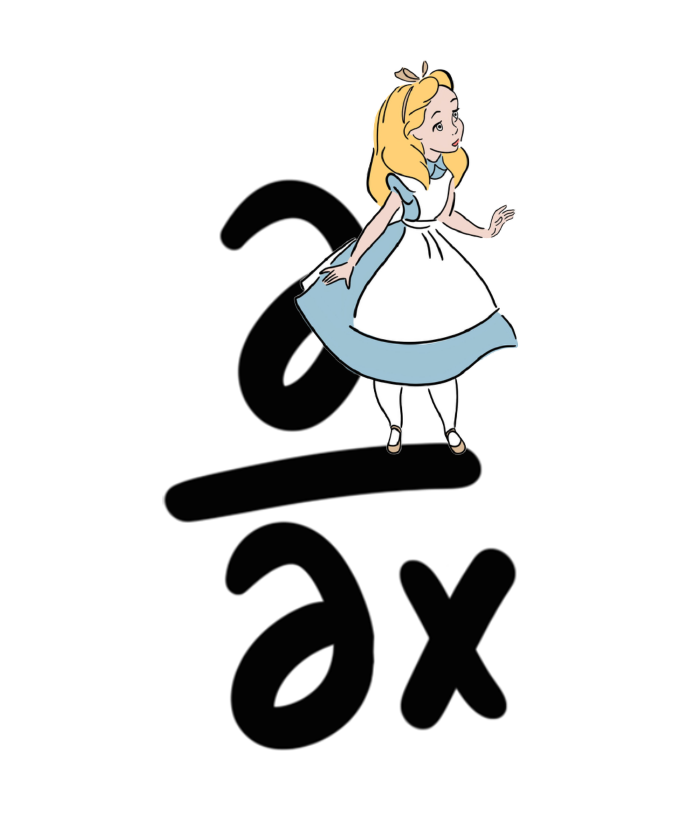}};

\end{titlepage}

\clearpage\null\newpage
\setstretch{1}

\vspace*{0.1\paperheight}

\begin{spacing}{1.2}
\tikz[remember picture,overlay] \node[opacity=1.0,inner sep=0pt] at (3, -9.5){\includegraphics[width=11cm]{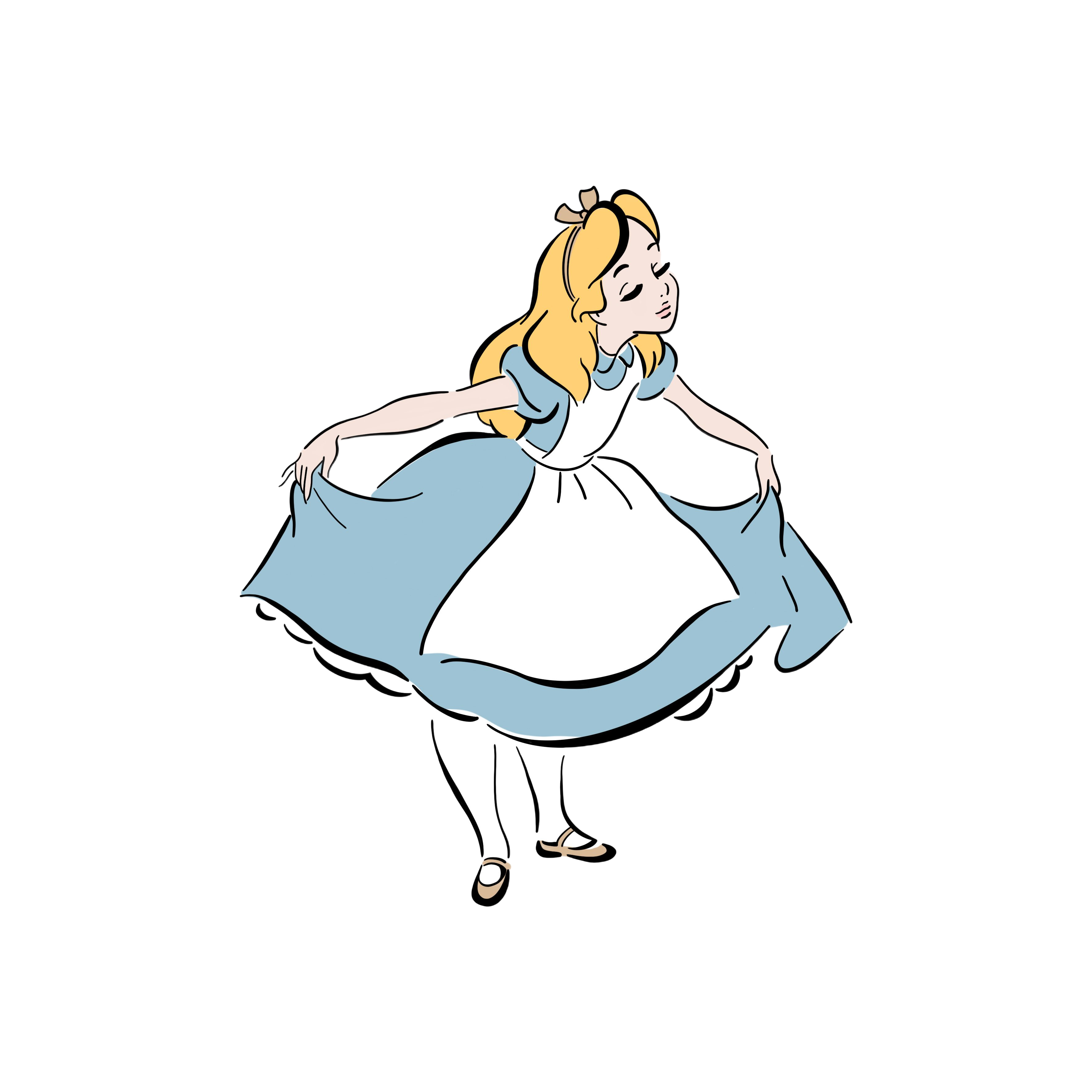}};
\begin{quote}\begin{flushright}\large“\textit{For, you see, so many out-of-the-way things had happened lately, that Alice had begun to think that very few things indeed were really impossible.}” \\\vspace{1em} — \textbf{Chapter 1, Down the Rabbit-Hole}\end{flushright}\end{quote}
\end{spacing}
\newpage\pagecolor{white}\null\newpage

\frontmatter

\chapter*{Foreword}
\markboth{}{}

\addcontentsline{toc}{chapter}{Foreword}

This book is an introduction to the topic of (deep) \textbf{neural networks}, the core technique at the heart of large language models, generative artificial intelligence - and many other applications. Because the term \textit{neural} comes with a lot of historical baggage, and because neural networks are simply compositions of differentiable primitives, I refer to them -- when feasible -- with the simpler term \textbf{differentiable models}.

In 2009, I stumbled almost by chance upon a paper by Yoshua Bengio on the power of `deep' networks \cite{bengio2009learning}, at the same time when automatic differentiation libraries like Theano \cite{al2016theano} were becoming popular. Like Alice, I had stumbled upon a strange programming realm - a \textit{differentiable} wonderland where simple things, such as selecting an element, were incredibly hard, and other things, such as recognizing cats, were amazingly simple.

I have spent more than ten years reading about, implementing, and teaching these ideas. This book is a rough attempt at condensing something of what I have learned in the process, with a focus on their design and most common components. Because the field is evolving quickly, I have tried to strike a good balance between theory and code, historical considerations and recent trends. I assume the reader has some exposure to machine learning and linear algebra, but I try to cover the preliminaries when necessary.

\vspace{3.5em}
\hfill%
\begin{minipage}{0.75\textwidth}\begin{flushright}\large
\textit{Gather round, friends: \\it's time for our beloved \\ {\color{drawred}{Alice's Adventures in a \\ differentiable wonderland}}}\end{flushright}
\end{minipage}
\tikz[remember picture,overlay] \node[opacity=1.0,inner sep=0pt] at (-9,-1.3){\includegraphics[width=6.5cm]{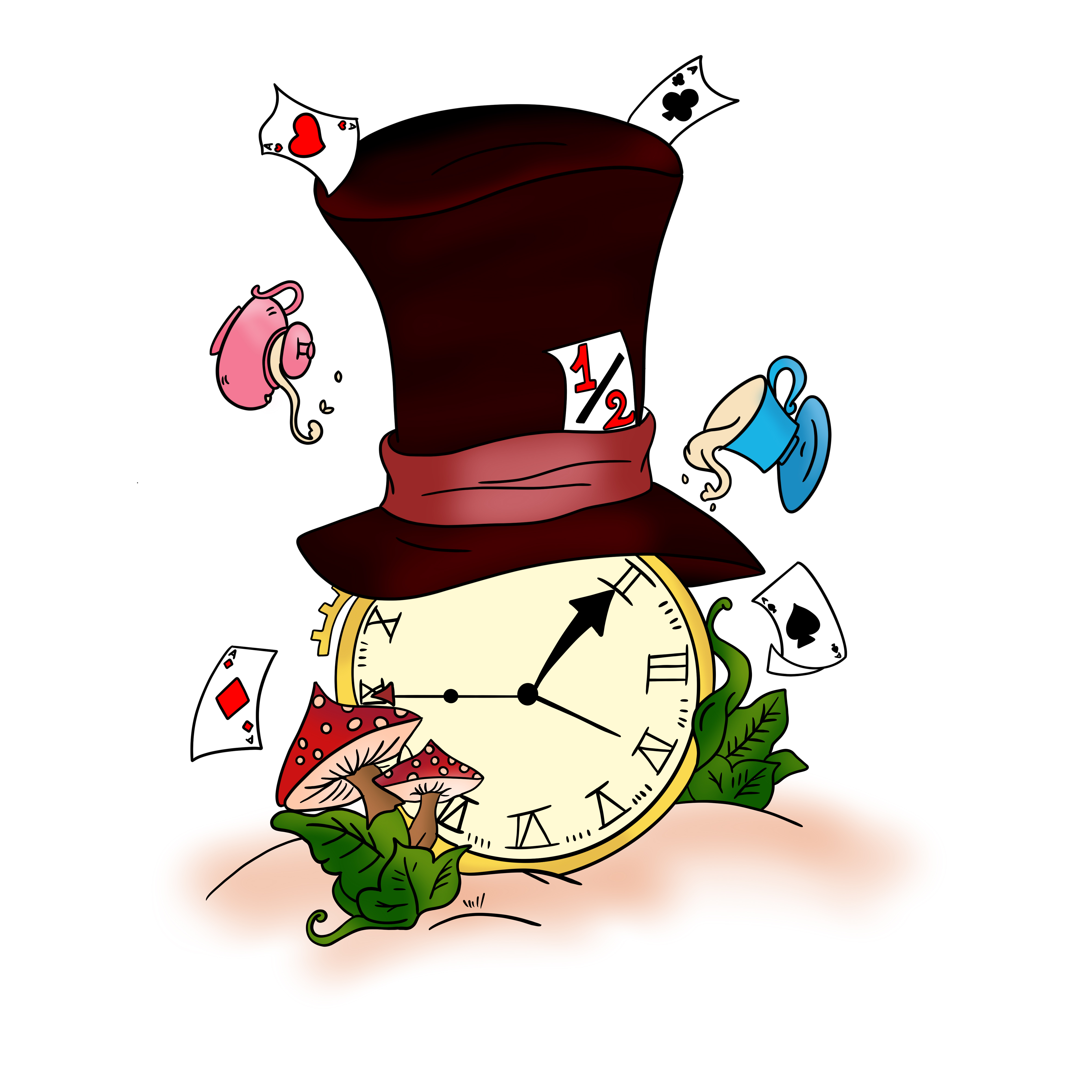}};

\newpage\newpage

\pagestyle{empty}
{
\hypersetup{linkcolor=drawred}
\setcounter{tocdepth}{1}
\tableofcontents
}
\clearpage
\pagestyle{fancy}

\mainmatter

\include{1_introduction}

\part[Compass and needle]{
\pagecolor{titlecolor}
\label{part:compass_and_needle}
\vspace*{1em}
\begin{centering}
\vspace{-2em}Compass and needle\end{centering}\vspace*{2em}
\begin{spacing}{1.2}
\hspace*{1.7em}\begin{minipage}[l]{10cm}
\begin{quote}\begin{flushright}
\normalfont\large\textit{“Would you tell me, please, which way I ought to go from here?” \\
“That depends a good deal on where you want to get to,” said  the Cat.\\
“I don’t much care where” said Alice.\\
“Then it doesn’t matter which way you go,” said the Cat.} \\\vspace{1em} — \textbf{Chapter 6, Pig and Pepper}
\end{flushright}\end{quote} 
\tikz[remember picture,overlay] \node[opacity=0.8,inner sep=0pt] at (-0.5,5.5){\includegraphics[width=5cm]{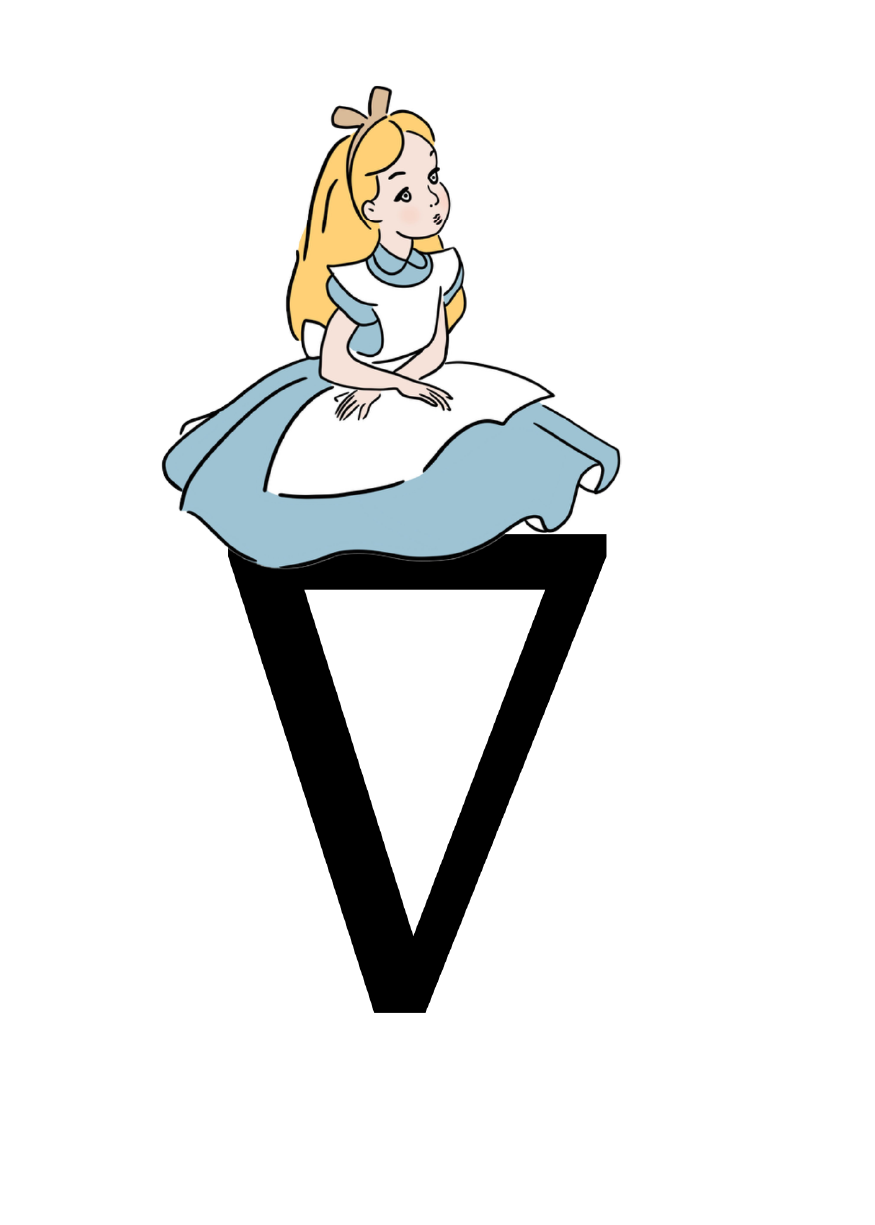}};
\end{minipage}\end{spacing} \newpage\pagecolor{white}
}

\include{2_preliminaries}
\include{3_datasets_and_loss_functions}
\include{4_linear_models}
\include{5_fully_connected_models}
\include{6_automatic_differentiation}

\part[A strange land]{\pagecolor{titlecolor}
\label{part:a_strange_land}
\vspace*{1em}
\vspace{-2em}\begin{center}A strange land\end{center}\vspace*{2em}
\begin{spacing}{1.2}
\hspace*{1.7em}\begin{minipage}[l]{10cm}
\begin{quote}\begin{flushright}
\normalfont\large\textit{“Curiouser and curiouser!” cried Alice (she was so much surprised, that for the moment she quite forgot \\ how to speak good English).} \\\vspace{1em} — \textbf{Chapter 2, The Pool of Tears}
\end{flushright}\end{quote} 
\tikz[remember picture,overlay] \node[opacity=0.8,inner sep=0pt] at (-0.5,3){\includegraphics[width=4.5cm]{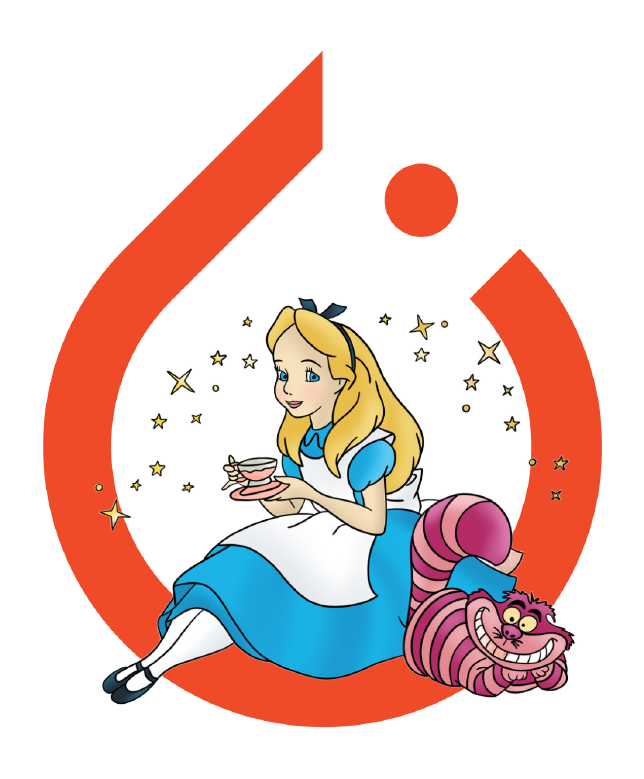}};
\end{minipage}\end{spacing}
\newpage
\pagecolor{white}
}

\include{7_convolutional_layers}
\include{8_convolutions_beyond_images}
\include{9_building_deep_cnns}

\part[Down the rabbit-hole]{\pagecolor{titlecolor}
\label{part:down_the_rabbit_hole}
\vspace*{1em}
\vspace{-2em}$\,$Down the rabbit-hole\vspace*{2em}
\begin{spacing}{1.2}
\hspace*{2.5em}\begin{minipage}[l]{8cm}
\begin{quote}\begin{flushright}
\normalfont\large\textit{“It would be so nice if something made sense for a change.”} \\\vspace{1em} \textbf{—Alice in Wonderland, 1951 movie}
\end{flushright}\end{quote} 
\tikz[remember picture,overlay] \node[opacity=1.0,inner sep=0pt] at (-1,3.5){\includegraphics[width=6.5cm]{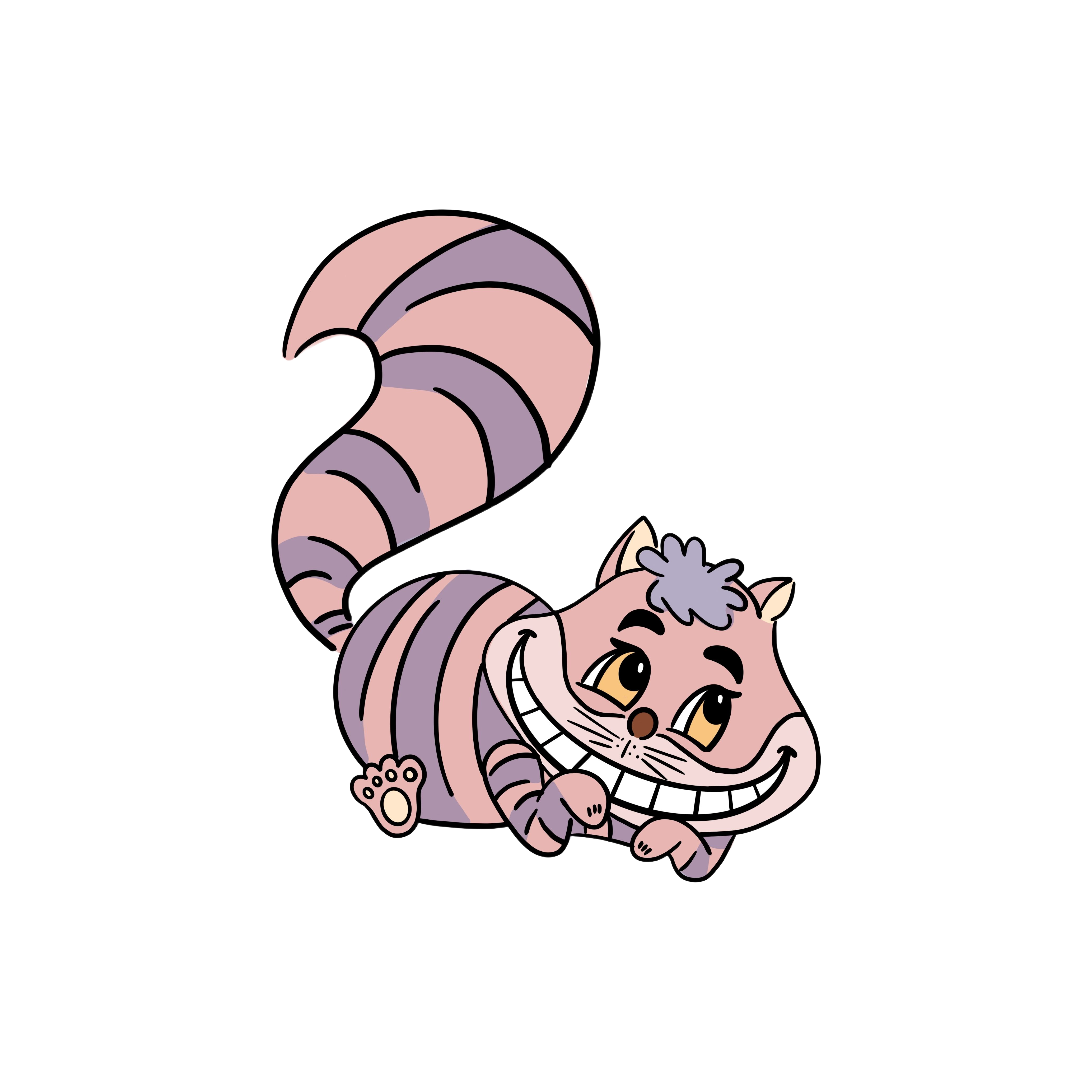}};
\end{minipage}\end{spacing}
\newpage\pagecolor{white}
}

\include{10_transformers}
\include{11_advanced_transformers}
\include{12_graph_models}
\include{13_recurrent_models}

\pagecolor{white}
\chapter*{Goodbye (for now)}

And so, Alice's first trip in this differentiable wonderland has come (for now) to an end. We only made a very broad tour, with a focus on the many ways layers can be designed and composed to create modern differentiable models (a.k.a., neural networks).

\begin{wrapfigure}{r}{5.5cm}
\includegraphics[width=5.5cm]{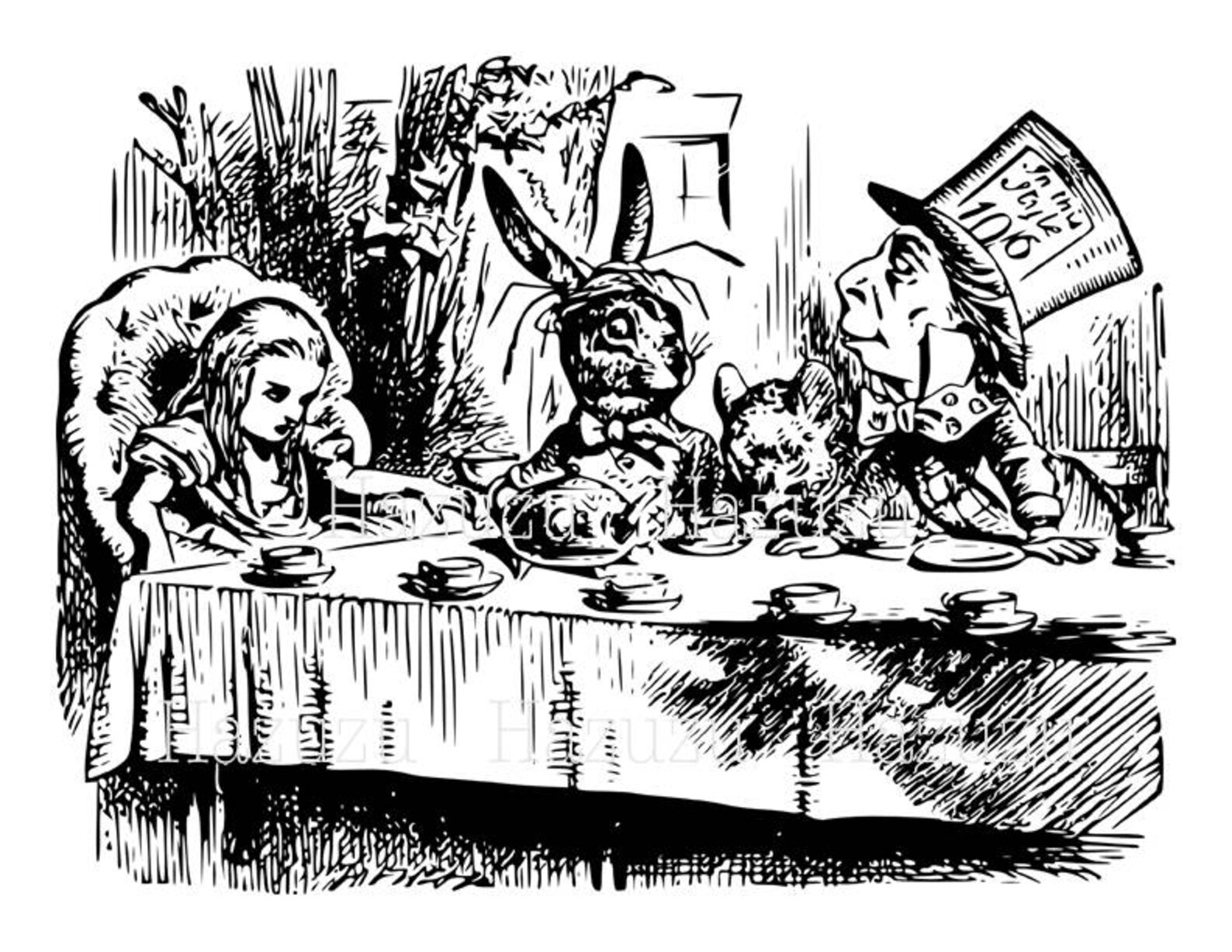}
\end{wrapfigure} 

There are many topics we discussed only briefly, including how we can \textit{use} these models in practice: from fine-tuning to generative modeling, continual learning, multimodality, explainability, and more. 

We also avoided pesky engineering aspects: training and serving large models is a huge engineering feat which requires, among other things, distributed training strategies, fast compilers, and DevOps techniques.\footnote{As an example, see \url{https://jax-ml.github.io/scaling-book/}.} And the emergence of LLMs has opened up new avenues for their use where knowledge of their inner workings is not even a prerequisite, from prompt engineering to model chaining and agentic behaviours.

This book has a companion website,\footnote{\url{https://sscardapane.it/alice-book}} where I hope to publish additional chapters that touch upon some of these topics. If time allows, some of them may be joined together in a new volume.

I hope you appreciated the journey! For comments, suggestions, and feedback on the book do not hesitate to contact me.

\newpage
\pagecolor{white}
\pagestyle{fancy}

\appendix
\fancyhead[LO]{\bfseries\normalfont {\color{black!40}\nouppercase{Appendix \thechapter: \leftmark}}}
\include{appendix_a}
\include{appendix_b}

\newlength{\bibitemsep}\setlength{\bibitemsep}{.2\baselineskip plus .05\baselineskip minus .05\baselineskip}
\newlength{\bibparskip}\setlength{\bibparskip}{0pt}
\let\oldthebibliography\thebibliography
\renewcommand\thebibliography[1]{%
  \oldthebibliography{#1}%
  \setlength{\parskip}{\bibitemsep}%
  \setlength{\itemsep}{\bibparskip}%
}

\bibliographystyle{alphaabbr}
\scriptsize
\bibliography{biblio}

\end{document}

%% file: 1_introduction.tex
\chapter{Introduction}
\markboth{\uppercase{Introduction}}{\uppercase{Introduction}}
\label{chap:introduction}

Neural networks have become an integral component of our everyday’s world, either openly (in the guise of \textbf{large language models}, LLMs), or hidden from view, by powering countless technologies and scientific discoveries including drones, cars, search engines, molecular design, and recommender systems \cite{wang2023scientific}. As we will see, all of this has been done by relying on a very small set of guiding principles and components, forming the core of this book, while the research focus has shifted to scaling them up to the limits of what is physically possible.

The power of scaling is embodied in the relatively recent concept of \textbf{neural scaling laws}, which in turn has been instrumental in driving massive investments in artificial intelligence (AI) \cite{kaplan2020scaling,ho2024algorithmic}: informally, for practically any task, simultaneously increasing data, compute power, and the size of the models -- almost always -- results in a \textit{predictable} increase in accuracy. Stated in another way, the compute power required to achieve a given accuracy for a task is decreasing by a constant factor per period of time \cite{ho2024algorithmic}. The tremendous power of combining simple, general-purpose tools with exponentially increased computational power in AI was called the \textit{bitter lesson} by R. Sutton.\footnote{\url{http://www.incompleteideas.net/IncIdeas/BitterLesson.html}.}

\begin{figure}
    \centering
    \includegraphics[width=\textwidth]{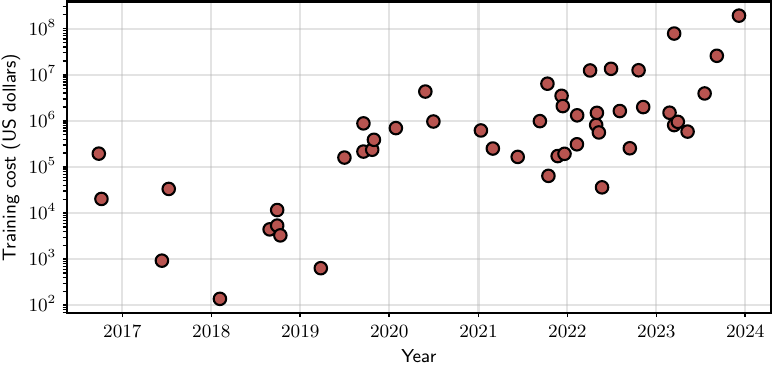}
    \caption[Training cost (in US dollars) of notable AI models released from 2016. Training cost is correlated to the three key factors of scaling laws: size of the datasets, compute power, and size of the models. As performance steadily increases, variations in modeling become asymptotically less significant \cite{ho2024algorithmic}. Data reproduced from the Stanford AI Index Report 2024.]{Training cost (in US dollars) of notable AI models released from 2016. Training cost is correlated to the three key factors of scaling laws: size of the datasets, compute power, and size of the models. As performance steadily increases, variations in modeling become asymptotically less significant \cite{ho2024algorithmic}. Data reproduced from the Stanford AI Index Report 2024.\footnotemark}
    \label{fig:compute_scaling}
\end{figure} \footnotetext{\url{https://hai.stanford.edu/research/ai-index-report}}

If we take scaling laws as given, we are left with an almost magical tool. In a nutshell, neural networks are optimized to approximate some probability distribution given data drawn from it. In principle, this approximation may fail: for example, modern neural networks are so large that they can easily memorize all the data they are shown \cite{zhang2021understanding} and transform into a trivial look-up table. Instead, trained models are shown to generalize well even to tasks that are not explicitly considered in the training data \cite{akyurek2022learning}. In fact, as the size of the datasets increases, the concept of what is \textit{in-distribution} and what is \textit{out-of-distribution} blurs, and large-scale models show hints of strong generalization capabilities and a fascinating low dependency on pure memorization, i.e., \textbf{overfitting} \cite{power2022grokking}. 

The emergence of extremely large models that can be leveraged for a variety of downstream tasks (sometimes called \textbf{foundation models}), coupled with a vibrant open-source community,\footnote{\url{https://huggingface.co/}} has also shifted how we interact with these models. Many tasks can now be solved by simply \textit{prompting} (i.e., interacting with text or visual instructions) a pre-trained model found on the web \cite{akyurek2022learning}, with the internals of the model remaining a complete black-box. From a high-level perspective, this is similar to a shift from having to programs your libraries in, e.g., C++, towards relying on open-source or commercial software whose source code is not accessible. The metaphor is not as far fetched as it may seems: nowadays, few teams worldwide have the compute and the technical expertise to design and release truly large-scale models such as the Llama LLMs \cite{touvron2023llama}, just like few companies have the resources to build enterprise CRM software.

And in the same way, just like open-source software provides endless possibilities for customizing or designing from scratch your programs, customer-grade hardware and a bit of ingenuity gives you a vast array of options to experiment with differentiable models, from \textbf{fine-tuning} them for your tasks \cite{liu2022few} to merging models \cite{ainsworth2022git}, quantizing them for low-power hardware, testing their robustness, or even designing completely new variants and ideas. For all of this, you need to look `under the hood'  and understand how these models process and manipulate data internally, with all their tricks and idiosincrasies that are born from experience and debugging. This book is an entry point into this world: if, like Alice, you are naturally curious, I hope you will appreciate the journey.

\section*{About this book}

\begin{figure}
    \centering
    \includegraphics[width=0.85\textwidth]{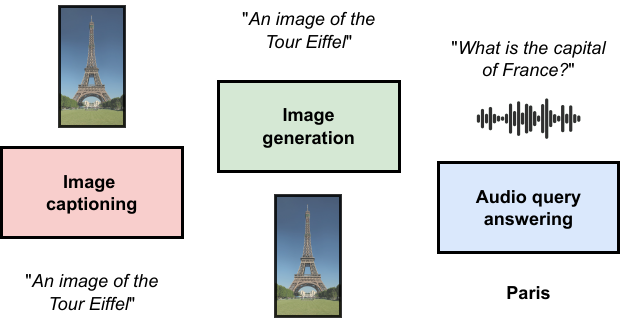}
    \caption{Most tasks can be categorized based on the desired input - output we need: {\color{drawgreen}image generation} wants an image (an \textit{ordered grid} of pixels) from a text (a \textit{sequence} of characters), while the inverse ({\color{drawred}image captioning}) is the problem of generating a caption from an image. As another example, {\color{drawblue}audio query answering} requires a text from an audio (another \textit{ordered} sequence, this time numerical). Fascinatingly, the design of the models follow similar specifications in all cases.}
    \label{fig:examples}
\end{figure}

We assume our readers are familiar with the basics of \textbf{machine learning} (ML), and more specifically \textbf{supervised learning} (SL). SL can be used to solve complex tasks by gathering data on a desired behavior, and `training' (optimizing) systems to approximate that behavior. This deceptively simple idea is extremely powerful: for example, image generation can be turned into the problem of collecting a sufficiently large collection of images with their captions; simulating the English language becomes the task of gathering a large collection of text and learning to predict a sentence from the preceding ones; and diagnosing an X-ray becomes equivalent to having a large database of scans with the associated doctors’ decision (Figure \ref{fig:examples}).

In general, learning is a \textbf{search} problem. We start by defining a program with a large number of \textit{degree-of-freedoms} (that we call parameters), and we manipulate the parameters until the model performance is satisfying. To make this idea practical, we need efficient ways of searching for the optimal configuration even in the presence of millions (or billions, or trillions) of parameters. As the name implies, \textbf{differentiable models} do this by restricting the selection of the model to differentiable components, i.e., mathematical functions that we can differentiate. Being able to compute a derivative of a high-dimensional function (a gradient) means knowing what happens if we slightly perturb their parameters, which in turn leads to automatic routines for their optimization (most notably, \textbf{automatic differentiation} and \textbf{gradient descent}). Describing this setup is the topic of the first part of the book (Part \ref{part:compass_and_needle}, \textbf{Compass and Needle}), going from Chapter \ref{chap:preliminaries} to Chapter \ref{chap:automatic_differentiation}.

\begin{figure}[t]
    \centering
    \hspace{1em}\includegraphics[width=0.8\textwidth]{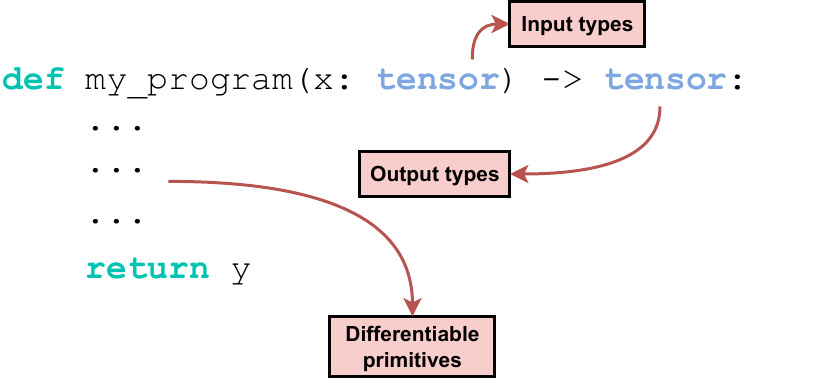}
    \caption{Neural networks are sequences of differentiable \textbf{primitives} which operate on structured arrays (\textbf{tensors}): each primitive can be categorized based on its input/output signature, which in turn defines the rules for composing them.}
    \label{fig:differentiable_programming}
\end{figure}

By viewing neural networks as simply compositions of differentiable primitives we can ask two basic questions (Figure \ref{fig:differentiable_programming}): first, what \textbf{data types} can we handle as inputs or outputs? And second, what sort of primitives can we use? Differentiability is a strong requirement that does not allow us to work directly with many standard data types, such as characters or integers, which are fundamentally \textit{discrete} and hence discontinuous. By contrast, we will see that differentiable models can work easily with more complex data represented as large arrays (what we will call \textbf{tensors}) of numbers, such as images, which can be manipulated algebraically by basic compositions of linear and nonlinear transformations. 

In the second part of the book we focus on a prototypical example of differentiable component, the \textbf{convolutional} operator (Part \ref{part:a_strange_land}, from Chapter \ref{chap:cnns} until Chapter \ref{chap:deep_cnns}). Convolutions can be applied whenever our data can be represented by an ordered sequence of elements: these include, among others, audio, images, text, and video. Along the way we also introduce a number of useful techniques to design \textit{deep} (a.k.a., composed of many steps in sequence) models, as well as several important ideas such as \textbf{text tokenization}, \textbf{autoregressive} generation of sequences, and \textbf{causal} modeling, which form the basis for state-of-the-art LLMs. 

The third part of the book (Part \ref{part:down_the_rabbit_hole}, \textbf{Down the Rabbit Hole}) continues our exploration of differentiable models by considering alternative designs for sets (most importantly \textbf{attention} layers and \textbf{transformer} models in Chapter \ref{chap:transformers} and \ref{chap:transformers_in_practice}), graphs (Chapter \ref{chap:gnns}), and finally recurrent layers for temporal sequences (Chapter \ref{chap:rnns}). 

The book is complemented by a website\footnote{\url{https://sscardapane.it/alice-book}} where I will (hopefully) collect additional chapters and material on topics of interest that do not focus on a specific type of data, including \textbf{generative modeling}, \textbf{conditional computation}, \textbf{transfer learning}, and \textbf{explainability}. These chapters are more research-oriented in nature and can be read in any order. In addition, I provide a series of guided lab sessions in notebook form, which cover a large part of the material from the book as well as advanced topics such as contrastive learning and model merging.\footnote{\url{http://tinyurl.com/guided-labs}}
\section*{In the land of differentiability}

Neural networks have a long and rich history. The name itself is a throwback to early attempts at modeling (biological) neurons in the 20th century, and similar terminology has remained pervasive: to be consistent with existing frameworks, in the upcoming chapters we may refer to \textit{neurons}, \textit{layers}, or, e.g., \textit{activations}. After multiple waves of interest, the period between 2012 and 2017 saw an unprecedented rise in complexity in the networks spurred by large-scale benchmarks and competitions, most notably the \textbf{ImageNet Large Scale Visual Recognition Challenge} (ILSVRC) that we cover in Chapter \ref{chap:deep_cnns}. A second major wave of interest came from the introduction of \textbf{transformers} (Chapter \ref{chap:transformers}) in 2017: just like computer vision was overtaken by convolutional models a few years before, natural language processing was overtaken by transformers in a very short period. Further improvements in these years were done for videos, graphs (Chapter \ref{chap:gnns}), and audio, culminating in the current excitement around LLMs, multimodal networks, and generative models.\footnote{This is not the place for a complete historical overview of modern neural networks; for the interested reader, I refer to \cite{metz2022genius} as a great starting point.}

This period paralleled a quick evolution in terminology, from the \textbf{connectionism} of the 80s \cite{rumelhart1986general} to the use of \textbf{deep learning} for referring to modern networks in opposition to the smaller, \textit{shallower} models of the past \cite{bengio2009learning,lecun2015deep}. Despite this, all these terms remain inexorably vague, because modern (artificial) networks retain almost no resemblance to biological neural networks and neurology \cite{zador2023catalyzing}. Looking at modern neural networks, their essential characteristic is being composed of differentiable blocks: for this reason, in this book I prefer the term \textbf{differentiable models} when feasible. Viewing neural networks as differentiable models leads directly to the wider topic of \textbf{differentiable programming}, an emerging discipline that blends computer science and optimization to study differentiable computer programs more broadly \cite{blondel2024elements}.\footnote{Like many, I was inspired by a `manifesto' published by Y. LeCun on Facebook in 2018: \url{https://www.facebook.com/yann.lecun/posts/10155003011462143}. For the connection between neural networks and open-source programming (and development) I am also thankful to a second manifesto, published by C. Raffel in 2021: {\url{https://colinraffel.com/blog/a-call-to-build-models-like-we-build-open-source-software.html}}.}

\begin{figure}
    \centering
    \includegraphics[width=0.8\textwidth]{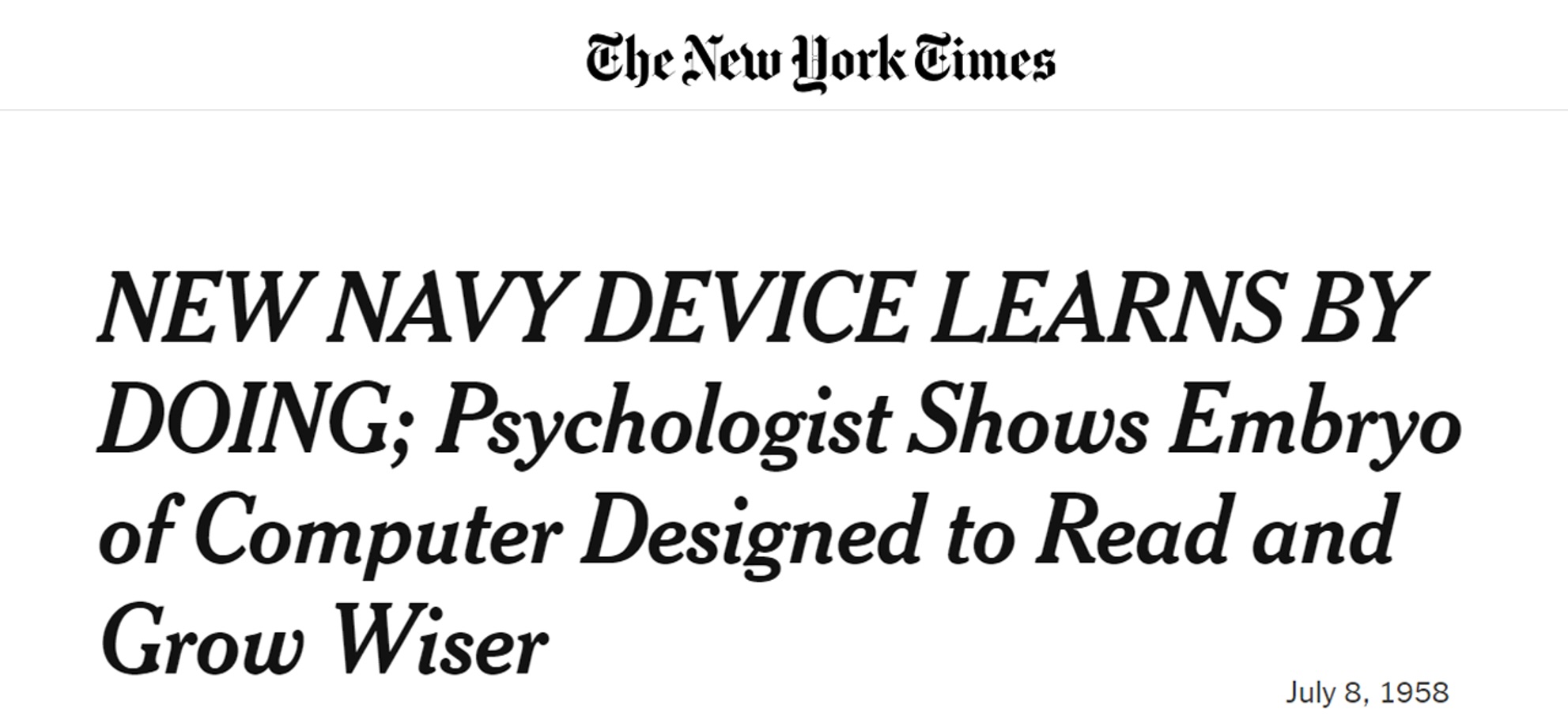}
    \caption{AI hype - except it is 1958, and the US psychologist Frank Rosenblatt has gathered up significant media attention with his studies on ``perceptrons'', one of the first working prototypes of neural networks.}
\end{figure}

As we travel through this land of differentiable models, we are also traveling through history: the basic concepts of numerical optimization of linear models by gradient descent (covered in Chapter \ref{chap:linear_models}) were known since at least the XIX century \cite{stigler1981gauss}; so-called ``fully-connected networks'' in the form we use later on can be dated back to the 1980s \cite{rumelhart1986general}; convolutional models were known and used already at the end of the 90s \cite{lecun1998gradient}.\footnote{For a history of NNs up to this period through interviews to some of the main characters, see \cite{anderson2000talking}; for a large opinionated history there is also an \textit{annotated history of neural networks} by J. Schmidhuber: \url{https://people.idsia.ch/~juergen/deep-learning-history.html}.} However, it took many decades to have sufficient data and power to realize how well they can perform given enough data and enough parameters.

While we do not have space to go in-depth on all possible topics (also due to how quickly the research is progressing), I hope the book provides enough material to allow the reader to easily navigate the most recent literature.

\section*{Notation and symbols}
\label{sec:notation}

\begin{figure}[t]
    \centering
    \hspace*{-0.5em}\includegraphics[width=1\textwidth]{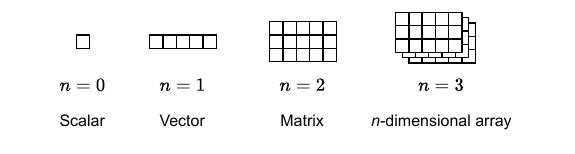}
    \caption{Fundamental data types: scalars, vectors, matrices, and generic $n$-dimensional arrays. We use the name \textbf{tensors} to refer to them. $n$ is called the \textbf{rank} of the tensor. We show the vector as a row for readability, but in the text we assume all vectors are \textit{column} vectors.}
\end{figure}

The fundamental data type when dealing with differentiable models is a \textbf{tensor},\footnote{In the scientific literature, tensors have a more precise definition as multilinear operators \cite{lim2021tensors}, while the objects we use in the book are simpler multidimensional arrays. Although a misnomer, the use of \textit{tensor} is so widespread that we keep this convention here.} which we define as an $n$ dimensional array of objects, typically real-valued numbers. With apologies to any mathematician reading us, we call $n$ the \textbf{rank} of the tensor. The notation in the book varies depending on $n$:

\begin{enumerate}
    \item A single-item tensor ($n=0$) is just a single value (a \textbf{scalar}). For scalars, we use lowercase letters, such as $x$ or $y$.\footnote{If you are wondering, scalars are named like this because they can be written as scalar multiples of one. Also, I promise to reduce the number of footnotes from now on.}
    \item Columns of values ($n=1$) are \textbf{vectors}. For vectors we use a lowercase bold font, such as $\mathbf{x}$. The corresponding row vector is denoted by $\mathbf{x}^\top$ when we need to distinguish them. We can also ignore the transpose for readability, if clear from context.
    \item Rectangular array of values ($n=2$) are \textbf{matrices}. We use an uppercase bold font, such as $\mathbf{X}$ or $\mathbf{Y}$.
    \item No specific notation is used for $n > 2$. We avoid calligraphic symbols such as $\mathcal{X}$, that we reserve for sets or probability distributions.
\end{enumerate}
For working with tensors, we use a variety of indexing strategies described better in Section \ref{sec:linear_algebra}. In most cases, understanding an algorithm or an operation boils down to understanding the shape of each tensor involved. To denote the shape concisely, we use the following notation:
$$
X \sim(b,h,w,3)
$$
This is a rank-$4$ tensor with shape $(b,h,w,3)$. Some dimensions can be pre-specified (e.g., $3$), while other dimensions can be denoted by variables. We use the same symbol to denote drawing from a probability distribution, e.g., $\varepsilon \sim \mathcal{N}(0,1)$, but we do this rarely and the meaning of the symbol should always be clear from context. Hence, $\mathbf{x} \sim (d)$ will substitute the more common $\mathbf{x} \in \mathbb{R}^d$, and similarly for $\mathbf{X} \sim (n,d)$ instead of $\mathbf{X} \in \mathbb{R}^{n \times d}$.
Finally, we may want to constrain the elements of a tensor, for which we use a special notation:
\begin{enumerate}
    \item $\mathbf{x} \sim \text{Binary}(c)$ denotes a tensor with only binary values, i.e., elements from the set $\left\{0,1\right\}$.
    \item $\mathbf{x} \sim \Delta(a)$ denotes a vector belonging to the so-called \textbf{simplex}, i.e., $x_i \ge 0$ and $\sum_i x_i = 1$. For tensors with higher rank, e.g., $\mathbf{X} \sim \Delta(n,c)$, we assume the normalization is applied with respect to the last dimension (e.g., in this case each row of $\mathbf{X}_i$ belongs to the simplex).
\end{enumerate}
Additional notation is introduced along each chapter when necessary. We also have a few symbols on the side: 

\begin{itemize}
\item  \addbottle A \textbf{bottle} to emphasize some definitions. We have many definitions, especially in the early chapters, and we use this symbol to visually discriminate the most important ones.
\item \addclock A \textbf{clock} for sections we believe crucial to understand the rest of the book -- please do not skip these!
\item \addteacup On the contrary, a \textbf{teacup} for more relaxed sections -- these are generally discursive and mostly optional in relation to the rest of the book.
\end{itemize}

\section*{Final thoughts before departing}

The book stems from my desire to give a coherent form to my lectures for \textbf{Neural Networks for Data Science Applications}, a course I teach in the Master Degree in Data Science at Sapienza University of Rome since many years. The core chapters of the book constitute the main part of the course, while the remaining chapters are topics that I cover on and off depending on the year. Some parts have been supplemented by additional courses I have taught (or I intend to teach), including parts of \textbf{Neural Networks} for Computer Engineering, an introduction to machine learning for Telecommunication Engineering, plus a few tutorials, PhD courses, and summer schools over the years.

There are already a number of excellent (and recent) books on the topic of modern, deep neural networks, including \cite{prince2023understanding, zhang2023dive,bishop2024deep,fleuret2023little,hardt2022patterns}. This book covers a similar content to all of these in the beginning, while the exposition and some additional parts (or a few sections in the advanced chapters) intersect less, and they depend mostly on my research interests. I hope I can provide an additional (and complementary) viewpoint on existing material.


As my choice of name suggests, understanding differentiable \textit{programs} comes from both theory and coding: there is a constant interplay between how we design models and how we implement them, with topics like automatic differentiation being the best example. The current resurgence of neural networks (roughly from 2012 onwards) can be traced in large part to the availability of powerful software libraries, going from Theano \cite{al2016theano} to Caffe, Chainer, and then directly to the modern iterations of TensorFlow, PyTorch, and JAX, among others. I try whenever possible to connect the discussion to concepts from existing programming frameworks, with a focus on PyTorch and JAX. The book is not a programming manual, however, and I refer to the documentation of the libraries for a complete introduction to each of them. 

Before moving on, I would like to list a few additional things this book \textit{is not}. First, I have tried to pick up a few concepts that are both (a) common today, and (b) general enough to be of use in the near future. However, I cannot foresee the future and I do not strive for completeness, and several parts of these chapters may be incomplete or outdated by the time you read them. Second, for each concept I try to provide a few examples of variations that exist in the literature (e.g., from batch normalization to layer normalization). However, keep in mind that hundreds more exist: I invite you for this to an exploration of the many pages of \href{http://paperswithcode.com}{Papers With Code}. Finally, this is a book on the fundamental components of differentiable models, but implementing them at scale (and making them work) requires both engineering sophistication and (a bit of) intuition. I cover little on the hardware side, and for the latter nothing beats experience and opinionated blog posts.\footnote{See for example this blog post by A. Karpathy: \url{http://karpathy.github.io/2019/04/25/recipe/}, or his recent \textbf{Zero to Hero} video series: \url{https://karpathy.ai/zero-to-hero.html}.}

\section*{Acknowledgments}
Equations' coloring is thanks to a beautiful LaTeX
package by ST John.\footnote{\url{https://github.com/st--/annotate-equations/tree/main}} Color images of Alice in Wonderland and the black and white symbols in the margin are all licensed from Shutterstock.com. The images of Alice in Wonderland in the figures from the main text are reproductions from the original John Tenniel illustrations, thanks to Wikimedia. I thank Roberto Alma for feedback on a previous draft of the book and for encouraging me to publish the book. I also thank Corrado Zoccolo, Emanuele Rodolà, Marcin Słaby, Konstantin Burlachenko, and Diego Sandoval for providing extensive corrections and suggestions to the current version, and everyone who sent me feedback via email.

\section*{License}

The book is released under CC BY-SA license.\footnote{\url{https://creativecommons.org/licenses/by-sa/4.0/}} This license enables ``\textit{reusers to distribute, remix, adapt, and build upon the material in any medium or format, so long as attribution is given to the creator. The license allows for commercial use. If you remix, adapt, or build upon the material, you must license the modified material under identical terms}''.

{\centering
\includegraphics[]{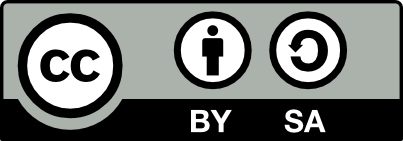}
}

%% file: 2_preliminaries.tex
\chapter[Mathematical preliminaries]{Mathematical \\ preliminaries}
\label{chap:preliminaries}

\begin{supportbox}{About this chapter}
We compress here the mathematical concepts required to follow the book. We assume prior knowledge on all these topics, focusing more on describing specific notation and giving a cohesive overview. When possible, we stress the relation between some of this material (e.g., tensors) and their implementation in practice. 
\end{supportbox}

\vspace{1em}
The chapter is composed of three parts that follow sequentially from each other, starting from \textbf{linear algebra}, moving to the definition of \textbf{gradients} for $n$-dimensional objects, and finally how we can \textbf{optimize} functions by exploiting such gradients. A self-contained overview of \textbf{probability theory} is given in Appendix \ref{chap:probability_theory}, with a focus on the \textbf{maximum likelihood} principle. 

This chapter is full of content and definitions: bear with me for a while!

\clearpage

\section{Linear algebra}
\label{sec:linear_algebra}

We recall here some basic concepts from linear algebra that will be useful in the following (and to agree on a shared notation). Most of the book revolves around the idea of a \textbf{tensor}.

\begin{definition}[Tensors] \addbottle

  A \textbf{tensor} $X$ is an $n$-dimensional array of elements of the same type. In the book we use:
  $$X \sim (s_1, s_2, \ldots, s_n)$$
  to quickly denote the \textbf{shape} of the tensor.
\end{definition}
For $n=0$ we obtain \textbf{scalars} (single values), while we have \textbf{vectors} for $n=1$, \textbf{matrices} for $n=2$, and higher-dimensional arrays otherwise. Recall that we use lowercase $x$ for scalars, lowercase bold $\mathbf{x}$ for vectors, uppercase bold $\mathbf{X}$ for matrices. Tensors in the sense described here are fundamental in deep learning because they are well suited to a massively-parallel implementation, such as using GPUs or more specialized hardware (e.g., TPUs, IPUs).

A tensor is described by the type of its elements and its \textit{shape}. Most of our discussion will be centered around tensors of floating-point values (the specific format of which we will consider later on), but they can also be defined for integers (e.g., in classification) or for strings (e.g., for text). Tensors can be \textbf{indexed} to get \textbf{slices} (subsets) of their values, and most conventions from NumPy indexing\footnote{If you want a refresher: \url{https://numpy.org/doc/stable/user/basics.indexing.html}. For readability in the book we index from $1$, not from $0$. See also the exercises at the end of the chapter.} apply. For simple equations we use pedices: for example, for a 3-dimensional tensor $X \sim (a, b, c)$ we can write $X_i$ to denote a slice of size $(b,c)$ or $X_{ijk}$ for a single scalar. We use commas for more complex expressions, such as $X_{i, :, j:k}$ to denote a slice of size $(b, k-j)$. When necessary to avoid clutter, we use a light-gray notation:
\begin{equation*}
    \idx{X}{ijk}
\end{equation*}
to visually split the indexing part from the rest, where the argument of $\idx{\bullet}{}$ can also be an expression.

\subsection{Common vector operations}
\label{subsec:common_vector_operations}

We are mostly concerned with models that can be written as composition of differentiable operations. In fact, the majority of our models will consist of basic compositions of sums, multiplications, and some additional non-linearities such as the exponential $\exp(x)$, sines and cosines, and square roots.

Vectors $\mathbf{x} \sim (d)$ are examples of 1-dimensional tensors. Linear algebra books are concerned with distinguishing between column vectors $\mathbf{x}$ and row vectors $\mathbf{x}^\top$, and we will try to adhere to this convention as much as possible. In code this is trickier, because row and column vectors correspond to $2$-dimensional tensors of shape $(1,d)$ or $(d,1)$, which are different from $1$-dimensional tensors of shape $(d)$. This is important to keep in mind because most frameworks implement broadcasting rules\footnote{In a nutshell, broadcasting aligns the tensors' shape from the right, and repeats a tensor whenever possible to match the two shapes: \url{https://numpy.org/doc/stable/user/basics.broadcasting.html}.} inspired by NumPy, giving rise to non-intuitive behaviors. See Box \ref{code:broadcasting} for an example of a very common error arising in implicit broadcasting of tensors' shapes.

\begin{mypy}{An example of (probably incorrect) broadcasting, resulting in a matrix output from an elementwise operation on two vectors due to their shapes. The same result can be obtained in practically any framework (NumPy, TensorFlow, JAX, ...).}{code:broadcasting}
import torch
x = torch.randn((4, 1))  # "Column"
y = torch.randn((4,))    # 1D tensor
print((x + y).shape)        
# [Out]: (4,4) (because of broadcasting!)
\end{mypy}

Vectors possess their own algebra (which we call a \textbf{vector space}), in the sense that any two vectors $\mathbf{x}$ and $\mathbf{y}$ of the same shape can be linearly combined $\mathbf{z} = a\mathbf{x} + b\mathbf{y}$ to provide a third vector:
$$
z_i=ax_i+by_i
$$
If we understand a vector as a point in $d$-dimensional Euclidean space, the sum is interpreted by forming a parallelogram, while the distance of a vector from the origin is given by the Euclidean ($\ell_2$) norm:
$$
\lVert \mathbf{x} \rVert=\sqrt{\sum_i x_i^2}
$$
The squared norm $\lVert \mathbf{x} \rVert^2$ is of particular interest, as it corresponds to the sum of the elements squared. The fundamental vector operation we are interested in is the \textbf{inner product} (or \textbf{dot product}), which is given by multiplying the two vectors element-wise, and summing the resulting values. \clearpage

\begin{definition}[Inner product]   \addbottle

The inner product between two vectors $\mathbf{x}, \mathbf{y} \sim (d)$ is given by the expression:
    \begin{equation}
            \langle \mathbf{x},\mathbf{y}\rangle=\mathbf{x}^\top\mathbf{y}=\sum_ix_iy_i
    \end{equation}
\end{definition}

The notation $\langle \bullet, \bullet \rangle$ is common in physics, and we use it sometimes for clarity. Importantly, the dot product between two vectors is a scalar. For example, if $\mathbf{x} = [0.1, 0, -0.3]$ and $\mathbf{y}=[-4.0, 0.05, 0.1]$:

\begin{equation*}
    \langle \mathbf{x}, \mathbf{y} \rangle = - 0.4 + 0 - 0.03 = - 0.43
\end{equation*}

A simple geometric interpretation of the dot product is given by its relation with the angle $\alpha$ between the two vectors:

\begin{equation}
    \mathbf{x}^\top\mathbf{y}=\lVert\mathbf{x}\rVert \lVert\mathbf{y}\rVert\cos(\alpha)
     \label{eq:dot_product_cosine}
\end{equation}

Hence, for two normalized vectors such that $\lVert \bullet \rVert = 1$, the dot product is equivalent to the cosine of their angle, in which case we call the dot product the \textbf{cosine similarity}. The cosine similarity $\cos(\alpha)$ oscillates between $1$ (two vectors pointing in the same direction) and $-1$ (two vectors pointing in opposite directions), with the special case of $\langle \mathbf{x}, \mathbf{y} \rangle=0$ giving rise to \textbf{orthogonal} vectors pointing in perpendicular directions. Looking at this from another direction, for two normalized vectors (having unitary norm), once we fix the first argument $\mathbf{x}$ we get:

\begin{equation}
    \mathbf{y}^* = \argmax \;\; \langle \mathbf{x}, \mathbf{y} \rangle = \mathbf{x}
    \label{eq:maximize_dot_product}
\end{equation}

where $\argmax$ denotes the operation of finding the value of $\mathbf{y}$ corresponding to the highest possible output value. From \eqref{eq:maximize_dot_product} we see that, to maximize the dot product, the second vector must equal the first one. This is important, because in the following chapters $\mathbf{x}$ will represent an input, while $\mathbf{w}$ will represent (adaptable) parameters, so that the dot product is maximized whenever $\mathbf{x}$ `resonates' with $\mathbf{w}$ (\textbf{template matching}).

We close with two additional observations that will be useful. First, we can write the sum of the elements of a vector as its dot product with a vector $\mathbf{1}$ composed entirely of ones, $\mathbf{1} = \left[1, 1, \ldots, 1 \right]^\top$:

$$
\langle\mathbf{x},\mathbf{1}\rangle=\sum_{i=1}^d x_i
$$

Second, the distance between two vectors can also be written in terms of their dot products:

$$
\lVert \mathbf{x} -\mathbf{y}\rVert^2 = \langle \mathbf{x},\mathbf{x}\rangle + \langle \mathbf{y},\mathbf{y}\rangle - 2 \langle \mathbf{x},\mathbf{y}\rangle
$$

The case $\mathbf{y}=\mathbf{0}$ gives us $\lVert \mathbf{x} \rVert^2 = \langle \mathbf{x}, \mathbf{x} \rangle$. Both equations can be useful when writing equations or in the code.

\subsection{Common matrix operations}

In the $2$-dimensional case we have matrices:

$$
\mathbf{X}=\begin{bmatrix} X_{11} & \cdots & X_{1d} \\ \vdots & \ddots & \vdots \\ X_{n1} & \cdots & X_{nd}\end{bmatrix} \sim (n,d)
$$

In this case we can talk about a matrix with $n$ rows and $d$ columns. Of particular importance for the following, a matrix can be understood as a \textbf{stack} of $n$ vectors $(\mathbf{x}_1, \mathbf{x}_2, \ldots, \mathbf{x}_n)$, where the stack is organized in a row-wise fashion:

$$
\mathbf{X} = \begin{bmatrix} \mathbf{x}_1^\top \\ \vdots \\ \mathbf{x}_n^\top \end{bmatrix}
$$

We say that $\mathbf{X}$ represents a \textbf{batch} of data vectors. As we will see, it is customary to define models (both mathematically and in code) to work on batched data of this kind. A fundamental operation for matrices is multiplication:

\begin{definition}[Matrix multiplication] \addbottle

For any two matrices $\mathbf{X}\sim $(a,b) and $\mathbf{Y}\sim (b,c)$ of compatible shape, matrix multiplication $\mathbf{Z} = \mathbf{X}\mathbf{Y}$, with $\mathbf{Z} \sim (a,c)$ is defined element-wise as:

\begin{equation}
    Z_{ij} = \langle \mathbf{X}_i, \mathbf{Y}^\top_j \rangle
    \label{eq:matrix_multiplication}
\end{equation}

\noindent i.e., the element $(i,j)$ of the product is the dot product between the $i$-th row of $\mathbf{X}$ and the $j$-th column of $\mathbf{Y}$.
\end{definition}

As a special case, if the second term is a vector we have a matrix-vector product:
\begin{equation}
\mathbf{z} = \mathbf{W}\mathbf{x}
\label{eq:matrix_vector_product}
\end{equation}
If we interpret $\mathbf{X}$ as a batch of vectors, matrix multiplication $\mathbf{X}\mathbf{W}^\top$ is a simple \textbf{vectorized} way of computing $n$ dot products as in \eqref{eq:matrix_vector_product}, one for each row of $\mathbf{X}$, with a single linear algebra operation. As another example, matrix multiplication of a matrix by its transpose, $\mathbf{X}\mathbf{X}^\top \sim(n,n)$, is a vectorized way to compute all possible dot products of pairs of rows of $\mathbf{X}$ simultaneously.

We close by mentioning a few additional operations on matrices that will be important. 

\begin{definition}[Hadamard multiplication] $\,$

\par For two matrices of the same shape, the \textbf{Hadamard multiplication} is defined element-wise as:

\begin{equation*}
\idx{\mathbf{X}\odot \mathbf{Y}}{ij} = {X}_{ij}{Y}_{ij}    
\end{equation*}

\end{definition}

While Hadamard multiplication does not have all the interesting algebraic properties of standard matrix multiplication, it is commonly used in differentiable models for performing \textit{masking} operations (e.g., setting some elements to zero) or scaling operations. Multiplicative interactions have also become popular in some recent families of models, as we will see next.

Sometimes we write expressions such as $\exp(\mathbf{X})$, which are to be interpreted as \textit{element-wise} applications of the operation:

\begin{equation}
\idx{\exp(\mathbf{X})}{ij} = \exp({X}_{ij})
\label{eq:elementwise_matrix_exp}
\end{equation}

By comparison, “true” matrix exponentiation is defined for a squared matrix as:

\begin{equation}
\text{mat-exp}(\mathbf{X})=\sum_{k=0}^\infty \frac{1}{k!}\mathbf{X}^k
\label{eq:true_matrix_exp}
\end{equation}

Importantly, \eqref{eq:elementwise_matrix_exp} can be defined for tensors of any shape, while \eqref{eq:true_matrix_exp} is only valid for (squared) matrices. This is why all frameworks, like PyTorch, have specialized modules that collect all matrix-specific operations, such as inverses and determinants. See Box \ref{code:exp} for an example.

\begin{mypy}{Difference between the element-wise exponential of a matrix and the matrix exponential as defined in linear algebra textbooks. Specialized linear algebra operations are generally encapsulated in their own sub-package.}{code:exp}
X = torch.randn((5, 5))
# Element-wise exponential
X = torch.exp(X)             
# Matrix exponential
X = torch.linalg.matrix_exp(X) 
\end{mypy}

Finally, we can write \textit{reduction} operations (sum, mean, ...) across axes without specifying lower and upper indices, in which case we assume that  the summation runs along the full axis:

$$
\sum_i {\mathbf{X}}_{i} = \sum_{{\color{drawred}i=1}}^{{\color{drawred}n}} {\mathbf{X}}_{i}
$$

In PyTorch and other frameworks, reduction operations correspond to methods having an \verb+axis+ argument:

{\small
\begin{center}\mintinline{python}{r = X.sum(axis=1)}\end{center}
} \clearpage

\begin{supportbox}{On the definition of matrix multiplication}
    Why is matrix multiplication defined as \eqref{eq:matrix_multiplication} and not as Hadamard multiplication? Consider a vector $\mathbf{x}$ and some generic function $f$ defined on it. The function is said to be \textbf{linear} if $f(\alpha \mathbf{x}_1 + \beta\mathbf{x}_2) =\alpha f(\mathbf{x}_1)+\beta f(\mathbf{x}_2)$. Any such function can be represented as a matrix $\mathbf{A}$ (this can be seen by extending the two vectors in a basis representation). Then, the matrix-vector product $\mathbf{A}\mathbf{x}$ corresponds to function application, $f(\mathbf{x})=\mathbf{A}\mathbf{x}$, and matrix multiplication $\mathbf{A}\mathbf{B}$ corresponds to function composition $f \circ g$, where $(f \circ g)(\mathbf{x}) = f(g(\mathbf{x}))$ and $g(\mathbf{x})=\mathbf{B}\mathbf{x}$.
\end{supportbox}

\subsection*{Computational complexity  \addteacup} 

I will use matrix multiplication to introduce the topic of \textit{complexity} of an operation.  Looking at \eqref{eq:matrix_multiplication}, we see that computing the matrix $\mathbf{Z} \sim (a,c)$ from the input arguments $\mathbf{X} \sim (a,b)$ and $\mathbf{Y} \sim (b,c)$ requires $ac$ inner products of dimension $b$ if we directly apply the definition (what we call the \textit{time} complexity), while the memory requirement for a sequential implementation is simply the size of the output matrix (what we call instead the \textit{space} complexity).

To abstract away from the specific hardware details, computer science focuses on the so-called big-$\mathcal{O}$ notation, from the German \textit{ordnung} (which stands for \textit{order} of approximation). A function $f(x)$ is said to be $\mathcal{O}(g(x))$, where we assume both inputs and outputs are non-negative, if we can find a constant $c$ and a value $x_0$ such that:
\begin{equation}
f(x) \le cg(x) \;\; \text{ for any } x \ge x_0
\end{equation}
meaning that as soon as $x$ grows sufficiently large, we can ignore all factors in our analysis outside of $g(x)$. This is called an \textbf{asymptotic} analysis. Hence, we can say that a naive implementation of matrix multiplication is $\mathcal{O}(abc)$, growing linearly with respect to all three input parameters. For two square matrices of size $(n,n)$ we say matrix multiplication is \textit{cubic} in the input dimension.

Reasoning in terms of asymptotic complexity is important (and elegant), but choosing an algorithm only in terms of big-$\mathcal{O}$ complexity does not necessarily translate to practical performance gains, which depends on many details such as what hardware is used, what parallelism is supported, and so on.\footnote{When you call a specific primitive in a linear algebra framework, such as matrix multiplication \mintinline{python}{A @ B} in PyTorch, the specific low-level implementation that is executed (the \textbf{kernel}) depends on the run-time hardware, through a process known as \textbf{dispatching}. Hence, the same code can run via a GPU kernel, a CPU kernel, a TPU kernel, etc. This is made even more complex by compilers such as XLA (\url{https://openxla.org/xla}), which can optimize code by fusing and optimizing operations with a specific target hardware in mind.} As an example, it is known that the best asymptotic algorithm for multiplying two square matrices of size $(n,n)$ scales as $\mathcal{O}(n^c)$ for a constant $c < 2.4$ \cite{coppersmith1982asymptotic}, which is much better than the cubic $\mathcal{O}(n^3)$ requirement of a naive implementation. However, these algorithms are much harder to parallelize efficiently on highly-parallel hardware such as GPUs, making them uncommon in practice.

Note that from the point of view of asymptotic complexity, having access to a parallel environment with $k$ processors has no impact, since it can only provide (at best) a constant $\frac{1}{k}$ speedup over a non-parallel implementation. In addition, asymptotic complexity does not take into consideration the time it takes to move data from one location to the other, which can become the major bottleneck in many situations.\footnote{\url{https://docs.nvidia.com/deeplearning/performance/dl-performance-gpu-background/index.html}} In these cases, we say the implementation is \textit{memory-bound} as opposed to \textit{compute-bound}. Practically, this can only be checked by running a profiler over the code. We will see that analyzing the complexity of an algorithm is far from trivial due to the interplay of asymptotic complexity and observed complexity.

\subsection{Higher-order tensor operations}

Vectors and matrices are interesting because they allow us to define a large number of operations which are undefined or complex in higher dimensions (e.g., matrix exponentials, matrix multiplication, determinants, ...). When moving to higher dimensions, most of the operations we are interested into are either batched variants of matrix operations, or specific combinations of matrix operations and reduction operations. 

As an example of the former, consider two tensors $X \sim (n,a,b)$ and $Y\sim (n,b,c)$. \textbf{Batched matrix multiplication} (BMM) is defined as:
\begin{equation}
\idx{\text{BMM}(X,Y)}{i} = {\mathbf{X}}_{i}{\mathbf{Y}}_{i} \sim (n, a, c)
\label{eq:bmm}
\end{equation}
Operations in most frameworks operate transparently on batched versions of their arguments, which are assumed like in this case to be \textit{leading dimensions} (the first dimensions). For example, batched matrix multiplication in PyTorch is the same as standard matrix multiplication, see Box \ref{code:bmm}.

\begin{mypy}{BMM in PyTorch is equivalent to standard matrix multiplication. Most operations can work on (possibly) batched inputs.}{code:bmm}
X = torch.randn((4, 5, 2))
Y = torch.randn((4, 2, 3))
(torch.matmul(X, Y)).shape # Or X @ Y 
# [Out]: (4, 5, 3)
\end{mypy}

As an example of a reduction operation, consider two tensors $X, Y \sim (a,b,c)$. A generalized version of the dot product (GDT) can be written as:

\begin{equation}
\text{GDT}(X,Y) =\sum_{i,j,k} \idx{X \odot Y}{ijk}
\label{eq:gdt}
\end{equation}

which is simply a dot product over the `flattened' versions of its inputs. This brief overview covers most of the tensor operations we will use in the rest of the book, with additional material introduced when necessary.

\subsection{Einstein's notation}

This is an optional section that covers {\footnotesize\mintinline{python}{einsum}},\footnote{\url{https://numpy.org/doc/stable/reference/generated/numpy.einsum.html}} \addteacup a set of conventions that allows the user to specify practically every tensor operation (including reductions, sums, multiplications) with a simple syntax based on text strings. 

\begin{mypy}{Examples of using einsum in PyTorch.}{code:einsum}
# Batched matrix multiply
M = torch.einsum('ijz,izk->ijk', A, B) 
# Generalized dot product
M = torch.einsum('ijk,ijk->', A, B)    
\end{mypy}

To introduce the notation, let us consider again the two examples shown before in \eqref{eq:bmm} and \eqref{eq:gdt}, writing down explicitly all the axes:

\begin{align}
M_{ijk}=\sum_z {A}_{ijz}{B}_{izk}\label{eq:bmm_idx}\\
M=\sum_i\sum_j\sum_k {X}_{ijk}{Y}_{ijk}\label{eq:gdt_idx}
\end{align}

In line with \textbf{Einstein’s notation},\footnote{The notation we use is a simplified version which ignores the distinction between upper and lower indices: \url{https://en.wikipedia.org/wiki/Einstein_notation}.} we can simplify the two equations by removing the sums, under the convention that any index appearing on the right but not on the left is summed over:
\begin{gather}
{M}_{ijk} ={A}_{ijz}{B}_{izk} \triangleq {\color{drawred}\sum_z} \; {A}_{ijz}{B}_{izk} \label{eq:einops_1}\\
M= {X}_{ijk}{Y}_{ijk} \triangleq {\color{drawred}\sum_i\sum_j\sum_k} \; {X}_{ijk}{Y}_{ijk}
\label{eq:einops_2}
\end{gather}

Then, we can condense the two definitions by isolating the indices in a unique string (where the operands are now on the left):

\begin{itemize}
    \item ‘\textit{ijz,izk→ijk}’ (batched matrix multiply);
    \item ‘\textit{ijk,ijk→}’ (generalized dot product).
\end{itemize}

There is a direct one-to-one correspondence between the definitions in \eqref{eq:einops_1}-\eqref{eq:einops_2} and their simplified string definition. This is implemented in most frameworks in the {\footnotesize\verb+einsum+} operation, see Box \ref{code:einsum}.

The advantage of this notation is that we do not need to remember the API of a framework to implement a given operation; and translating from one framework to the other is transparent because the einsum syntax is equivalent. For example, PyTorch has several matrix multiplication methods, including {\footnotesize\verb+matmul+} and {\footnotesize\verb+bmm+}, with different broadcasting rules and shape constraints, and einsum provides a uniform syntax for all of them. In addition, the einsum definition of our batched matrix multiplication is identical to, e.g., the definition in JAX, see Box \ref{code:einsum_pt}.

\begin{mypy}{Example of using einsum in JAX - compare with Box \ref{code:einsum}.}{code:einsum_pt}
M = jax.numpy.einsum('ijz,izk->ijk', A, B)
\end{mypy}

Working with transposed axes is also simple. For example, for $A \sim (n,a,b)$ and $B \sim (n, c, b)$, a batched multiplication of $\idx{\mathbf{A}}{i}$ times $\idx{\mathbf{B}^\top}{i}$ is obtained by switching the corresponding axes in the einsum definition:

{\footnotesize
\noindent\mintinline{python}{M = torch.einsum('ijz,ikz->ijk', A, B)}
}

\noindent Because of these reasons, einsum and its generalizations (like the popular {\footnotesize\verb+einops+}\footnote{\url{http://einops.rocks}} package) have gained a wide popularity recently.

\section{Gradients and Jacobians}
\label{sec:gradients_and_jacobians}

As the name \textit{differentiable} implies, gradients play a pivotal role in the book, providing a way to optimize our models through semi-automatic mechanisms deriving from gradient descent. We recall here some basic definitions and concepts concerning differentiation of multi-valued functions. We focus on properties that will be essential for later, partially at the expense of mathematical precision.

\subsection{Derivatives of scalar functions}

Starting from a simple function $y=f(x)$ with a scalar input and a scalar output, its derivative is defined as follows.

\begin{definition}[Derivative] 

The \textbf{derivative} of $f(x)$ is defined as:

\begin{equation}
 f'(x)= \lim_{h\rightarrow 0}\frac{f(x+h)-f(x)}{h}
\end{equation}

\end{definition}

We use a variety of notation to denote derivatives: $\partial$ will denote generically derivatives and gradients of any dimension (vectors, matrices); $\partial_x$ or $\frac{\partial}{\partial x}$ to highlight the input argument we are differentiating with respect to (when needed); while $f^\prime(x)$ is specific to scalar functions and it is sometimes called \textit{Lagrange's notation}. 

We are not concerned here about the existence of the derivative of the function (which is not guaranteed everywhere even for a continuous function), which we assume as given. We will only touch upon this point when discussing derivatives of non-smooth functions, such as $f(x) = \lvert x \rvert$ in $0$ later on in Chapter \ref{chap:automatic_differentiation}.

Derivatives of simple functions can be obtained by direct application of the definition, e.g., the derivatives of a polynomial, logarithm, or sine should be familiar:
\begin{align*}
\partial x^p = px^{p-1} \\
\partial \log(x) = \frac{1}{x} \\
\partial \sin(x) = \cos(x)
\end{align*}

\begin{SCfigure}
    \centering
    \hspace{1em}\includegraphics[width=0.6\textwidth]{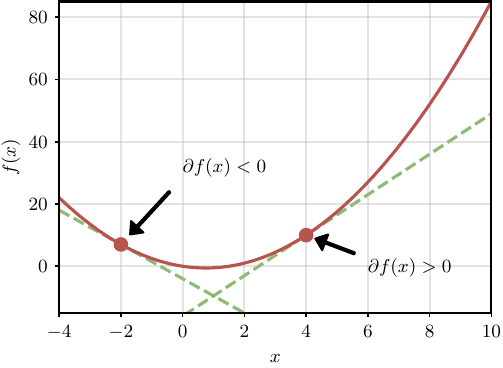}
    \caption{Plot of the function $f(x)=x^2 -1.5x$, shown along with the derivatives on two separate points.}
    \label{fig:derivative}
\end{SCfigure}

Geometrically, the derivative can be understood as the slope of the tangent passing through a point, or equivalently as the best first-order approximation of the function itself in that point, as shown in Figure \ref{fig:derivative}. This is a fundamental point of view, because the slope of the line tells us how the function is evolving in a close neighborhood: for a positive slope, the function is increasing on the right and decreasing on the left (again, for a sufficiently small interval), while for a negative slope the opposite is true. As we will see, this insight extends to vector-valued functions.

We recall some important properties of derivatives that extend to the multi-dimensional case:
\begin{itemize}
    \item \textbf{Linearity}: the derivative is linear, so the derivative of a sum is the sum of derivatives:
    $$
    \partial \Big[{\color{drawred}f(x)} + {\color{drawgreen}g(x)}\Big] = {\color{drawred}{f}^\prime(x)} + {\color{drawgreen}g^\prime(x)} \,.
    $$
    \item \textbf{Product rule}:
    $$
    \partial\Big[ {\color{drawred}f(x)}{\color{drawgreen}g(x)}\Big] = {\color{drawred}{f}^\prime(x)}{\color{drawgreen}g(x)} + {\color{drawred}f(x)}{\color{drawgreen}g'(x)} \,,
    $$
    \item \textbf{Chain rule}: the derivative of function composition is given by multiplying the corresponding derivatives:
    \begin{equation}
        \partial \Big[{\color{drawred}f(}{\color{drawgreen}g(x)}{\color{drawred})}\Big] = {\color{drawred}{f^\prime}(}{\color{drawgreen}g(x)}{\color{drawred})}{\color{drawgreen}g'(x)} \,
    \end{equation}
\end{itemize}

\subsection{Gradients and directional derivatives}

Consider now a function $y = f(\mathbf{x})$ taking a vector $\mathbf{x} \sim(d)$ as input. Talking about infinitesimal perturbations here does not make sense unless we specify the direction of this perturbation (while in the scalar case we only had “left” and “right”, in this case we have infinite possible directions in the Euclidean space). In the simplest case, we can consider moving along the $i$-th axis, keeping all other values fixed:
\begin{equation}
\partial_{x_i} f(\mathbf{x}) = \frac{\partial y}{\partial x_i} = \lim_{h \rightarrow 0} \frac{f(\mathbf{x} + h\mathbf{e}_i) - f(\mathbf{x})}{h} \,,
\label{eq:partial_derivative}
\end{equation}
where $\mathbf{e}_i \sim (d)$ is the $i$-th basis vector (the $i$-th row of the identity matrix):
\begin{equation}
\idx{\mathbf{e}_i}{j} = \begin{cases} 1 & \text{ if } i =j \\ 0 & \text{otherwise} \end{cases}
\label{eq:basis_vector}
\end{equation}
\eqref{eq:partial_derivative} is called a \textbf{partial derivative}. Stacking all partial derivatives together gives us a $d$-dimensional vector called the \textbf{gradient} of the function.

\begin{definition}[Gradient] \addbottle

The \textbf{gradient} of a function $y = f(\mathbf{x})$ is given by:
\begin{equation}
    \nabla f(\mathbf{x}) = \partial f(\mathbf{x})=\begin{bmatrix} \partial_{x_1} f(\mathbf{x}) \\ \vdots \\ \partial_{x_d} f(\mathbf{x}) \end{bmatrix}
\end{equation}
\end{definition}

Because gradients are fundamental, we use the special notation $\nabla f(\mathbf{x})$ to distinguish them. What about displacements in a general direction $\mathbf{v}$? In this case we obtain the \textbf{directional derivative}:
\begin{equation}
\mathrm{D}_{\mathbf{v}}f(\mathbf{x}) = \lim_{h \rightarrow 0} \frac{f(\mathbf{x} + h\mathbf{v}) - f(\mathbf{x})}{h} \,,
\label{eq:directional_derivative}
\end{equation}
Movement in space can be decomposed by considering individual displacements along each axis, hence it is easy to prove that the directional derivative is given by the dot product of the gradient with the displacement vector $\mathbf{v}$:
\begin{equation}
\mathrm{D}_\mathbf{v} f(\mathbf{x})=\langle\nabla f(\mathbf{x}),\mathbf{v}\rangle = \sum_i \eqnmarkbox[drawred]{node}{\partial_{x_i} f(\mathbf{x})v_i}
\label{eq:directional_derivative_dot_product}
\end{equation}
\annotate[yshift=-1em]{below,left}{node}{Displacement on the $i$-th axis}

Hence, knowing how to compute the gradient of a function is enough to compute all possible directional derivatives.

\subsection{Jacobians} \addclock 

Let us now consider the generic case of a function $\mathbf{y} = f(\mathbf{x})$ with a vector input $\mathbf{x} \sim (d)$ as before, and this time a \textit{vector} output $\mathbf{y} \sim(o)$. As we will see, this is the most general case we need to consider. Because we have more than one output, we compute a gradient for each output, and their stack provides an $(o, d)$ matrix called the \textbf{Jacobian} of $f$. 

\begin{definition}[Jacobian] 

The \textbf{Jacobian} matrix of a function $\mathbf{y} = f(\mathbf{x})$, $\mathbf{x} \sim (d)$, $\mathbf{y} \sim (o)$ is given by:
\begin{equation}
\partial f(\mathbf{x}) = \begin{pmatrix}				\frac{\partial y_1}{\partial x_1} & \dots & \frac{\partial y_1}{\partial x_d} \\				\vdots & \ddots & \vdots \\				\frac{\partial y_o}{\partial x_1} & \dots & \frac{\partial y_o}{\partial x_d} \\			\end{pmatrix} \sim (o,d)
\end{equation}
\end{definition}

We recover the gradient for $o=1$, and the standard derivative for $d=o=1$. Jacobians inherit all the properties of derivatives: importantly, the Jacobian of a composition of functions is now a \textit{matrix multiplication} of the corresponding individual Jacobians:
\begin{equation}
\partial\left[f(g(\mathbf{x}))\right] = \left[\partial f(\bullet)\right]\partial g(\mathbf{x})
\label{eq:jacobian_chain_rule}
\end{equation}
where the first derivative is evaluated in $g(\mathbf{x}) \sim (h)$. See \cite[Chapter 2]{petersen2008matrix} for numerical examples of worked out gradients and Jacobians. Like in the scalar case, gradients and Jacobians can be understood as linear functions tangent to a specific point. In particular, the gradient is the best “first-order approximation” in the following sense. For a point $\mathbf{x}_0$, the best linear approximation in an infinitesimal neighborhood of $f(\mathbf{x}_0)$ is given by:

\vspace{1em}
$$
\widetilde{f}(\mathbf{x})= f(\mathbf{x}_0)+\langle \eqnmarkbox[drawred]{node}{\partial f(\mathbf{x}_0)}, \eqnmarkbox[drawgreen]{node2}{\mathbf{x}-\mathbf{x}_0} \rangle
$$
\annotate[yshift=1em]{above,right}{node}{Slope of the line}
\annotate[yshift=-1em]{below,left}{node2}{Displacement from $\mathbf{x}_0$}

\begin{SCfigure}
    \centering
    \hspace{1em}\includegraphics[width=0.6\textwidth]{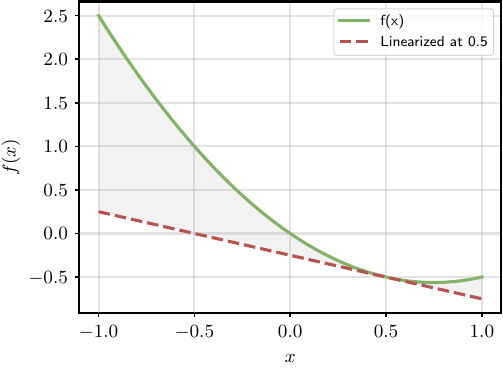}
    \caption{The function $f(x)=x^2-1.5x$ and its first-order approximation shown in $0.5$.}
    \label{fig:taylor_approximation}
\end{SCfigure}

\vspace{0.5em}
This is called \textbf{Taylor's theorem}. See Box \ref{code:taylor_approximation} and Figure \ref{fig:taylor_approximation} for a visualization in the scalar case $f(x) = x^2 - 1.5x$.

\begin{mypy}{Example of computing a first-order approximation (scalar case). The result is plotted in Figure \ref{fig:taylor_approximation}.}{code:taylor_approximation}
# Generic function
f = lambda x: x**2-1.5*x

# Derivative (computed manually for now) 
df = lambda x: 2*x-1.5 

# Linearization at 0.5 
x = 0.5
f_lin = lambda h: f(x) + df(x)*(h-x) 

# Numerical check
print(f(x + 0.01))     # -0.5049
print(f_lin(x + 0.01)) # -0.5050
\end{mypy}

\subsection*{On the dimensionality of the Jacobians \addteacup}

We close with a pedantic note on dimensionality that will be useful in the following.   Consider the following function:

$$
\mathbf{y} = \mathbf{W}\mathbf{x}
$$

When viewed as a function of $\mathbf{x}$, the derivative is, as before, an $(o, d)$ matrix, and it can be shown that:

$$
\partial_{\mathbf{x}}\left[\mathbf{W}\mathbf{x}\right] =\mathbf{W}
$$

When viewed as a function of $\mathbf{W}$, instead, the input is itself an $(o, d)$ matrix, and the “Jacobian” in this case has shape $(o,o,d)$ (see box in the following page). However, we can always imagine an identical (isomorphic) function taking as input the vectorized version of $\mathbf{W}$, $\text{vect}(\mathbf{W})  \sim(od)$, in which case the Jacobian will be a matrix of shape $(o, od)$.

\begin{supportbox}{Working out the Jacobian}
To compute the Jacobian $\partial_{\mathbf{W}} \mathbf{W}\mathbf{x}$, we can rewrite the expression element-wise as:
$$
y_i=\sum_j W_{ij}x_j
$$
from which we immediately find that:
\begin{equation}
\frac{\partial y_i}{\partial W_{ij}}=x_j
\label{eq:partial_yi_wij}
\end{equation}
Note that to materialize the Jacobian explicitly (store it in memory), we would need a lot of repeated values. As we will see in Chapter \ref{chap:automatic_differentiation}, this can be avoided because, in practice, we only care about the application of the Jacobian on another tensor.
\label{supportbox:jacobian}
\end{supportbox}

This quick example clarifies what we mean by our statement that working with vector inputs and outputs is enough from a notational point of view. However, it will be important to keep this point in mind in Chapter \ref{chap:automatic_differentiation}, when we will use matrix Jacobians for simplicity of notation (in particular, to avoid the proliferation of indices), but the sizes of these Jacobians may “hide” inside the actual shapes of the inputs and the outputs, most importantly the batch sizes. We will see in Chapter \ref{chap:automatic_differentiation} that explicit computation of Jacobians can be avoided in practice by considering the so-called \textbf{vector-Jacobian products}. This can also be formalized by viewing Jacobians as abstract linear maps - see \cite{blondel2024elements} for a formal overview of this topic.

\section{Gradient descent} 

\addclock To understand the usefulness of having access to gradients, consider the problem of  minimizing a generic function $f(\mathbf{x})$, with $\mathbf{x} \sim (d)$: 
\begin{equation}
\mathbf{x}^* = \underset{\mathbf{x}}{\arg\min} \; f(\mathbf{x})
\label{eq:optimization}
\end{equation}

where, similarly to $\argmax$, $\arg\min \; f(\mathbf{x})$ denotes the operation of finding the value of $\mathbf{x}$ corresponding to the lowest possible value of $f(\mathbf{x})$. We assume the function has a single output (\textbf{single-objective optimization}), and that the domain over which we are optimizing $\mathbf{x}$ is unconstrained.

In the rest of the book $\mathbf{x}$ will encode the parameters of our model, and $f$ will describe the performance of the model itself on our data, a setup called \textbf{supervised learning} that we introduce in the next chapter. We can consider minimizing instead of maximizing with no loss of generality, since minimizing $f(\mathbf{x})$ is equivalent to maximizing $-f(\mathbf{x})$ and vice versa (to visualize this, think of a function in 1D and rotate it across the $x$-axis, picturing what happens to its low points).

In very rare cases, we may be able to express the solution in closed-form (we will see one example in the context of least-squares optimization in Section \ref{subsec:least_squares}). In general, however, we are forced to resort to iterative procedures. Suppose we start from a random guess $\mathbf{x}_0$ and that, for every iteration, we take a step, that we decompose in terms of its magnitude $\eta_t$ (the length of the step) and the direction $\mathbf{p}_t$:

\vspace{0.5em}
\begin{equation}
\mathbf{x}_t = \eqnmarkbox[drawred]{node}{\mathbf{x}_{t-1}} + \eqnmarkbox[drawblue]{node2}{\eta_t\mathbf{p}_t}
\label{eq:iterative_descent}
\annotate[yshift=1em]{above,left}{node}{Guess at iteration $t$}
\annotate[yshift=-1em]{below,left}{node2}{Displacement at iteration $t$}
\end{equation}

\vspace{1em}
We call $\eta_t$ the \textbf{step size} (or, in machine learning terminology, the \textbf{learning rate}, for reasons that will become clear in the next chapter). A direction $\mathbf{p}_t$ for which there exists an $\eta_t$ such that $f(\mathbf{x}_t) \le f(\mathbf{x}_{t-1})$ is called a \textbf{descent direction}. If we can select a descent direction for every iteration, and if we are careful in the choice of step size, the iterative algorithm in \eqref{eq:iterative_descent} will converge to a minimum in a sense to be described shortly.

For differentiable functions, we can precisely quantify all descent directions by using the directional derivative from \eqref{eq:directional_derivative}, as they can be defined as the directions inducing a negative change with respect to our previous guess $\mathbf{x}_{t-1}$:
$$
\mathbf{p}_t \text{ is a descent direction} \;\;\Rightarrow\;\; \mathrm{D}_{\mathbf{p}_t} f(\mathbf{x}_{t-1}) \le 0
$$
Using what we learned in Section \ref{sec:gradients_and_jacobians} and the definition of the dot product in terms of cosine similarity from \eqref{eq:dot_product_cosine} we get:
$$
\mathrm{D}_{\mathbf{p}_t} f(\mathbf{x}_{t-1})=\langle \nabla f(\mathbf{x}_{t-1}), \mathbf{p}_t\rangle=\lVert\nabla f(\mathbf{x}_{t-1})\rVert \lVert\mathbf{p}_t\rVert \cos(\alpha)
$$
where $\alpha$ is the angle between $\mathbf{p}_t$ and $\nabla f(\mathbf{x}_{t-1})$. Considering the expression on the right, the first term is a constant with respect to $\mathbf{p}_t$. Because we have assumed $\mathbf{p}_t$ only encodes the direction of movement, we can also safely restrict it to $\lVert\mathbf{p}_t\rVert = 1$, rendering the second term another constant. Hence, by the properties of the cosine we deduce that any $\mathbf{p}_t$ whose angle is between $\pi/2$ and $3\pi/2$ with $\nabla f(\mathbf{x}_{t-1})$ is a descent direction. Among these, the direction $\mathbf{p}_t = -\nabla f(\mathbf{x}_{t-1})$ (with an angle of $\pi$) has the lowest possible directional derivative, and we refer to it as the \textbf{steepest descent direction}. 

Putting together this insight with the iterative procedure in \eqref{eq:iterative_descent} gives us an algorithm to minimize any differentiable function, that we call \textbf{(steepest) gradient descent}.

\begin{definition}[Steepest gradient descent] \addbottle

 Given a differentiable function $f(\mathbf{x})$, a starting point $\mathbf{x}_0$, and a step size sequence $\eta_t$, \textbf{gradient descent} proceeds as:
\begin{equation}
\mathbf{x}_{t}=\mathbf{x}_{t-1}-\eta_t\nabla f(\mathbf{x}_{t-1})
\label{eq:gradient_descent}
\end{equation}
\end{definition}

We will not be concerned with the problem of finding an appropriate step size, which we will just assume “small enough” so that the gradient descent iteration provides a reduction in $f$. In the next section we focus on what points are obtained by running gradient descent from a generic initialization. Note that gradient descent is as efficient as the procedure we use to compute the gradient: we introduce a general algorithm to this end in Chapter \ref{chap:automatic_differentiation}.

\subsection{Convergence of gradient descent}

When discussing the convergence of gradient descent, we need to clarify what we mean by “a minimizer” of a function. If you do not care about convergence and you trust gradient descent, proceed with no hesitation to the next section.

\begin{definition}[Minimum] $\,$

A \textbf{local minimum} of $f(\mathbf{x})$ is a point $\mathbf{x}^+$ such that the following is true for some $\varepsilon > 0$:
$$
f(\mathbf{x}^+)\le f(\mathbf{x}) \;\;\; \forall \mathbf{x} \; :\; \eqnmarkbox[drawred]{node}{\lVert \mathbf{x}-\mathbf{x}^+ \rVert <\varepsilon}
$$
\annotate[yshift=-1em]{below,left}{node}{Ball of size $\varepsilon$ centered in $\mathbf{x}^+$}

\end{definition}

In words, the value of $f(\mathbf{x}^+$) is a minimum if we consider a sufficiently small neighborhood of $\mathbf{x}^+$. Intuitively, in such a point the slope of the tangent will be $0$, and the gradient everywhere else in the neighborhood of $\mathbf{x}^+$ will point upwards. We can formalize the first idea by the concept of \textbf{stationary points}.

\begin{definition}[Stationary points] $\,$

A point $\mathbf{x}^+$ is called a \textbf{stationary point} of $f(\mathbf{x})$ if $\nabla f(\mathbf{x}^+)=0$.
\end{definition}

\begin{SCfigure}
    \centering
    \hspace{1em}\includegraphics[width=0.6\textwidth]{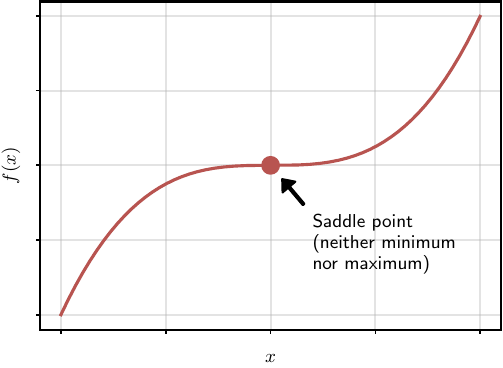}
    \caption{Simple example of a saddle point (try visualizing the tangent line in that point to see it is indeed stationary).}
    \label{fig:saddle_point}
\end{SCfigure}

Stationary points are not limited to minima: they can be maxima (the minima of $-f(\mathbf{x})$) or \textbf{saddle points}, which are inflexion points where the curvature of the function is changing (see Figure \ref{fig:saddle_point} for an example). In general, without any constraint on $f$, gradient descent can only be proven to converge to a generic stationary point depending on its initialization.

Can we do better? Picture a parabola: in this case, the function does not have any saddle points, and it only has a single minimum. This minimum is also special, in the sense that the function in that point attains its lowest possible value across the entire domain: we say this is a \textbf{global minimum}.

\begin{definition}[Global minimum] $\,$
    
    A \textbf{global minimum} of $f(\mathbf{x})$ is a point $\mathbf{x}^*$ such that $f(\mathbf{x}^*) \le f(\mathbf{x})$ for any possible input $\mathbf{x}$.
\end{definition}

Intuitively, gradient descent will converge to this global minimum if run on a parabola (from any possible initialization) because all gradients will point towards it. We can generalize this idea with the concept of \textbf{convexity} of a function. There are many possible definitions of convexity, we choose the one below for simplicity of exposition.

\begin{definition}[Convex function] $\,$

A function $f(\mathbf{x})$ is convex if for any two points $\mathbf{x}_1$ and $\mathbf{x}_2$ and $\alpha \in \left[0,1\right]$ we have:

    \vspace{1em}
    \begin{equation}
        f(\eqnmarkbox[drawblue]{node}{\alpha \mathbf{x}_1 + (1-\alpha)\mathbf{x}_2})\le \eqnmarkbox[drawred]{node2}{\alpha f(\mathbf{x}_1)+(1-\alpha)f(\mathbf{x}_2)}
        \label{eq:convex_function}
    \end{equation}
    \annotate[yshift=-1em]{below,right}{node}{Interval from $\mathbf{x}_1$ to $\mathbf{x}_2$}
    \annotate[yshift=1em]{above,left}{node2}{Line segment from $f(\mathbf{x}_1)$ to $f(\mathbf{x}_2)$}
\end{definition}

\vspace{1em}
The left-hand side in \eqref{eq:convex_function} is the value of $f$ on any point inside the interval ranging from $\mathbf{x}_1$ to $\mathbf{x}_2$, while the right-hand side is the corresponding value on a line connecting $f(\mathbf{x}_1)$ and $f(\mathbf{x}_2)$. If the function is always below the line joining any two points, it is convex (as an example, a parabola pointing upwards is convex). 

Convexity qualifies the simplicity of optimizing the function, in the following sense \cite{jain2017non}:
\begin{enumerate}
    \item For a generic \textit{non-convex} function, gradient descent converges to a \textit{stationary point}. Nothing more can be said unless we look at higher-order derivatives (derivatives of the derivatives).
    \item For a \textit{convex} function, gradient descent will converge to a \textit{global minimum}, irrespective of initialization.
    \item  If the inequality in \eqref{eq:convex_function} is satisfied in a strict way (\textbf{strict convexity}), the global minimizer will also be \textit{unique}.
\end{enumerate}

This is a hard property: to find a global minimum in a non-convex problem with gradient descent, the only solution is to run the optimizer infinite times from any possible initialization, turning it into an NP-hard task \cite{jain2017non}.

This discussion has a strong historical significance. As we will see in Chapter \ref{chap:fully_connected_models}, any non-trivial model is non-convex, meaning that its optimization problem may have several stationary points. This is in contrast to alternative algorithms for supervised learning, such as support vector machines, which maintain non-linearity while allowing for convex optimization. Interestingly, complex differentiable models seem to work well even in the face of such restriction, in the sense that their optimization, when started from a reasonable initialization, converge to points with good empirical performance.

\subsection{Accelerating gradient descent}
The negative gradient describes the direction of steepest descent, but only in an infinitesimally small neighborhood of the point. As we will see in Chapter \ref{chap:fully_connected_models} (where we introduce stochastic optimization), these directions can be extremely noisy, especially when dealing with large models. A variety of techniques have been developed to accelerate convergence of the optimization algorithm by selecting better descent directions. For computational reasons, we are especially interested in methods that do not require higher-order derivatives (e.g., the Hessian), or multiple calls to the function.

\begin{figure}[t]
    \centering
    \includegraphics[width=0.7\textwidth]{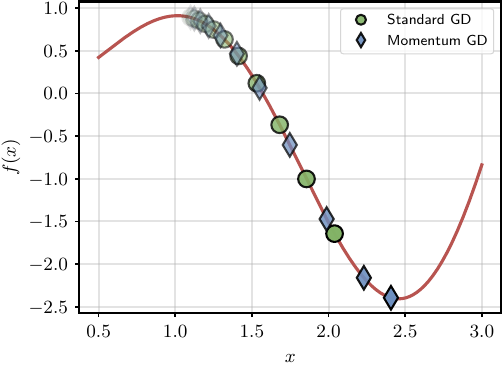}
    \caption{GD and GD with momentum when minimizing the function $x\sin(2x)$ starting from $x=1+\varepsilon$, with $\lambda=0.3$.}
    \label{fig:momentum}
\end{figure}

We describe here one such technique, \textbf{momentum}, and we refer to \cite[Chapter 12]{zhang2023dive}, for a broader introduction.\footnote{See also this 2016 blog post by S. Ruder: \url{https://www.ruder.io/optimizing-gradient-descent/}.} If you picture gradient descent as a ball “rolling down a hill”, the movement is relatively erratic, because each gradient can point in a completely different direction (in fact, for a perfect choice of step size and a convex loss function, any two gradients in subsequent iterations will be orthogonal). We can smooth this behavior by introducing a “momentum” term that conserves some direction from the previous gradient iteration:

\begin{align*}
\mathbf{g}_t& =- \eqnmarkbox[drawred]{node}{\eta_t\nabla f(\mathbf{x}_{t-1})} + \eqnmarkbox[drawblue]{node2}{\lambda\mathbf{g}_{t-1}} \\ 
\mathbf{x}_{t}& =\mathbf{x}_{t-1}+\mathbf{g}_t
\annotate[yshift=1em]{above,left}{node}{Steepest descent}
\annotate[yshift=1em]{above,right}{node2}{Momentum term}
\end{align*}

where we initialize $\mathbf{g}_0 = \mathbf{0}$. See Figure \ref{fig:momentum} for an example. The coefficient $\lambda$ determines how much the previous term is dampened. In fact, unrolling two terms:

\begin{gather*}
\mathbf{g}_t=-\eta_t\nabla f(\mathbf{x}_{t-1}) +\lambda(-\eta_t\nabla f(\mathbf{x}_{t-2}) +\lambda\mathbf{g}_{t-2}) \\ = -\eta_t\nabla f(\mathbf{x}_{t-1}) -\lambda\eta_t\nabla f(\mathbf{x}_{t-2}) +\lambda^2\mathbf{g}_{t-2}
\end{gather*}

Generalizing, the iteration at time $t-n$ gets dampened by a factor $\lambda^{n-1}$. Momentum can be shown to accelerate training by smoothing the optimization path \cite{sutskever2013importance}. Another common technique is adapting the step size for each parameter based on the gradients’ magnitude \cite{zhang2023dive}. A common optimization algorithm combining several of these ideas is Adam \cite{kingma2014adam}. One advantage of Adam is that it is found to be relatively robust to the choice of its \textbf{hyper-parameters},\footnote{A hyper-parameter is a parameter which is selected by the user, as opposed to being learnt by gradient descent.} with the default choice in most frameworks being a good starting point in the majority of cases. Designing novel optimizers that can `unseat' Adam (or its variations, such as AdamW \cite{loshchilov2018fixing}) from its place as \textit{de-facto} default optimizer in deep learning remains an open research problem, e.g., see \cite{bernstein2024old} for recent work on designing customized optimizers for neural networks from first principles.

One disadvantage of using accelerated optimization algorithms can be increased storage requirements: for example, momentum requires us to store the previous gradient iteration in memory, doubling the space needed by the optimization algorithm (although in most cases, the memory required to compute the gradient is the most influential factor in terms of memory, as we will see in Section \ref{sec:reverse_mode_automatic_differentiation}).

\clearpage

\section*{From theory to practice}

\begin{supportbox}{About the exercises}
This book does not have classical end-of-chapter exercises, which are covered in many existing textbooks. Instead, I propose a self-learning path to help you explore two frameworks (JAX and PyTorch) as you progress in the book. Solutions to the exercises will be published on the book's website.\footnote{\url{https://www.sscardapane.it/alice-book}} These sections are full of URLs linking to online material -- they might be expired or moved by the time you search for them.
\end{supportbox}

\subsection*{Starting from the basics}

\begin{wrapfigure}{R}{3.0cm}
\vspace{-4em}\includegraphics[width=3.0cm]{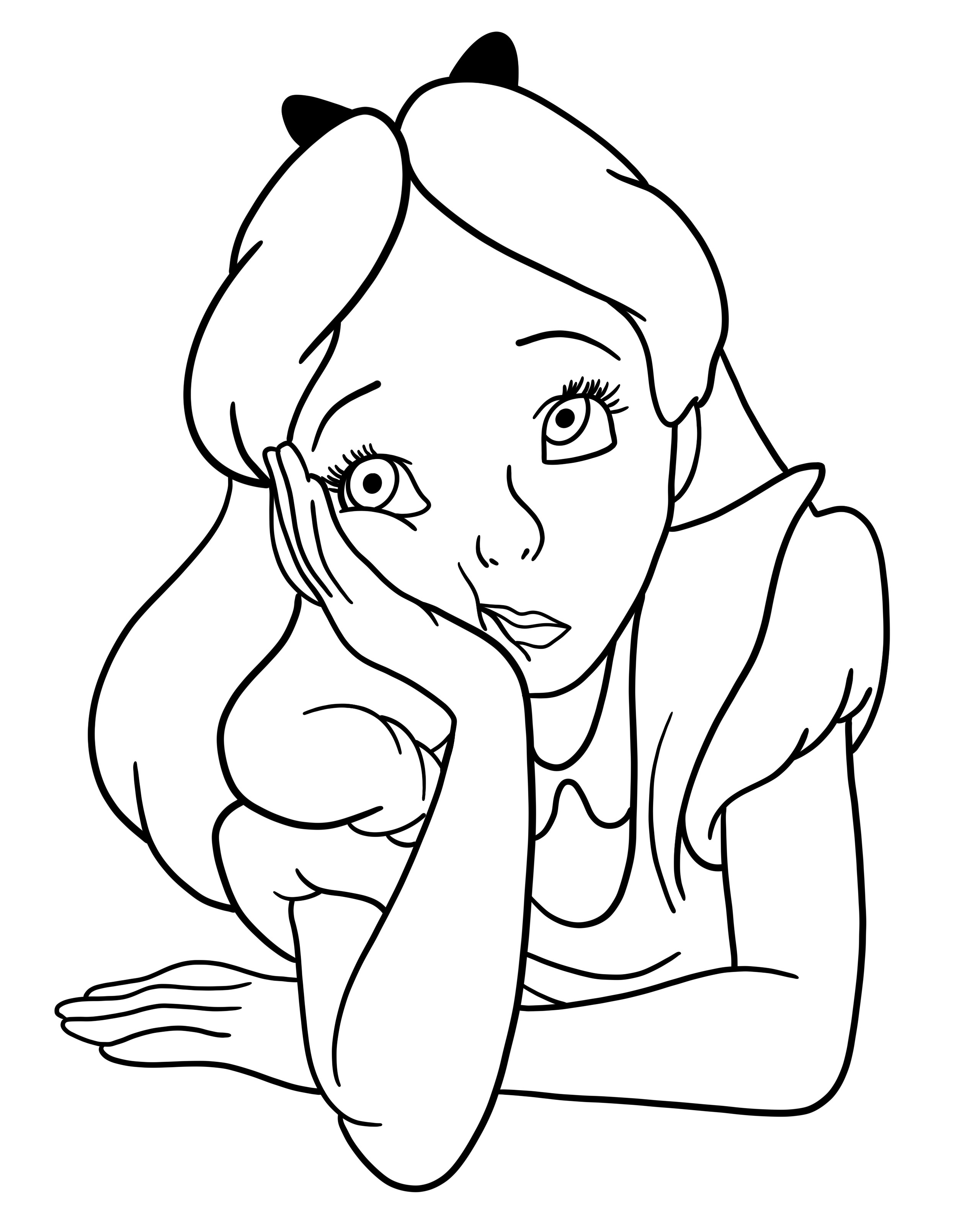}
\vspace{-2em}
\end{wrapfigure} 

The starting block for any designer of differentiable models is a careful study of NumPy. NumPy implements a generic set of functions to manipulate multidimensional arrays (what we call \textit{tensors} in the book), as long as functions to index and transform their content. You can read more on the library's quick start.\footnote{\url{https://numpy.org/doc/stable/user/quickstart.html}} You should feel comfortable in handling arrays in NumPy, most notably for their indexing: the {\footnotesize\verb+rougier/numpy-100+}\footnote{\url{https://github.com/rougier/numpy-100}} repository provides a nice, slow-paced series of exercises to test your knowledge.

\subsection*{Moving to a realistic framework}

Despite its influence, NumPy is limited in his support for parallel hardware such as GPUs (unless additional libraries are used), and for his lack of automatic differentiation (introduced in Chapter \ref{chap:automatic_differentiation}). JAX replicates the NumPy's interface while adding extended hardware support, the automatic computation of gradients, and additional transformations such as the \textbf{vectorized map} (\mintinline{python}{jax.vmap}). Frameworks such as PyTorch also implement a NumPy-like interface at their core, but they make minor adjustments in nomenclature and functionality and they add  high-level utilities for building differentiable models. Take your time to skim the documentation of \mintinline{python}{jax.numpy.array} and \mintinline{python}{torch.tensor} to understand how much they have in common with NumPy. For now, you can ignore high-level modules such as \mintinline{python}{torch.nn}. We will have more to say about how these frameworks are designed in Chapter \ref{chap:automatic_differentiation}, after we introduce their gradient computation mechanism.

\subsection*{Implementing a gradient descent algorithm}

To become proficient with all three frameworks (NumPy, JAX, PyTorch), I suggest to replicate the exercise below thrice -- each variant should only take a few minutes if you know the syntax.  Consider a 2D function $f(\mathbf{x})$, $\mathbf{x} \sim (2)$, where we take the domain to be $[0,10]$:\footnote{I asked ChatGPT to generate a nice function with several minima and maxima. Nothing else in the book is LLM-generated, which I feel is becoming an important disclaimer to make.}
\begin{equation*}
    f(\mathbf{x}) = \sin(x_1) \cos(x_2) + \sin(0.5x_1) \cos(0.5x_2)
\end{equation*}
Before proceeding in the book, repeat this for each framework:
\begin{enumerate}
\item Implement the function in a \textbf{vectorized} way, i.e., given a matrix $\mathbf{X} \sim (n,2)$ of $n$ inputs, it should return a vector $f(\mathbf{X}) \sim (n)$ where $\idx{f(\mathbf{X})}{i} = f(\mathbf{X}_i)$.
\item Implement another function to compute its gradient (hard-coded -- we have not touched automatic differentiation yet).
\item Write a basic gradient descent procedure and visualize the paths taken by the optimization process from multiple starting points.
\item Try adding a momentum term and visualizing the norm of the gradients, which should converge to zero as the algorithm moves towards a stationary point.
\end{enumerate}
If you are using JAX or PyTorch to solve the exercise, point (3) is a good place to experiment with \mintinline{python}{vmap} for vectorizing a function.

%% file: 3_datasets_and_loss_functions.tex
\chapter{Datasets and losses}
\label{chap:supervised_learning}

\begin{supportbox}{About this chapter}
This chapter formalizes the supervised learning scenario. We introduce the concepts of datasets, losses,  empirical risk minimization, and the basic assumptions made in supervised learning. We close by providing a probabilistic formulation of supervised learning built on the notion of maximum likelihood. This short chapter serves as the backbone for the rest of the book.
\end{supportbox}

\section{What is a dataset?}
\label{sec:dataset}

We consider a scenario in which manually coding a certain function is unfeasible (e.g., recognizing objects from real-world images), but gathering \textbf{examples} of the desired behaviour is sufficiently easy. Examples of this abound, ranging from speech recognition to robot navigation. We formalise this idea with the following definition.

\clearpage

\begin{definition}[Dataset] \addbottle

A \textbf{supervised dataset} $\mathcal{S}_n$ of size $n$ is a set of $n$ pairs $\mathcal{S}_n = \left\{(x_i, y_i)\right\}_{i=1}^n$, where each $(x_i, y_i)$ is an example of an input-output relationship we want to model. We further assume that each example is an \textbf{identically} and \textbf{independently} distributed (i.i.d.) draw from some unknown (and unknowable) probability distribution $p(x,y)$.
\end{definition}

See Appendix \ref{chap:probability_theory} if upon reading the definition you want to brush up on probability theory. The last assumption appears technical, but it is there to ensure that the relationship we are trying to model is meaningful. In particular, samples being \textbf{identically distributed} means that we are trying to approximate something which is sufficiently stable and unchanging through time. As a representative example, consider the task of gathering a dataset to recognise car models from photos. This assumption will be satisfied if we collect images over a short time span, but it will be invalid if collecting images from the last few decades, since car models will have changed over time. In the latter case, training and deploying a model on this dataset will fail as it will be unable to recognise new models or will have sub-optimal performance when used.

Similarly, samples being \textbf{independently distributed} means that our dataset has no bias in its collection, and it is sufficiently representative of the entire distribution. Going back to the previous example, gathering images close to a Tesla dealership will be invalid, since we will collect an overabundance of images of a certain type while loosing on images of other makers and models. Note that the validity of these assumptions depends on the context: a car dataset collected in Italy may be valid when deploying our model in Rome or Milan, while it may be invalid when deploying our model in Tokyo or in Taiwan. The i.i.d. assumption should always be checked carefully to ensure we are applying our supervised learning tools to a valid scenario. Interestingly, modern LLMs are trained on such large distributions of data that even understanding what tasks are truly \textit{in-distribution} against what is \textit{out-of-distribution} (and how much the models are able to generalize) becomes blurred \cite{yuan2024revisiting}.

\vspace{1em}
\begin{supportbox}{More on the i.i.d. property}
Importantly, ensuring the i.i.d. property is not a one-shot process, and it must be checked constantly during the lifetime of a model. In the case of car classification, if unchecked, subtle changes in the distribution of cars over time will degrade the performance of a machine  learning model, an example of \textbf{domain shift}. As another example, a recommender system will change the way users interact with a certain app, as they will start reacting to suggestions of the recommender system itself. This creates \textbf{feedback loops} \cite{cinus2022effect} that require constant re-evaluation of the performance of the system and of the app.
\end{supportbox}

\begin{figure}[p]
    \centering
    \includegraphics[width=0.95\textwidth]{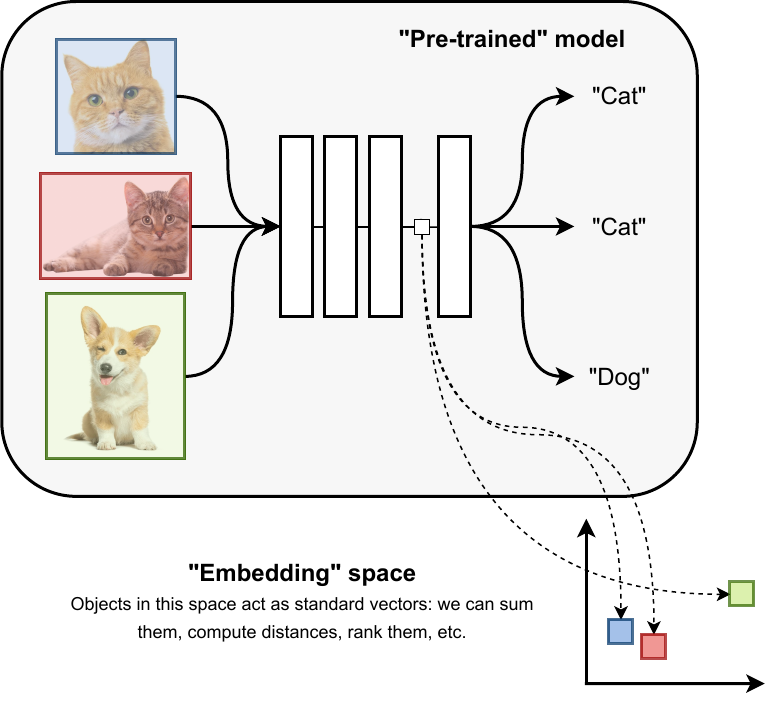}
    \caption{Differentiable models process data by transforming it sequentially via linear algebra operations. In many cases, after we optimize these programs, the internal representations of the input data of the model (what we call a \textbf{pre-trained} model) have geometric properties: for example, semantically similar images are projected to points that are close in this “latent” space. Transforming data from a non-metric space (original input images) to a metric space (bottom right) is called \textbf{embedding} the data.}
    \label{fig:embedding}
\end{figure}

\subsection{Variants of supervised learning}
\label{subsec:variants_supervised_learning}

There exists many variations on the standard supervised learning scenario, although most successful applications make use of supervised learning in some form or another. For example, some datasets may not have available \textbf{targets} $y_i$, in which case we  talk about \textbf{unsupervised} learning. Typical applications of unsupervised learning are \textbf{clustering} algorithms, in which we want to aggregate our input data into \textit{clusters} such that points in a cluster are similar and points between clusters are dissimilar \cite{hastie2009elements}. As another example, in a \textbf{retrieval} system we may want to search a large database for the top-$k$ most similar elements to a user-given query.

When dealing with complex data such as images, this is non-trivial because distances on images are ill-defined if we operate on pixels (i.e., even small perturbations can modify millions of pixels). However, assume we have available some differentiable model that we have already optimized for some other task which we assume sufficiently generic, e.g., image classification. We call it a  \textbf{pre-trained} model. As we will see, the internal states of this model can be interpreted as vectors in a high-dimensional space. In many cases, these vectors are shown to have useful geometrical properties, in the sense that objects that are semantically similar are sent (\textbf{embedded}) into points that are close in these representations. Hence, we can use these latent representations with standard clustering models, such as Gaussian mixture models \cite{huang2014deep}. See Figure \ref{fig:embedding} for a high-level overview of this idea.

What if we do not have access to a pre-trained model? A common variation of unsupervised learning is called \textbf{self-supervised} learning (SSL, \cite{zbontar2021barlow}). The aim of SSL is to automatically find some supervised objective from a generic unsupervised dataset, in order to optimize a model that can be used in a large set of downstream tasks. For example, if we have access to a large corpus of text, we can always optimize a program to predict how a small piece of text is likely to continue \cite{radford2019language}. The realization that neural networks can also perform an efficient embedding of text  when pre-trained in a self-supervised way had a profound impact on the community \cite{mikolov2013distributed}.\footnote{Large-scale web datasets are also full of biases, profanity, and vulgar content. Recognizing that models trained on this data internalize these biases was another important realization \cite{bolukbasi2016man} and it is one of the major criticisms of closed-source foundation models \cite{bender2021dangers}.}

\begin{figure}[t]
    \centering
    \includegraphics[width=\textwidth]{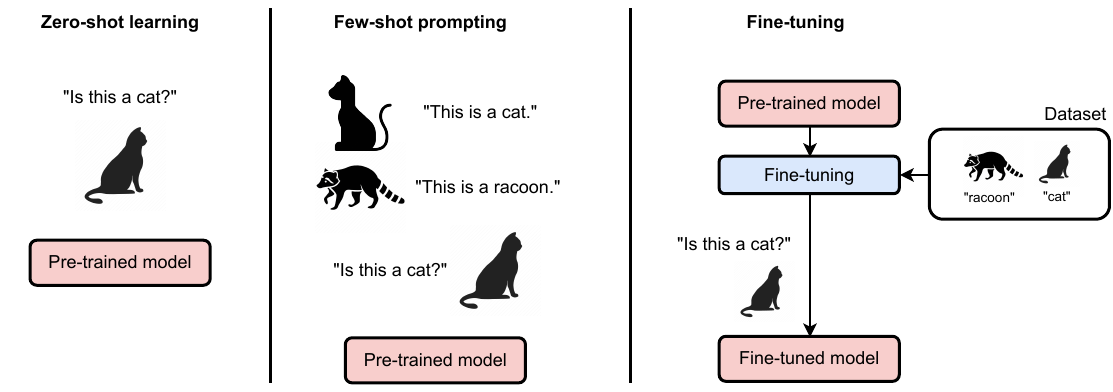}
    \caption{Three ways of using trained models. \textbf{Zero-shot}: a question is directly given to the model. This can be achieved with generative language models (introduced in Chapter \ref{chap:convolutions_beyond_images}). \textbf{Few-shot prompting} is similar, but a few examples are provided as input. Both techniques can be employed only if the underlying model shows a large amount of generalization capabilities. \textbf{Fine-tuning}: the model is optimized via gradient descent on a small dataset of examples. This proceeds similarly to training the model from scratch.}
    \label{fig:using_models}
\end{figure}

As we will see in Chapter \ref{chap:convolutions_beyond_images} and Chapter \ref{chap:transformers}, LLMs can be seen as modern iterations on this basic idea, since optimizing models such as GPT or Llama \cite{touvron2023llama} always start by a basic self-supervised training in terms of next-token prediction. These models are sometimes called \textbf{foundation models}. In the simplest case, they can be used out-of-the-box for a new task, such as answering a query: in this case, we say they are used in a \textbf{zero-shot} fashion. For LLMs, it is also possible to provide a small number of examples of a new task as input prompt, in which case we talk about \textbf{few-shot prompting}. In the most general case, we can take a pre-trained foundation model and optimize its parameters by gradient descent on a new task: this is called \textbf{fine-tuning} the model. See Figure \ref{fig:using_models} for a comparison of the three approaches. In this book we focus on building models from scratch, but fine-tuning can be done by similar means.

Fine-tuning is made especially easy by the presence of large open-source repositories online.\footnote{\url{https://huggingface.co/models}} Fine-tuning can be done on the full set of parameters of the starting model, or by considering only a smaller subset or a small number of additional parameters: this is called \textbf{parameter-efficient fine-tuning} (PEFT) \cite{lialin2023scaling}.\footnote{Few-shot learning can also be done by fine-tuning the model. In cases in which fine-tuning is not needed, we say the model is performing \textbf{in-context learning} \cite{akyurek2022learning}.} We will consider PEFT techniques in the next volume.

Many other variations of supervised learning are possible, which we do not have space to list in detail here except for some generic hints. If only parts of a dataset are labeled, we have a \textbf{semi-supervised} scenario \cite{belkin2006manifold}. We will see some examples of semi-supervised learning in Chapter \ref{chap:gnns}. Additionally, we can have scenarios with multiple datasets belonging to “similar” distributions, or the same distribution over different period of times, giving rise to countless problems depending on the order in which the tasks or the data are provided, including \textbf{domain adaptation}, \textbf{meta-learning} \cite{finn2017model}, \textbf{continual learning} \cite{parisi2019continual,biesialska2020continual}, \textbf{metric learning}, \textbf{unlearning}, etc. Some of these will be treated in the next volume.

\clearpage

\section{Loss functions}
\label{sec:loss_functions}

\addclock Once data has been gathered, we need to formalize our idea of “approximating” the desired behavior, which we do by introducing the concept of \textbf{loss functions}.

\begin{definition}[Loss function] \addbottle

Given a desired target $y$ and the predicted value $\hat{y}=f(x)$ from a model $f$, a \textbf{loss function} $l(y, \hat{y}) \in \mathbb{R}$ is a scalar, differentiable function whose value correlates with the performance of the model, i.e., $l(y, \hat{y}_1) < l(y, \hat{y}_2)$ means that the prediction $\hat{y}_1$ is better than the prediction $\hat{y}_2$ when considering the reference value (target) $y$.
\end{definition}

A loss function embeds our understanding of the task and our preferences in the solutions’ space on a real-valued scale that can be exploited in an optimization algorithm. Being differentiable, it allows us to turn our learning problem into a mathematical optimization problem that can be solved via gradient descent by minimizing the average loss on our dataset.

To this end, given a dataset $\mathcal{S}_n = \left\{(x_i, y_i)\right\}$ and a loss function $l(\cdot, \cdot)$, a sensible optimization task to solve is the minimum average loss on the dataset achievable by any possible \textit{differentiable} model $f$:

\vspace{1em}
\begin{equation}
    f^* = \underset{f}{\arg\min} \;\; \eqnmarkbox[drawred]{node}{\frac{1}{n}\sum_{i=1}^n} l(y_i, \eqnmarkbox[drawblue]{node2}{f(x_i)})
    \label{eq:empirical_risk_minimization}
\end{equation}
\annotate[yshift=1em]{above,right}{node}{Average over the dataset}
\annotate[yshift=-1.5em]{below,left}{node2}{Prediction on the $i$-th  sample}

\vspace{1em}
For historical reasons, \eqref{eq:empirical_risk_minimization} is referred to as \textbf{empirical risk minimization} (ERM), where \textit{risk} is used as a generic synonym for \textit{loss}. See also the box in the next page for more on the origin of the term.

In \eqref{eq:empirical_risk_minimization} we are implicitly assuming that we are minimizing across the space of all possible functions defined on our input $x$. We will see shortly that our models can always be parameterized by a set of tensors $w$ (called \textbf{parameters} of the model), and minimization is done by searching for the optimal value of these parameters via numerical optimization, which we denote by $f(x,w)$. Hence, given a dataset $\mathcal{S}_n$, a loss function $l$, and a model space $f$, we can \textbf{train} our model by optimizing the empirical risk \eqref{eq:empirical_risk_minimization} via gradient descent \eqref{eq:gradient_descent}:

\begin{equation}
    {\color{drawred}w^*} = \underset{w}{\arg\min} \;\; \frac{1}{n}\sum_{i=1}^n l(y_i, f(x_i, {\color{drawred}w}))
    \label{eq:empirical_risk_minimization_2}
\end{equation}

where the minimization is now done with respect to the parameter's tensor $w$.
\subsection*{On the differentiability of the loss}
\label{subsec:differentiability_loss}

 Before proceeding, we make two observations on the ERM framework. \addteacup First, note that the differentiability requirement on $l$ is fundamental. Consider a simple \textbf{binary classification} task (that we will introduce properly in the next chapter), where $y \in \left\{-1,+1\right\}$ can only take two values, $-1$ or $1$. Given a real-valued model $f(x) \in \mathbb{R}$, we can equate the two decisions with the sign of $f$ -- which we denote as $\text{sign}(f(x))$ -- and define a \textbf{0/1 loss} as:

\begin{equation}
l(y, \hat{y})=\begin{cases} 0 & \text{ if } \text{sign}(\hat{y}) = y \\ 1 & \text{ otherwise } \end{cases}
\label{eq:01_loss}
\end{equation}

While this aligns with our intuitive notion of “being right”, it is useless as loss function since its gradient will almost always be zero (except when the sign of $f$ switches), and any gradient descent algorithm will remain stuck at initialization. A less intuitive quantity in this case is the \textbf{margin} $y\hat{y}$, which is positive [negative] depending on whether the sign of the model aligns [or does not align] with the desired one, but it varies continuously differently from 0/1 loss in \eqref{eq:01_loss}. A possible loss function in this case is the \textbf{hinge loss} $l(y,\hat{y}) = \max(0, 1 - y\hat{y})$, which is used to train support-vector models. Details apart, this shows the inherent tension between designing loss functions that encode our notion of performance while at the same time being useful for numerical optimization.

\begin{supportbox}{Risk and loss}
Empirical and expected risk minimization framed in this way are generally associated with the work of the Russian computer scientist V. Vapnik \cite{vapnik2013nature}, which gave rise to the field of \textit{statistical learning theory} (SLT). SLT is especially concerned with the behaviour of \eqref{eq:empirical_risk_minimization} when seen as a finite-sample approximation of \eqref{eq:expected_risk} under some restricted class of functions $f$ and measure of underlying complexity \cite{poggio2003mathematics,shalev2014understanding,mohri2018foundations}. The counter-intuitive properties of modern neural networks (such as strong generalization long after overfitting should have been expected) have opened many new avenues of research in SLT \cite{poggio2020theoretical}. See also the introduction of Chapter \ref{chap:deep_cnns}.
\end{supportbox}

\subsection{Expected risk and overfitting}

As a second observation, note that the empirical risk is always trivial to minimize, by defining:

\vspace{1em}
\begin{equation}
f(x) = \begin{cases} y & \text{ if } \eqnmarkbox[drawred]{node}{(x,y) \in \mathcal{S}_n} \\ \eqnmarkbox[drawblue]{node2}{\bar{y}} & \text{ otherwise} \end{cases} \;.
\label{eq:lookup_table}
\end{equation}
\annotate[yshift=1em]{above,right}{node}{$x$ is in the training set}
\annotate[yshift=-1em]{below,right}{node2}{Default value, e.g., $0$}

\vspace{1em}
This is a look-up table that returns a prediction $y$ if the pair $(x,y)$ is contained in the dataset, while it defaults to some constant prediction $\bar{y}$ (e.g., 0) otherwise. Assuming that the loss is lower-bounded whenever $y = \hat{y}$, this model will always achieve the lowest possible value of empirical risk, while providing no actual practical value. 

This shows the difference between \textbf{memorization} and \textbf{learning} (optimization). Although we search for a model by optimizing some average loss quantity on our training data, as in \eqref{eq:empirical_risk_minimization}, our true objective is minimizing this quantity on some unknown, future input yet to be seen. The elements of our training set are only a proxy to this end. We can formalize this idea by defining the \textbf{expected risk minimization} problem.

\begin{definition}[Expected risk] $\,$

Given a probability distribution $p(x,y)$ and a loss function $l$, the \textbf{expected risk} (ER) is defined as:
\begin{equation}
\textnormal{ER}[f] = \mathbb{E}_{p(x,y)}\left[ l(y, f(x)) \right]
\label{eq:expected_risk}
\end{equation}
\end{definition}

Minimizing \eqref{eq:expected_risk} can be interpreted as minimizing the average (expected) loss across \textit{all possible input-output pairs} (e.g., all possible emails) that our model could see. Clearly, a model with low expected risk would be guaranteed to work correctly. However, the quantity in \eqref{eq:expected_risk} is unfeasible to compute in practice, as enumerating and labeling all data points is impossible. The empirical risk provides an estimate of the expected risk under the choice of a given dataset and can be seen as a Monte Carlo approximation of the ER term.

The difference in loss between the expected and the empirical risk is called the \textbf{generalization gap}: a pure memorization algorithm like \eqref{eq:lookup_table} will have poor generalization or, in other terms, it will \textbf{overfit} to the specific training data we provided. Generalization can be tested in practice by keeping a separate \textbf{test dataset} $\mathcal{T}_m$ with $m$ data points never used during training, $\mathcal{S}_n \cap \mathcal{T}_m = \emptyset$. Then, the difference in empirical loss between $\mathcal{S}_n$ and $\mathcal{T}_m$ can be used as an approximate measure of overfitting.

\subsection{How to select a valid loss function?} 
\label{subsec:how_to_select_a_loss}

\begin{tcolorbox}
If you have not done so already, this is a good time to study (or skim) the material in Appendix \ref{chap:probability_theory}, especially probability distributions, sufficient statistics, and maximum likelihood estimation.
\end{tcolorbox}

As we will see in the next chapters, the loss encodes our a priori knowledge on the task to be solved, and it has a large impact on performance. In some cases, simple considerations on the problem are enough to design valid losses (e.g., as done for the hinge loss in Section \ref{subsec:differentiability_loss}).

However, it is possible to work in a more principled fashion by reformulating the entire training process in purely probabilistic terms, as we show now. This formulation provides an alternative viewpoint on learning, which may be more intuitive or more useful in certain scenarios. It is also the preferred viewpoint of many books \cite{bishop2024deep}. We provide the basic ideas in this section, and we consider specific applications later on in the book.

The key observation is the following. In Section \ref{sec:dataset}, we started by assuming that our examples come from a distribution $p(x, y)$. By the product rule of probability, we can decompose $p(x,y)$ as $p(x,y) = p(x)p(y \mid x)$, such that $p(x)$ depends on the probability of observing each input $x$, and the conditional term $p(y \mid x)$ describes the probability of observing a certain output $y$ given an input $x$.\footnote{We can also decompose it as $p(x, y) = p(x \mid y)p(y)$. Methods that require to estimate $p(x)$ or $p(x \mid y)$ are called \textbf{generative}, while methods that estimate $p(y \mid x)$ are called \textbf{discriminative}. Apart from language modeling, in this book we focus on the latter case. We consider generative modeling more broadly in the next volume.} Approximating $p(y \;\vert\; x)$ with a function $f(x)$ makes sense if we assume that the probability mass is mostly concentrated around a single point $y$, i.e., $p(y \;\vert\; x)$ is close to a so-called Dirac delta function, and it drastically simplifies the overall problem formulation.

However, we can relax this by assuming that our model $f(x)$ does not provide directly the prediction, but it is used instead to parameterize the sufficient statistics of a conditional probability distribution $p(y \mid f(x))$ over possible outputs. For example, consider a classification problem where $y \in \left\{1,2,3\right\}$ can take three possible values. We can assume our model has three outputs that parameterize a categorical distribution over these classes, such that:
$$
p(\mathbf{y} \;\vert\; f(x)) = \prod_{i=1}^3 {f_i(x)}^{y_i}
$$
where $\mathbf{y} \sim \text{Binary}(3)$ is the one-hot encoding of the class $y$\footnote{Given an integer $i$, its one-hot representation is a vector of all zeros except the $i$-th element, which is $1$. This is introduced formally in Section \ref{sec:linear_models_for_classification}.} and $f(x) \sim \Delta(3)$ are the predicted probabilities for each class. As another example, assume we want to predict a single scalar value $y \in \mathbb{R}$ (\textbf{regression}). We can model this with a two-valued function $f(x) \sim (2)$ such that the prediction is a Gaussian with appropriate mean and variance:

\begin{equation}
p(y \;\vert\; f(x))=\mathcal{N}(y \mid f_1(x), \eqnmarkbox[drawred]{node}{f_2^2(x)})
\label{eq:gaussian_conditional_distribution}
\end{equation}
\annotate[yshift=-1em]{below,left}{node}{Squared to ensure positivity}

\vspace{1em}
where the second output of $f(x)$ is squared to ensure that the predicted variance remains positive. As can be seen, this is a very general setup that subsumes our previous discussion, and it provides more flexibility to the designer, as choosing a specific parameterization for $p(y \;\vert\; x)$ can be easier than choosing a specific loss function $l(y, \hat{y})$. In addition, this framework provides a more immediate way to model uncertainty, such as the variance in \eqref{eq:gaussian_conditional_distribution}.
\subsection{Maximum likelihood}
How can we train a probabilistic model? Remember that we assumed the samples in our dataset $\mathcal{S}_n$ to be i.i.d. samples from a probability distribution $p(x,y)$. Hence, given a model $f(x)$, the probability assigned to the dataset itself by a specific choice of function $f$ is given by the product of each sample in the dataset:
$$
p(\mathcal{S}_n \;\vert\; f)=\prod_{i=1}^n p(y_i \;\vert\; f(x_i))
$$

The quantity $p(\mathcal{S}_n  \;\vert\; f)$ is called the \textbf{likelihood} of the dataset. For a random choice of $f(x)$, the model will assign probabilities more or less at random across all possible inputs and outputs, and the likelihood of our specific dataset will be small. A sensible strategy, then, is to select the model such that the likelihood of the dataset is instead maximized. This is a direct application of the maximum likelihood approach (see Section \ref{sec:maximum_likelihood_estimation} in Appendix \ref{chap:probability_theory}).

\begin{definition}[Maximum likelihood] \addbottle

Given a dataset $\mathcal{S}_n = \left\{(x_i, y_i)\right\}$ and a family of probability distributions $p(y \;\vert\; f(x))$ parameterized by $f(x)$, the \textbf{maximum likelihood} solution is given by:
$$
f^* = \underset{f}{\arg\max}\prod_{i=1}^n p(y_i \;\vert\; f(x_i)) \,.
$$
\end{definition}

While we are again left with an optimization problem, it now follows directly from the laws of probability once all probability distributions are chosen, which is in contrast to before, where the specific loss was part of the design space. The two viewpoints, however, are closely connected. Working in log-space and switching to a minimization problem we obtain:

\clearpage

\begin{align}
\underset{f}{\arg\max} \left\{ \log \prod_{i=1}^n p(y_i \;\vert\; f(x_i)) \right\} = \nonumber\\ \underset{f}{\arg\min} \left\{ \sum_{i=1}^n -\log(p(y_i \;\vert\; f(x_i)) \right\}
\end{align}

Hence, the two formulations are identical if we identify $-\log(p(y \;\vert\; f(x))$ as a “pseudo-loss” to be optimized. As we will see, all loss functions used in practice can be obtained under the ML principle for specific choices of this term. Both viewpoints are interesting, and we urge readers to keep them in mind as we progress in the book.

\section{Bayesian learning}
\label{sec:bayesian_learning}

\addteacup We discuss here a further generalization of the probabilistic formulation called \textbf{Bayesian neural networks} (BNNs), which is of interest in the literature. We only provide the general idea and we refer the reader to one of many in-depth tutorials, e.g., \cite{jospin2022hands}, for more details.

By designing a probability function $p(y \;\vert\; f(x))$ instead of $f(x)$ directly, we can handle situations where more than one prediction is of interest (i.e., the probability function has more than a single mode). However, our procedure still returns a \textit{single function} $f(x)$ out of the space of all possible functions, while it may happen than more than a single parameterization across the entire model’s space is valid. In this case, it could be useful to have access to all of them for a more faithful prediction.

Once again, we can achieve this objective by designing another probability distribution and then letting the rules of probability guide us. Since we are now planning to obtain a distribution across all possible functions, we start by defining a \textbf{prior probability distribution} $p(f)$ over all possible functions (recall than in practice $f$ is described by a finite set of parameters, in which case the prior $p(f)$ becomes a prior over these weights). For example, we will see that in many situations functions with smaller norm are preferred (as they are more stable), in which case we could define a prior $p(f) \propto \frac{1}{\lVert f \rVert}$ for some norm $\lVert f \rVert$ of $f$. 

Once a dataset is observed, the probability over $f$ shifts depending on the prior and the likelihood, and the update is given by \textbf{Bayes’ theorem}:

\vspace*{1em}
\begin{equation}
\eqnmarkbox[drawblue]{node2}{p(f \;\vert\; \mathcal{S}_n)}=\frac{p(\mathcal{S}_n \;\vert\; f)\eqnmarkbox[drawred]{node}{p(f)}}{p(\mathcal{S}_n)}
\label{eq:posterior_distribution}
\end{equation}
\annotate[yshift=1em]{above,left}{node}{Prior (\textit{before} observing the dataset)}
\annotate[yshift=-1em]{below,right}{node2}{Posterior (\textit{after} observing the dataset)}

\vspace{1em}
The term $p(f \;\vert\; \mathcal{S}_n)$ is called the \textbf{posterior distribution function}, while the term $p(\mathcal{S}_n)$ in the denominator is called the \textbf{evidence} and it is needed to ensure that the posterior is properly normalized. Assume for now that we have access to the posterior. Differently from before, the distribution can encode preference for more than a single function $f$, which may provide better predictive power. Given an input $x$, we can make a prediction by averaging all possible models based on their posterior’s weight:
\clearpage

\vspace{1em}
\begin{align}
p(y \;\vert\; x) & =\int_f \eqnmarkbox[drawred]{node}{p(y \;\vert\; f(x))}\eqnmarkbox[drawblue]{node2}{p(f \;\vert\; \mathcal{S}_n)} \\ & \approx \eqnmarkbox[drawgreen]{node3}{\frac{1}{k}\sum_{i=1}^k} p(y \;\vert\; f_i(x))p(f_i \;\vert\; \mathcal{S}_n)
\label{eq:predictive_distribution}
\end{align}
\annotate[yshift=1em]{above,left}{node}{Prediction of $f(x)$}
\annotate[yshift=1.5em]{above,right}{node2}{Weight assigned to $f$}
\annotate[yshift=-1em]{below,right}{node3}{Monte Carlo approximation}

\vspace{1em}
where in \eqref{eq:predictive_distribution} we have approximated the integral with a Monte Carlo average over $k$ random samples from the posterior distribution $f_k \sim p(f \;\vert\; \mathcal{S}_n)$. The overall beauty of this setup is marred by the fact that the posterior is in general impossible to compute in closed-form, except for very specific choices of prior and likelihood \cite{bishop2006pattern}. Lacking this, one is forced to approximated solutions, either by \textbf{Markov chain Monte Carlo} or by \textbf{variational inference} \cite{jospin2022hands}. We will see in Section \ref{subsec:dropout} one example of Bayesian treatment of the model's parameters called \textbf{Monte Carlo dropout}.

We remark on two interesting facts about the posterior before closing this section. First, suppose we are only interested about the function having highest posterior density. In this case, the evidence term can be ignored and the solution decomposes into two separate terms:
\begin{gather}
f^*=\underset{f}{\arg\max} \; p(\mathcal{S}_n \;\vert\; f)p(f) = \\ \underset{f}{\arg\max} \; \left\{\eqnmarkbox[drawred]{node}{\log p(\mathcal{S}_n \;\vert\; f)} + \eqnmarkbox[drawblue]{node2}{\log p(f)}\right\}
\end{gather}
\annotate[yshift=-1em]{below,left}{node}{Likelihood term}
\annotate[yshift=-1em]{below,right}{node2}{Regularization term}

\clearpage
This is called the \textbf{maximum a posteriori} (MAP) solution. If all functions have the same weight a priori (i.e., $p(f)$ is uniform over the function’s space), then the second term is a constant and the problem reduces to the maximum likelihood solution. In general, however, the MAP solution will impose a penalty to functions deviating too much from our prior distribution. We will see this is a useful idea to combat overfitting and impose specific constraints on the function $f$. The term $\log p(f)$ is generally called a \textbf{regularizer} over the function’s space as it pushes the solution towards the basin of attraction defined by the prior distribution.\footnote{The difference between maximum likelihood and maximum a posteriori solutions is loosely connected to the difference between the \textbf{frequentist} and \textbf{Bayesian} interpretation of probability \cite{bishop2006pattern}, i.e., probabilities as frequency of events or probabilities as a measure of uncertainty. From a very high-level point of view, ML sees the parameters as an unknown fixed term and the data as a random sample, while a Bayesian treatment sees the data as fixed and the parameters as random variables.} 

Second, the full Bayesian treatment provides a simple way to incorporate new data, e.g., a new dataset $\mathcal{S}^\prime_n$ from the same distribution. To do that, we replace the prior function in \eqref{eq:posterior_distribution} with the posterior distribution that we computed on the first portion of the dataset, which now represents the starting assumption on the possible values of $f$ which gets updated
by looking at new data.\footnote{Think of the original prior function as the distribution on $f$ after having observed an initial \textit{empty set} of values.} This can mitigate issues when training models online, most notably the so-called \textit{catastrophic forgetting} of old information \cite{kirkpatrick2017overcoming}.

%% file: 4_linear_models.tex
\chapter{Linear models}
\label{chap:linear_models}

\begin{supportbox}{About this chapter}
Programming is done by choosing the appropriate sequence of primitive operations to solve a task. By analogy, building a model is done by choosing the correct sequence of \textit{differentiable} blocks. In this chapter we introduce the simplest block, linear models, which assume that inputs act additively on the output via a weighted average. In a sense, all differentiable models are smart variations and compositions of linear blocks.
\end{supportbox}

\section{Least-squares regression}

Summarizing the previous chapter, a supervised learning problem can be defined by choosing the input type $x$, the output type $y$, the model $f$, and the loss function $l$. In this chapter we consider the simplest possible choices for all of them, namely:
\begin{itemize}
    \item The input is a vector $\mathbf{x} \sim (c)$, corresponding to a set of features (e.g., $c$ personal features of a client of a bank). We use the scalar $c$ (short for \textit{channels}) to denote the number of features to be consistent with the following chapters.
    \item The output is a single real value $y \in \mathbb{R}$. In the unconstrained case, we say this is a \textbf{regression} task. If $y$ can only take one out of $m$ possible values, i.e., $y \in \left\{1, \ldots, m\right\}$, we say this is a \textbf{classification} task. In the special case of $m=2$, we say this is a \textbf{binary classification}  task.
    \item We take $f$ to be a linear model, providing us with simple closed form solutions in some cases, most notably \textbf{least-squares regression} (Section \ref{subsec:least_squares}).
\end{itemize}
The basic shapes to remember are summarized in Table \ref{tab:shapes}. We begin by discussing the choice of loss in the regression case. We start from the regression case since, as we show later, classification can be solved by small modifications to the regression case.

\begin{table}[t]
\centering
\caption{Basic shapes to remember for this chapter. For uniformity, we will use the same letters as much as possible throughout the book.}
\begin{tabular}{@{}cl@{}}
\toprule
 $n$ & \textbf{size of the dataset} \\ $c$ & \textbf{features} \\ $m$ & \textbf{classes}\\ \midrule
\end{tabular}
\label{tab:shapes}
\end{table}

\subsection{The squared loss and variants}

Finding a loss for regression is relatively simple, since the prediction error $e = (\hat{y} - y)$ between the predicted output of the model $\hat{y} = f(x)$ and the true desired output $y$ is a well-defined target, being a continuous function of the model’s output that decreases monotonically. Since in general we do not care about the sign of the prediction error, a common choice is the \textbf{squared loss}:

\begin{equation}
    l(\hat{y},y)=(\hat{y}-y)^2
    \label{eq:squared_loss}
\end{equation}

Here and in the following we use the symbol $\hat{y}$ to denote the prediction of a model. As we will see, working with \eqref{eq:squared_loss} grants several benefits to our solution. Among others, the gradient of the squared loss is a linear function of the model’s output, allowing us to solve it in closed form for the optimal solution. 

Recalling the maximum likelihood principle (Section \ref{subsec:how_to_select_a_loss}), the squared loss can be obtained by assuming that the outputs of the model follow a Gaussian distribution centered in $f(\mathbf{x})$ and with a constant variance $\sigma^2$:

$$
p(y \;\vert\; f(\mathbf{x}))=\mathcal{N}(y\;\vert\;f(\mathbf{x}), \sigma^2)
$$

In this case the log-likelihood (for a single point) can be written as:\footnote{Recalling that $\log(ab)=\log(a)+\log(b)$ and $\log(a^b)=b\log(a)$.}
\begin{align}
\log(p(y \mid f(\mathbf{x}), \sigma^2)) = \nonumber \\ -\log(\sigma) - \frac{1}{2}\log(2\pi) - \frac{1}{2\sigma^2}(y - f(\mathbf{x}))^2
\label{eq:ls_maximum_likelihood}
\end{align}
Minimizing \eqref{eq:ls_maximum_likelihood} for $f$, we see that the first two terms on the right-hand side are constant, and the third reverts to the squared loss. Minimizing for $\sigma^2$ can be done independently from the optimization of $f$, with a simple closed form solution (see below, equation \eqref{eq:ls_sigma}).

Coming up with variants to the squared loss is also easy. For example, one drawback of the squared loss is that higher errors will be penalized with a strength that grows quadratically in the error, which may provide undue influence to \textbf{outliers}, i.e., points that are badly mislabeled. Other choices that diminish the influence of outliers can be the \textbf{absolute value loss} $l(\hat{y}, y) = \lvert \hat{y} - y \rvert$ or the Huber loss (a combination of the squared loss and the absolute loss):
\begin{equation}
L(y, \hat{y}) = \begin{cases} \frac{1}{2}\left(y - \hat{y}\right)^2 & \text{ if } \lvert y - \hat{y} \rvert \le 1 \\ \left(\lvert y - \hat{y} \rvert - \frac{1}{2}\right) & \text{ otherwise } \end{cases}
\end{equation}
\begin{figure}[t]
    \centering
    \includegraphics[width=0.7\textwidth]{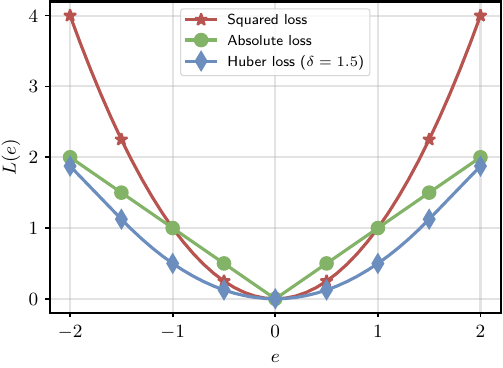}
    \caption{Visualization of the squared loss, the absolute loss, and the Huber loss with respect to the prediction error $e = (\hat{y} - y)$.}
    \label{fig:losses_regression}
\end{figure}

which is quadratic in the promixity of $0$ error, and linear otherwise (with the $-\frac{1}{2}$ term added to ensure continuity). See Figure \ref{fig:losses_regression} for a visualization of these losses with respect to the prediction error. 

The absolute loss seems an invalid choice in our context, since it has a point of non-differentiability in $0$ due to the absolute value. We will see later that functions with one (or a small number) of points of this form are not truly problematic. Mathematically, they can be handled by the notion of \textbf{subgradient} (a slight generalization of the derivative). Practically, you can imagine that if we start from a random initialization, gradient descent will never reach these points with perfect precision, and the derivatives of $\lvert \varepsilon \rvert$ for any $\varepsilon > 0$ is always defined.

\subsection{The least-squares model}
\label{subsec:least_squares}

\addclock With a loss function in hand, we consider the following model (a linear model) to complete the specification of our first supervised learning problem.
\begin{definition}[Linear models] $\,$

A \textbf{linear model} on an input $\mathbf{x}$ is defined as:
$$
f(\mathbf{x})=\mathbf{w}^\top\mathbf{x} + b
$$
where $\mathbf{w} \sim (c)$ and $b \in \mathbb{R}$ (the \textit{bias}) are trainable parameters.
\end{definition}

The intuition is that the model assigns a fixed weight $w_i$ to each input feature $x_i$, and provides a prediction by linearly summing all the effects for a given input $\mathbf{x}$, reverting to a default prediction equal to $b$ whenever $\mathbf{x} = \mathbf{0}$. Geometrically, the model defines a line for $d=1$, a plane for $d=2$, and a generic hyperplane for $d > 1$. From a notational perspective, we can sometimes avoid writing  a bias term by assuming a constant term of $1$ as the last feature of $\mathbf{x}$:
$$
f\left(\begin{bmatrix}\mathbf{x}\\1\end{bmatrix}\right) = \mathbf{w}^\top\begin{bmatrix}\mathbf{x}\\1\end{bmatrix} = \mathbf{w}_{1:c}^\top\mathbf{x}+w_{c+1}
$$

Combining the linear model, the squared loss, and an empirical risk minimization problem we obtain the \textbf{least-squares optimization problem}.

\begin{definition}[Least-squares] \addbottle $\,$

The \textbf{least-squares} optimization problem is given by:

\begin{equation}
\mathbf{w}^*, b^* = \underset{\mathbf{w}, b}{\arg\min} \;\;\frac{1}{n}\sum_{i=1}^n \left(y_i - \mathbf{w}^\top \mathbf{x}_i - b\right)^2
\label{eq:least_squares}
\end{equation}
\end{definition}

Before proceeding to the analysis of this problem, we rewrite the least-squares in a \textbf{vectorized} form that only involves matrix operations (matrix products and norms). This is useful because, as already stated, modern code for training differentiable models is built around $n$-dimensional arrays, with optimized hardware to perform matrix operations on them. To this end, we first stack all the inputs and outputs of our training set into an \textbf{input matrix}:

$$
\mathbf{X} = \begin{bmatrix} \mathbf{x}_1^\top \\ \vdots \\ \mathbf{x}_n^\top \end{bmatrix} \sim (n,c)
$$

and a similar \textbf{output vector} $y = \left[ y_1, \ldots, y_n\right]^\top$. We can write a batched model output (the model output for a mini-batch of values) as:

\begin{equation}
f(\mathbf{X})=\mathbf{X}\mathbf{w} + \eqnmarkbox[drawblue]{node}{\mathbf{1}b}
\label{eq:batched_linear_model}
\end{equation}
\annotate[yshift=-1em]{below,left}{node}{Same bias $b$ for all $n$ predictions}

\vspace{1em}
Equations like \eqref{eq:batched_linear_model} can be replicated almost line-by-line in code - see Box \ref{code:linear_model} for an example in PyTorch. 

\begin{mypy}{Computing a batched linear model as in \protect\eqref{eq:batched_linear_model}. For clarity, we are showing the array dimensions as type hints using {\footnotesize \texttt{jaxtyping} (\url{https://docs.kidger.site/jaxtyping/})}.}{code:linear_model}
def linear_model(w: Float[Tensor, "c"],
                 b: Float,
                 X: Float[Tensor, "n c"]) 
                 -> Float[Tensor, "n"]:
  return X @ w + b
\end{mypy}

Of only marginal interest for now but of more importance for later, we note that the row ordering of the input matrix and of the output vector are fundamentally arbitrary, in the sense that permuting their rows will only result in a corresponding permutation of the rows of $f(\mathbf{X})$. This is a simple example of a phenomenon called \textbf{permutation equivariance} that will play a much more important role later on.

The least-squares optimization problem written in a vectorized form becomes:

\begin{equation}
\text{LS}(\mathbf{w},b) =  \frac{1}{n} \left\lVert \mathbf{y} - \mathbf{X}\mathbf{w} - \mathbf{1}b \right\rVert^2 
\label{eq:ls_vectorized}
\end{equation}

where we recall that the norm of a vector is defined as $\lVert \mathbf{e} \rVert^2 = \sum_i e_i^2$.

\subsection{Solving the least-squares problem}

To solve the least-squares problem through gradient descent, we need the equation for its gradient. Although we will soon develop a general algorithmic framework to compute these gradients automatically (Chapter \ref{chap:automatic_differentiation}), it is instructive to look at the gradient itself in this simple scenario. Ignoring the bias (for the reasons stated above, we can incorporate it in the weight vector), and other constant terms we have:
$$
\nabla \text{LS}(\mathbf{w}) = \mathbf{X}^\top\left( \mathbf{X}\mathbf{w} - \mathbf{y} \right)
$$
The LS problem is convex in the weights of the model, as can be understood informally by noting that the equations describe a paraboloid in the space of the weights (a quadratic function). The global minima are then described by the equations:
$$
\mathbf{X}^\top\left( \mathbf{X}\mathbf{w} - \mathbf{y} \right) = 0 \;\;\Rightarrow\;\; \mathbf{X}^\top\mathbf{X}\mathbf{w} = \mathbf{X}^\top\mathbf{y}
$$

\begin{mypy}{Solving the least-squares problem with the closed form solution. The numerically stable variant calls a solver specialized for systems of linear equations.}{code:ls_closed_form}
from torch import linalg
def ls_solve(X: Float[Tensor, "n c"],
             y: Float[Tensor, "n"],
             numerically_stable = True) \
             -> Float[Tensor, "c"]:
  # Explicit solution
  if not numerically_stable:
    return linalg.inv(X.T @ X) @ X.T @ y 
  else:
    return linalg.solve(X.T @ X, X.T @ y)
\end{mypy}

These are called the \textbf{normal equations}. Importantly, the normal equations describe a linear system of equations in $\mathbf{w}$,\footnote{That is, we can write them as $\mathbf{A}\mathbf{w}=\mathbf{b}$, with $\mathbf{A} = \mathbf{X}^\top\mathbf{X}$ and $\mathbf{b} = \mathbf{X}^\top\mathbf{y}$.} meaning that under the appropriate conditions (corresponding to the invertibility of the matrix $\mathbf{X}^\top\mathbf{X}$) we can solve for the optimal solution as:
\begin{equation}
\mathbf{w}_*=\left(\mathbf{X}^\top\mathbf{X}\right)^{-1}\mathbf{X}^\top\mathbf{y}
\label{eq:ls_closed_form_solution}
\end{equation}

\vspace{1em}
\begin{supportbox}{Tidbits of information}
The matrix $\mathbf{X}^\dagger=\left(\mathbf{X}^\top\mathbf{X}\right)^{-1}\mathbf{X}^\top$ is called the \textbf{pseudoinverse} (or \textbf{Moore-Penrose inverse}) of the non-square matrix $\mathbf{X}$, since $\mathbf{X}^\dagger\mathbf{X}=\mathbf{I}$. Performing the inversion in \eqref{eq:ls_closed_form_solution} is not always possible: for example, if one feature is a scalar multiple of the other, the matrix $\mathbf{X}$ does not have full rank (this is called \textbf{collinearity}). Finally, note that the predictions of the least-squares model can be written as $\hat{\mathbf{y}}=\mathbf{M}\mathbf{y}$, with $\mathbf{M} = \mathbf{X}\mathbf{X}^\dagger$. Hence, least-squares can also be interpreted as performing a weighted average of the training labels, where the weights are given by a projection on the column space induced by $\mathbf{X}$. This is called the \textbf{dual} formulation of least-squares. Dual formulations provide an intrinsic level of debugging of the model, as they allow to check which inputs were the most relevant for a prediction by checking the corresponding dual weights \cite{irie2022dual}. 
\end{supportbox}

\begin{SCfigure}
    \centering
    \hspace{1em} \includegraphics[width=0.5\textwidth]{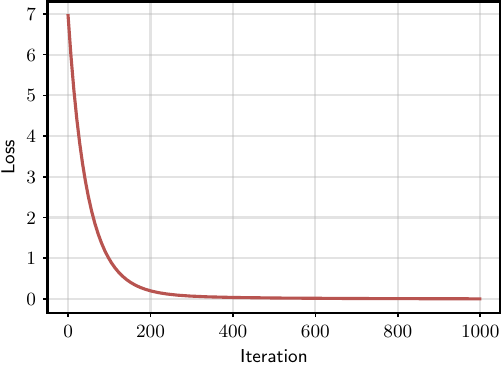}
    \caption{An example of running code from Box \ref{code:ls_closed_form}, where the data is composed of $n=10$ points drawn from a linear model $\mathbf{w}^\top \mathbf{x} + \varepsilon$, with $w_i \sim \mathcal{N}(0, 1)$ and $\varepsilon \sim \mathcal{N}(0, 0.01)$. Details apart, note the very smooth descent: each step provides a decrease in loss.}
    \label{fig:loss_evolution}
\end{SCfigure}

This is the only case in which we will be able to express the optimal solution in a closed form way, and it is instructive to compare this solution with the gradient descent one. To this end, we show in Box \ref{code:ls_closed_form} an example of solving the least-squares in closed form using \eqref{eq:ls_closed_form_solution}, and in Box \ref{code:ls_gradient_descent} the equivalent gradient descent formulation. A prototypical evolution of the loss in the latter case is plotted in Figure \ref{fig:loss_evolution}. Since we selected a very small learning rate, each step in the gradient descent procedure provides a stable decrease in the loss, until convergence. Practically, convergence could be checked by numerical means, e.g., by evaluating the difference in norm between two iterations for some numerical threshold $\varepsilon > 0$:

\begin{equation}
    \lVert \mathbf{w}_{t+1} - \mathbf{w}_t \rVert^2 < \varepsilon
\end{equation}

As we will see, understanding when more complex models have converged will be a more subtle task.

\begin{mypy}{Same task as Box \protect\ref{code:ls_closed_form}, solved with a naive implementation of gradient descent with a fixed learning rate that defaults to $\eta = 0.001$.}{code:ls_gradient_descent}
def ls_gd(X: Float[Tensor, "n c"],
          y: Float[Tensor, "n 1"],
          lr=1e-3) \
          -> Float[Tensor, "c"]:
    # Initializing the parameters
    w = torch.randn((X.shape[1], 1))

    # Fixed number of iterations
    for i in range(15000):
      # Note the sign (why?)
      w = w + lr * X.T @ (y - X @ w)

    return w
\end{mypy}

Considering again the Gaussian log-likelihood in \eqref{eq:ls_maximum_likelihood}, we can also optimize the term with respect to $\sigma^2$ once the weights have been trained, obtaining:

\begin{equation}
\sigma_*^{2} = \frac{1}{n}\sum_{i=1}^n (y_i - \mathbf{w}_*^{\top}\mathbf{x}_i)^2 \,.
\label{eq:ls_sigma}
\end{equation}

which has the intuitive meaning that the variance of the model is constant (by definition) and given by the average squared prediction error on our training data. More sophisticated probabilistic models can be obtained by assuming the variance itself is predicted by the model (\textbf{heteroscedastic} models), see \cite{bishop2006pattern}.

\subsection{Some computational considerations}

\addteacup Even if the inverse can be computed, the quality of the solution will depend on the condition number of $\mathbf{X}^\top\mathbf{X}$, and large numerical errors can occur for poorly conditioned matrices.\footnote{The condition number of a matrix $\mathbf{A}$ is defined as $\kappa(\mathbf{A}) = \lVert\mathbf{A}\rVert\lVert\mathbf{A}^{-1}\rVert$ for some choice of matrix norm $\lVert \bullet \rVert$. Large conditions number can make the inversion difficult, especially if the floating-point precision is not high.} In addition, the computational cost of solving \eqref{eq:ls_closed_form_solution} may be prohibitive. The matrix inversion will scale, roughly, as $\mathcal{O}(c^3)$. As for the matrix multiplications, the algorithm requires a multiplication of a $c \times n$ matrix with another $n \times c$ one, and a multiplication between a $c \times c$ matrix and a $c \times n$ one. Both these operations will scale as $\mathcal{O}(c^2n)$.

In general, we will always prefer algorithms that scale linearly both in the feature dimension $c$ and in the batch size $n$, since super-linear algorithms will become quickly impractical (e.g., a batch of $32$ RGB images of size $1024 \times 1024$ has $c \approx 1e^{7}$). We can avoid a quadratic complexity in the equation of the gradient by computing the multiplications in the correct order, i.e., computing the matrix-vector product $\mathbf{X}\mathbf{w}$ first. Hence, pure gradient descent is linear in both $c$ and $n$, but only if proper care is taken in the implementation: generalizing this idea is the fundamental insight for the development of \textbf{reverse-mode automatic differentiation}, a.k.a. \textbf{back-propagation} (Section \ref{sec:reverse_mode_automatic_differentiation}).

\subsection{Regularizing the least-squares solution}
\label{subsec:regularizing_least_squares}

Looking again at the potential instability of the inversion operation, suppose we have a dataset for which the matrix is almost singular, but we still wish to proceed with the closed form solution. In that case, it is possible to slightly modify the problem to achieve a solution which is “as close as possible” to the original one, while being feasible to compute. For example, a known trick is to add a small multiple, $\lambda > 0$, of the identity matrix to the matrix being inverted:
$$
\mathbf{w}^*=\left(\mathbf{X}^\top\mathbf{X} {\color{drawred}+\lambda\mathbf{I}}\right)^{-1}\mathbf{X}^\top\mathbf{y}
$$
This pushes the matrix to be “more diagonal” and improves its condition number. Backtracking to the original problem, we note this is the closed form solution of a modified optimization problem:
$$
\text{LS-Reg}(\mathbf{w}) =  \frac{2}{n} \left\lVert \mathbf{y} - \mathbf{X}\mathbf{w} \right\rVert^2 {\color{drawred}+ \frac{\lambda}{2}\lVert \mathbf{w} \rVert^2}
$$
This problem is called \textbf{regularized least-squares} (or \textbf{ridge regression}), and the red part in the loss is an instance of $\ell_2$-regularization (or, more generally, regularization). Note that regularization does not depend on the dataset, as it simply encodes a preference for a certain type of solution (in this case, low-norm weights), where the strength of the preference itself is defined by the hyper-parameter $\lambda$. From a Bayesian perspective (Section \ref{sec:bayesian_learning}), the regularized least-squares corresponds to a MAP solution when defining a Gaussian prior over the weights centered in zero with constant variance.

\section{Linear models for classification}
\label{sec:linear_models_for_classification}

We now move to classification, in which $y_i \in \left\{1, \ldots, m\right\}$, where $m$ defines the number of \textbf{classes}. As we will see later, this is a widely influential problem, encompassing a range of tasks in both computer vision (e.g., image classification) and natural language processing (e.g., next-token prediction). We can tackle this problem by slight variations with respect to the regression case.

While we can solve the task by regressing directly on the integer value $y_i$, it is instructive to consider why this might not be a good idea. First, it is difficult for a model to directly predict an integer value, since this requires some thresholding that would render its gradient zero almost everywhere. Instead, we could regress on a real value $\widetilde{y}_i \in \left[1, m\right]$ inside the interval from $1$ to $m$ (as we will show, bounding the output of the model inside an interval can be done easily). During inference, given the output $\hat{y}_i = f(\mathbf{x}_i)$, we map back to the original domain by rounding:
$$
\text{Predicted class} =\text{round}(\hat{y}_i)
$$
For example, $\hat{y}_i = 1.3$ would be mapped to class $1$, while $\hat{y}_i = 3.7$ would be mapped to class $4$. Note that this is a post-hoc processing of the values that is only feasible at inference time. The reason this is not a good modeling choice is that we are introducing a spurious ordering of the classes which might be exploited by the model itself, where class $2$ is “closer” to class $3$ than it is to class $4$. We can avoid this by moving to a classical \textbf{one-hot encoded} version of $y$, which we denote by $\mathbf{y}^{\text{oh}} \sim \text{Binary}(m)$:
$$
\idx{\mathbf{y}^{\text{oh}}}{j}=\begin{cases} 1 &\text{ if } y = j \\ 0 & \text{ otherwise} \end{cases}
$$
For example, in the case of three classes, we would have $\mathbf{y}^{\text{oh}} = \left[1 \;\; 0 \;\; 0 \right]^\top$ for class $1$, $\mathbf{y}^{\text{oh}} = \left[0 \;\; 1 \;\; 0 \right]^\top$ for class $2$, and $\mathbf{y}^{\text{oh}} = \left[0 \;\; 0 \;\; 1 \right]^\top$ for class $3$ (this representation should be familiar to readers with some background in machine learning, as it is a standard representation for categorical variables). 

One-hot vectors are unordered, in the sense that given two generic outputs $\mathbf{y}_1^{\text{oh}}$ and $\mathbf{y}_2^{\text{oh}}$, their Euclidean distance is either $0$ (same class) or $\sqrt{2}$ (different classes). While we can perform a multi-valued regression directly on the one-hot encoded outputs, with the mean-squared error known as the \textbf{Brier score} in this case, we show below that a better and more elegant solution exists, in the form of \textbf{logistic regression}. \clearpage

\subsection{The softmax function}
\label{sec:softmax}

We cannot train a model to directly predict a one-hot encoded vector (for the same reasons described above), but we can achieve something similar by a slight relaxation. To this end, we re-introduce the \textbf{probability simplex}. 

\begin{definition}[Probability simplex]$\,$

The \textbf{probability simplex} $\Delta_n$ is the set of vectors $\mathbf{x} \sim \Delta(n)$ such that:

$$
x_i \ge 0, \; \sum_i x_i=1
$$

\end{definition}

Geometrically, you can picture the set of one-hot vectors as the vertices of an $n$-dimensional polytope, and the simplex as its convex hull: values inside the simplex, such as $[0.2, 0.05, 0.75]$, do not precisely correspond to a vertex, but they allow for gradient descent because we can smoothly move inside the polytope. Given a value $\mathbf{x} \in \Delta_n$, we can project to its closest vertex (the predicted class) as:
$$
\underset{i}{\arg\max} \left\{ \mathbf{x}_i \right\}
$$
As the name implies, we can interpret values inside the simplex as probability distributions, and projection on the closest vertex as finding the mode (the most probable class) in the distribution. In this interpretation, a one-hot encoded vector is a “special case” where all the probability mass is concentrated on a single class (which we know to be the correct one).

In order to predict a value in this simplex, we need two modifications to the linear model from \eqref{eq:least_squares}: first, we need to predict an entire vector simultaneously; and second, we need to constrain the outputs to lie in the simplex. As a first step, we modify the linear model to predict an $m$-dimensional vector:
\begin{equation}
\mathbf{y} =\mathbf{W}\mathbf{x}+\mathbf{b}
\label{eq:multi_valued_linear_model}
\end{equation}
where $\mathbf{W} \sim (m,c)$ can be interpreted as $m$ linear regression models running in parallel, and $\mathbf{b} \sim(m)$. This output is unconstrained and it is not guaranteed to be in the simplex. The core idea of logistic regression is to combine the linear model in \eqref{eq:multi_valued_linear_model} with a simple, parameter-free transformation that projects inside the simplex, called the \textbf{softmax} function.

\begin{definition}[Softmax function] \addbottle $\,$

The \textbf{softmax} function is defined for a generic vector $\mathbf{x} \sim(m)$ as:

\begin{equation}
\idx{\textnormal{softmax}(\mathbf{x})}{i} = \frac{\exp(x_i)}{\sum_j\exp(x_j)}
\label{eq:softmax}
\end{equation}

\end{definition}

To understand what is happening, we decompose the terms in \eqref{eq:softmax} by introducing two intermediate terms. First, the numerator of the softmax converts each number to a positive value $h_i$ by exponentiation:

\begin{equation}
h_i =\exp(x_i)
\end{equation}

Second, we compute a normalization factor $Z$ as the sum of these new (non-negative) values:

\begin{equation}
 Z = \sum_jh_j 
 \end{equation}

The output of the softmax is then given by dividing $h_i$ by $Z$, thus ensuring that the new values sum to $1$:

\begin{equation}
y_i = \frac{h_i}{Z}
\end{equation}

Another perspective comes from considering a more general version of the softmax, where we add an additional hyper-parameter $\tau > 0$ called the \textbf{temperature}:

$$
\text{softmax}(\mathbf{x};\tau)=\text{softmax}(\mathbf{x}/\tau)
$$

The softmax keeps the relative ordering among the values of $x_i$ for all values of $\tau$, but their absolute distance is increased or decreased based on the temperature. In particular, we have the following two limiting cases:

\begin{gather}
\lim_{\tau \rightarrow \infty} \text{softmax}(\mathbf{x};\tau)=1/c \\ 
\lim_{\tau \rightarrow 0} \text{softmax}(\mathbf{x};\tau)=\underset{i}{\arg\max} \;\; \mathbf{x}
\end{gather}

\begin{figure}[t]
    \centering
    \begin{subfigure}[b]{0.24\textwidth}
    \includegraphics[width=\textwidth]{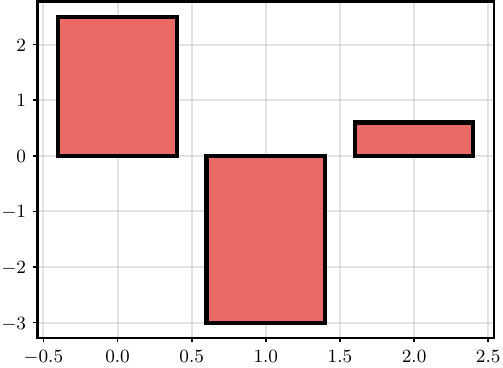}
    \caption{Inputs}
    \end{subfigure}
    \hfill
    \begin{subfigure}[b]{0.24\textwidth}
    \includegraphics[width=1.0\textwidth]{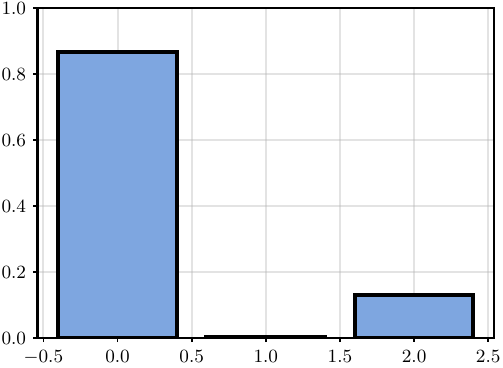}
    \caption{$\tau=1$}
    \end{subfigure}
    \hfill
    \begin{subfigure}[b]{0.24\textwidth}
    \includegraphics[width=1.0\textwidth]{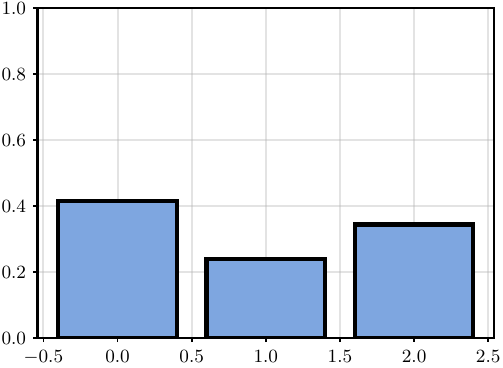}
    \caption{$\tau=10$}
    \end{subfigure}
    \hfill
    \begin{subfigure}[b]{0.24\textwidth}
    \includegraphics[width=1.0\textwidth]{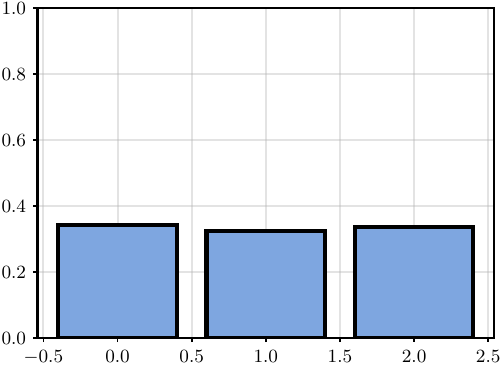}
    \caption{$\tau=100$}
    \end{subfigure}
    \hfill
    \caption{Example of softmax applied to a three-dimensional vector (a), with temperature set to 1 (b), 10 (c), and 100 (d). As the temperature increases, the output converges to a uniform distribution. Note that inputs can be both positive or negative, but the outputs of the softmax are always constrained in $[0,1]$.}
    \label{fig:softmax}
\end{figure}

For infinite temperature, relative distances will disappear and the output reverts to a uniform distribution. At the contrary, at $0$ temperature the softmax reverts to the (poorly differentiable) argmax operation. Hence, softmax can be seen as a simple differentiable approximation to the argmax, and a better name should be \textbf{softargmax}. However, we will retain the most standard name here. See Figure \ref{fig:softmax} for a visualization of a softmax applied on a generic three-dimensional vector with different temperature values.

\subsection{The logistic regression model}
\label{subsec:logistic_regression}

\addclock We can summarize our previous discussion by combining the softmax in \eqref{eq:softmax} with the linear model in \eqref{eq:multi_valued_linear_model} to obtain a linear model for classification:

$$
\hat{\mathbf{y}}=\text{softmax}\left(\mathbf{W}\mathbf{x} + \mathbf{b}\right)
$$

The pre-normalized outputs $\mathbf{h} = \mathbf{W}\mathbf{x}+\mathbf{b}$ are called the \textbf{logits} of the model, a name that will be discussed in more detail in the next section. 

We now need a loss function. Considering the probabilistic viewpoint from Section \ref{subsec:how_to_select_a_loss}, because our outputs are restricted to the probability simplex, we use them as the parameters of a categorical distribution:

\vspace{1em}
\begin{equation*}
p(\eqnmarkbox[drawblue]{node}{\mathbf{y}^{\text{oh}}}\;\vert\;\hat{\mathbf{y}}) = \prod_i \hat{y}_i^{\eqnmarkbox[drawred]{node2}{y_i^{\text{oh}}}}
\annotate[yshift=1em]{above,left}{node2}{Exponent is always either $0$ or $1$}
\annotate[yshift=-1em]{below,right}{node}{One-hot encoded class}
\end{equation*}

\vspace{1em}
Computing the maximum likelihood solution in this case (try it) gets us the \textbf{cross-entropy loss}.

\begin{definition}[Cross-entropy loss] \addbottle $\,$

The \textbf{cross-entropy} loss function between $\mathbf{y}^{\text{oh}}$ and $\hat{\mathbf{y}}$ is given by:

\begin{equation}
\textnormal{CE}(\mathbf{y}^{\textnormal{oh}},\hat{\mathbf{y}})= - \sum_i y_i^{\textnormal{oh}}\log(\hat{y}_i)
\label{eq:cross_entropy}
\end{equation}
\end{definition}

The loss can also be derived as the KL divergence between the two probability distributions. While unintuitive at first, it has a very simple interpretation by noting that only one value of $\mathbf{y}^{\text{oh}}$ will be non-zero, corresponding to the true class $y = \underset{i}{\arg\max} \left\{ y^{\text{oh}}_i \right\}$. We can then simplify the loss as:

\begin{equation}
\text{CE}(y, \hat{\mathbf{y}}) = - \log(\eqnmarkbox[drawred]{node}{\hat{y}_y})
\label{eq:ce_single_term}
\end{equation}
\annotate[yshift=-1em]{below,left}{node}{Probability assigned to the true class}

\vspace{1em}
From \eqref{eq:ce_single_term}, we see that the effect of minimizing the CE loss is to maximize the output probability corresponding to the true class. This works since, due to the denominator in the softmax, any increase in one output term will automatically lead to a decrease of the other terms. Putting everything together, we obtain the logistic regression optimization problem:

$$
\text{LR}(\mathbf{W},\mathbf{b})=\frac{1}{n}\sum_{i=1}^n \text{CE}\left( \mathbf{y}_i^{\text{oh}}, \text{softmax}(\mathbf{W}\mathbf{x}_i+\mathbf{b}) \right) \,.
$$

Differently from least-squares, we cannot compute a closed form solution anymore, but we can still proceed with gradient descent. We will show in the next section an example of gradient in this case, and in Section \ref{sec:reverse_mode_automatic_differentiation} a generic technique to compute gradients in cases such as this one.

\section{More on classification}

\subsection{Binary classification}

Consider now the specific case of $m=2$. In this case we have $y \in \left\{0,1\right\}$, and the problem reduces to \textbf{binary classification}, sometimes called \textbf{concept learning} (as we need to learn whether a certain binary “concept” is present or absent in the input). With a standard logistic regression, this would be modelled by a function having two outputs. However, because of the softmax denominator, the last output of a logistic regression is always redundant, as it can be inferred knowing that the outputs must sum to $1$:

$$
f_m(\mathbf{x}) = \sum_{i=1}^{m-1}f_i(\mathbf{x})
$$

Based on this, we can slightly simplify the formulation by considering a scalar model with a single output $f(\mathbf{x}) \in [0,1]$, such that:

$$
\text{Predicted class} = \text{round}(f(\mathbf{x}))= \begin{cases} 0 & \text{ if } f(\mathbf{x}) \le 0.5 \\ 1 & \text{ otherwise } \end{cases} 
$$

To achieve the desired normalization in $[0,1]$, the first output of a two-valued softmax can be rewritten as $\frac{\exp(x_1)}{1+\exp(x_1)}$, and we can further simplify it by dividing both sides by $\exp(x_1)$. The result is the \textbf{sigmoid} function.

\begin{definition}[Sigmoid function] \addbottle $\,$

The \textbf{sigmoid} function $\sigma(s) : \mathbb{R} \rightarrow [0,1]$ is given by:
$$
\sigma(s)=\frac{1}{1+\exp(-s)}
$$
\end{definition}

\begin{SCfigure}
    \centering
    \hspace{1em}\includegraphics[width=0.6\textwidth]{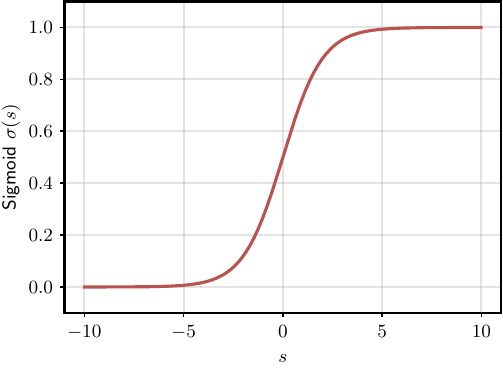}
    \caption{Plot of the sigmoid function. Note that $\sigma(0)=0.5$.}
    \label{fig:sigmoid}
\end{SCfigure}

The sigmoid provides a generic transformation projecting any real value to the $[0,1]$ interval (with the two extremes being reached only asymptotically). Its graph is shown in Figure \ref{fig:sigmoid}.

The \textbf{binary logistic regression} model is obtained by combining a one-dimensional linear model with a sigmoid rescaling of the output:
$$
f(\mathbf{x})=\sigma\left(\mathbf{w}^\top\mathbf{x}+b\right)
$$
The cross-entropy similarly simplifies to:

\vspace{1em}
\begin{equation}
\text{CE}(\hat{y},y) = \eqnmarkbox[drawred]{node}{-y\log(\hat{y})} \eqnmarkbox[drawblue]{node2}{-(1-y)\log(1-\hat{y})}
\end{equation}
\annotate[yshift=1em]{above,left}{node}{Loss for class $1$}
\annotate[yshift=1em]{above,right}{node2}{Loss for class $2$}

Hence, in the binary classification case we can solve the problem with two equivalent approaches: (a) a two-valued model with the standard softmax, or (b) a simplified one-valued output with a sigmoid output transformation. 

As an interesting side-note, consider the gradient of the binary logistic regression model with respect to $\mathbf{w}$ (a similar gradient can also be written for the standard multi-class case):

$$
\nabla \text{CE}(f(\mathbf{x}),y) = (f(\mathbf{x})-y)\mathbf{x}
$$

Note the similarity with the gradient of a standard linear model for regression. This similarity can be further understood by rewriting our model as:

\vspace{1em}
\begin{equation}
\eqnmarkbox[drawred]{node}{\mathbf{w}^\top \mathbf{x}+b}=\eqnmarkbox[drawblue]{node2}{\log\left(\frac{y}{1-y}\right)}
\end{equation}
\annotate[yshift=1em]{above,left}{node}{Logits}
\annotate[yshift=1em]{above,right}{node2}{Sigmoid inverse: $\sigma^{-1}(y)$}

This clarifies why we were referring to the model as a “linear model” for classification: we can always rewrite it as a purely linear model in terms of a non-linear transformation of the output (in this case, the inverse of the sigmoid, also known as the \textbf{log-odds}). In fact, the logistic regression model is part of a broader family of models extending this idea, called \textbf{generalized linear models}. For the curious reader, the name \textit{logits} can be understood in this context in reference to the probit function.\footnote{\url{https://en.wikipedia.org/wiki/Probit}}

\subsection{The logsumexp trick} 

\addteacup This is a more technical subsection that clarifies an implementation aspect of what we described up to now. Looking at frameworks like TensorFlow or PyTorch, we can find multiple existing implementations of the cross-entropy loss, based on whether the output is described as an integer or as a one-hot encoded vector. This can be understood easily, as we have already seen that we can formulate the cross-entropy loss in both cases. However, we can also find variants that accept logits instead of the softmax-normalized outputs, as shown in Box \ref{code:cross_entropy}.

\begin{mypy}{Cross entropy losses in PyTorch. Some losses are only defined starting from the logits of the model, instead of the post-softmax output. These are the \textit{functional} variants of the losses - equivalent object-oriented variants are also present in most frameworks.}{code:cross_entropy}
from torch.nn import functional as F
# Binary cross-entropy
F.binary_cross_entropy
# Binary cross-entropy accepting logits
F.binary_cross_entropy_with_logits
# Cross-entropy, but from logits
F.cross_entropy
# Cross-entropy with log f(x) as inputs
F.nll_loss
\end{mypy}

To understand why we would need this, consider the $i$-th term of the cross-entropy in terms of the logits $\mathbf{p}$:
$$
- \log\left( \frac{\exp{p_i}}{\sum_j \exp{p_j}} \right) \,.
$$
This term can give rise to several numerical issues, notably due to the interplay between the (potentially unbounded) logits and the exponentiation. To solve this, we first rewrite it as:
$$
- \log\left( \frac{\exp{p_i}}{\sum_j \exp{p_j}} \right) = -p_i + \underbrace{\log\left(\sum_j \exp p_j\right)}_{\triangleq \; \text{logsumexp}(\mathbf{p})}
$$
The first term does not suffer from instabilities, while the second term (the \textbf{logsumexp} of the logits) is a function of the entire logits’ vector, and it can be shown to be invariant for a given scalar $c \ge 0$ in the following sense:\footnote{\url{https://gregorygundersen.com/blog/2020/02/09/log-sum-exp/}}
$$
\text{logsumexp}(\mathbf{p})=\text{logsumexp}(\mathbf{p} - c) +c
$$
Note that $\nabla \text{softmax}(\bullet) =  \text{logsumexp}(\bullet)$. By taking $c = \max(\mathbf{p})$ we can prevent numerical problems by bounding the maximum logit value at $0$. However, this is only possible if we have access to the original logits, which is why numerically stable variants of the cross-entropy require them as inputs. This creates a little amount of ambiguity, in that the softmax can now be included as either part of the model, or as part of the loss function.

\subsection{Calibration and classification}

We close the chapter by briefly discussing the important topic of \textbf{calibration} of the classifier. To understand it, consider the following fact: although our model provides an entire distribution over the possible classes, our training criterion only targets the maximization of the true class. Hence, the following sentence is justified:

\begin{quote}The predicted class of $f(\mathbf{x})$ is $\underset{i}{\arg\max} \; \idx{f(\mathbf{x})}{i}$.\end{quote}

Instead, this more general sentence might not be correct:

\begin{quote} The probability of $\mathbf{x}$ being of class $i$ is $\idx{f(\mathbf{x})}{i}$.\end{quote}

When the confidence scores of the network match the probability of a given prediction being correct, we say the network’s outputs are \textbf{calibrated}.

\begin{definition}[Calibration] $\,$

A classification model $f(\mathbf{x})$ giving in output the class probabilities is said to be calibrated if the following holds for any possible prediction:
$$
\idx{f(\mathbf{x})}{i} = p(y=i \;\vert\; \mathbf{x})
$$
\end{definition}

Although the cross entropy should recover the conditional probability distribution over an unrestricted class of models and in the limit of infinite data \cite{hastie2009elements}, in practice the mismatch between the two may be high \cite{blasiok2024does}, especially for the more complex models we will introduce later on.

To understand the difference between accuracy and calibration, consider these two scenarios. First, consider a binary classification model that has perfect accuracy, but always predicts the true class with $0.8$ confidence. In this case, the model is clearly \textit{underconfident} in its predictions, since by looking at the confidence we may assume that $20\%$ of them would be incorrect. Second, consider a $4$ class problem with perfectly balanced classes, with a model that always predict $[0.25, 025, 0.25, 0.25]$. In this case, the model is perfectly calibrated, but useless from the point of view of accuracy.

Having access to a calibrated model is very important in situations in which different predictions may have different costs. This can be formalized by defining a so-called \textit{cost matrix} assigning a cost $C_{ij}$ for any input of class $i$ predicted as class $j$. A standard example is a binary classification problem having the matrix of costs shown in Table \ref{tab:cost_matrix}.

\begin{table}[h]
\centering
\caption{Example of cost matrix for a classification problem having asymmetric costs of misclassification.}
\label{tab:cost_matrix}
\begin{tabular}{@{}lcc@{}}
\toprule
 & True class 0 & True class 1 \\ \midrule
Predicted class 0 & 0 & 10 \\
Predicted class 1 & 1 & 0 \\ \bottomrule
\end{tabular}
\end{table}

We can interpret Table \ref{tab:cost_matrix} as follows: making a correct prediction incurs no cost, while making a false negative mistake (0 instead of 1) is 10 times more costly than making a false positive mistake. As an example, an incorrect false negative mistake in a medical diagnosis is much worse than a false positive error, in which a further test may correct the mistake. A calibrated model can help us in better estimating the average risk of its deployment, and to fine-tune our balance of false positive and false negative mistakes.

To see this, denote by $\mathbf{C} \sim (m, m)$ the generic matrix of costs for a multiclass problem (like the $2\times2$ matrix in Table \ref{tab:cost_matrix}). The rational choice is to select a class which minimizes the expected cost based on the scores assigned by our model:
$$
\underset{i}{\arg\min} \sum_{j=1}^m C_{ij} \idx{f(\mathbf{x})}{j}
$$
If $C_{ij} = 1$ whenever $i \neq j$ and $0$ otherwise, this reduces to selecting the argmax of $f$, but for a general matrix of costs the choice of predicted class will be influenced by the relative costs of  making specific mistakes. This is a simple example of \textbf{decision theory} \cite{bishop2006pattern}.

\subsection{Estimating the calibration error}

To estimate whether a model is calibrated we can bin its predictions, and compare its calibration to the accuracy in each bin. To this end, suppose we split the interval $[0,1]$ into $b$ equispaced bins, each of size $1/b$. Take a validation set of size $n$, and denote by $\mathcal{B}_i$ the elements whose confidence falls into bin $i$. For each bin, we can further compute the average confidence $p_i$ of the model (which will be, approximately, in the middle of the bin), and the average accuracy $a_i$. Plotting the set of pairs $(a_i, p_i)$ on an histogram is called a \textbf{reliability diagram}, as shown in Figure \ref{fig:calibration_plot}. To have a single, scalar metric of calibration we can use, for example, the \textbf{expected calibration error} (ECE):

\begin{figure}[t]
    \centering
    \includegraphics[width=0.6\textwidth]{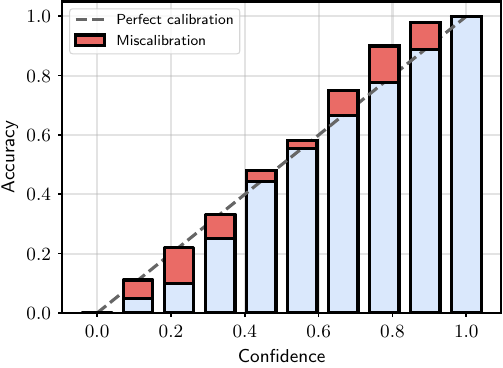}
    \caption{An example of reliability plot with $b=10$ bins. The blue bars show the average accuracy of the model on that bin, while the red bars show the miscalibration for the bin, which can be either under-confident (below the diagonal) or over-confident (above the diagonal). The weighted sum of the red blocks is the ECE in \eqref{eq:ece}.}
    \label{fig:calibration_plot}
\end{figure}

\begin{equation}
\displaystyle\text{ECE}= \sum_i \eqnmarkbox[drawred]{node2}{\frac{\lvert\mathcal{B}_i\rvert}{n}} \eqnmarkbox[drawblue]{node}{\left|a_i - p_i\right|}
\label{eq:ece}
\end{equation}
\annotate[yshift=1em]{above,right}{node}{Calibration for bin $i$}
\annotate[yshift=-1em]{below,right}{node2}{Fraction falling into bin $i$}

\vspace{1em}
Other metrics, such as the maximum over the bins, are also possible. If the model is found to be uncalibrated, modifications need to be made. Examples include rescaling the predictions via temperature scaling \cite{guo2017calibration} or optimizing with a different loss function such as the focal loss \cite{mukhoti2020calibrating}.

We close by mentioning an alternative to direct calibration of the model, called \textbf{conformal prediction}, which has become popular recently \cite{angelopoulos2021gentle}. Suppose we fix a threshold $\gamma$, and we take the set of classes predicted by the model whose corresponding probability is higher than $\gamma$:
\begin{equation}
\mathcal{C}(\mathbf{x}) = \left\{ i \mid \idx{f(\mathbf{x})}{i} > \gamma \right\}
\label{eq:support_set}
\end{equation}

\begin{figure}[t]
    \centering
    \includegraphics[width=0.6\textwidth]{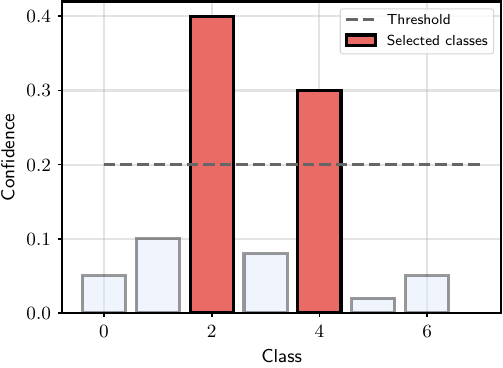}
    \caption{Calibration by turning the model's output into a set: we return all classes whose predicted probability exceeds a given threshold. By properly selecting the threshold we can bound the probability of the true class being found in the output set.}
    \label{fig:conformal_prediction}
\end{figure}

i.e., the answer of the model is now a \textit{set} $\mathcal{C}(\mathbf{x})$ of potential classes. An example is shown in Figure \ref{fig:conformal_prediction}. The idea of conformal prediction is to select the minimum $\gamma$ such that the probability of finding the correct class $y$ in the set is higher than a user-defined error $\alpha$:\footnote{Note that it is always possible to satisfy this property by selecting $\gamma = 0$, i.e., including all classes in the output set.}
\begin{equation}
p(y \in \mathcal{C}(\mathbf{x})) \ge 1 -\alpha
\label{eq:conformal_prediction}
\end{equation}
Intuitively, there is an inversely proportional relation between $\gamma$ and $\alpha$. Conformal prediction provides automatic algorithms to guarantee \eqref{eq:conformal_prediction} at the cost of not having a single class in output anymore.

\clearpage

\section*{From theory to practice}

\begin{wrapfigure}{r}{3.0cm}
\vspace{-3em}\includegraphics[width=3.0cm]{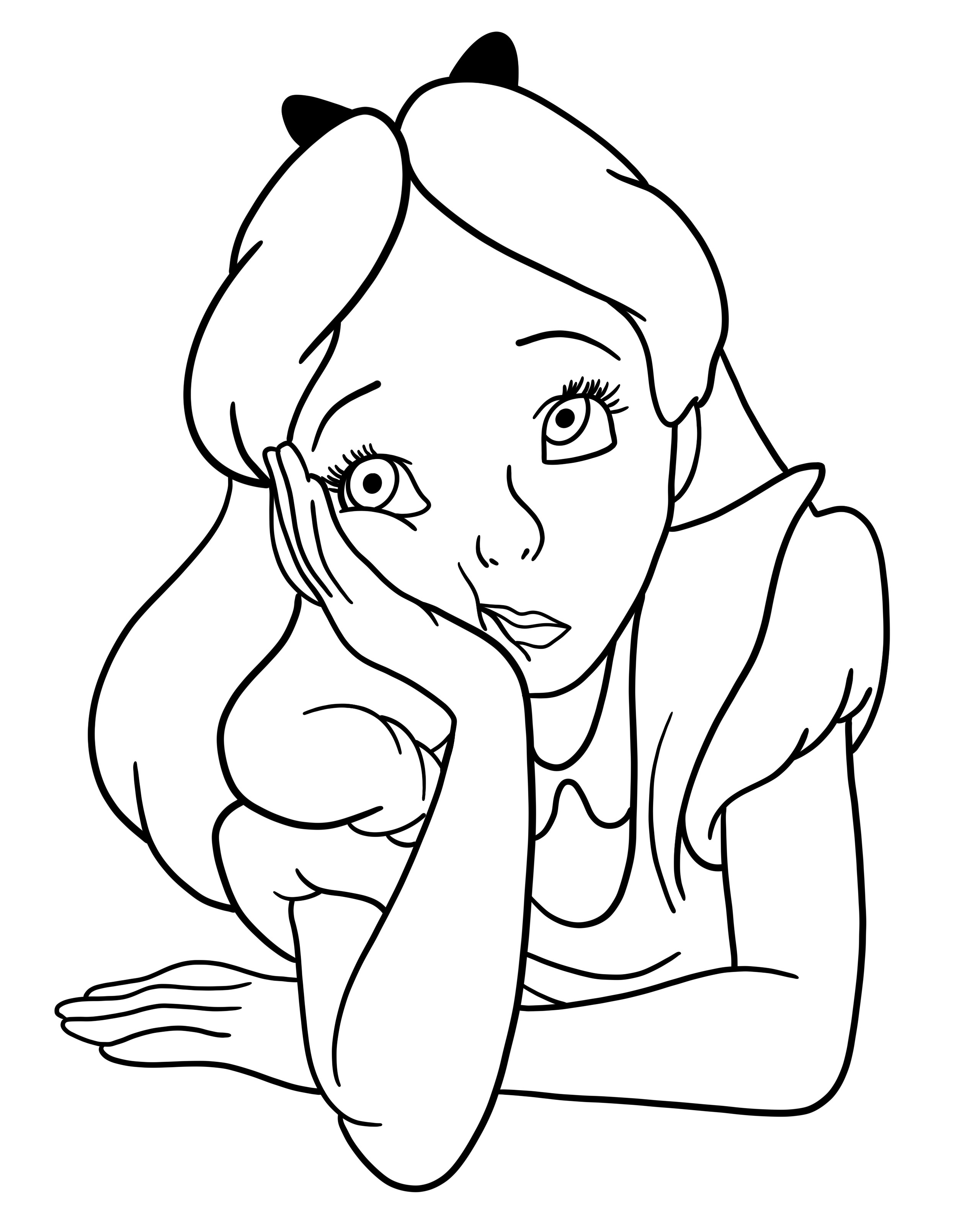}
\vspace{-3em}
\end{wrapfigure}

From Chapter \ref{chap:preliminaries} you should have a good grasp of NumPy, JAX, and PyTorch's \mintinline{python}{torch.tensor}. This is all that is needed for this chapter, and nothing else is required. From the next chapter we will progress to their higher-level APIs.

I suggest a short exercise to let you train your first differentiable model from scratch:

\begin{enumerate}
\item Load a toy dataset: for example, one of those contained in scikit-learn datasets module.\footnote{\url{https://scikit-learn.org/stable/datasets/toy_dataset.html}}
\item Build a linear model (for regression or classification depending on the dataset). Think about how to make the code as modular as possible: as we will see, you will need at least two functions, one for initializing the parameters of the model and one for computing the model's predictions.
\end{enumerate}

\begin{enumerate}\addtocounter{enumi}{2}
\item Train the model via gradient descent. For now you can compute the gradients manually: try to imagine how you can make also this part modular, i.e., how do you change the gradient's computation if you want to dynamically add or remove the bias from a model?
\item Plot the loss function and the accuracy on an independent test set. If you know some standard machine learning, you can compare the results to other supervised learning models, such as a decision tree or a $k$-NN, always using scikit-learn.
\end{enumerate}

%% file: 5_fully_connected_models.tex
\chapter{Fully-connected models}
\label{chap:fully_connected_models}

\begin{supportbox}{About this chapter}
In this chapter we show how differentiable models can be built by composing a sequence of so-called \textit{fully-connected layers}. For historical reasons, these models are also known as multilayer perceptrons (MLPs). MLPs interleave linear blocks (similar to Chapter \ref{chap:linear_models}) with non-linear functions, sometimes called \textit{activation functions}.
\end{supportbox}

\section{The limitations of linear models}

Linear models are fundamentally limited, in the sense that by definition they cannot model non-linear relationships across features. As an example, consider two input vectors $\mathbf{x}$ and $\mathbf{x}^\prime$, which are identical except for a single feature indexed by $j$:
$$
x_i^\prime=\begin{cases} x_i & \text{ if } i \neq j \\ 2x_i & \text{ otherwise } \end{cases}
$$
For example, this can represent two clients of a bank, which are identical in all aspects except for their income, with $\mathbf{x}^\prime$ having double the income of $\mathbf{x}$. If $f$ is a linear model (with no bias) we have:

\vspace{1em}
\begin{equation*}
f(\mathbf{x}^\prime) = \eqnmarkbox[drawred]{node}{f(\mathbf{x})} + \eqnmarkbox[drawblue]{node2}{w_jx_j}
\end{equation*}
\annotate[yshift=1em]{above,left}{node}{Original output}
\annotate[yshift=-1em]{below,left}{node2}{Change induced by $x_j^\prime = 2x_j$}

\vspace{1em}
Hence, the only consequence of the change in input is a small linear change of output dictated by $w_j$. Assuming we are scoring the users, we may wish to model relationships such as “\textit{an income of 1500 is low, except if the age < 30}”.\footnote{You probably shouldn't do credit scoring with machine learning anyways.} Clearly, this cannot be done with a linear model due to the analysis above. 

\begin{figure}[t]
    \centering
    \hspace{1em}\includegraphics[width=0.7\textwidth]{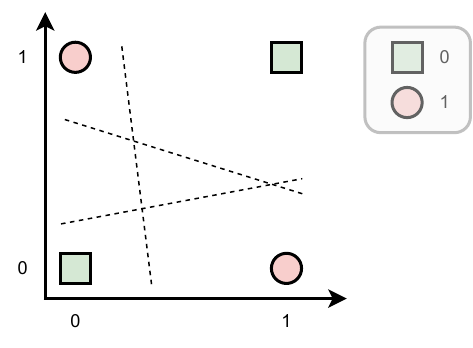}
    \caption{Illustration of the XOR dataset: green squares are values of one class, red circles are values of another class. No linear model can separate them perfectly (putting all squares on one side and all circles on the other side of the decision boundary). We say that the dataset is not \textbf{linearly separable}.}
    \label{fig:xor}
\end{figure}

The prototypical example of this is the XOR dataset, a two-valued dataset where each feature can only take values in $\left\{0, 1\right\}$. Hence, the entire dataset is given by only 4 possibilities:
$$
f([0,0])=0 \;,\; f([0,1])=1\;,\;f([1,0])=1\;,\;f([1,1])=0
$$
where the output is positive whenever \textit{only one} of the two inputs is positive. Despite its simplicity, this is also \textbf{non-linearly separable}, and cannot be solved with 100\% accuracy by a linear model - see Figure \ref{fig:xor} for a visualization.

\section{Composition and hidden layers}

\addclock A powerful idea in programming is decomposition, i.e., breaking down a problem into its constituent parts recursively, until each part can be expressed in simple, manageable operations. Something similar can be achieved in our case by imagining that our model $f$ is, in fact, the composition of two trainable operations:

$$
f(\mathbf{x})=(f_2 \circ f_1)(\mathbf{x})
$$

where $f_2 \circ f_1$ is the composition of the two functions: $(f_2 \circ f_1)(\mathbf{x}) = f_2(f_1(\mathbf{x}))$, and we assume that each function instantiates its own set of trainable parameters. We can keep subdividing the computations:

$$
f(\mathbf{x})=(f_l\circ f_{l-1}\circ \cdots\circ f_2\circ f_1)(\mathbf{x})
$$

where we now have a total of $l$ functions that are being composed. Note that as long as each $f_i$ does not change the “type” of its input data, we can chain together as many of these transformations as we want, and each one will add its own set of trainable parameters. 

For example, in our case the input $\mathbf{x}$ is a vector, hence any vector-to-vector operation (e.g., a matrix multiplication $f_i(\mathbf{x}) = \mathbf{W}\mathbf{x})$ can be combined together an endless number of times. However, some care must be taken. Suppose we chain together two different linear projections:
\begin{gather}
\mathbf{h} = f_1(\mathbf{x}) = \mathbf{W}_1\mathbf{x} +\mathbf{b}_1 \\y=f_2(\mathbf{h})=\mathbf{w}^\top_2\mathbf{h} + b_2
\end{gather}
It is easy to show that the two projections “collapse” into a single one:
$$
y = \underbrace{(\mathbf{w}^\top_2\mathbf{W}_1)}_{\triangleq\;\mathbf{A}}\mathbf{x} + \underbrace{(\mathbf{w}_2^\top\mathbf{b}_1 + b_2)}_{\triangleq \;\mathbf{c}} = \mathbf{A}\mathbf{x}+\mathbf{c}
$$
The idea of \textbf{fully-connected} (FC) models, also known as \textbf{multi-layer perceptrons} (MLPs) for historical reasons, is to insert a simple elementwise non-linearity $\phi : \mathbb{R} \rightarrow \mathbb{R}$ in-between projections to avoid the collapse:

\vspace{0.5em}

\begin{equation}
\mathbf{h} = f_1(\mathbf{x}) = \eqnmarkbox[drawred]{node}{\phi}\left(\mathbf{W}_1\mathbf{x} +\mathbf{b}_1\right)
\label{eq:mlps_single_hidden_layer_1}
\end{equation}%
\annotate[yshift=0.7em]{above,right}{node}{Element-wise non-linearity}
\begin{equation}
y=f_2(\mathbf{h})=\mathbf{w}^\top_2\mathbf{h} + b_2
\label{eq:mlps_single_hidden_layer_2}
\end{equation}

\clearpage

The second block can be linear, as in \eqref{eq:mlps_single_hidden_layer_2}, or it can be wrapped into another non-linearity depending on the task (e.g., a softmax function for classification). The function $\phi$ can be any non-linearity, e.g., a polynomial, a square root, or the sigmoid function $\sigma$. As we will see in the next chapter, choosing it has a strong effect on the gradients of the model and, consequently, on optimization, and the challenge is to select a $\phi$ which is “non-linear enough” to prevent the collapse while staying as close as possible to the identity in its derivative. A good default choice is the so-called \textbf{rectified linear unit} (ReLU). 

\begin{definition}[Rectified linear unit] \addbottle $\,$

The \textbf{rectified linear unit} (ReLU) is defined elementwise as:
\begin{equation}
\textnormal{ReLU}(s)=\max(0,s)
\label{eq:relu}
\end{equation}
\end{definition}

We will have a lot more to say on the ReLU in the next chapter. With the addition of $\phi$, we can now chain as many transformations as we want:
\begin{equation}
y = \mathbf{w}_l^\top\phi\left( \mathbf{W}_{l-1}\left(\phi\left( \mathbf{W}_{l-2}\phi\left(\cdots \right)   +\mathbf{b}_{l-2} \right)\right)+\mathbf{b}_{l-1} \right)
\label{eq:mlps_multiple_hidden_layer}
\end{equation}
In the rest of the chapter we focus on analyzing training and approximation properties of this class of models. First, however, a brief digression on naming conventions.

\subsection*{On neural network terminology}

As we already mentioned, neural networks have a long history and a long baggage of terminology, which we briefly summarize here. Each $f_i$ is called a \textbf{layer} of the model, with $f_l$ being the \textbf{output layer}, $f_{i}, i=1,\ldots,l-1$ the \textbf{hidden layers} and, with a bit of notational overloading, $\mathbf{x}$ being the \textbf{input layer}. With this terminology, we can restate the definition of the \textbf{fully-connected layer} in batched form below.

\begin{definition}[Fully-connected layer] \addbottle $\,$

For a batch of $n$ vectors, each of size $c$, represented as a matrix $\mathbf{X} \sim (n,c)$, a \textbf{fully-connected} (FC) layer is defined as:
\begin{equation}
\textnormal{FC}(\mathbf{X}) = \phi\left(\mathbf{X}\mathbf{W} + \mathbf{b}\right)
\label{eq:fully_connected_layer}
\end{equation}
The parameters of the layer are the matrix $\mathbf{W} \sim (c,c^\prime)$ and the bias vector $\mathbf{b} \sim (c^\prime)$, for a total of $(c+1)c^\prime$ parameters (assuming $\phi$ does not have parameters). Its hyper-parameters are the width $c^\prime$ and the non-linearity $\phi$.
\end{definition}

The outputs $f_i(\mathbf{x})$ are called the \textbf{activations} of the layer, where we can sometimes distinguish between the \textbf{pre-activation} and the \textbf{post-activation} (before and after the non-linearity). The non-linearity $\phi$ itself can be called the \textbf{activation function}. Each output of $f_i$ is called a \textbf{neuron}. Although much of this terminology is outdated, it is still pervasive and we will use it when needed.

The size of the each layer (the shape of the output) is an hyperparameter that can be selected by the user, as it only influences the input shape of the next layer, which is known as the \textbf{width} of the layer. For a large number of  layers, the number of hyperparameters grows linearly and their selection becomes a combinatorial task. We will return on this point in Chapter \ref{chap:deep_cnns}, when we discuss the design of models with dozens (or hundreds) of layers.

\begin{mypy}{The FC layer in \eqref{eq:fully_connected_layer} implemented as an object in PyTorch. We require a special syntax to differentiate trainable parameters, such as $\mathbf{W}$, from other non-trainable tensors: in PyTorch, this is obtained by wrapping the tensors in a {\footnotesize\mintinline{python}{Parameter}} object. PyTorch also has its collection of layers in {\footnotesize\mintinline{python}{torch.nn}}, including the FC layer (implemented as {\footnotesize\mintinline{python}{torch.nn.Linear}}).}{code:fully_connected_layer}
class FullyConnectedLayer(nn.Module):
  def __init__(self, c: int, cprime: int):
    super().__init__()
    # Initialize the parameters
    self.W = nn.Parameter(
                torch.randn(c, cprime))
    self.b = nn.Parameter(
                torch.randn(1, cprime))

  def forward(self, x):
    return relu(x @ self.W + self.b)
\end{mypy}

The layer concept is also widespread in common frameworks. A layer such as \eqref{eq:fully_connected_layer} can be defined as an object having two functions: an initialization function that randomly initializes all parameters of the model based on the selected hyper-parameters, and a call function that provides the output of the layer itself. See Box \ref{code:fully_connected_layer} for an example. Then, a model can be defined by chaining together instances of such  layers. For example, in PyTorch this can be achieved by the {\footnotesize\mintinline{python}{Sequential}} object:

\vspace{1em}
{\begin{center}\footnotesize
\begin{minted}{python}
model = nn.Sequential(
    FullyConnectedLayer(3, 5), 
    FullyConnectedLayer(5, 4)
)
\end{minted}
\end{center}}
 
Note that from the point of view of their input-output signature, there is no great difference between a layer as defined in Box \ref{code:fully_connected_layer} and a model as defined above, and we could equivalently use \mintinline{python}{model} as a layer of a larger one. This compositionality is a defining characteristic of differentiable models.

\subsection{Approximation properties of MLPs} \addteacup

Training MLPs proceeds similarly to what we discussed for linear models. For example, for a regression task, we can minimize the mean-squared error:
$$
\underset{\left\{\mathbf{W}_k, \mathbf{b}_k\right\}_{k=1}^l}{\min} \;\; \frac{1}{n}\sum_{i} \left(y_i - f(\mathbf{x}_i)\right)^2
$$
where the minimization is now done on all parameters of the model simultaneously. We will see in the next chapter a general procedure to compute gradients in this case. 

For now, we note that the main difference with respect to having a linear model is that adding an hidden layer makes the overall optimization problem non-convex, with multiple local optima depending on the initialization of the model. This is an important aspect historically, as alternative approaches to supervised learning (e.g., support vector machines \cite{hofmann2008kernel}) provide non-linear models while remaining convex. However, the results of the last decade show that highly non-convex models can achieve significantly good performance in many tasks.\footnote{The reason differentiable models generalize so well is an interesting, open research question, to which we return in Chapter \ref{chap:deep_cnns}. Existing explanations range from an implicit bias of (stochastic) gradient descent \cite{pesme2021implicit} to intrinsic properties of the architectures themselves \cite{arpit2017closer,teney2024neural}.}

From a theoretical perspective, we can ask what is the significance of having added hidden layers, i.e., if linear models can only solve tasks which are linearly separable, what is instead the class of functions that can be approximated by adding hidden layers? As it turns out, having a single hidden layer is enough to have \textbf{universal approximation} capabilities. A seminal result in this sense was proved by G. Cybenko in 1989 \cite{cybenko1989approximation}.

\begin{theorem}[Universal approximation of MLPs]

Given a continuous function $g: \mathbb{R}^d \rightarrow \mathbb{R}$, we can always find a model $f(\mathbf{x})$ of the form \eqref{eq:mlps_single_hidden_layer_1}-\eqref{eq:mlps_single_hidden_layer_2} (an MLP with a single hidden layer) and sigmoid activation functions, such that for any $\varepsilon > 0$:

$$
\lvert f(\mathbf{x}) - g(\mathbf{x})\rvert\le\varepsilon \;,\;\forall \mathbf{x}
$$

where the result holds over a compact domain. Stated differently, one-hidden-layer MLPs are “dense” in the space of continuous functions.

\end{theorem}

The beauty of this theorem should not distract from the fact that this is purely a theoretical construct, that makes use of the fact that the width of the hidden layer of the model can grow without bounds. Hence, for any $\mathbf{x}$ for which the previous inequality does not hold, we can always add a new unit to reduce the approximation error (see Appendix \ref{sec:universal_approximation}). In fact, it is possible to devise classes of functions on which the required number of hidden neurons grows exponentially in the number of input features \cite{bengio2009learning}.\footnote{One of these problems, the \textit{parity} problem, is closely connected to the XOR task: \url{https://blog.wtf.sg/posts/2023-02-03-the-new-xor-problem/}.}

Many other authors, such as \cite{hornik1991approximation}, have progressively refined this result to include models with fundamentally any possible activation function, including ReLUs. In addition, universal approximation can also be proved for models having finite \textit{width} but possibly infinite \textit{depth} \cite{lu2017expressive}. A separate line of research has investigated the approximation capabilities of \textit{overparameterized} models, in which the number of parameters exceeds the training data. In this case, training to a global optimum can be proved in many interesting scenarios \cite{du2018gradient,allen2019learning} (informally, for sufficiently many parameters, the model can achieve the minimum of the loss on each training sample and, hence, the global minimum of the optimization problem).  See Appendix \ref{sec:universal_approximation} for a one-dimensional visualization of Cybenko's theorem.

Approximation and learning capabilities of differentiable models are immense fields of study, with countless books devoted to them, and we have only mentioned some significant results here. In the rest of the book, we will be mostly concerned with the effective design of the models themselves, whose behavior can be more complex and difficult to control (and design) than these theorems suggest.

\section{Stochastic optimization}

To optimize the models we can perform gradient descent on the corresponding empirical risk minimization problem. However, this can be hard to achieve when $n$ (the size of the dataset) grows very large. We will see in the next chapter that computing the gradient of the loss requires a time linear in the number of examples, which becomes unfeasible or slow for $n$ in the order of $10^4$ or more, especially for large models (memory issues aside).

Fortunately, the form of the problem lends itself to a nice approximation, where we use subsets of the data to compute a descent direction. To this end, suppose that for iteration $t$ of gradient descent we sample a subset $\mathcal{B}_t \subset \mathcal{S}_n$ of $r$ points (with $r \ll n$) from the dataset, which we call a \textbf{mini-batch}. We can compute an approximated loss by only considering the mini-batch as:
\begin{equation}
\widetilde{L}_t=\eqnmarkbox[drawred]{node}{\frac{1}{r} {\sum_{(x_i, y_i) \in \mathcal{B}_t}}} l(y_i,f(x_i)) \approx \eqnmarkbox[drawblue]{node2}{\frac{1}{n} {\sum_{(x_i, y_i) \in \mathcal{S}_n}}} l(y_i,f(x_i))
\end{equation}
\annotate[yshift=-1em]{below,left}{node}{Mini-batch}
\annotate[yshift=-1em]{below,right}{node2}{Full dataset}

If we assume the elements in the mini-batch are sampled i.i.d. from the dataset, $\widetilde{L}_t$ is a Monte Carlo approximation of the full loss, and the same holds for its gradient. However, its computational complexity grows only with $r$, which can be controlled by the user. Roughly speaking, lower dimensions $r$ of the mini-batch result in faster iterations with higher gradient variance, while higher $r$ results in slower, more precise iterations. For large models, memory is in general the biggest bottleneck, and the mini-batch size $r$ can be selected to fill up the available hardware for each iteration.

\begin{figure}
    \centering
    \includegraphics[width=0.85\textwidth]{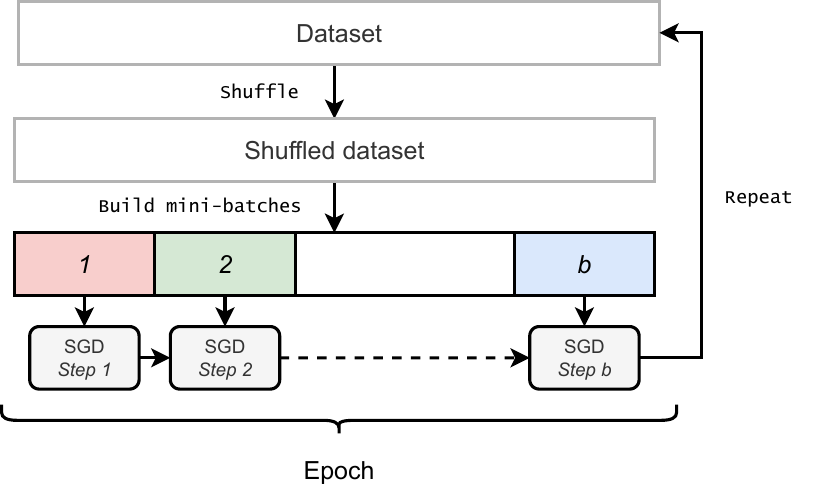}
    \caption{Building the mini-batch sequence: after shuffling, stochastic optimization starts at mini-batch $1$, which is composed of the first $r$ elements of the dataset. It proceeds in this way to mini-batch $b$ (where $b = \frac{n}{r}$, assuming the dataset size is perfectly divisible by $r$). After one such \textit{epoch}, training proceed with mini-batch $b+1$, which is composed of the first $r$ elements of the shuffled dataset. The second epoch ends at mini-batch $2b$, and so on.}
    \label{fig:building_mini_batches}
\end{figure}

Gradient descent applied on mini-batches of data is an example of \textbf{stochastic gradient descent} (SGD). Due to the properties discussed above, SGD can be proven to converge to a minimum in expectation, and it is the preferred optimization strategy when training differentiable models.

The last remaining issue is how to select the mini-batches. For large datasets, sampling elements at random can be expensive, especially if we need to move them back and forth from the GPU memory. An intermediate solution that lends itself to easier optimization is the following:

\begin{enumerate}
    \item Begin by shuffling the dataset.
    \item Then, subdivide the original dataset into mini-batches of $r$ \textit{consecutive} elements and process each of them sequentially. Assuming a dataset of size $n=rb$, this results in $b$ mini-batches and hence $b$ steps of SGD. If we are executing the code on a GPU, this step includes sending the mini-batch to the GPU memory.
    \item After completing all mini-batches constructed in this way, return to point 1 and iterate.
\end{enumerate}

\begin{mypy}{Building the mini-batch sequence with PyTorch's data loader: all frameworks provide similar tools.}{code:data_loader}
# A dataset composed by two tensors
dataset = torch.utils.data.TensorDataset(
    torch.randn(1000, 3), 
    torch.randn(1000, 1))

# The data loader provides 
# shuffling and mini-batching
from torch.utils.data import DataLoader
dataloader = DataLoader(dataset, 
                shuffle=True, 
                batch_size=32)

for xb, yb in dataloader:
  # Iterating over mini-batches (one epoch)
  # xb has shape (32, 3)
  # yb has shape (32, 1)
\end{mypy}

One complete loop of this process is called an \textbf{epoch} of training, and it is a very common hyper-parameter to specify (e.g., for a dataset of $1000$ elements and mini-batches of $20$ elements, “\textit{training for 5 epochs}” means training for $250$ iterations). The expensive shuffling operation is only done once per epoch, while in-between an epoch mini-batches can be quickly pre-fetched and optimized by the framework. This is shown schematically in Figure \ref{fig:building_mini_batches}. Most frameworks provide a way to organize the dataset into elements that can be individually  indexed, and a separate interface to build the mini-batch sequence. In PyTorch, for example, this is done by the {\footnotesize\mintinline{python}{Dataset}} and {\footnotesize\mintinline{python}{DataLoader}} interfaces, respectively - see Box \ref{code:data_loader}.

\begin{figure}[t]
    \centering
    \includegraphics[width=\textwidth]{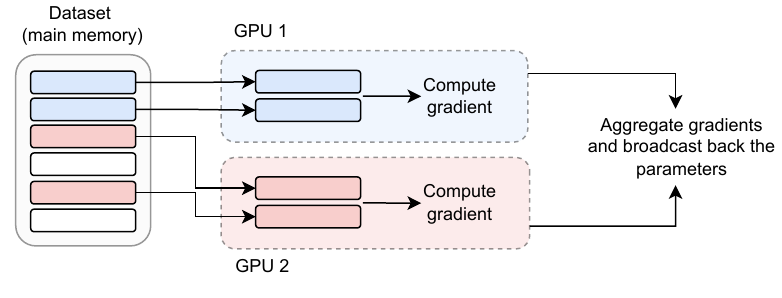}
    \caption{A simple form of distributed stochastic optimization: we process one mini-batch per available machine or GPU (by replicating the weights on each of them) and sum or average
    the corresponding gradients before broadcasting back the result (which is valid due to the linearity of the gradient operation). This requires a synchronization mechanism across the machines or the GPUs.}
    \label{fig:mini_batch}
\end{figure}

This setup also leads itself to a simple form of parallelism across GPUs or across machines. If we assume each machine is large enough to hold an entire copy of the model's parameters, we can process different mini-batches in parallel over the machines and then sum their local contributions for the final update, which is then broadcasted back to each machine. This is called a \textbf{data parallel} setup in PyTorch,\footnote{\url{https://pytorch.org/tutorials/intermediate/ddp_tutorial.html}} and it is shown visually in Figure \ref{fig:mini_batch}. More complex forms of parallelism, such as \textbf{tensor parallelism}, are also possible, but we do not cover them in this book.

\section{Activation functions}
\label{sec:activation_functions}

We close the chapter by providing a brief overview on the selection of activation functions. As we stated in the previous section, almost any element-wise non-linearity is theoretically valid. However, not all choices have good performance. As an example, consider a simple polynomial function, for some user-defined positive integer $p$:

$$
\phi(s)=s^p
$$

\clearpage

For large $p$, this will grow rapidly on both sides, compounding across layers and resulting in models which are hard to train and with numerical instabilities. 

Historically, neural networks were introduced as approximate models of biological neurons (hence, the name \textit{artificial NNs}). In this sense, the weights $\mathbf{w}^\top$ in the dot product $\mathbf{w}^\top \mathbf{x}$ were simple models of synapses, the bias $b$ was a threshold, and the neuron was “activated” when the cumulative sum of the inputs surpassed the threshold:

$$
s = \mathbf{w}^\top\mathbf{x}-b \;,\; \phi(s)= \mathbb{I}_{s \ge 0}
$$

where $\mathbb{I}_{b}$ is an indicator function which is $1$ when $b$ is true, $0$ otherwise. Because this activation function is non-differentiable, the sigmoid $\sigma(s)$ can be used as a soft-approximation. In fact, we can define a generalized sigmoid function with a tunable slope $a$ as $\sigma_a(s) = \sigma(as)$, and we have:

$$
\lim_{a \rightarrow \infty}\sigma_a(s)=\mathbb{I}_{s \ge 0}
$$

Another common variant was the hyperbolic tangent, which is a scaled version of the sigmoid in $[-1,+1]$:

$$
\tanh(s)=2\sigma(s)-1
$$

Modern neural networks, popularized by AlexNet in 2012 \cite{krizhevsky2012imagenet}, have instead used the ReLU function in \eqref{eq:relu}. The relative benefits of ReLU with respect to sigmoid-like functions will be discussed in the next chapter. We note here that ReLUs have several counter-intuitive properties. For example, they have a point of non-differentiability in $0$, and they have a large output sparsity since all negative inputs are set to $0$. This second property can result in what is known as “dead neurons”, wherein certain units have a constant $0$ output for all inputs. This can be solved by a simple variant of ReLU, known as \textbf{Leaky ReLU}:

\begin{equation}
\text{LeakyReLU}(s) = \begin{cases} s & \text{ if } s \ge 0 \\ \alpha s & \text{ otherwise } \end{cases} 
\label{eq:leaky_relu}
\end{equation}

for a very small $\alpha$, e.g., $\alpha = 0.01$. We can also train a different $\alpha$ for each unit (as the function is differentiable with respect to $\alpha$). In this case, we call the AF a \textbf{parametric ReLU} (PReLU) \cite{he2015delving}. Trainable activation functions are, in general, an easy way to add a small amount of flexibility with a minor amount of parameters -- in the case of PReLU, one per neuron.

Fully-differentiable variants of ReLU are also available, such as the \textbf{softplus}:

\begin{equation}
\text{softplus}(s)=\log(1+\exp(s))
\label{eq:softplus}
\end{equation}

The softplus does not pass through the origin and it is always greater than $0$. Another variant, the \textbf{exponential linear unit} (ELU), preserves the passage at the origin while switching the lower bound to $-1$:

\begin{equation}
\text{ELU}(s)=\begin{cases} s & \text{ if } s \ge 0 \\ \exp(s)-1 & \text{ otherwise } \end{cases}
\label{eq:elu}
\end{equation}

Yet another class of variants can be defined by noting the similarity of ReLU with the indicator function. We can rewrite the ReLU as:

$$
\text{ReLU}(s)=s \cdot \mathbb{I}_{s \ge 0}
$$

Hence, ReLU is identical to the indicator function on the negative quadrant, while replacing $1$ with $s$ on the positive quadrant. We can generalize this by replacing the indicator function with a weighting factor $\beta(s)$:

$$
\text{GeneralizedReLU}(s)=s \cdot \beta(s)
$$

\begin{figure}[t]
    \centering
    \captionsetup[subfigure]{labelformat=empty}
    \begin{subfigure}[b]{0.18\textwidth}
    \includegraphics[width=\textwidth]{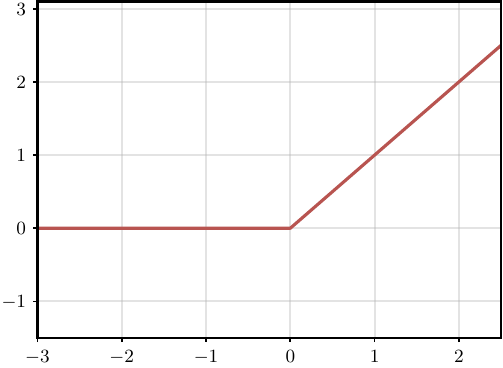}
    \caption{\footnotesize ReLU}
    \end{subfigure}
    \hfill
    \begin{subfigure}[b]{0.18\textwidth}
    \includegraphics[width=1.0\textwidth]{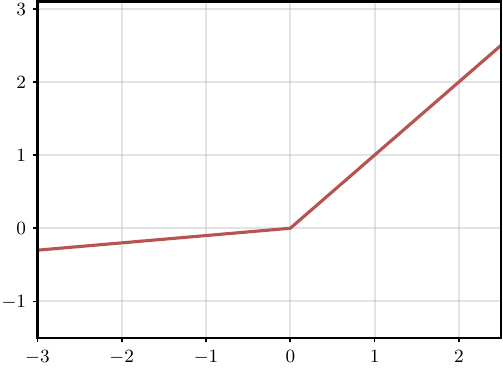}
    \caption{\footnotesize LeakyReLU}
    \end{subfigure}
    \hfill
    \begin{subfigure}[b]{0.18\textwidth}
    \includegraphics[width=1.0\textwidth]{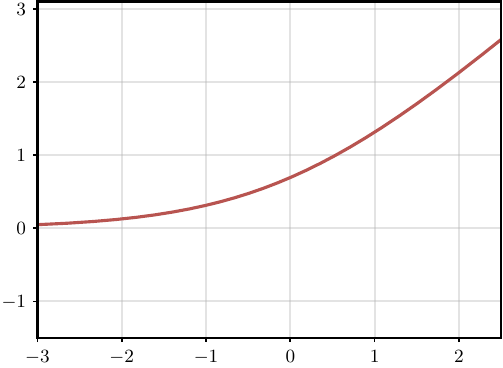}
    \caption{\footnotesize Softplus}
    \end{subfigure}
    \hfill
    \begin{subfigure}[b]{0.18\textwidth}
    \includegraphics[width=1.0\textwidth]{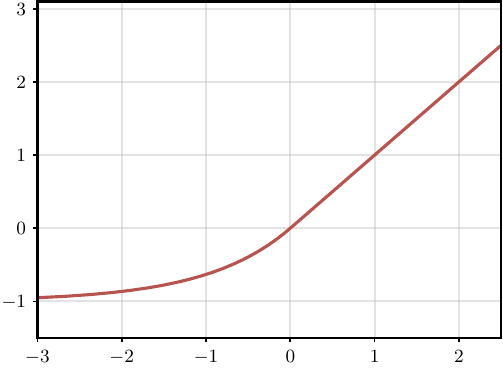}
    \caption{\footnotesize ELU}
    \end{subfigure}
    \hfill
    \begin{subfigure}[b]{0.18\textwidth}
    \includegraphics[width=1.0\textwidth]{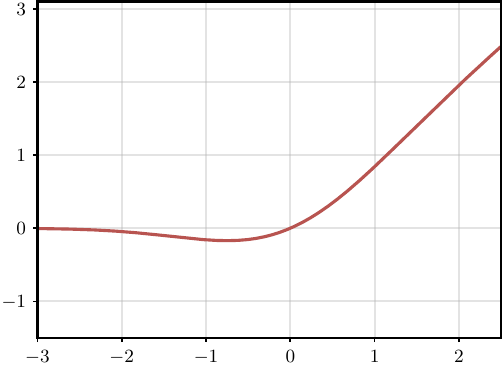}
    \caption{\footnotesize GELU}
    \end{subfigure}
    \hfill
    \caption{Visual comparison of ReLU and four variants: LeakyReLU \eqref{eq:leaky_relu}, Softplus \eqref{eq:softplus}, ELU \eqref{eq:elu}, and GELU. LeakyReLU is shown with $\alpha=0.1$ for better visualization, but in practice $\alpha$ can be closer to $0$ (e.g., $0.01$)..}
    \label{fig:activation_functions}
\end{figure}

Choosing $\beta(s)$ as the cumulative Gaussian distribution function, we obtain the \textbf{Gaussian ELU} (GELU) \cite{hendrycks2016gaussian}, while for $\beta(s) = \sigma(s)$ we obtain the \textbf{sigmoid linear unit} (SiLU)  \cite{hendrycks2016gaussian}, also known as the \textbf{Swish} \cite{ramachandran2017searching}. We plot some of these AFs in Figure \ref{fig:activation_functions}. Apart from some minor details (e.g., monotonicity in the negative quadrant), they are all relatively similar, and it is in general very difficult to obtain a significant boost in performance by simply replacing the activation function.

Multiple trainable variants of each function can be obtained by adding trainable parameters to the functions. For example, a common trainable variant of the Swish with four parameters $\left\{a,b,c,d\right\}$ is obtained as:
\begin{equation}
\text{Trainable-Swish}(s)=\sigma(as+b)(cs+d)
\label{eq:trainable_swish}
\end{equation}
We can also design \textbf{non-parametric} activation functions, in the sense of activation functions that do not have a fixed number of trainable parameters. For example, consider a generic set of (non-trainable) scalar functions $\phi_i$ indexed by an integer $i$. We can build a fully flexible activation function as a linear combination of $n$ such bases:
\begin{equation}
\phi(s) = \sum_{i=1}^n \alpha_i \phi_i(s) \,
\label{eq:non_parametric_af}
\end{equation}
where $n$ is an hyper-parameter, while the coefficients $\alpha_i$ are trained by gradient descent. They can be the same for all functions, or different for each layer and/or neuron. Based on the choice of $\phi_i$ we obtain different classes of functions: if each $\phi_i$ is a ReLU we obtain the \textbf{adaptive piecewise linear} (APL) function \cite{agostinelli2014learning}, while for more general kernels we obtain the \textbf{kernel activation function} (KAF) \cite{marra2018learning,scardapane2019kafnets}. Even more general models can be obtained by considering functions with multiple inputs and multiple outputs \cite{li2023generalized}. See \cite{apicella2021survey} for a survey.

In general, there is no answer to the question of “what is the best AF”, as it depends on the task, dataset, and architecture. ReLU is a common choice because it performs well, is highly optimized in code, and it has a minor cost overhead. It is important to consider the fundamental computational trade-off that, for a given budget, more complex AFs can result in having smaller width or smaller depth, potentially hindering the performance of the entire architecture. For this reason, AFs with a lot of trainable parameters are less common.

\topskip1pt
\vspace*{\fill}
\begin{supportbox}{Design variants}
Not every layer fits into the framework of \textit{linear projections} and \textit{element-wise} non-linearities, and we describe here three common variants. First, the \textbf{gated linear unit} (GLU) \cite{dauphin2017language} combines the structure of \eqref{eq:trainable_swish} with multiplicative (Hadamard) interactions:
\begin{equation}
f(\mathbf{x}) = \sigma\left(\mathbf{W}_1\mathbf{x}\right)\odot\left(\mathbf{W}_2\mathbf{x}\right)
\label{eq:glu}
\end{equation}
where $\mathbf{W}_1$ and $\mathbf{W}_2$ are trained. Another possibility, the SwiGLU, replaces the sigmoid in \eqref{eq:glu} with a Swish function \cite{shazeer2020glu}. Gated MLPs are in fact a popular choice in modern LLMs, see Chapter \ref{chap:transformers_in_practice}. 

\vspace{1em}
Second, in a \textbf{maxout} network \cite{goodfellow2013maxout} each unit produces the maximum of $k$ (hyper-parameter) different projections. Finally, replacing the linear projection $\mathbf{W}$ with a matrix of trainable non-linearities $W_{ij} \rightarrow \phi_{ij}(x_j)$ of the form \eqref{eq:non_parametric_af} has also been proposed recently under the name of \textbf{Kolmogorov-Arnold networks} (KANs, \cite{liu2024kan}):

\begin{equation*}
h_i = \sum_j \phi_{ij}(x_j)
\end{equation*}

\end{supportbox}

\vspace*{\fill}\clearpage

\section*{From theory to practice}

\begin{wrapfigure}{r}{3.0cm}
\vspace{-3em}\includegraphics[width=3.0cm]{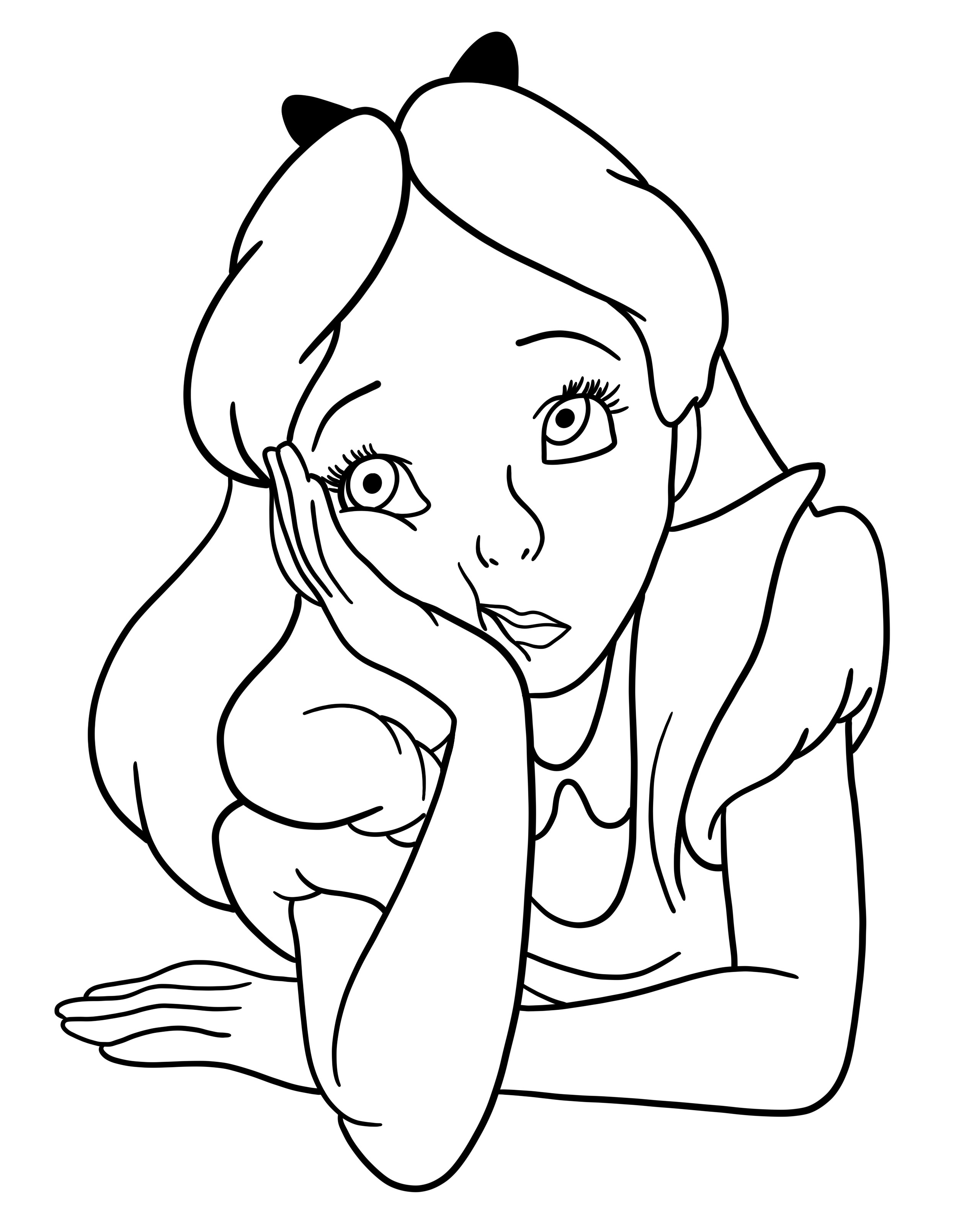}
\vspace{-3em}
\end{wrapfigure}

This chapter has introduced two key requirements for any general-purpose framework for training differentiable models:

\begin{enumerate}
\item A way to handle large datasets that need to be shuffled, separated into mini-batches, and moved back and forth from the GPU. In PyTorch, most of this is implemented via the {\mintinline{python}{Dataset}} and \mintinline{python}{DataLoader} interfaces, as in Box \ref{code:data_loader}.
\end{enumerate}

\begin{enumerate}\addtocounter{enumi}{1}
\item A mechanism to build models from the combination of basic blocks, known as \textit{layers}. In PyTorch, layers are implemented in the \mintinline{python}{torch.nn} module, and they can be composed via the \mintinline{python}{Sequential} interface or by subclassing the \mintinline{python}{Module} class, as in Box \ref{code:fully_connected_layer}.
\end{enumerate}

I suggest you now try to replicate one of the many quick guides available on the documentation of PyTorch.\footnote{\url{https://pytorch.org/tutorials/beginner/basics/quickstart_tutorial.html}} Everything should be reasonably clear, apart from the gradient computation mechanism, introduced in the next chapter. This is also a good time to investigate Hugging Face Datasets, which combines a vast repository of datasets with a framework-agnostic interface to process and cache them backed by Apache Arrow.\footnote{\url{https://huggingface.co/docs/datasets/en/quickstart}}

JAX does not provide high-level utilities. For data loading you can use any existing tool, including PyTorch's data loaders and Hugging Face Datasets. For building models, the easiest way is to rely on an external library. Because JAX is fully functional, object-oriented abstractions like Box \ref{code:fully_connected_layer} are not possible. My personal suggestion is Equinox \cite{kidger2021equinox}, which provides a class-like experience by combining the basic data structure of JAX (the \mintinline{python}{pytree}) with callable nodes.

%% file: 6_automatic_differentiation.tex
\chapter{Automatic differentiation}
\label{chap:automatic_differentiation}

\begin{supportbox}{About this chapter}
The previous chapter highlighted the need for an efficient, automatic procedure to compute gradients of any possible sequence of operations. In this chapter we describe such a method, called \textbf{back-propagation} in the neural network’s literature or \textbf{reverse-mode automatic differentiation} in the computer science one. Its analysis has several insights, ranging from the model's choice to the memory requirements for optimizing it.
\end{supportbox}

\section{Problem setup}

We consider the problem of efficiently computing gradients of generic \textit{computational graphs}, such as those induced by optimizing a scalar loss function on a fully-connected model, a task called \textbf{automatic differentiation} (AD) \cite{baydin2018automatic}. You can think of a computational graph as the set of atomic operations (which we call \textbf{primitives}) obtained by running the program itself. We will consider sequential graphs for brevity, but everything can be easily extended to acyclic and even more generic computational graphs \cite{griewank2008evaluating,blondel2024elements} .

The problem may seem trivial, since the chain rule of Jacobians (Section \ref{sec:gradients_and_jacobians}, \eqref{eq:jacobian_chain_rule}) tells us that the gradient of function composition is simply the matrix product of the corresponding Jacobian matrices. However, efficiently implementing this process is the key challenge of this chapter, and the resulting algorithm (\textbf{reverse-mode AD} or \textbf{backpropagation}) is a cornerstone of neural networks and differentiable programming in general \cite{griewank2008evaluating,blondel2024elements}. Understanding it is also key to understanding the design (and the differences) of most frameworks for implementing and training such programs (such as TensorFlow or PyTorch or JAX). A brief history of the algorithm can be found in \cite{griewank2012invented}.

To setup the problem, we assume we have at our disposal a set of \textbf{primitives}:
$$
\mathbf{y} =f_i(\mathbf{x}, \mathbf{w}_i)
$$
Each primitive represents an operation on an input vector $\mathbf{x} \sim ({c_i})$, parameterized by the vector $\mathbf{w}_i \sim (p_i)$ (e.g., the weights of a linear projection), and giving as output another vector $\mathbf{y} \sim (c^\prime_i)$.

There is a lot of flexibility in our definition of primitives, which can represent basic linear algebra operations (e.g., matrix multiplication), layers in the sense of Chapter \ref{chap:fully_connected_models} (e.g., a fully-connected layer with an activation function), or even larger blocks or models. This recursive composability is a key property of programming and extends to our case.

We only assume that for each primitive we know how to compute the partial derivatives with respect to the two input arguments, which we call the \textbf{input Jacobian} and the \textbf{weight Jacobian} of the operation:
\begin{eqnarray*}
\textbf{Input Jacobian: } & \partial_{\mathbf{x}}\left[f(\mathbf{x},\mathbf{w})\right] \sim (c^\prime,c) \\ 
\textbf{Weight Jacobian: } & \partial_{\mathbf{w}}\left[f(\mathbf{x},\mathbf{w})\right] \sim (c^\prime,p) 
\end{eqnarray*}

These are reasonable assumptions since we restrict our analysis to differentiable models. Continuous primitives with one or more points of non-differentiability, such as the ReLU, can be made to fit into this framework with the use of \textbf{subgradients} (Section \ref{subsec:subdifferentiability}). Non differentiable operations such as sampling or thresholding can also be included by finding a relaxation of their gradient or an equivalent estimator \cite{niculae2023discrete}. We cover the latter case in the next volume.
\subsection*{On our notation and higher-order Jacobians}

\addteacup We only consider vector-valued quantities for readability, as all resulting gradients are matrices. In practice, existing primitives may have inputs, weights, or outputs of higher rank. For example, consider a basic fully-connected layer on a mini-batched input:
$$
f(\mathbf{X},\mathbf{W})=\mathbf{X}\mathbf{W}+\mathbf{b}
$$
In this case, the input $\mathbf{X}$ has shape $(n,c)$, the weights have shape $(c,c’)$ and $(c^\prime)$ (with $c^\prime$ a hyper-parameter), and the output has shape $(n, c^\prime)$. Hence, the input Jacobian has shape $(n,c^\prime, n, c)$, and the weight Jacobian has shape $(n,c^\prime, c, c^\prime)$, both having rank $4$. 

In our notation, we can consider the equivalent flattened vectors $\mathbf{x} = \text{vect}(\mathbf{X})$ and $\mathbf{w} = \left[\text{vect}(\mathbf{W}); \mathbf{b}\right]$, and our resulting “flattened” Jacobians have shape $(nc^\prime, nc)$ and $(nc^\prime, cc^\prime)$ respectively. This is crucial in the following, since every time we refer to “\textit{the input size $c$}” we are referring to “\textit{the product of all input shapes}”, including eventual mini-batching dimensions. This also shows that, while we may know how to compute the Jacobians, we may not wish to fully \textit{materialize} them in memory due to their large dimensionality.

As a final note, our notation aligns with the way these primitives are implemented in a functional library, such as JAX. In an object-oriented framework (e.g., TensorFlow, PyTorch), we saw
that layers are implemented as objects (see Box \ref{code:fully_connected_layer} in the previous chapter), with the parameters being a property of the object, and the function call being replaced by an object’s method. This style simplifies certain practices, such as deferred initialization of all parameters until the input shapes are known (\textbf{lazy initialization}), but it adds a small layer of abstraction to consider to translate our notation into workable code. As we will see, these differences are reflected in turn in the way AD is implemented in the two frameworks.
\subsection{Problem statement}
With all these details out of the way, we are ready to state the AD task. Consider a sequence of $l$ primitive calls, followed by a final summation:

\clearpage

\begin{eqnarray*}
\mathbf{h}_1 & = & f_1(\mathbf{x}, \mathbf{w}_1) 
\\ \mathbf{h}_2 & = & f_2(\mathbf{h}_1, \mathbf{w}_2) \\ 
& \vdots &  \\ 
\mathbf{h}_{l} & = & f_{l}(\mathbf{h}_{l-1}, \mathbf{w}_{l}) 
\\ y & = & \sum \mathbf{h}_{l} 
\end{eqnarray*}

This is called an \textbf{evaluation trace} of the program. Roughly, the first $l-1$ operations can represent several layers of a differentiable model, operation $l$ can be a per-input loss (e.g., cross-entropy), and the final operation sums the losses of the mini-batch. Hence, the output of our program is always a scalar, since we require it for numerical optimization. We abbreviate the previous program as $F(\mathbf{x})$. 

\begin{definition}[Automatic differentiation] \addbottle $\,$

Given a program $F(\mathbf{x})$ composed of a sequence of differentiable primitives, \textbf{automatic differentiation} (AD) refers to the task of \textit{simultaneously} and \textit{efficiently} computing all weight Jacobians of the program given knowledge of the computational graph and all individuals input and weight Jacobians:

$$
\text{AD}(F(\mathbf{x}))=\left\{\partial_{\mathbf{w}_i} y\right\}_{i=1}^{l}
$$

\end{definition}

As we will see, there are two major classes of AD algorithms, called \textbf{forward-mode} and \textbf{backward-mode}, corresponding to a different ordering in the composition of the individual operations. We will also see that the backward-mode (called \textbf{back-propagation} in the neural networks’ literature) is significantly more efficient in our context. While we focus on a simplified scenario, it is relatively easy to extend our derivation to acyclic graphs of primitives (as already mentioned), and also to situations where parameters are shared across layers (\textbf{weight sharing}). We will see an example of weight sharing in Chapter \ref{chap:rnns}.

\subsection{Numerical and symbolic differentiation}

\addteacup Before moving on to forward-mode AD, we comment on the difference between AD and other classes of algorithms for differentiating functions. First, we could directly apply the definition of gradients (Section \ref{sec:gradients_and_jacobians}) to obtain a suitable numerical approximation of the gradient. This process is called \textbf{numerical differentiation}. However, each scalar value to be differentiated requires $2$ function calls in a naive implementation, making this approach unfeasible except for numerical checks over the implementation.

Second, consider this simple function:
$$
f(x)=a\sin(x)+bx\sin(x)
$$
We can ask a symbolic engine to pre-compute the full, symbolic equation of the derivative. This is called \textbf{symbolic differentiation} and shown in Python in Box \ref{code:symbolic_differentiation}.

In a realistic implementation, the intermediate value $h=\sin(x)$ would be computed only once and stored in an intermediate variable, which can also be reused for the corresponding computation in the gradient trace (and a similar reasoning goes for the $\cos(x)$ term in the derivative). This is less trivial than it appears: finding an optimal implementation for the Jacobian which avoids any unnecessary computation is an NP-complete task (\textbf{optimal Jacobian accumulation}). However, we will see that we can exploit the structure of our program to devise a suitably efficient implementation of AD that is significantly better than a symbolic approach like the above (and it is, in fact, equivalent to a symbolic approach allowing for the presence of subsequences \cite{laue2019equivalence}).

\begin{mypy}{Symbolic differentiation in Python using SymPy.}{code:symbolic_differentiation}
import sympy as sp
x, a, b = sp.symbols('x a b')
y = a*sp.sin(x) + b*x*sp.sin(x)
sp.diff(y, x) 
# [Out]: acos(x)+bxcos(x)+bsin(x)
\end{mypy}

\section{Forward-mode differentiation}

We begin by recalling the chain rule of Jacobians. Consider a combination of two primitive functions:

$$
\mathbf{h}=f_1(\mathbf{x)} \;,\; \mathbf{y}=f_2(\mathbf{h})
$$

In terms of their gradients, we have:

$$
\partial_{\mathbf{x}}\,\mathbf{y} = {\color{drawgreen}\partial_{\mathbf{h}} \,\mathbf{y}} \,\cdot\,{\color{drawred}\partial_{\mathbf{x}}\,\mathbf{h}}
$$

If $\mathbf{x}$, $\mathbf{h}$, and $\mathbf{y}$ have dimensions $a$, $b$, and $c$ respectively, the previous Jacobian requires the multiplication of a $c \times b$ matrix (in green) with a $b \times a$ one (in red). We can interpret the rule as follows: if we have already computed $f_1$ and its Jacobian (red term), once we apply $f_2$ we can “update” the gradient by multiplying with the corresponding Jacobian (green term).

We can immediately apply this insight to obtain a working algorithm called \textbf{forward-mode automatic differentiation} (F-AD). The idea is that every time we apply a primitive function, we initialize its corresponding weight Jacobian (called \textbf{tangent} in this context), while simultaneously updating all previous tangent matrices. Let us see a simple worked-out example to illustrate the main algorithm.

Consider the first instruction, $\mathbf{h}_1 = f_1(\mathbf{x}, \mathbf{w}_1)$, in our program. Because nothing has been stored up to now, we initialize the tangent matrix for $\mathbf{w}_1$ as its weight Jacobian:

$$
\widehat{\mathbf{W}}_1 = \partial_{\mathbf{w}_1} \, \mathbf{h}_1
$$

We now proceed to the second instruction, $\mathbf{h}_2 = f_2(\mathbf{h}_1, \mathbf{w}_2)$. We update the previous tangent matrix while simultaneously initializing the second one:

\begin{gather*}
\eqnmarkbox[drawgreen]{node2}{\widehat{\mathbf{W}}_1} \leftarrow \eqnmarkbox[drawred]{node}{\left[\partial_{\mathbf{h}_1} \mathbf{h}_2\right]} \widehat{\mathbf{W}}_1 \\
\widehat{\mathbf{W}}_2 = \partial_{\mathbf{w}_2}\,\mathbf{h}_2
\end{gather*}
\annotate[yshift=1em]{above,right}{node}{Input Jacobian of $f_2$}
\annotate[yshift=-2.5em]{below,right}{node2}{Updated tangent matrix for $\mathbf{w}_1$}

The update requires the input Jacobian of the primitive, while the second term requires the weight Jacobian of the primitive. Abstracting away, consider the generic $i$-th primitive given by $\mathbf{h}_i = f_i(\mathbf{h}_{i-1}, \mathbf{w}_i)$. We initialize the tangent matrix for $\mathbf{w}_i$ while simultaneously updating \textit{all} previous matrices:

\begin{gather*}
\widehat{\mathbf{W}}_j \leftarrow \left[\eqnmarkbox[drawred]{node}{\partial_{\mathbf{h}_{i-1}} \mathbf{h}_i}\right] \widehat{\mathbf{W}}_j \;\; \forall j <i \\
\widehat{\mathbf{W}}_i = \eqnmarkbox[drawgreen]{node2}{\partial_{\mathbf{w}_i}}\,\mathbf{h}_i
\end{gather*}
\annotate[yshift=1em]{above,right}{node}{Input Jacobian of $f_i$}
\annotate[yshift=-1em]{below,right}{node2}{Weight Jacobian of $f_i$}

\vspace{1em}
There are $i-1$ updates in the first row (one for each tangent matrix we have already stored in memory), with the red term -- the input Jacobian of the $i$-th operation -- being shared for all previous tangents. The last operation in the program is a sum, and the corresponding gradient gives us the output of the algorithm:\footnote{To be fully consistent with notation, the output of \eqref{eq:last_step_fad} is a row vector, while we defined the gradient as a column vector. We will ignore this subtle point for simplicity until it is time to define vector-Jacobian products later on the in the chapter.}

\begin{equation}
\nabla_{\mathbf{w}_i}y=  \mathbf{1}^\top\widehat{\mathbf{W}}_i \;\; \forall i
\label{eq:last_step_fad}
\end{equation}

Done! Let us analyze the algorithm in more detail. First, all the operations we listed can be easily \textit{interleaved} with the original program, meaning that the space complexity will be roughly proportional to the space complexity of the program we are differentiating.

On the negative side, the core operation of the algorithm (the update of $\widehat{\mathbf{W}}_i$) requires a multiplication of two matrices, generically shaped $(c^\prime_i, c_i)$ and $(c_i, p_j)$, where $c_i, c^\prime_i$ are input/output shapes, and $p_j$ is the shape of $\mathbf{w}_j$. This is an extremely expensive operation: for example, assume that inputs and outputs are both shaped $(n,d)$, where $n$ is the mini-batch dimension and $d$ represents the input/output features. Then, the matrix multiplication will have complexity $\mathcal{O}(n^2d^2p_j)$, which is quadratic in both mini-batch size and feature dimensionality. This can easily become unfeasible, especially for high-dimensional inputs such as images.

We can obtain a better trade-off by noting that the last operation of the algorithm is a simpler matrix-vector product, which is a consequence of having a scalar output. This is explored in more detail in the next section.

\section{Reverse-mode differentiation}
\label{sec:reverse_mode_automatic_differentiation}

\addclock To proceed, we unroll the computation of a single gradient term corresponding to the $i$-th weight matrix:
\begin{equation}
\nabla_{\mathbf{w}_i}\,y = \mathbf{1}^\top {\color{drawred}\left[ \partial_{\mathbf{h}_{l-1}}\,\mathbf{h}_{l}\right] \cdots \left[ \partial_{\mathbf{h}_{i}}\,\mathbf{h}_{i+1}\right]} {\color{drawgreen}\left[ \partial_{\mathbf{w}_{i}}\,\mathbf{h}_{i}\right]}
\label{eq:backward_pass}
\end{equation}
Remember that, notation apart, \eqref{eq:backward_pass} is just a potentially long series of matrix multiplications, involving a constant term (a vector $\mathbf{1}$ of ones), a series of input Jacobians (the red term) and a weight Jacobian of the corresponding weight matrix (the green term). Let us define a shorthand for the red term:
\begin{equation}
    \widetilde{\mathbf{h}}_i = \mathbf{1}^\top \prod_{j=i+1}^{l} \partial_{\mathbf{h}_{j-1}}\,\mathbf{h}_{j}
    \label{eq:recursive_h}
\end{equation}
Because matrix multiplication is associative, we can perform the computations in \eqref{eq:backward_pass} in any order. In F-AD, we proceeded from the right to the left, since it corresponds to the ordering in which the primitive functions were executed. However, we can do better by noting two interesting aspects:

\begin{enumerate}
\item The leftmost term in \eqref{eq:backward_pass} is a product between a vector and a matrix (which is a consequence of having a scalar term in output), which is computationally better than a product between two matrices. Its output is also another vector.
\item The term in \eqref{eq:recursive_h} (the product of all input Jacobians from layer $i$ to layer $l$) can be computed recursively starting from the \textit{last} term and iteratively multiplying by the input Jacobians in the reverse order.
\end{enumerate}

We can put together these observations to develop a second approach to automatic differentiation, that we call \textbf{reverse-mode automatic differentiation} (R-AD), which is outlined next.

\begin{enumerate}
\item Differently from F-AD, we start by executing the \textit{entire} program to be differentiated, storing all intermediate outputs.
\item We inizialize a vector $\widetilde{\mathbf{h}} =\mathbf{1}^\top$, which corresponds to the leftmost term in \eqref{eq:backward_pass}.
\item Moving in reverse order, i.e., for an index $i$ ranging in $l, l-1, l-2,\ldots, 1$, we first compute the gradient with respect to the $i$-th weight matrix as:
$$
\partial_{\mathbf{w}_i}\, y  = \widetilde{\mathbf{h}} \left[\partial_{\mathbf{w}_i} \mathbf{h}_{i}\right]
$$
which is the $i$-th gradient we need. Next, we update our “back-propagated” input Jacobian as:
$$
\widetilde{\mathbf{h}} \leftarrow \widetilde{\mathbf{h}} \left[\partial_{\mathbf{h}_{i-1}}\mathbf{h}_{i}\right]
$$
\end{enumerate}

Steps (1)-(3) describe a program which is roughly symmetrical to the original program, that we call the \textbf{dual} or \textbf{reverse} program. The terms $\widetilde{\mathbf{h}}$ are called the \textbf{adjoints} and they store (sequentially) all the gradients of the output with respect to the variables $\mathbf{h}_1, \mathbf{h}_2, \ldots, \mathbf{h}_{l}$ in our program.\footnote{Compare this with F-AD, where the tangents represented instead the gradients of the $\mathbf{h}_i$ variables with respect to the weights.}

In the terminology of neural networks, we sometimes say that the original (\textbf{primal}) program is a \textbf{forward pass} (not to be confused with forward-mode), while the reverse program is a \textbf{backward pass}. Differently from F-AD, in R-AD the full primal program must be executed before the reverse program can be run, and we need specialized mechanisms to store all intermediate outputs to “unroll” the computational graph. Different frameworks implement this differently, as outlined next.

\begin{figure}
    \centering
    \begin{subfigure}[b]{0.23\textwidth}
    \includegraphics[width=1.0\textwidth]{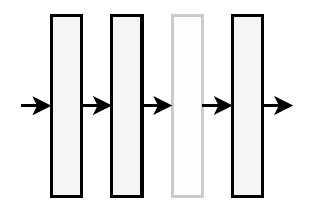}
    \caption{}
    \end{subfigure}
    \hfill
    \begin{subfigure}[b]{0.23\textwidth}
    \includegraphics[width=1.0\textwidth]{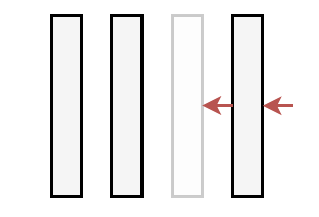}
    \caption{}
    \end{subfigure}
    \hfill
    \begin{subfigure}[b]{0.23\textwidth}
    \includegraphics[width=1.0\textwidth]{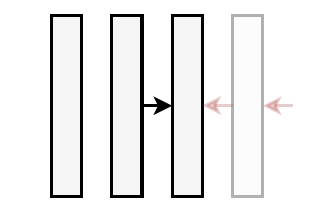}
    \caption{}
    \end{subfigure}
    \hfill
    \begin{subfigure}[b]{0.23\textwidth}
    \includegraphics[width=1.0\textwidth]{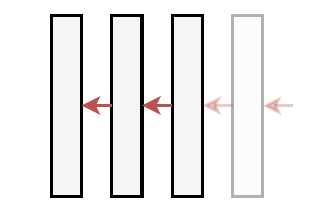}
    \caption{}
    \end{subfigure}
    \hfill
    \caption{An example of \textbf{gradient checkpointing}. (a) We execute a forward pass, but we only store the outputs of the first, second, and fourth blocks (\textbf{checkpoints}). (b) The backward pass (red arrows) stops at the third block, whose activations are not available. (c) We run a second forward pass starting from the closest checkpoint to materialize again the activations. (d) We complete the forward pass. Compared to a standard backward pass, this requires 1.25x more computations. In general, the less checkpoints are stored, the higher the computational cost of the backward pass.}
    \label{fig:gradient_checkpointing}
\end{figure}

Computationally, R-AD is significantly more efficient than F-AD. In particular, both operations in step (3) of R-AD are vector-matrix products scaling only linearly in all shape quantities. The tradeoff is that executing R-AD requires a large amount of memory, since all intermediate values of the primal program must be stored on disk with a suitable strategy. Specific techniques, such as \textbf{gradient checkpointing}, can be used to improve on this tradeoff by increasing computations and partially reducing the memory requirements. This is done by only storing a few intermediate outputs (called \textbf{checkpoints}) while recomputing the remaining values during the backward pass. See Figure \ref{fig:gradient_checkpointing} for a visualization.

\section{Practical considerations}

\subsection{Vector-Jacobian products}

Looking at step (3) in the R-AD algorithm, we can make an interesting observation: the only operation we need is a product between a row vector $\mathbf{v}$ and a Jacobian of $f$ (either the input or the weight Jacobian). We call these two operations the \textbf{vector-Jacobian products} (VJPs) of $f$.\footnote{By contrast, F-AD can be formulated entirely in terms of the transpose of the VJP, called a \textbf{Jacobian-vector product} (JVP). For a one-dimensional output, the JVP is the directional derivative \eqref{eq:directional_derivative} from Section \ref{sec:gradients_and_jacobians}. Always by analogy, the VJP represents the application of a linear map connected to infinitesimal variations of the \textit{output} of the function, see \cite{blondel2024elements}.} In the next definition we restore dimensional consistency by adding a transpose to the vector.

\begin{definition}[Vector-Jacobian product (VJP)] \addbottle $\,$

    Given a function $\mathbf{y}=f(\mathbf{x})$, with $\mathbf{x} \sim (c)$ and $\mathbf{y} \sim (c^\prime)$, its VJP is another function defined as:
    
    \begin{equation}
    \textnormal{vjp}_f(\mathbf{v})=\mathbf{v}^\top \partial f(\mathbf{x})
    \end{equation}
    
    where $\mathbf{v}\sim (c^\prime)$. If $f$ has multiple parameters $f(\mathbf{x}_1, \ldots, \mathbf{x}_n)$, we can define $n$ individual VJPs denoted as $\textnormal{vjp}_{f,\mathbf{x}_1}(\mathbf{v})$, ..., $\textnormal{vjp}_{f, \mathbf{x}_n}(\mathbf{v})$.
    
\end{definition}

In particular, in our case we can define two types of VJPs, corresponding to the partial derivative of the primitive with respect to the input and the weight arguments:
\begin{gather}
\text{vjp}_{f,\mathbf{x}}(\mathbf{v})=\mathbf{v}^\top\partial_\mathbf{x}f(\mathbf{x},\mathbf{w}) \label{eq:input_jacobian}\\ \text{vjp}_{f, \mathbf{w}}(\mathbf{v})=\mathbf{v}^\top\partial_\mathbf{w}f(\mathbf{x},\mathbf{w})\label{eq:weight_jacobian}
\end{gather}

We can now rewrite the two operations in step (3) of the R-AD algorithm as two VJP calls of the primitive function with the adjoint values (ignoring the $i$ indices for readability), corresponding to the adjoint times the weight VJP, and the adjoint times the input VJP:
\begin{gather}
\partial_{\mathbf{w}} \, y = \text{vjp}_{f,\mathbf{w}}\left(\widetilde{\mathbf{h}}\right) \label{eq:r_ad_1}\\ \widetilde{\mathbf{h}}  \gets \text{vjp}_{f, \mathbf{h}}\left(\widetilde{\mathbf{h}}\right) \label{eq:r_ad_2}
\end{gather}

\begin{figure}[t]
    \centering
    \includegraphics[width=0.7\textwidth]{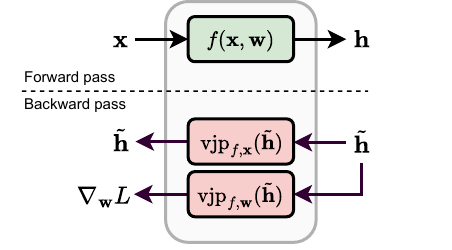}
    \caption{For performing R-AD, primitives must be augmented with two VJP operations to be able to perform a backward pass, corresponding to the input VJP \eqref{eq:input_jacobian} and the weight VJP \eqref{eq:weight_jacobian}. One call for each is sufficient to perform the backward pass through the primitive, corresponding to \eqref{eq:r_ad_1}-\eqref{eq:r_ad_2}.}
    \label{fig:backward_pass}
\end{figure}

Hence, we can implement an entire automatic differentiation system by first choosing a set of primitives operations, and then augmenting them with the corresponding VJPs, without having to materialize the Jacobians in memory at any point. This is shown schematically in Figure \ref{fig:backward_pass}. 

In fact, we can recover the Jacobians' computation by repeatedly calling the VJPs with the basis vectors $\mathbf{e}_1, \ldots, \mathbf{e}_n$, to generate them one row at a time, e.g., for the input Jacobian we have:
$$
\partial_{\mathbf{x}}f(\mathbf{x},\mathbf{w})=\begin{bmatrix} \text{vjp}_{f,\mathbf{x}}(\mathbf{e}_1) \\ \text{vjp}_{f,\mathbf{x}}(\mathbf{e}_2) \\ \vdots\\ \text{vjp}_{f,\mathbf{x}}(
\mathbf{e}_n) \end{bmatrix}
$$
To understand why this reformulation can be convenient, let us look at the VJPs of a fully-connected layer, which is composed of linear projections and (elementwise) non-linearities. First, consider a simple linear projection with no bias:
$$
f(\mathbf{x}, \mathbf{W})=\mathbf{W}\mathbf{x}
$$
The input Jacobian here is simply $\mathbf{W}$, but the weight Jacobian is a rank-3 tensor (Section \ref{sec:gradients_and_jacobians}). By comparison, the input VJP has no special structure:
\begin{equation}
\text{vjp}_{f, \mathbf{x}}(\mathbf{v})=\mathbf{v}^\top\mathbf{W}^\top = \left[\mathbf{W}\mathbf{v}\right]^\top
\label{eq:vjp_x_matrix_multiplication}
\end{equation}
The weight VJP, instead, turns out to be a simple outer product, which avoids rank-3 tensors completely:
\begin{equation}
\text{vjp}_{f,\mathbf{w}}(\mathbf{v}) = \mathbf{v}\mathbf{x}^\top
\label{eq:vjp_w_matrix_multiplication}
\end{equation}
\begin{supportbox}{Working out the VJP}
To compute \eqref{eq:vjp_w_matrix_multiplication}, we can write $y=\mathbf{v}^\top \mathbf{W}\mathbf{x} =\sum_i\sum_j W_{ij}v_ix_j$, from which we immediately get $\frac{\partial y}{\partial W_{ij}} = v_ix_j$, which is the elementwise definition of the outer product.
\end{supportbox}
Hence, every time we apply a linear projection in the forward pass, we modify the back-propagated gradients by the transpose of its weights, and we perform an outer product to compute the gradient of $\mathbf{W}$. 

Consider now an element-wise activation function with no trainable parameters, e.g., the ReLU:
$$
f(\mathbf{x},\left\{\right\})=\phi(\mathbf{x})
$$
Because we have no trainable parameters, we need only consider the input VJP. The gradient is a diagonal matrix having as elements the derivatives of $\phi$:
$$
\idx{\partial_{\mathbf{x}} \phi(\mathbf{x})}{ii}=\phi^\prime(x_i)
$$
The input VJP is a multiplication of a diagonal matrix by a vector, which is equivalent to an Hadamard product (i.e., a scaling operation):
\begin{equation}
\text{vjp}_{\mathbf{x}}(f,\mathbf{v})=\mathbf{v}\odot \phi^\prime(\mathbf{x})
\label{eq:backward_pass_activation_function}
\end{equation}
Interestingly, also in this case we can compute the VJP without having to materialize the full diagonal matrix.

\subsection{Implementing a R-AD system}
\label{subsec:implementing_rad}
There are many ways to implement the R-AD system, ranging form Wengert lists (as done in TensorFlow) to source-to-source code transformations \cite{griewank2008evaluating}. Here, we discuss briefly some common implementations in existing frameworks. 

First, describing primitives as functions with two arguments $f(\mathbf{x}, \mathbf{w})$ aligns with functional frameworks such as JAX, where everything is a function. Consider a function $f(\mathbf{x})$ with a $c$-dimensional input and a $c^\prime$-dimensional output. From this point of view, a VJP can be implemented as a higher-order function with signature:

\begin{mypy}{Gradient computation as a higher-order function. The \mintinline{python}{torch.func} interface replicates the JAX API. In practice, the function can be \textit{traced} (e.g., with {\footnotesize\mintinline{python}{torch.compile}}) to generate an optimized computational graph.}{code:functional_grad}
# Original function (sum-of-squares)
def f(x: Float[Array, "c"]):
  return (x**2).sum()

grad_f = func.grad(f)
print(grad_f(torch.randn(10)).shape) 
# [Out]: torch.Size([10])
\end{mypy}

\begin{equation}
(\mathbb{R}^c \rightarrow \mathbb{R}^{c^\prime}) \rightarrow \mathbb{R}^c \rightarrow (\mathbb{R}^{c^\prime} \rightarrow \mathbb{R}^c)
\end{equation}
i.e., given a function $f$ and an input $\mathbf{x}^\prime$, a VJP returns another function that can be applied to a $c^\prime$-dimensional vector $\mathbf{v}$ to return $\mathbf{v}^\top \partial f(\mathbf{x}^\prime)$. Similarly, the gradient for a one-dimensional function can be implemented as another higher-order function with signature:
\begin{equation}
(\mathbb{R}^c \rightarrow \mathbb{R}) \rightarrow (\mathbb{R}^c \rightarrow \mathbb{R}^c)
\end{equation}
taking as input the function $f(\mathbf{x})$ and returning another function that computes $\nabla f(\mathbf{x})$. In JAX, these ideas are implemented in the functions {\footnotesize\mintinline{python}{jax.grad}} and {\footnotesize\mintinline{python}{jax.jvp}} respectively, which is also replicated in PyTorch in the {\footnotesize\mintinline{python}{torch.func}} module - see Box \ref{code:functional_grad} for an example.\footnote{Many operations, such as computing an Hessian, can be achieved by smartly composing JVPs and VJPs based on their signatures: \url{https://jax.readthedocs.io/en/latest/notebooks/autodiff_cookbook.html}.}

As we mentioned, in practice our models are implemented as compositions of objects whose parameters are encapsulated as properties (Box \ref{code:fully_connected_layer}). One possibility is to ``purify'' the object to turn it into a pure function, e.g.:\footnote{\url{https://sjmielke.com/jax-purify.htm}}

{\footnotesize
\begin{minted}{python}
# Extract the parameters
params = dict(model.named_parameters())
# Functional call over the 
# model's forward function
y = torch.func.functional_call(
                model, params, x
                )
\end{minted}
}

\begin{figure}
    \centering
    \includegraphics[width=\textwidth]{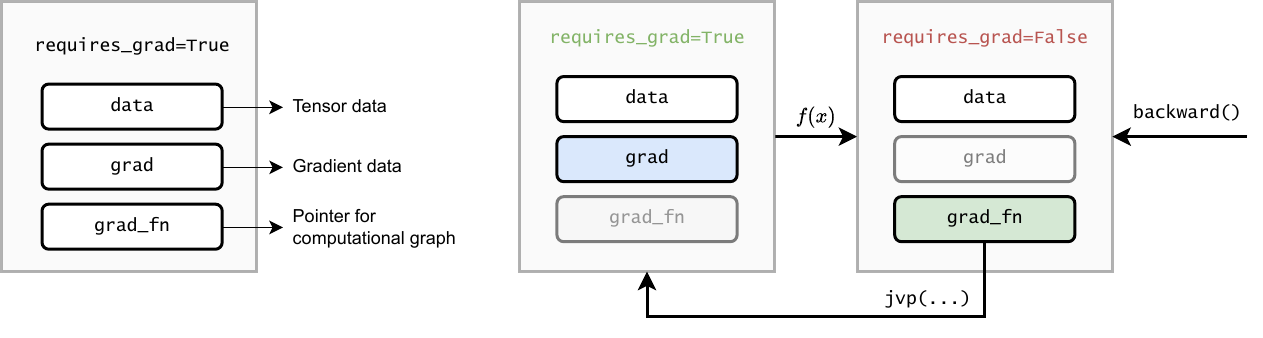}
    \caption{Left: in PyTorch, a tensor is augmented with information about its gradient (empty at initialization), and about the operation that created it. Right: during a backward pass, the {\footnotesize\mintinline{python}{grad_fn}} property is used to traverse the computational graph in reverse, and gradients are stored \textit{inside} the tensor's {\footnotesize\mintinline{python}{grad}} property whenever {\footnotesize\mintinline{python}{requires_grad}} is explicitly set to {\footnotesize\mintinline{python}{True}} (to avoid consumming unnecessary memory).}
    \label{fig:pytorch_wrapper}
\end{figure}

More in general, frameworks like PyTorch are augmented with techniques to handle this scenario directly, without introducing intermediate operations. In PyTorch, for example, tensors' objects are augmented with information about the operation that generated them (Figure \ref{fig:pytorch_wrapper}, left). Whenever a {\footnotesize\mintinline{python}{backward()}} call is requested on a scalar value, these properties are used to traverse the computational graph in reverse, storing the corresponding gradients \textit{inside} the tensors that requires them (Figure \ref{fig:pytorch_wrapper}, right). 

This is just a high-level overview of how these systems are implemented in practice, and we are leaving behind many details, for which we refer to the official documentations.\footnote{I definitely suggest trying to implement an R-AD system from scratch: many didactical implementations can be found online, such as \url{https://github.com/karpathy/micrograd}.}

\subsection{Choosing an activation function}

Coincidentally, we can now motivate why ReLU is a good choice as activation function. A close look at \eqref{eq:backward_pass_activation_function} tells us that every time we add an activation function in our model, the adjoints in the backward pass are scaled by a factor of $\phi^\prime(\mathbf{x})$. For models with many layers, this can give rise to two pathological behaviors:

\begin{enumerate}
\item If $\phi^\prime(\cdot) < 1$ everywhere, there is the risk of the gradient being shrank to 0 exponentially fast in the number of layers. This is called the \textbf{vanishing gradient} problem.
\item Conversely, if $\phi^\prime(\cdot) > 1$ everywhere, the opposite problem appears, with the gradients exponentially converging to infinity in the number of layers. This is called the \textbf{exploding gradient} problem.
\end{enumerate}

These are serious problems in practice, because libraries represent floating point numbers with limited precision (typically 32 bits or lower), meaning that underflows or overflows can manifest quickly when increasing the number of layers.

\begin{supportbox}{Linear non-linear models}
Surprisingly, a stack of linear layers implemented in floating point precision is not fully linear because of small discontinuities at machine precision! This is generally not an issue, but it can be exploited to train fully-linear deep neural networks.\footnote{\url{https://openai.com/research/nonlinear-computation-in-deep-linear-networks}}
\end{supportbox}

\begin{figure}
    \centering
    \begin{subfigure}[b]{0.48\textwidth}
    \includegraphics[width=1.0\textwidth]{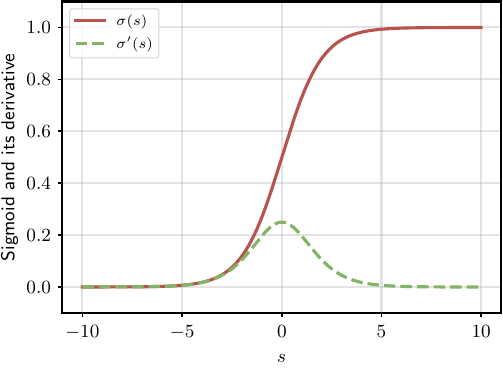}
    \caption{Sigmoid}
    \label{fig:sigmoid_and_derivative}
    \end{subfigure}
    \hfill
    \begin{subfigure}[b]{0.48\textwidth}
    \includegraphics[width=1.0\textwidth]{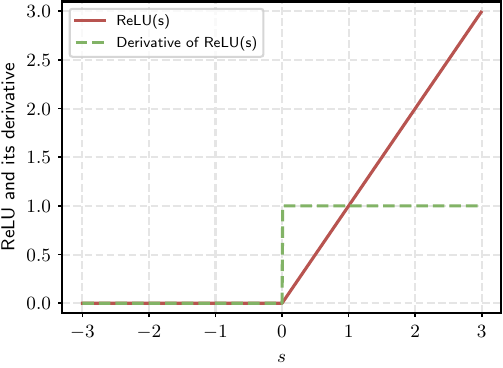}
    \caption{ReLU}
    \label{fig:relu_and_derivative}
    \end{subfigure}
    \caption{(a) Plot of the sigmoid function ({\color{drawred}red}) and its derivative ({\color{drawgreen}green}). (b) Plot of ReLU ({\color{drawred}red}) and its derivative ({\color{drawgreen}green}).}
\end{figure}

As an example of how vanishing gradients can appear, consider the sigmoid function $\sigma(s)$. We already mentioned that this was a common AF in the past, due to it being a soft approximation to the step function. We also know that $\sigma^\prime(s)=\sigma(s)(1-\sigma(s))$. Combined with the fact that $\sigma(s) \in \left[0,1\right]$, we obtain that:
$$
\sigma^\prime(s)\in\left[0,0.25\right]
$$
Hence, the sigmoid is a prime candidate for vanishing gradient issues: see Figure \ref{fig:sigmoid_and_derivative}.

Designing an AF that never exhibits vanishing or exploding gradients is non trivial, since the only function having $\phi^\prime(s)=1$ everywhere is the identity function. We then need a function which is “linear enough” to avoid gradient issues, but “non-linear” enough to separate the linear layers. The ReLU ends up being a good candidate since:
$$
\partial_s \text{ReLU}(s)=\begin{cases} 0 & s < 0 \\ 1 & s >0 \end{cases}
$$
The gradient is either zeroed-out, inducing sparsity in the computation, or multiplied by $1$, avoiding scaling issues - this is shown in Figure \ref{fig:relu_and_derivative}.

As a side note, the ReLU’s gradient is identical irrespective of whether we replace the input to the ReLU layer with its output (since we are only masking the negative values while keeping the positive values untouched). Hence, another benefit of using ReLU as activation function is that we can save a small bit of memory when performing R-AD, by overwriting the layer’s input in the forward pass without impacting the correctness of the AD procedure: this is done in PyTorch, for example, by setting the {\footnotesize\verb+in_place+} parameter.\footnote{\url{https://pytorch.org/docs/stable/generated/torch.nn.ReLU.html}}

\subsection{Subdifferentiability and AD}
\label{subsec:subdifferentiability}

\addteacup There is a small detail we avoided discussing until now: the ReLU is non-differentiable in $0$, making the overall network non-smooth. What happens in this case? The “pragmatic” answer is that, by minimizing with stochastic gradient descent from a random (non-zero) initialization, the probability of ending up exactly in $s=0$ is practically null, while the gradient is defined in $\text{ReLU}(\varepsilon)$ for any $\lvert\varepsilon\rvert>0$.

For a more technical answer, we can introduce the concept of \textbf{subgradient} of a function.

\begin{definition}[Subgradient] $\,$

Given a convex function $f(x)$, a subgradient in $x$ is a point $z$ such that, for all $y$:

$$
f(y) \ge f(x)+z(y-x)
$$

\end{definition}

Note the similarity with the definition of convexity: a subgradient is the slope of a line “tangent” to $f(x)$, such that the entire $f$ is lower bounded by it. If $f$ is differentiable in $x$, then only one such line exists, which is the derivative of $f$ in $x$. In a non-smooth point, multiple subgradients exists, and they form a set called the \textbf{subdifferential} of $f$ in $x$:
$$
\partial_x f(x)=\left\{z \,\vert\, z \text{ is a subgradient of } f(x)\right\}
$$

With this definition in hand, we can complete our analysis of the gradient of ReLU by replacing the gradient with its subdifferential in $0$:
$$
\partial_s \text{ReLU}(s)=\begin{cases} \left\{0\right\} & s < 0 \\ \left\{1\right\} & s >0 \\ \left[0,1\right] & s=0 \end{cases}
$$
Hence, any value in $[0,1]$ is a valid subgradient in $0$, with most implementations in practice favoring $\text{ReLU}^\prime(0)=0$. Selecting subgradients at every step of an iterative descent procedure is called \textbf{subgradient descent}. 

In fact, the situation is even more tricky, because the subgradient need not be defined for non-convex functions. In that case, one can resort to generalizations that relax the previous definition to a local neighborhood of $x$, such as the Clarke subdifferential.\footnote{\url{https://en.wikipedia.org/wiki/Clarke_generalized_derivative}} Subdifferentiability can also create problems in AD, where different implementations of the same functions can provide different (possibly invalid) subgradients, and more refined concepts of chain rules must be considered for a formal proof \cite{kakade2018provably,bolte2020mathematical}.\footnote{Consider this example reproduced from \cite{bolte2020mathematical}: define two functions, $\text{ReLU}_2(s) = \text{ReLU}(-s)+s$ and $\text{ReLU}_3(s) = 0.5(\text{ReLU}(s)+\text{ReLU}_2(s))$. They are both equivalent to ReLU, but in PyTorch a backward pass in $0$ returns $0.0$ for ReLU, $1.0$ for ReLU$_2$, and $0.5$ for ReLU$_3$.}

\clearpage

\section*{From theory to practice}

\begin{wrapfigure}{r}{3.0cm}
\vspace{-3em}\includegraphics[width=3.0cm]{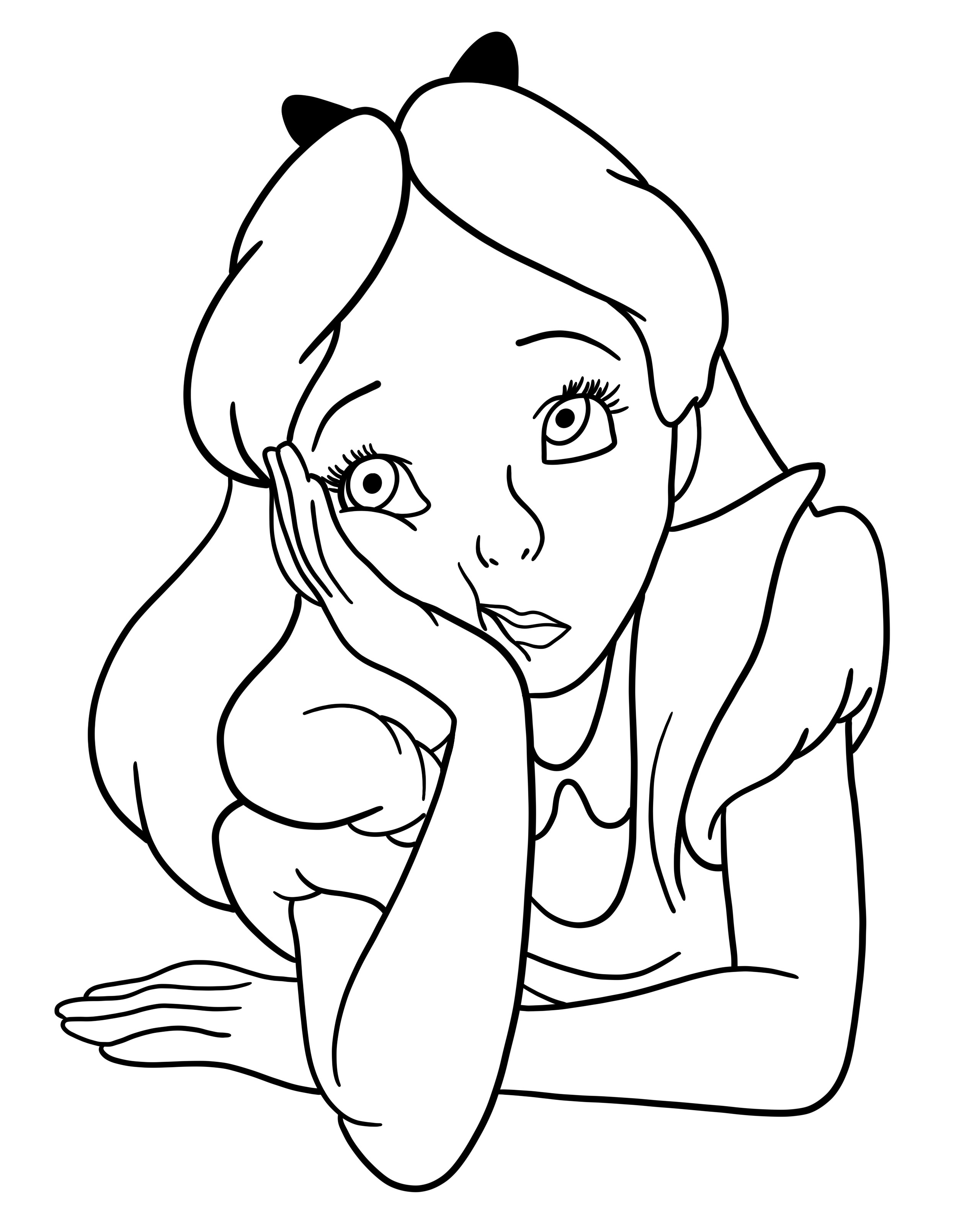}
\vspace{-3em}
\end{wrapfigure}

If you followed the exercises in Chapter \ref{chap:fully_connected_models}, you already saw an application of R-AD in both PyTorch and JAX, and this chapter (especially Section \ref{subsec:implementing_rad}) should have clarified their implementation.

It is a good idea to try and re-implement a simple R-AD system, similar to the one of PyTorch. For example, focusing on scalar-valued quantities, the {\footnotesize\texttt{micrograd}} repository\footnote{\url{https://github.com/karpathy/micrograd}} is a very good didactical implementation. The only detail we do not cover is that, once you move to a general acyclic graph, an ordering of the variables in the computational graph before the backward pass is essential to avoid creating wrong backpropagation paths. In micrograd, this is achieved via a non-expensive topological sorting of the variables.

It is also interesting to try and implement a new primitive (in the sense used in this chapter) in PyTorch, which requires specifying its forward pass along with its VJPs.\footnote{\url{https://pytorch.org/docs/master/notes/extending.html}} One example can be one of the trainable activation functions from Section \ref{sec:activation_functions}. This is a didactical exercise, in the sense that this can be implemented equivalently by subclassing \mintinline{python}{nn.Module} and letting PyTorch's AD engine work out the backward pass.

All these steps can also be replicated in JAX:
\begin{itemize}
\item Implement a didactic version of JAX with {\footnotesize\texttt{autodidax}}.\footnote{ \url{https://jax.readthedocs.io/en/latest/autodidax.html}}
\item Write out a new primitive by implementing the corresponding VJP.\footnote{ \url{https://jax.readthedocs.io/en/latest/notebooks/Custom_derivative_rules_for_Python_code.html}}
\item Read the \textbf{JAX Autodiff} Cookbook\footnote{\url{https://jax.readthedocs.io/en/latest/notebooks/autodiff_cookbook.html}} to discover advanced use-cases for the automatic differentiation engine, such as higher-order derivatives, Hessians, and more. 
\end{itemize}

%% file: 7_convolutional_layers.tex
\chapter{Convolutional layers}
\label{chap:cnns}

\begin{supportbox}{About this chapter}
In this chapter we introduce our second core layer, the \textbf{convolutional layer}, which is designed to work with images (or, more in general, sequential data of any kind) by exploiting two  key ideas that we call \textit{locality} and \textit{parameter sharing}.
\end{supportbox}

Fully-connected layers are important historically, but less so from a practical point of view: on unstructured data (what we also call \textbf{tabular} data, as it can be easily represented as a table) MLPs are generally outperformed by other alternatives, such as random forests or well tuned support vector machines \cite{grinsztajn2022tree}. This is not true, however, as soon as we consider other types of data, having some structure that can be exploited in the design of the layers and of the model.

In this chapter we  consider the image domain, while in the next chapters we also consider applications to time series, audio, graphs, and videos. In all these cases, the input has a sequential structure (either temporal, spatial, or of other type) that can be leveraged to design layers that are both performant, easily composable, and highly efficient in terms of parameters. Interestingly, we will see that possible solutions can be designed by taking as starting point a fully-connected layer, and then suitably restricting or generalizing it based on the properties of the input.
\section{Towards convolutional layers}
\label{sec:towards_convolutive_layers}
\subsection{Fully-connected layers are not enough}
An image can be described by a tensor $X \sim(h,w,c)$, where $h$ is the height of the image, $w$ the width of the image, and $c$ is the number of channels (which can be $1$ for black and white images, $3$ for color images, or higher for, e.g., hyper-spectral images). Hence, a mini-batch of images will generally be of rank $4$ with an additional leading batch dimension $(b, h, w, c)$. The three dimensions are not identical, since $h$ and $w$ represent a spatial arrangement of \textit{pixels}, while the channels $c$ do not have a specific ordering, in the sense that storing images in an RGB or a GBR format is only a matter of convention. 

\begin{supportbox}{On notation, channels, and features}
We use the same symbol we used for features in the tabular case ($c$) because it will play a similar role in the design of the models, i.e., we can think of each pixel as described by a generic set of $c$ \textit{features} which are updated in parallel by the layers of the model. Hence, the convolutional layer will return a generic tensor $(h,w,c^\prime)$ with an embedding of size $c^\prime$ for each of the $hw$ pixels.
\end{supportbox}

In order to use a fully-connected layer, we would need to “flatten” (vectorize) the image:
\begin{equation}
\mathbf{h} =\phi(\mathbf{W}\cdot \eqnmarkbox[drawred]{node}{\text{vect}(X)})
\label{eq:basic_image_layer}
\end{equation}
\annotate[yshift=-1em]{below,right}{node}{Flattened image}

\vspace{1em}
where $\text{vect}(x)$ is equivalent to {\footnotesize\mintinline{python}{x.reshape(-1)}} in PyTorch, and it returns for a generic rank-$n$ tensor $x \sim (i_1, i_2, \ldots, i_n)$ an equivalent tensor $\mathbf{x} \sim \left(\prod_{j=1}^n i_j\right)$.

Although it should be clear this is an inelegant approach, it is worth emphasizing some of its disadvantages. First, we have lost a very important property from the previous section, namely, \textbf{composability}: our input is an image, while our output is a vector, meaning we cannot concatenate two of these layers. We can recover this by reshaping the output vector to an image:
\begin{equation}
H = \text{unvect}(\phi(\mathbf{W}\cdot\text{vect}(X)))
\end{equation}
where we assume that the layer does not modify the number of pixels, and $\text{unvect}$ reshapes the output to a $(h,w,c^\prime)$ tensor, with $c^\prime$ an hyper-parameter. 

This leads directly to the second issue, which is that the layer has a \textit{huge} number of parameters. Considering, for example, a (1024, 1024) image in RGB, keeping the same dimensionality in output results in $(1024*1024*3)^2$ parameters (or $(hw)^2cc^\prime$ in general), which is in the order of $10^{13}$! We can interpret the previous layer as follows: for each pixel, every channel in the output is a weighted combination of \textit{all} channels of \textit{all} pixels in the input image. As we will see, we can obtain a more efficient solution by restricting this computation.

\begin{supportbox}{More on reshaping}

In order to flatten (or, more in general, reshape) a tensor, we need to decide an ordering in which to process the values. In practice, this is determined by the way the tensors are stored in memory: in most frameworks, the tensor's data is stored sequentially in a contiguous block of memory, in what is called a \textbf{strided layout}. Consider the following example:

{
\vspace{1em}
\mintinline[breaklines]{python}{torch.randn(32, 32, 3).stride()       # (96, 3, 1)}
\vspace{1em}
}

The stride is the number of steps that must be taken in memory to move of $1$ position along that axis, i.e., the last dimension of the tensor is contiguous, while to move of one position in the first dimension we need $96$ ($32*3$) steps. This is called a \textbf{row-major} ordering or, in image analysis, a \textbf{raster} order.\footnote{\url{https://en.wikipedia.org/wiki/Raster_scan}} Every reshaping operation works by moving along this strided representation.

\end{supportbox}

\vspace{1em}
As a running example to visualize what follows, consider a 1D sequence (we will consider 1D sequences more in-depth later on; for now, you can think of this as “\textit{4 pixels with a single channel}”):

$$
\mathbf{x} = \left[x_1, x_2, x_3, x_4\right]
$$

In this case, we do not need any reshaping operations, and the previous layer (with $c^\prime = 1$) can be written as: \clearpage

$$
\begin{bmatrix} h_1 \\ h_2 \\ h_3 \\ h_4 \end{bmatrix}=\begin{bmatrix}W_{11} & W_{12} & W_{13} & W_{14} \\ W_{21} & W_{22} & W_{23} & W_{24} \\ W_{31} & W_{32} & W_{33} & W_{34} \\ W_{41} & W_{42} & W_{43} & W_{44} \end{bmatrix} \begin{bmatrix} x_1 \\ x_2 \\ x_3 \\ x_4 \end{bmatrix}
$$

\begin{SCfigure}
    \centering
    \hspace{1em}\includegraphics[width=0.4\textwidth]{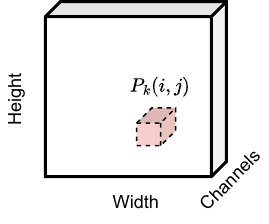}
    \caption{Given a tensor $(h,w,c)$ and a maximum distance $k$, the \textbf{patch} $P_k(i,j)$ (shown in red) is a $(2k+1,2k+1,c)$ tensor collecting all pixels at distance at most $k$ from the pixel in position $(i,j)$.}
    \label{fig:patch}
\end{SCfigure}

\subsection{Local layers}

The spatial arrangement of pixels introduces a metric (a distance) between the pixels. While there are many valid notions of “distance”, we will find it convenient to work with the following definition, which defines the distance between pixel $(i,j)$ and $(i^\prime, j^\prime)$ as the maximum distance across the two axes:
\begin{equation}
d((i,j), (i^\prime,j^\prime))=\max(\lvert i-i^\prime \rvert,\lvert j-j^\prime\rvert)
\label{eq:pixel_distance}
\end{equation}
How can we exploit this idea in the definition of a layer? Ideally, we can imagine that the influence of a pixel on another one decreases with a factor inversely proportional to their distance. Pushing this idea to its extreme, we can assume that the influence is effectively zero for a distance larger than some threshold. To formalize this insight, we introduce the concept of a \textbf{patch}.

\begin{definition}[Image patch] \addbottle $\,$

Given an image $X$, we define the \textbf{patch} $P_{k}(i,j)$ as the sub-image centered at $(i,j)$ and containing all pixels at distance equal or lower than $k$:

$$
P_{k}(i,j) = \idx{X}{i-k:i+k,j-k:j+k,:}
$$

where distance is defined as in \eqref{eq:pixel_distance}. This is shown visually in Figure \ref{fig:patch}.

\end{definition}

The definition is only valid for pixels which are at least $k$ steps away from the borders of the image: we will ignore this point for now and return to it later. Each patch is of shape $(s,s,c)$, where $s=2k+1$, since we consider $k$ pixels in each direction along with the central pixel. For reasons that will be clarified later on, we call $s$ the \textbf{filter size} or \textbf{kernel size}.

Consider a generic layer $H = f(X)$ taking as input a tensor of shape $(h,w,c)$ and returning a tensor of shape $(h,w,c^\prime)$. If the output for a given pixel only depends on a patch of predetermined size, we say that the layer is \textbf{local}.

\begin{definition}[Local layer] $\,$

Given an input image $X \sim(h,w,c)$, a layer $f(X) \sim(h,w,c^\prime)$ is \textbf{local} if there exists a $k$ such that:

$$
\idx{f(X)}{ij} = f(P_{k}(i, j))
$$

This has to hold for all pixels of the image.

\end{definition}

We can transform the layer \eqref{eq:basic_image_layer} into a local layer by setting to $0$ all weights belonging to pixels outside the influence region (\textbf{receptive field}) of each pixel:

\vspace{1em}
$$
H_{ij} =\phi\left(\eqnmarkbox[drawgreen]{node2}{\mathbf{W}_{ij}}\cdot\eqnmarkbox[drawred]{node}{\text{vect}(P_{k}(i,j))}\right)
$$
\annotate[yshift=1em]{above,left}{node}{Flattened patch (of shape $s^2 c$)}
\annotate[yshift=-1em]{below,right}{node2}{Position-dependent weight matrix}
\vspace{1em}

We call this class of layers \textbf{locally-connected}. Note that we have a different weight matrix $\mathbf{W}_{ij} \sim({c^\prime, ssc})$ for each output pixel, resulting in $hw(s^2cc^\prime)$ parameters. By comparison, we had $(hw)^2cc^\prime$ parameters in the initial layer, for a reduction factor of $\frac{s^2}{hw}$ in the number of parameters.

Considering our toy example, assuming for example $k=1$ (hence $s=3$) we can write the resulting operation as:

$$
\begin{bmatrix} h_1 \\ h_2 \\ h_3 \\ h_4 \end{bmatrix}=\begin{bmatrix}W_{12} & W_{13} & {\color{drawred}0} & {\color{drawred}0} \\ W_{21} & W_{22} & W_{23} & {\color{drawred}0} \\ {\color{drawred}0} & W_{31} & W_{32} & W_{33} \\ {\color{drawred}0} & {\color{drawred}0} & W_{41} & W_{42} \end{bmatrix} \begin{bmatrix} x_1 \\ x_2 \\ x_3 \\ x_4 \end{bmatrix}
$$

Our operation is not defined for $x_1$ and $x_4$, in which case we have considered a “shortened” filter by removing the weights corresponding to undefined operations. Equivalently, you can think of adding $0$ on the border whenever necessary: \clearpage

$$
\begin{bmatrix} h_1 \\ h_2 \\ h_3 \\ h_4 \end{bmatrix}=\begin{bmatrix}W_{11} & W_{12} & W_{13} & 0 & 0 & {\color{drawred}0}\\ {\color{drawred}0} & W_{21} & W_{22} & W_{23} & 0 & {\color{drawred}0}\\ {\color{drawred}0} & 0 & W_{31} & W_{32} & W_{33} & {\color{drawred}0} \\ {\color{drawred}0} & 0 & 0 & W_{41} & W_{42} & W_{43} \end{bmatrix} \begin{bmatrix} {\color{drawred}0} \\ x_1 \\ x_2 \\ x_3 \\ x_4 \\ {\color{drawred}0} \end{bmatrix}
$$

This technique is called \textbf{zero-padding}. In an image, for a kernel size $2k+1$ we need exactly $k$ rows and columns of $0$ on each side to ensure that the operation is valid for each pixel. Otherwise, the output cannot be computed close to the borders, and the output tensor will have shape $(h-2k, w-2k, c^\prime)$. Both are valid options in most frameworks.

\vspace{1em}
\begin{supportbox}{On our definition of patches}
The definition of convolutions using the idea of patches is a bit unconventional, but I find it to greatly simplify the notation. I provide a more conventional, signal processing oriented definition later on. The two definitions are equivalent and can be used interchangeably. The patch-oriented definition requires an \textbf{odd} kernel size and does not allow for \textbf{even} kernel sizes, but these are uncommon in practice.
\end{supportbox}

\subsection{Translation equivariance and the convolutional layer}

\addclock In a locally-connected layer, two identical patches can result in different outputs based on their location: some content on pixel $(5,2)$, for example, will be processed differently than the same content on pixel $(39, 81)$ because the two matrices $\mathbf{W}_{5,2}$ and $\mathbf{W}_{39,81}$ are different. For the most part, however, we can assume that this information is irrelevant: informally, “a horse is a horse”, irrespective of its positioning on the input image. We can formalize this with a property called \textbf{translation equivariance}.

\begin{definition}[Translation equivariance] $\,$

We say that a layer $H = f(X)$ is \textbf{translation equivariant} if translations of the inputs imply an equivalent translation of the output:

$$
\eqnmarkbox[drawred]{node}{P_{k}(i,j) = P_{k}(i^\prime, j^\prime)} \;\; \textnormal{implies} \;\;  \eqnmarkbox[drawgreen]{node2}{f(P_{k}(i,j)) = f(P_{k}(i^\prime, j^\prime))}
$$
\annotate[yshift=1em]{above,right}{node}{Identical patches}
\annotate[yshift=-1em]{below,left}{node2}{Identical outputs}

\end{definition}

\vspace{0.5em}
To understand the nomenclature, note that we can interpret the previous definition as follows: whenever an object moves (translates) on the image from position $(i,j)$ to position $(i^\prime, j^\prime)$, the output $f(P_{k}(i,j))$ that we we had in $(i,j)$ will now be found in $f(P_{k}(i^\prime,j^\prime))$. Hence, the activations of the layer are moving with the same (\textit{èqui} in Latin) translational movement as the input. We will define more formally equivariance and invariance later on.

A simple way to achieve translation equivariance is given by \textbf{weight sharing}, i.e., letting every position share the same set of weights:
$$
H_{ij} =\phi(\eqnmarkbox[drawred]{node}{\mathbf{W}}\cdot\text{vect}(P_{k}(i,j)))
$$
\annotate[yshift=-1em]{below,right}{node}{Weight matrix independent of $(i,j)$}

This is called a \textbf{convolutional layer}, and it is extremely efficient in terms of parameters: we only have a single weight matrix $\mathbf{W}$ of shape $(c^\prime, ssc)$, which is independent from the resolution of the original image (once again, contrast this with a layer which is only locally-connected with $hw(s^2c^\prime c)$ parameters: we have reduced them by another factor $\frac{1}{hw}$). We can write a variant with biases by adding $c^\prime$ additional parameters in the form of a bias vector $\mathbf{b} \sim (c^\prime)$. Because of its importance, we restate the full definition of the layer below.

\begin{definition}[Convolutional layer] \addbottle $\,$

Given an image $X \sim (h,w,c)$ and a kernel size $s=2k+1$, a \textbf{convolutional layer} $H=\textnormal{Conv2D}(X)$ is defined element-wise by:

\begin{equation}
H_{ij} = \mathbf{W} \cdot \textnormal{vect}(P_{k}(i,j)) + \mathbf{b}
\label{eq:convolutive_layer}
\end{equation}

The trainable parameters are $\mathbf{W} \sim (c^\prime, ssc)$ and $\mathbf{b} \sim (c^\prime)$. The hyper-parameters are $k$, $c^\prime$, and (eventually) whether to apply zero-padding or not. In the former case the output has shape $(h,w,c^\prime)$, in the latter case it has shape $(h-2k,w-2k,c^\prime)$. 

\end{definition}

\begin{mypy}{Convolution in PyTorch. Note that the channel dimension is -- by default -- the first one after the batch dimension. The kernel matrix is organized as a $(c^\prime, c, k, k)$ tensor. Padding can be specified as an integer or a string (`same' meaning that the output must have the same shape as the input, `valid' meaning no padding).}{code:convolution}
from torch.nn import functional as F
x = torch.randn(16, 3, 32, 32)
w = torch.randn(64, 3, 5, 5)
F.conv2d(x, w, padding='same').shape 
# [Out]: torch.Size([16, 64, 32, 32])
\end{mypy}

See Box \ref{code:convolution} for a code example. The equivalent object-oriented implementation can be found in {\footnotesize\mintinline{python}{torch.nn.Conv2D}}. By comparison, our toy example can be refined as follows:

\begin{equation}
\begin{bmatrix} h_1 \\ h_2 \\ h_3 \\ h_4 \end{bmatrix}=\begin{bmatrix}{\color{drawred}W_2} & {\color{drawred}W_3} & 0 & 0 \\ {\color{drawred}W_1} & {\color{drawred}W_2} & {\color{drawred}W_3} & 0 \\ 0 & {\color{drawred}W_1} & {\color{drawred}W_2} & {\color{drawred}W_3} \\ 0 & 0 & {\color{drawred}W_1} & {\color{drawred}W_2} \end{bmatrix} \begin{bmatrix} x_1 \\ x_2 \\ x_3 \\ x_4 \end{bmatrix}
\label{eq:convolution_example}
\end{equation}

where we now have only three weights $\mathbf{W} = \left[W_1, W_2, W_3\right]^\top$ (the zero-padded version is equivalent to before and we omit it for brevity). This weight matrix has a special structure, where each element across any diagonal is a constant (e.g., on the main diagonal we only find $W_2$). We call these matrices \textbf{Toeplitz matrices},\footnote{\url{https://en.wikipedia.org/wiki/Toeplitz_matrix}} and they are fundamental to properly implement a convolutional layer on modern hardware. Toeplitz matrices are an example of \textbf{structured} dense matrices \cite{qiu2024compute}. Equation \eqref{eq:convolution_example} should also clarify that a convolution remains a \textit{linear} operation, albeit with a highly restricted weight matrix compared to a fully-connected one.

\subsection*{Convolutions and terminology}

\addteacup Our terminology comes (mostly) from signal processing. We can understand this by rewriting the output of the convolutional layer in a more standard form. To this end, we first rearrange the weight matrix into an equivalent weight tensor $W$ of shape $(s,s,c,c^\prime)$, similar to the PyTorch implementation in Box \ref{code:convolution}. For convenience, we also define a function that converts an integer $i^\prime$ from the interval $\left[1, \ldots, 2k+1\right]$ to the interval $\left[ i - k, \ldots, i + k\right]$:
\begin{equation}
t(i) = i - k - 1
\label{eq:convolutive_offset}
\end{equation}
where $k$ is left implicit in the arguments of $t(\bullet)$. We now rewrite the output of the layer with explicit summations across the axes:
\begin{equation}
H_{ijz} = \sum_{i^\prime=1}^{2k+1}\sum_{j^\prime = 1}^{2k+1}\sum_{d=1}^{c} \idx{W}{i^\prime, j^\prime, z, d}\idx{X}{i^\prime+t(i),j^\prime+t(j), d}
\label{eq:convolution_full_indexing}
\end{equation}
Check carefully the indexing: for a given pixel $(i,j)$ and output channel $z$ (a free index running from $1$ to $c^\prime$), on the spatial dimensions $W$ must be indexed along $1, 2, \ldots, 2k+1$, while $X$ must be indexed along $i-k, i-k+1, \ldots, i+k-1,i+k$. The index $d$ runs instead over the input channels.

From the point of view of signal processing, equation \eqref{eq:convolution_full_indexing} corresponds to a filtering operation on the input signal $X$ through a set of \textbf{finite impulse response} (FIR) filters \cite{uncini2015fundamentals}, implemented via a discrete convolution (apart from a sign change). Each filter here corresponds to a slice $W_{:,:,:,i}$ of the weight matrix. In standard signal processing, these filters can be manually designed to perform specific operations on the image. As an example, a $3 \times 3$ filter to detect ridges can be written as:\footnote{\url{https://en.wikipedia.org/wiki/Kernel_(image_processing)}}
$$
W =\begin{bmatrix}-1&-1&-1\\-1&8&-1\\-1&-1&-1\end{bmatrix}
$$
In convolutional layers, instead, these filters are initialized randomly and trained via gradient descent. We consider the design of \textbf{convolutional models} built on convolutional layers in the next section. An interesting aspect of convolutional layers is that the output maintains a kind of “spatial consistency” and it can be plotted: we call a slice $H_{:,:,i}$ of the output an \textbf{activation map} of the layer, representing how much the specific filter was “activated” on each input region. We will consider in more detail the exploration of these maps in the next volume.

\section{Convolutional models}

\subsection{Designing convolutional “blocks”}

With the definition of a convolutional layer in hand, we now turn to the task of building \textbf{convolutional models}, also called \textbf{convolutional neural networks} (CNNs). We consider the problem of image classification, although a lot of what we say can be extended to other cases. To begin with, we formalize the concept of \textbf{receptive field}.

\begin{definition}[Receptive field] $\,$

Denote by $X$ an image, and by $H = g(X)$ a generic intermediate output of a convolutional model, e.g., the result of applying 1 or more convolutional layers. The \textbf{receptive field} $R(i,j)$ of pixel $(i,j)$ is the subset of $X$ which contributed to its computation:

$$
 \idx{g(X)}{ij} = g(R(i,j)), \;\;\; R(i,j) \subseteq X
$$

\end{definition}

For a single convolutional layer, the receptive field of a pixel is equal to a patch: $R(i,j) = P_{k}(i,j)$. However, it is easy to prove that for two convolutional layers in sequence with identical kernel size, the resulting receptive field is $R(i,j) = P_{2k}(i,j)$, then $P_{3k}(i,j)$ for three layers, and so on. Hence, the receptive field increases \textit{linearly} in the number of convolutional layers. This motivates our notion of locality: even if a single layer is limited in its receptive field by the kernel size, a sufficiently large stack of them results in a \textit{global} receptive field.

Consider now a sequence of two convolutional layers:
$$
H=\text{Conv}(\text{Conv}(X))
$$
Because convolution is a linear operation (see previous section), this is equivalent to a single convolution with a larger kernel size (as per the above). We can avoid this “collapse” in a similar way to fully-connected layers, by interleaving them with activation functions:
\begin{equation}
H = (\phi \circ \text{Conv}\circ\ldots\circ\phi\circ\text{Conv})(X)
\label{eq:convolutional_block}
\end{equation}
To continue with our design, we note that in \eqref{eq:convolutional_block} the channel dimension will be modified by each convolutional layer, while the spatial dimensions will remain of the same shape (or will be slightly reduced if we avoid zero-padding). However, it can be advantageous in practice to eventually reduce this dimensionality if our aim is something like image classification.

Consider again the example of a horse appearing in two different regions across two different images. The translation equivariance property of convolutional layers guarantees that every feature found in region 1 in the first image will be found, correspondingly, in region 2 of the second image. However, if our aim is “horse classification”, we eventually need one or more neurons activating for an horse \textit{irrespective of where it is found} in the image itself: if we only consider shifts, this property is called \textbf{translation invariance}.

Many operations that reduce over the spatial dimensions are trivially invariant to translations, for example:
$$
H^\prime=\sum_{i,j}H_{ij} \;\text{ or }\; H^\prime=\max_{i,j}(H_{ij})
$$
In the context of CNNs, this is called a \textbf{global pooling}. However, this destroys all spatial information present in the image. We can obtain a slightly more efficient solution with a partial reduction, called \textbf{max-pooling}. 

\begin{definition}[Max-pooling layer] \addbottle $\,$

Given a tensor $X \sim (h,w,c)$, a max-pooling layer, denoted as $\text{MaxPool(X)} \sim (\frac{h}{2}, \frac{w}{2}, c)$, is defined element-wise as:

$$
\idx{\textnormal{MaxPool}(X)}{ijc} = \max\left(\eqnmarkbox[drawred]{node}{\idx{X}{2i-1:2i, 2j-1:2j,c}}\right)
$$
\annotate[yshift=-1em]{below,left}{node}{$2 \times 2$ image patch}

\end{definition}

\begin{SCfigure}
    \centering
    \hspace{1em}\includegraphics[width=0.5\textwidth]{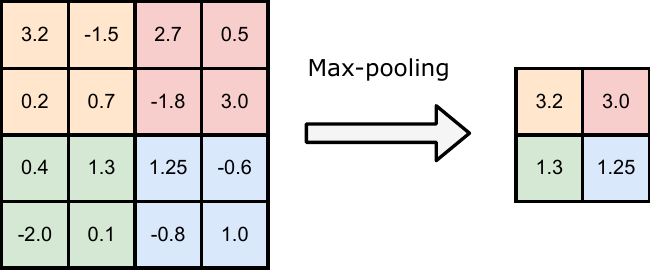}
    \caption{Visualization of 2x2 max-pooling on a (4,4,1) image. For multiple channels, the operation is applied independently on each channel.
}
    \label{fig:max_pooling}
\end{SCfigure}

\vspace{1em}
Hence, we take $2\times 2$ windows of the input, and we compute the maximum value independently for each channel (this is generalized trivially to larger windows). Max-pooling effectively halves the spatial resolution while leaving the number of channels untouched. An example is shown in Figure \ref{fig:max_pooling}.

\begin{SCfigure}
    \centering
    \hspace{1em}\includegraphics[width=0.6\textwidth]{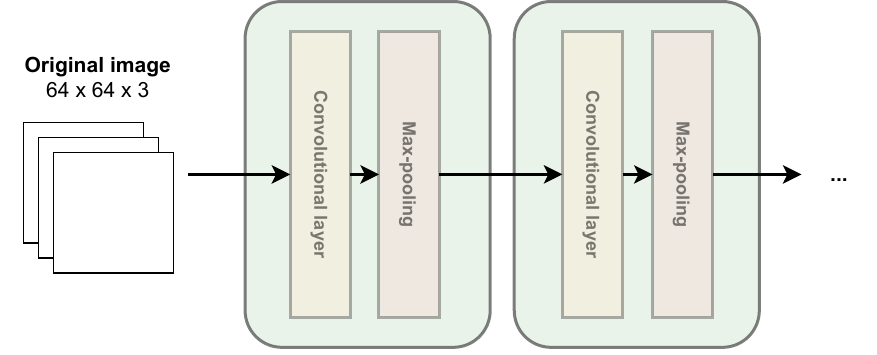}
    \caption{Abstracting away from “layers” to “blocks” to simplify the design of differentiable models.}
    \label{fig:cnn_blocks}
\end{SCfigure}

We can build a convolutional “block” by stacking several convolutional layers with a max-pooling operation  (see Figure \ref{fig:cnn_blocks}):
$$
\text{ConvBlock}(X)= (\text{MaxPool} \circ \phi \circ \text{Conv}\circ\ldots\circ\phi\circ\text{Conv})(X)
$$
Proceeding iteratively, we define a more complex network by stacking together multiple such blocks:
\begin{equation}
H = (\text{ConvBlock}\circ\text{ConvBlock}\circ\ldots\circ\text{ConvBlock})(X)
\label{eq:convolutive_backbone}
\end{equation}
This design has a large number of hyper-parameters: the output channels of each layer, the kernel size of each layer, etc. It is common to drastically reduce the search space for the design by making some simplifying assumptions. For example, the VGG design \cite{szegedy2015going} popularized the idea of maintaining the filter size constant in each layer (e.g., $k=3$), while keeping the number of channels constant in each block and doubling them in-between every block.

An alternative way for reducing the dimensionality is to downsample the output of a convolutional layer: this is called the \textbf{stride} of the convolution. For example, a convolution with stride $1$ is a normal convolution, while a convolution with stride $2$ will compute only one output pixel every $2$, a convolution with stride $3$ will compute one output every $3$ pixels, and so on. Large strides and max-pooling can also be combined together depending on how the entire model is designed.

\vspace{1em}
\begin{supportbox}{Invariance and equivariance}
Informally, if $T$ is a transformation on $x$ from some set (e.g., all possible shifts), we say a function $f$ is equivariant if $f(Tx)=Tf(x)$, and invariant if $f(Tx)=f(x)$. The space of all transformations form a group \cite{bronstein2017geometric}, and the matrix corresponding to a specific transformation is called a \textbf{representation} for that group. Convolutional layers are (roughly) equivariant to translations by design, but other strategies can be found for more general forms of symmetries, such as averaging over the elements of the group (\textbf{frame averaging}, \cite{puny2021frame}). We will see other types of layers' equivariances in Chapter \ref{chap:transformers} and Chapter \ref{chap:gnns}.
\end{supportbox}

\subsection{Designing the complete model}

We can now complete the design of our model. By stacking together multiple convolutional blocks as in \eqref{eq:convolutive_backbone}, the output $H$ will be of shape $(h^\prime, w^\prime, c^\prime)$, where $w^\prime$ and $h^\prime$ depend on the number of max-pooling operations (or on the stride of the convolutional layers), while $c^\prime$ will depend only on the hyper-parameters of the last convolutional layer in the sequence. Note that each element $H_{ij}$ will correspond to a “macro-region” in the original image, e.g., if $h^\prime, w^\prime = 2$, $H_{11}$ will correspond to the “top-left” quadrant in the original image. We can remove this spatial dependency by performing a final global pooling operation before classification. 

\begin{figure}
    \centering
    \includegraphics[width=0.9\textwidth]{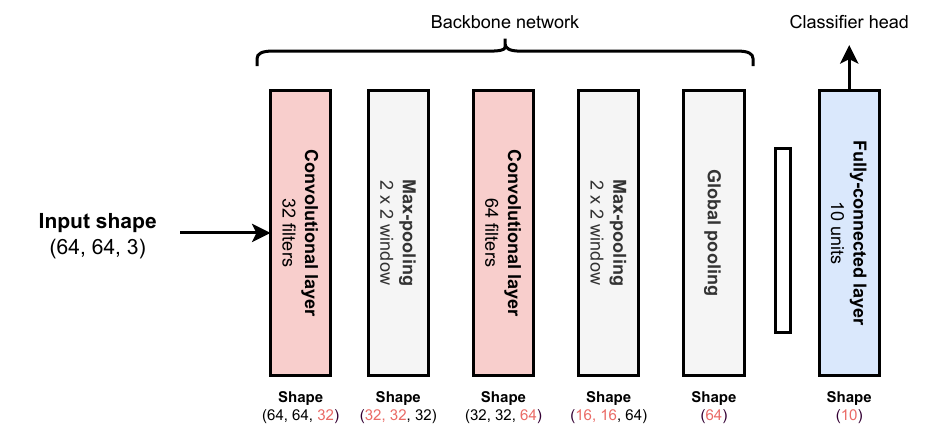}
    \caption{Worked-out design of a very simple CNN for image classification (assuming 10 output classes). We show the output shape for each layer on the bottom. The global pooling operation can be replaced with a flattening operation. The last (\textbf{latent}) representation before the classification head is very useful when fine-tuning large-scale pre-trained models -- it is an \textbf{embedding} of the image in the sense of Section \ref{subsec:variants_supervised_learning}.}
    \label{fig:cnn_architecture}
\end{figure}

The complete model, then, can be decomposed as three major components: a series of convolutional blocks, a global average pooling, and a final block for classification.
\begin{align} 
H = (\text{ConvBlock}\circ\ldots\circ\text{ConvBlock})(X) \label{eq:conv_blocks} \\
\mathbf{h}= \frac{1}{h^\prime w^\prime}\sum_{i,j}H_{ij}  \label{eq:global_avg_pooling} \\ 
y=\text{MLP}(\mathbf{h}) \label{eq:classification_head}
\end{align}
where $\text{MLP}(\mathbf{h})$ is a generic sequence of fully-connected layers (a flattening operation can also be used in place of the global pooling). This is a prototypical example of a CNN. See Figure \ref{fig:cnn_architecture} for a worked-out example.

This design has a few interesting properties we list here:

\begin{enumerate}
\item It can be trained like the models described in Chapter \ref{chap:linear_models} and Chapter \ref{chap:fully_connected_models}. For example, for classification, we can wrap the output in a softmax and train all parameters by minimizing the cross-entropy. The same rules of back-propagation described in Chapter \ref{chap:automatic_differentiation} apply here.
\item Because of the global pooling operation, it does not depend on a specific input resolution. However, it is customary to fix this during training and inference to simplify mini-batching (more on variable length inputs in the next chapter).
\item \eqref{eq:global_avg_pooling} can be thought of as a “feature extraction” block, while \eqref{eq:classification_head} as the “classification block”. This interpretation will be very useful when we consider transfer learning in the next volume. We call the feature extraction block the \textbf{backbone} of the model, and the classification block the \textbf{head} of the model.
\end{enumerate}

\subsection*{Notable types of convolution}

We close the chapter by mentioning two instances of convolutional layers that are common in practice. 

First, consider a convolutional layer with $k=0$, i.e., a so-called $1 \times 1$ convolution. This corresponds to updating each pixel’s embedding by a weighted sum of its channels, disregarding all other pixels:
$$
H_{{\color{drawred}ij}z} = \sum_{t=1}^c W_{zt}X_{{\color{drawred}ij}t}
$$
It is a useful operation for, e.g., modifying the channel dimension (we will see an example when dealing with residual connections in Chapter \ref{chap:deep_cnns}). In this case, the parameters can be compactly represented by a matrix $\mathbf{W} \sim (c^\prime, c)$. This is equivalent to a fully-connected layer applied on each pixel independently.

Second, consider an “orthogonal” variant to $1 \times 1$ convolutions, in which we combine pixels in a small neighborhood, but disregarding all channels except one:
$$
H_{ij{\color{drawred}c}} = \sum_{i^\prime=1}^{2k+1}\sum_{j^\prime = 1}^{2k+1} W_{i^\prime, j^\prime,{\color{drawred}c}}X_{i^\prime +t(i),j^\prime+t(j),{\color{drawred}c}}
$$
where $t(\bullet)$ is the offset defined in \eqref{eq:convolutive_offset}. In this case we have a rank-$3$ weight matrix $W$ of shape $(s, s, c)$, and each output channel $H_{:,:,c}$ is updated by considering only the corresponding input channel $X_{:,:,c}$. This is called a \textbf{depthwise convolution}, and it can be generalized by considering groups of channels, in which case it is called a \textbf{groupwise convolution} (with the depthwise convolution being the extreme case of a group size equal to $1$).

We can also combine the two ideas and have a convolution block made of alternating $1 \times 1$ convolutions (to mix the channels) and depthwise convolutions (to mix the pixels). This is called a \textbf{depthwise separable} convolution and it is common in CNNs targeted for low-power devices \cite{howard2017mobilenets}. Note that in this case, the number of parameters for a single block (compared to a standard convolution) is reduced from $sscc^\prime$ to $ssc + cc^\prime$. We will see later how these decompositions, where the input is processed alternatively across separate axes, are fundamental for other types of architectures, such as transformers, in Chapter \ref{chap:transformers}. \clearpage

\section*{From theory to practice}

\begin{wrapfigure}{r}{3.0cm}
\vspace{-3em}\includegraphics[width=3.0cm]{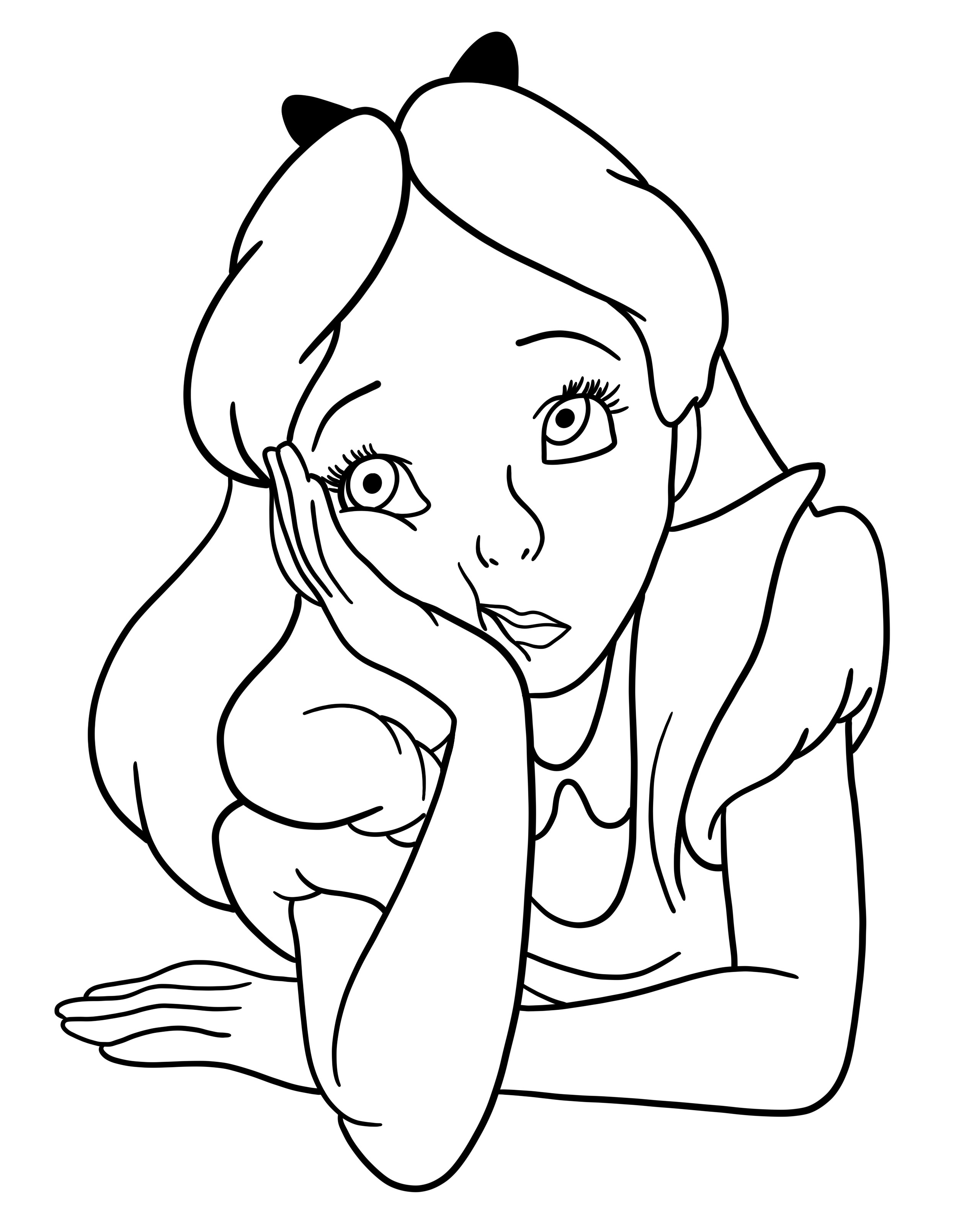}
\vspace{-2em}
\end{wrapfigure}

All the layers introduced in this chapter (convolution, max-pooling) are implemented in the \mintinline{python}{torch.nn} module. The torchvision library provides datasets and functions to load images, as well as interfaces to apply transformations to the images that will be very useful in the next chapter.\footnote{\url{https://pytorch.org/vision/stable/transforms.html}} 

Before proceeding, I suggest you follow and re-implement one of the many online tutorials on image classification in torchvision, which should now be relatively easy to follow.\footnote{As an example from the official documentation: \url{https://pytorch.org/tutorials/beginner/blitz/cifar10_tutorial.html}} Toy image datasets abound, including MNIST (digit classification) and CIFAR-10 (general image classification). Combining the torchvision loader with the layers in Equinox allows you to replicate the same tutorial in JAX, e.g.:

\url{https://docs.kidger.site/equinox/examples/mnist/}.

Implementing a convolution from scratch is also an interesting exercise, whose complexity depends on the level of abstraction. One possibility is to use the {\footnotesize\texttt{fold}/\texttt{unfold}} operations from PyTorch to extract the patches.\footnote{See for example: \url{https://github.com/loeweX/Custom-ConvLayers-Pytorch}} Premade kernels for convolutions will always be significantly faster, making this a purely didactic exercise.

If you have some signal processing background, you may know that convolution can also be implemented as multiplication by moving to the frequency domain. This is impractical for the small kernels we tend to use, but it can be useful for very large (also known as \textit{long}) convolutions, e.g.:

\url{https://github.com/fkodom/fft-conv-pytorch}

PyTorch also provides a differentiable Fast Fourier transform that you can use as a starting point.

%% file: 8_convolutions_beyond_images.tex
\chapter{Convolutions beyond images}
\label{chap:convolutions_beyond_images}

\begin{supportbox}{About this chapter}
Convolutional models are a powerful baseline model in many applications, going far beyond image classification. In this chapter we provide an overview of several such extensions, including the use of convolutional layers for 1D and 3D data, text modeling, and autoregressive generation. Several of the concepts we introduce (e.g., masking, tokenization) are fundamental in the rest of the book and for understanding modern LLMs.
\end{supportbox}

\section{Convolutions for 1D and 3D data}
\subsection{Beyond images: time series, audio, video, text}

\begin{figure}
    \centering
    \includegraphics[width=1.0\textwidth]{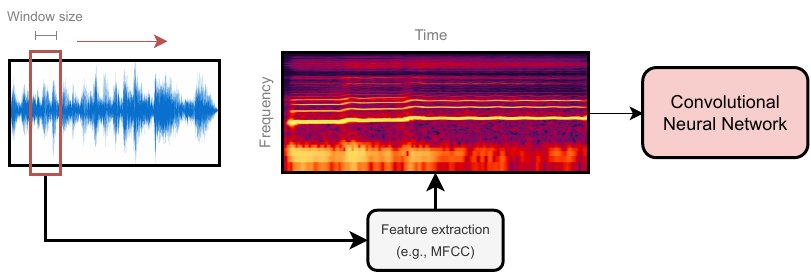}
    \caption{Audio can be represented as either a 1D sequence (left), or a 2D image in a time-frequency domain (middle). In the second case, we can apply the same techniques described in the previous chapter.}
    \label{fig:audio_analysis_frequency}
\end{figure}

In the previous chapter we focused exclusively on images. However, many other types of data share similar characteristics, i.e., one or more “ordered” dimensions representing time or space, and one dimension representing features (the channels in the image case). Let us consider some examples:

\begin{enumerate}
\item \textbf{Time series} are collections of measurements of one or more processes (e.g., stocks prices, sensor values, energy flows). We can represent a time series as a matrix $\mathbf{X} \sim (t,c)$, where $t$ is the length of the time series, and $\mathbf{X}_i \sim (c)$ are the $c$ measurements at time $t$ (e.g., $c$ sensors from an EEG scan, or $c$ stock prices). Each time instant is equivalent to a pixel, and each measurement is equivalent to a channel.
\item \textbf{Audio} files (speech, music) can also be described by a matrix $\mathbf{X} \sim (t,c)$, where $t$ is the length of the audio signal, while $c$ are the channels of the recording ($1$ for a mono audio, $2$ for a stereo signal, etc.). 

\begin{supportbox}{Frequency-analysis}
    Audios can also be converted to an image-like format via frequency analysis (e.g., extracting the MFCC coefficients over small windows), in which case the resulting \textit{time-frequency} images represent the evolution of the frequency content over the signal - see Figure \ref{fig:audio_analysis_frequency} for an example. With this preprocessing we can use standard convolutional models to process them.
\end{supportbox}
    
    \item \textbf{Videos} can be described by a rank-$4$ tensor $X \sim (t, h, w, c)$, where $t$ is the number of \textit{frames} of the video, and each frame is an image of shape $(h,w,c)$. Another example is a volumetric scan in medicine, in which case $t$ is the volume depth.
\end{enumerate}

Time series, audio signals, and videos can be described by their \textbf{sampling rate}, which denotes how many samples are acquired per unit of time, sometimes expressed in samples per second, or hertz (Hz). For example, classical EEG units acquire signals at 240 Hz, meaning 240 samples each second. A stock can be checked every minute, corresponding to 1/60 Hz. By contrast, audio is acquired with very high frequency to ensure fidelity: for example, music can be acquired at $44.1e^3$ Hz (or $44.1$ kHz). Typical acquisition \textbf{frame rates} for video are instead around $24$ frames per second (fps) to ensure smoothness to the human eye.

Image resolution, audio sampling rate, and video frame rates all play similar roles in determining the precision with which a signal is acquired. For an image, we can assume a fixed resolution a priori (e.g., $1024 \times 1024$ pixels). This is reasonable, since images can always be reshaped to a given resolution while maintaining enough consistency, except for very small resolutions. By contrast, audio and video durations can vary from input to input (e.g., a song of 30 seconds vs. a song of 5 minutes), and they cannot be reshaped to a common dimension,\footnote{In the sense of having the same duration \textit{and} resolution.} meaning that our datasets will be composed of \textbf{variable-length} data. In addition, audio resolution can easily grow very large: with a $44.1$ kHz sampling rate, a $3$-minute audio will have $\approx 8M$ samples.

We also note that the dimensions in these examples can be roughly categorized as either “spatial dimensions” (e.g., images) or “temporal dimensions” (e.g., audio resolution). While images can be considered symmetric along their spatial axes (in many cases, an image flipped along the width is another valid image), time is \textit{asymmetric}: an audio sample inverted on its temporal axis is in general invalid, and an inverted time series represents a series evolving from the future towards its past. Apart from exploiting this aspect in the design of our models (\textbf{causality}), we can also be interested in \textit{predicting} future values of the signal: this is called \textbf{forecasting}.

Finally, consider a text sentence, such as “\textit{the cat is on the table}”. There are many ways to split this sentence into pieces. For example, we can consider its individual syllables: [”\textit{the}”, “\textit{cat}”, “\textit{i}”, “\textit{s}”, “\textit{on}”, “\textit{the}”, “\textit{ta}”, \textit{ble}”]. This is another example of a sequence, except that each element of the sequence is now a categorical value (the syllable) instead of a numerical encoding. Hence, we need some way of encoding these values into features that can be processed by the model: splitting a text sequence into components is called \textbf{tokenization}, while turning each token into a vector is called \textbf{embedding} the tokens.

In the next sections we consider all these aspects (variable-length inputs, causality, forecasting, tokenization, and embedding) in turn, to see how we can build convolutional models to address them. Some of the techniques we introduce, such as masking, are very general and are useful also for other types of models, such as transformers. Other techniques, such as dilated convolutions, are instead specific to convolutional models.

\subsection{1D and 3D convolutional layers}

Let us consider how to define convolutions for 1D signals (e.g., time series, audio) and their extension to 3D signals (e.g., videos). Note that the dimensionality refers only to the number of dimensions along which we convolve (spatial or time), and does not include the channel dimension. Recall that, in the 1D case, we can represent the input as a single matrix:

\vspace{1em}
\begin{equation*}
\mathbf{X} \sim (\eqnmarkbox[drawred]{node}{t}, \eqnmarkbox[drawgreen]{node2}{c})
\end{equation*}
\annotate[yshift=1em]{above,left}{node}{Length of the sequence}
\annotate[yshift=1em]{above,right}{node2}{Features}

We now replicate the derivation from Chapter \ref{chap:cnns}. Given a patch size $s=2k+1$, we define $P_{k}(i) \sim (s,c)$ as the subset of rows in $\mathbf{X}$ at distance at most $k$ from $i$ (ignoring border elements for which zero-padding can be used). A 1D convolutional layer $\mathbf{H} = \text{Conv1D}(\mathbf{X})$ outputs a matrix $\mathbf{H} \sim (t, c^\prime)$, with $c^\prime$ an hyper-parameter that defines the output dimensionality, defined row-wise as:

\begin{equation}
\idx{\text{Conv1D}(X)}{i} = \phi(\mathbf{W} \cdot\text{vect}(P_{k}(i)) + \mathbf{b})
\label{eq:1d_convolution}
\end{equation}

with trainable parameters $\mathbf{W} \sim (c^\prime,sc)$ and $\mathbf{b} \sim (c^\prime)$. Like in the 2D case, this layer is local (for a properly modified definition of locality) and equivariant to translations of the sequence. 

In the 2D case, we also discussed an alternative notation with all indices explicitly summed over:

\begin{equation}
H_{ijz} = \sum_{i^\prime=1}^{2k+1}\sum_{j^\prime = 1}^{2k+1}\sum_{d=1}^{c} \idx{W}{i^\prime, j^\prime,z,d}\idx{X}{i^\prime+t(i),j^\prime+t(j),d}
\label{eq:conv_again}
\end{equation}

where $t(i)=i+k-1$ as in \eqref{eq:convolutive_offset}. Recall that we use $t$ to index $i^\prime$ and $j^\prime$ differently for the two tensors: from $1$ to $2k+1$ for $W$, and from $i-k$ to $i+k$ for $X$. The equivalent variant for \eqref{eq:1d_convolution} is obtained trivially by removing one summation index:
\begin{equation}
H_{iz} = \sum_{i^\prime=1}^{2k+1}  \sum_{d=1}^{c} \idx{W}{i^\prime,z,d}\idx{X}{i^\prime+t(i),d}
\label{eq:1d_convolution_sum}
\end{equation}
where the parameters $W \sim (s, c^\prime, c)$ are now organized in a rank-$3$ tensor. By contrast, the 3D variant is obtained by adding a new summation over the third dimension with index $p$:
$$
H_{{\color{drawred}p}ijz} = {\color{drawred}\sum_{p^\prime=1}^{2k+1}}\sum_{i^\prime=1}^{2k+1}\sum_{j^\prime = 1}^{2k+1},\sum_{d=1}^{c} \idx{W}{{\color{drawred}p^\prime}, i^\prime, j^\prime,z,d}\idx{X}{{\color{drawred}p^\prime+t(p)},i^\prime+t(i),j^\prime+t(j),d}
$$
We assume that the kernel size is identical across all dimensions for simplicity. With similar reasonings we can derive a vectorized 3D variant of convolution, and also 1D and 3D variants of max pooling.
\section{1D and 3D convolutional models}

We now consider the design of convolutional models in the 1D case, with a focus on how to handle variable-length inputs and how to deal with text sequences. Several of the ideas we introduce are fairly generic for all differentiable models.

\subsection{Dealing with variable-length inputs}

Consider two audio files (or two time series, or two texts), described by their corresponding input matrices $\mathbf{X}_1 \sim (t_1, c)$ and $\mathbf{X}_2 \sim (t_2, c)$. The two inputs share the same number of channels $c$ (e.g., the number of sensors), but they have different lengths, $t_1$ and $t_2$. Remember from our discussion in Section \ref{sec:towards_convolutive_layers} that convolutions can handle (in principle) such \textbf{variable-length} inputs. In fact, denote by $g$ a generic composition of 1D convolutions and max-pooling operations, corresponding to the feature extraction part of the model. The output of the block are two matrices:

$$
\mathbf{H}_1=g(\mathbf{X}_1)\,,\,\mathbf{H}_2=g(\mathbf{X}_2)
$$

having the same number of columns but a different number of rows (depending on how many max-pooling operations or strided convolutions are applied on the inputs). After global average pooling, the dependence on the length disappears:

$$
\mathbf{h}_1=\sum_i\mathbf{H}_{1i} \,,\,\mathbf{h}_2=\sum_i\mathbf{H}_{2i}
$$

and we can proceed with a final classification on the vectors $\mathbf{h}_1$ and $\mathbf{h}_2$. However, while this is not a problem at the level of the model, it is a problem in practice, since mini-batches cannot be built from matrices of different dimensions, and thus operations cannot be easily vectorized. This can be handled by zero-padding the resulting mini-batch to the maximum dimension across the sequence length. Assuming for example, without lack of generality, $t_1 > t_2$, we can build a “padded” mini-batch as:

$$
X=\text{stack}\left(\mathbf{X}_1,\begin{bmatrix}\mathbf{X}_2\\ \mathbf{0}\end{bmatrix}\right)
$$

where $\text{stack}$ operates on a new leading dimension, and the resulting tensor $X$ has shape $(2, t_1,  c)$. We can generalize this to any mini-batch by considering the largest length with respect to all elements of the mini-batch. For a convolution, this is not very different from zero-padding, and operating on the padded input will not influence significantly the operation (e.g., in audio, zero-padding is equivalent to adding silence at the end).  See Box \ref{code:mini_batch_padding} for an example of building a padded mini-batch. 

\begin{mypy}{A padded mini-batch from three sequences of variable length (with $c=8$). When using a {\footnotesize\mintinline{python}{DataLoader}}, padding can be achieved by overwriting the default {\footnotesize\mintinline{python}{collate_fn}}, which describes how the loader concatenates the individual samples.}{code:mini_batch_padding}
# Sequences with variable length
# (3, 5, 2, respectively)
X1, X2, X3 = torch.randn(3, 8), 
             torch.randn(5, 8), 
             torch.randn(2, 8)

# Pad into a single mini-batch
X = torch.nn.utils.rnn.pad_sequence(
                    [X1, X2, X3], 
                    batch_first=True)
print(X.shape) 
# [Out]: torch.Size([3, 5, 8])
\end{mypy}

Alternatively, we can build a masking matrix describing valid and invalid indexes in the mini-batched tensor:

$$
\mathbf{M}=\begin{bmatrix} \mathbf{1}_{t_1} \\ \mathbf{1}_{t_2} \;\;\mathbf{0}_{t_1-t_2} \end{bmatrix}
$$

where the index denotes the size of the vectors. These masking matrices can be helpful to avoid invalid operations on the input tensor.

\subsection{CNNs for text data}

Let us consider now the problem of dealing with text data. As we mentioned previously, the first step in dealing with text is \textbf{tokenization}, in which we divide the text (a string) into a sequence of known symbols (also called \textbf{tokens} in this context). There are multiple types of tokenizers:
\begin{enumerate}
\item \textbf{Character tokenizer}: each character becomes a symbol.
\item \textbf{Word tokenizer}: each (allowed) word becomes a symbol.
\item \textbf{Subword tokenizer}: intermediate between a character tokenizer and a word tokenizer, each symbol is possibly larger than a character but also smaller than a word.
\end{enumerate}

\begin{figure}[t]
    \centering
    \includegraphics[width=0.8\textwidth]{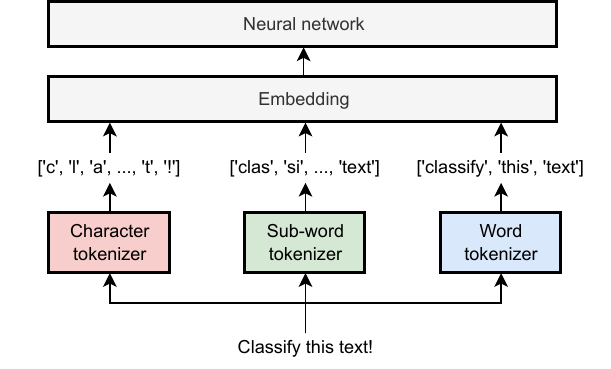}
    \caption{Starting from a text, multiple types of tokenizers are possible. In all cases, symbols are then embedded as vectors and processed by a generic 1D model.}
    \label{fig:text_tokenization}
\end{figure}

This is shown schematically in Figure \ref{fig:text_tokenization}. In all three cases, the user has to define a \textbf{dictionary} (\textbf{vocabulary}) of allowed tokens, such as all ASCII characters for a character tokenizer. In practice, one can select a desired size of the dictionary, and then look at the most frequent tokens in the text to fill it up, with every other symbol going into a special “out-of-vocabulary” (OOV) token. Subword tokenizers have many specialized algorithms to this end, such as byte-pair encoding (BPE) \cite{shibata1999byte}.\footnote{This is a short exposition focused on differentiable models, and we are ignoring many preprocessing operations that can be applied to text, such as removing stop words, punctuation, “stemming”, and so on. As the size of the models has grown, these operations have become less common.}

\begin{SCfigure}
    \centering
    \hspace{1em}\includegraphics[width=0.35\textwidth]{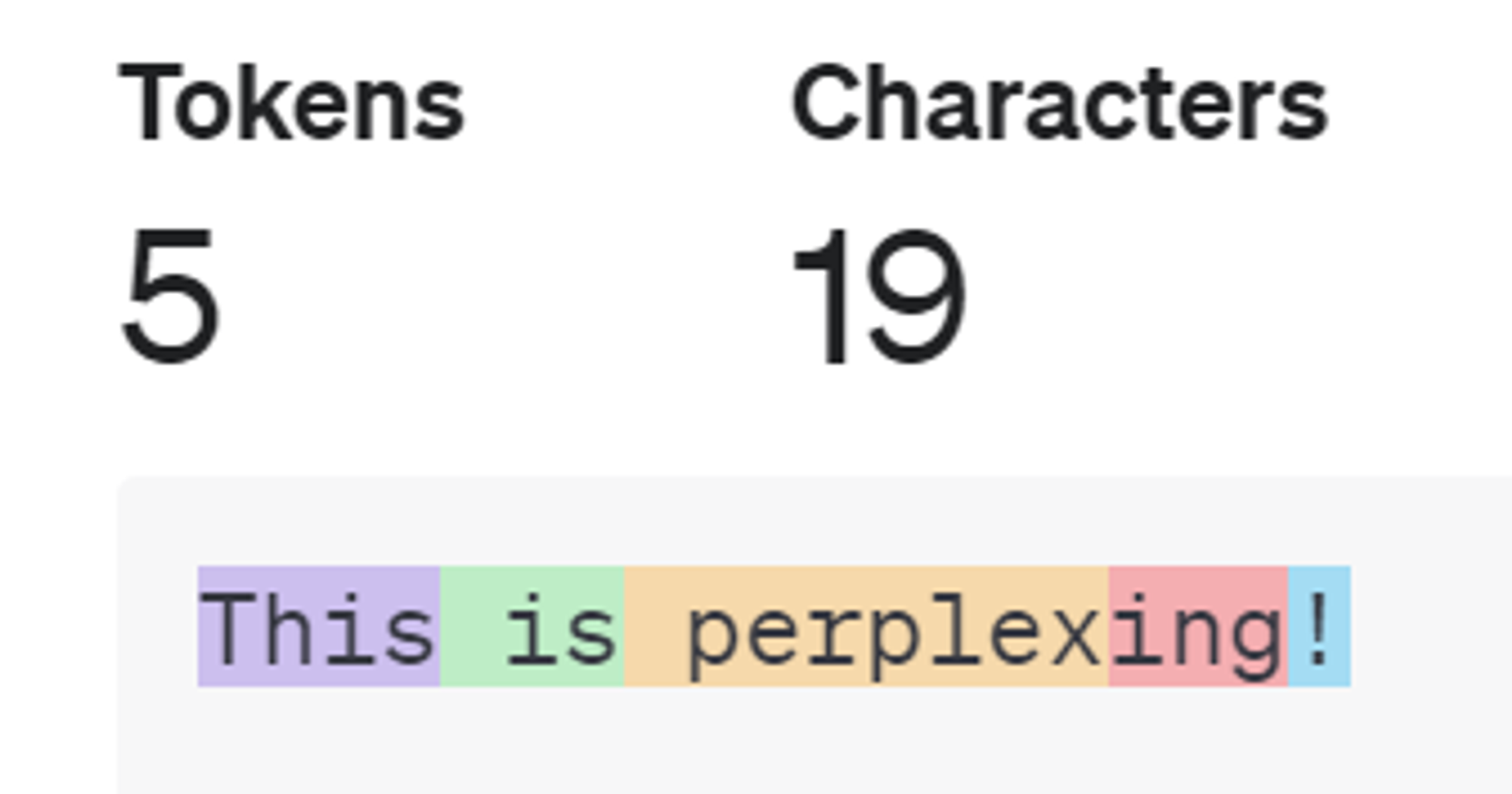}
    \caption{Example of applying the tiktoken tokenizer to a sentence.}
    \label{fig:tiktoken}
\end{SCfigure}

Because large collections of text can have a wide variability, pre-trained subword tokenizers are a standard choice nowadays.  As a concrete example, OpenAI has released an open-source version of its own tokenizer,\footnote{\url{https://github.com/openai/tiktoken}} which is a subword model consisting of approximately 100k subwords (at the time of writing). Consider for example the encoding of “\textit{This is perplexing!}” with this tokenizer, shown in Figure \ref{fig:tiktoken}. Some tokens correspond to entire words (e.g., “\textit{This}”), some to pieces of a word (e.g, “\textit{perplex}”), while others to punctuation marks. The sequence can be equivalently represented by a sequence of integers:
\begin{equation}
[2028, 374, 74252, 287, 0]
\label{eq:list_of_indices}
\end{equation}
Each integer spans between $0$ and the size of the vocabulary (in this case, roughly 100k), and it uniquely identifies the token with respect to that vocabulary. In practice, nothing prevents us from adding “special” tokens to the sequence, such as tokens representing the beginning of the sentence (sometimes denoted as [BOS]), OOV tokens, or anything else. The [BOS] token will be of special significance in the next section.

Subword tokenization with very large dictionaries can be counter-intuitive at times: for example, common digits such as $52$ have their unique token, while digits like $2512$ can be split into a “251” token and a “2” token, so that visualizing the tokenization process is always important to debug the models' behaviour.\footnote{For applications where processing numbers is important, specialized numerical tokenizers can be applied \cite{golkar2023xval}.} Given the importance of the tokenization step, this is a very active research topic -- we mention here, for example, byte-level tokenizers \cite{pagnoni2024byte} and tokenizers that can be trained end-to-end  \cite{hwang2025dynamic}.

After the tokenization step, the tokens must be \textbf{embedded} into vectors to be used as inputs for a CNN. A simple one-hot encoding strategy here works poorly, since vocabularies are large and the resulting vectors would be significantly sparse. Instead, we have two alternative strategies: the first is to use \textit{pretrained} networks that perform the embedding for us; we will consider this option later on, when we introduce transformers. In order to build some intuition for it, we consider here the second alternative, \textit{training} the embeddings together with the rest of the network.

Suppose we fix an embedding dimension $e$ as a hyper-parameter. Since the size $n$ of the dictionary is also fixed, we can initialize a matrix of embeddings $\mathbf{E} \sim (n, e)$. We now define a look-up operation that replaces each integer with the corresponding row in $\mathbf{E}$. Denoting by $x$ the sequence of IDs we have:

\vspace*{1em}
$$
\text{LookUp}(x) =\mathbf{X} = \begin{bmatrix}   \eqnmarkbox[drawred]{node}{\mathbf{E}_{x_1}} \\\mathbf{E}_{x_2} \\ \vdots  \\\mathbf{E}_{x_{m}}\end{bmatrix}
$$
\annotate[yshift=1em]{above,left}{node}{Row $x_1$ in the embedding matrix}

\begin{figure}[t]
    \centering
    \hspace{1em}\includegraphics[width=0.8\textwidth]{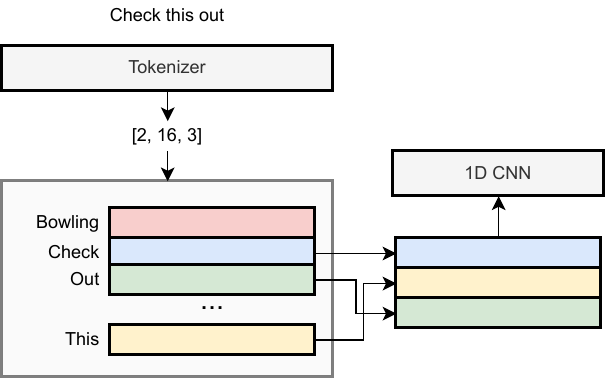}
    \caption{A lookup table to convert a sequence of tokens' IDs to their curresponding embeddings: the input is a list, the output is a matrix. The embeddings (shown inside the box) can be trained together with all the other parameters via gradient descent. We assume the size of the vocabulary is $n=16$.}
    \label{fig:custom_embeddings}
\end{figure}

\begin{mypy}{A 1D CNN with trainable embeddings. $n$ is the size of the dictionary, $e$ is the size of each embedding. We use two convolutional layers with $32$ and $64$ output channels. The shape of the output for each operation in the forward pass is shown as a comment.}{code:custom_embeddings}
class TextCNN(nn.Module):
  def __init__(self, n, e):
    super().__init__()
    self.emb = nn.Embedding(n, e)
    self.conv1 = nn.Conv1d(e, 32, 5, 
                        padding='same')
    self.conv2 = nn.Conv1d(32, 64, 5, 
                        padding='same')
    self.head = nn.Linear(64, 10)

  def forward(self, x):      # (*, m)
    x = self.emb(x)          # (*, m, e)
    x = x.transpose(1, 2)    # (*, e, m)
    x = relu(self.conv1(x))  # (*, 32, m)
    x = max_pool1d(x, 2)     # (*, 32, m/2)
    x = relu(self.conv2(x))  # (*, 64, m/2)
    x = x.mean(2)            # (*, 64)
    return self.head(x)      # (*, 10)
\end{mypy}

The resulting input matrix $\mathbf{X}$ will have shape $(m, e)$, where $m$ is the length of the sequence. We can now apply a generic 1D convolutional model for, e.g., classifying the text sequence:
$$
\hat{y}=\text{CNN}(\mathbf{X})
$$
This model can be trained in a standard way depending on the task, except that gradient descent will be performed jointly on the parameters of the model and the embedding matrix $\mathbf{E}$. This is shown visually in Figure \ref{fig:custom_embeddings}, and an example of model's definition is given in Box \ref{code:custom_embeddings}.

This idea is extremely powerful, especially because in many cases we find that the resulting embeddings can be manipulated algebraically as vectors, e.g., by looking at the closest embeddings in an Euclidean sense to find “semantically similar” words or sentences. This idea is at the core of the use of differentiable models in many sectors that necessitate retrieval or search of documents. \clearpage

\vspace{1em}
\begin{supportbox}{Differentiable models and embeddings}
Once again, the idea of embedding is very general: any procedure that converts an object into a vector with algebraic characteristics is an embedding. For example, the output of the backbone of a trained CNN after global pooling can be understood as a high-level embedding of the input image, and it can be used to retrieve “similar” images by comparing it to all other embeddings.
\end{supportbox}

\subsection{Dealing with long sequences}

Many of the sequences described before can be very long. In this case, the locality of convolutional layers can be a drawback, because we need a linearly increasing number of layers to process larger and larger receptive fields. We will see in the next chapters that other classes of models (e.g., transformers) can be designed to solve this problem. For now we remain in the realm of convolutions and we show one interesting solution, called \textbf{dilated} (or \textbf{atrous}, from the French \textit{à trous}) convolutions, popularized in the WaveNet model for speech generation \cite{oord2016wavenet}.

\begin{figure}[t]
    \centering
    \includegraphics[width=0.8\textwidth]{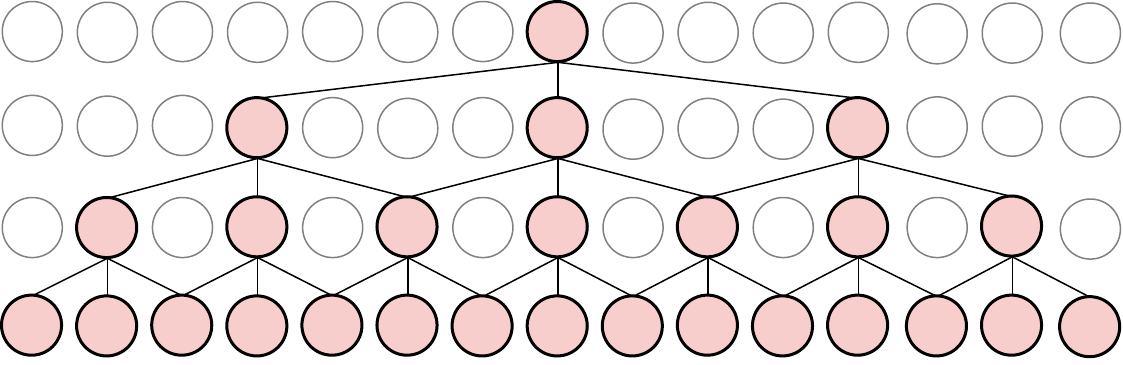}
    \caption{Convolutional layers with increasing dilation rates. Elements selected for the convolution are in red, the others are greyed out. We show the receptive field for a single output element.}
    \label{fig:convolution_with_dilation}
\end{figure}

We introduce an additional hyper-parameter called the \textbf{dilation rate}. A convolution with dilation rate of $1$ is a standard convolution. For a dilation rate of $2$, we modify the convolution operation to select elements for our patch by skipping one out of two elements in the sequence. Similarly, for a dilation rate of $4$, we skip three elements over four, etc. We stack convolutional layers with exponentially increasing dilation rates, as shown in Figure \ref{fig:convolution_with_dilation}.
The number of parameters does not change, since the number of neighbors remains constant irrespective of the dilation rate. However, it is easy to show that the resulting receptive field in this case grows \textit{exponentially fast} in the number of layers.

\section{Forecasting and causal models}
\subsection{Forecasting sequences}

One important aspect of working with sequences is that we can build a model to predict future elements, e.g., energy prices, turbulence flows, call center occupations, etc. Predicting tokens is also the fundamental building block for large language models and other recent breakthroughs. In a very broad sense, much of the current excitement around neural networks revolves around the question of how much a model can be expected to infer from next-token prediction on large corpora of text, and how much this setup can be replicated across different modalities (e.g., videos) and dynamics \cite{wang2023scientific}. Formally, predicting the next element of a sequence is called \textbf{forecasting} in statistics and time series analysis. From now on, to be consistent with modern literature, we will use the generic term \textbf{token} to refer to each element of the sequence, irrespective of whether we are dealing with an embedded text token or a generic vector-valued input.

The reason forecasting is an important problem is that we can train a forecasting model by just having access to a set of sequences, with no need for additional target labels: in modern terms, this is also called a \textbf{self-supervised learning} task, since the targets can be automatically extracted from the inputs. 

\begin{supportbox}{Stationarity and forecasting}
Just like text processing, forecasting real-world time series has a number of associated problems (e.g., the possible non-stationarity of the time series, trends and seasonalities) that we do not consider here.\footnote{\url{https://filippomb.github.io/python-time-series-handbook/}} In practice, audio, text, and many other sequences of interest can be considered stationary and do not need special preprocessing. Like for text, for very large forecasting datasets and correspondingly large models, the impact of preprocessing tend to diminish \cite{ansari2024chronos}.
\end{supportbox}

To this end, suppose we fix a user-defined length $t$, and we extract all possible subsequences of length $t$ from the dataset (e.g., with $t=12$, all consecutive windows of $12$ elements, or all sentences composed of $12$ tokens, etc.). In the context of LLMs, the size of the input sequence is called the \textbf{context} of the model. We associate to each subsequence a target value which is the next element in the sequence itself. Thus, we build a set of pairs $(\mathbf{X}, \mathbf{y}), \mathbf{X} \sim (t, c) \,,\, y \sim (c)$ and our forecasting model is trained in a supervised way over this dataset:
$$
f(\mathbf{X})\approx \mathbf{y}
$$
Note that a standard 1D convolutional model can be used as forecasting model, trained with either mean-squared error (for continuous time series) or cross-entropy (for categorical sequences, such as text). While the model is trained to predict a single step-ahead, we can easily use it to generate as many steps as we want by what is called an \textbf{autoregressive} approach, meaning that the model is predicting (\textit{regressing}) on its own outputs. Suppose we predict a single step, $\widehat{\mathbf{y}} = f(\mathbf{X})$, and we create a “shifted” input by adding our predicted value to the input (removing the first element to avoid exceeding $t$ elements):

\vspace{1em}
\begin{equation}
\mathbf{X}^\prime = \begin{bmatrix} \eqnmarkbox[drawred]{node}{\mathbf{X}_{2:t}} \\ \eqnmarkbox[drawgreen]{node2}{\widehat{\mathbf{y}}} \end{bmatrix}
\label{eq:shifted_input}
\end{equation}
\annotate[yshift=1em]{above,left}{node}{Window of $t-1$ input elements}
\annotate[yshift=-1em]{below,left}{node2}{Predicted value at time $t+1$}

\vspace{1.5em}
\begin{supportbox}{Forecasting discrete sequences}
For a continuous time series this is trivial. For a time series with discrete values, $f$ will return a probability vector over the possible values (i.e., possible tokens), and we can obtain $\widehat{\mathbf{y}}$ by taking its $\arg\max$, i.e., the token associated to the highest probability. Alternatively, we can sample a token proportionally to the predicted probabilities: see Section \ref{subsec:probabilistic_formulation_generative_models}.
\end{supportbox}

\vspace{0.5em}
We can now run $f(\mathbf{X}^\prime)$ to generate the next input value in the sequence, and so on iteratively, by always updating our buffered input in a FIFO fashion. This approach is extremely powerful, but it requires us to fix a priori the input sequence length, which limits its applicability. To overcome this limitation, we need only a minor modification to our models.

\clearpage

\subsection{Causal models} \addclock

Suppose we only have available a short sequence of 4 elements collected into a matrix $\mathbf{X} \sim (4, c)$, but we have trained a forecasting model on longer sequences with $t=6$. In order to run the model on the shorter sequence, we can zero-pad the sequence with two zero vectors $\mathbf{0}$ at the beginning, but these will be interpreted by the model as actual values of the time series unless we mask its operations. Luckily, there is a simpler and more elegant approach in the form of \textbf{causal} models.

\begin{definition}[Causal layer] \addbottle $\,$

A layer $\mathbf{H} = f(\mathbf{X})$ is \textbf{causal} if $\mathbf{H}_i = f(\mathbf{X}_{:i})$, i.e., the output value corresponding to the $i$-th element of the sequence depends only on elements “from its past”.
\end{definition}

\begin{figure}[t]
    \centering
    \hspace{1em}\includegraphics[width=0.7\textwidth]{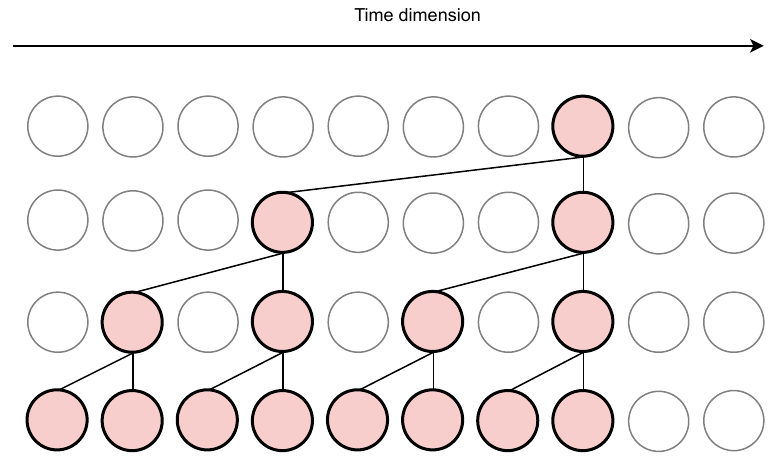}
    \caption{Overview of a 1D \textit{causal} convolutional layer with (original) kernel size of $3$ and exponentially increasing dilation rates. Zeroed out connections are removed, and we show the receptive field for a single output element.}
    \label{fig:causal_convolutions}
\end{figure}

A model composed only of causal layers will, of course, be causal itself. For example, a convolutional layer with kernel size $1$ is causal, since each element is processed considering only itself. However, a convolutional layer with kernel size $3$ is not causal, since it is processed considering in addition one element to the left and one element to the right. We can convert any convolution into a causal variant by partially zero masking the weights corresponding to non-causal connections:
$$
\mathbf{h}_i=\phi\left(\left[\eqnmarkbox[drawred]{node}{\mathbf{W}\odot \mathbf{M}}\right]\text{vect}(P_{k}(i)) + \mathbf{b}\right)
$$
\annotate[yshift=-1em]{below,right}{node}{Masked weight matrix}

where $M_{ij} = 0$ if the weight corresponds to an element in the input such that $j > i$, $1$ otherwise. Causal 1D convolutions can be combined with dilated kernels to obtain autoregressive models for audio, such as in the WaveNet model \cite{oord2016wavenet} - see Figure \ref{fig:causal_convolutions} for an example.

Masking is easier to understand in the case of a single channel, in which case $\mathbf{M}$ is simply a lower-triangular binary matrix. The masking operation effectively reduces the number of parameters from $(2k+1)cc^\prime$ to $(k+1)cc^\prime$. 

By stacking several causal convolutional layers, we can obtain a causal 1D model variant. Suppose we apply it on our input sequence, with a model that has no max-pooling operations. In this case, the output sequence has the same length as the input sequence:
$$
\widehat{\mathbf{Y}} = f_{\text{causal}}(\mathbf{X})
$$
In addition, any element in the output only depends on input elements in the same position or preceding it. Hence, we can define a more sophisticated forecasting model by predicting a value \textit{for each element of the input sequence}. Practically, consider now a matrix output defined as:
$$
\mathbf{Y} = \begin{bmatrix} \mathbf{X}_{2:t} \\\mathbf{y} \end{bmatrix}
$$

\begin{figure}[t]
    \centering
    \begin{subfigure}[b]{0.45\textwidth}
    \includegraphics[width=1.0\textwidth]{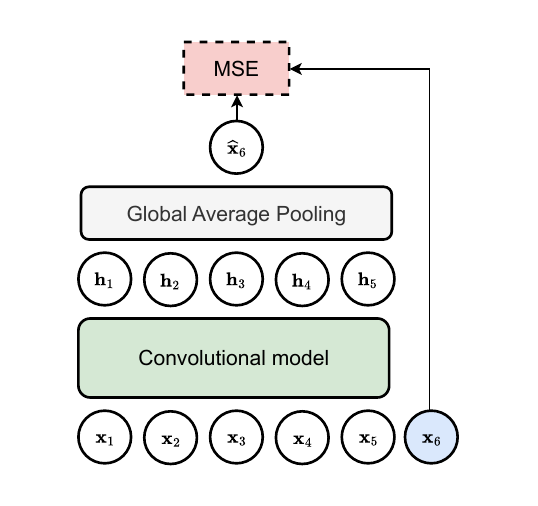}
    \caption{Non-causal model}
    \label{fig:forecasting_a}
    \end{subfigure}
    \begin{subfigure}[b]{0.48\textwidth}
    \includegraphics[width=1.0\textwidth]{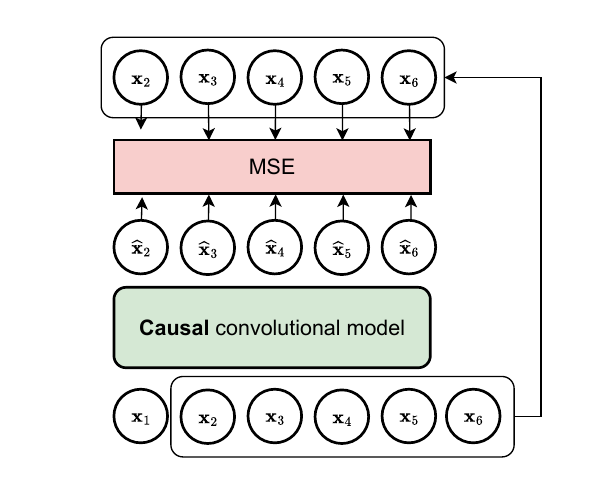}
    \caption{Causal model}
    \label{fig:forecasting_b}
    \end{subfigure}
    \caption{Comparison between (a) a non-causal model for forecasting (predicting only a single element for the entire input sequence) and (b) a causal model trained to predict one output element for each input element in the sequence.}
    \label{fig:forecasting}
\end{figure}

This is similar to the shifted input from \eqref{eq:shifted_input}, except that we are adding the true value as last element of the sequence. We can train this model by minimizing a loss on all elements, e.g., a mean-squared error:
\begin{equation}
l(\widehat{\mathbf{Y}},\mathbf{Y})=\lVert \widehat{\mathbf{Y}} - \mathbf{Y} \rVert^2=\sum_{i=1}^t \eqnmarkbox[drawred]{node}{\lVert \widehat{\mathbf{Y}}_i - \mathbf{Y}_i \rVert^2}
\label{eq:mse_forecasting_multiple}
\end{equation}
\annotate[yshift=-1em]{below,left}{node}{Loss when predicting $\mathbf{X}_{i+1}$}

We simultaneously predict the second element based on the first one, the third one based on the first two, etc. For a single input window, we have $t$ separate loss terms, greatly enhancing the gradient propagation. A comparison between the two approaches is shown in Figure \ref{fig:forecasting}: in Figure \ref{fig:forecasting_a} we show a non-causal convolutional model trained to predict the next element in the sequence, while in Figure \ref{fig:forecasting_b} we show a causal model trained according to \eqref{eq:mse_forecasting_multiple}.

\begin{figure}
    \centering
    \hspace*{-2em}\includegraphics[width=1.1\textwidth]{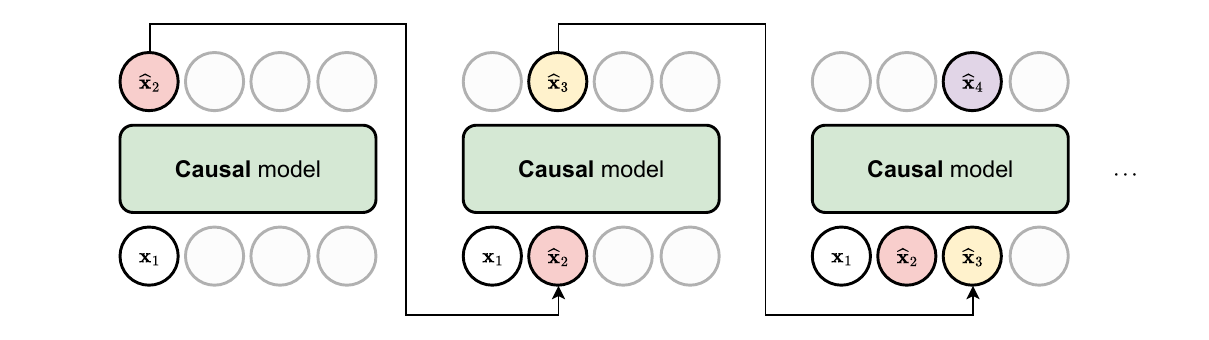}
    \caption{Inference with a causal CNN, generating a sequence step-by-step in an autoregressive way. Unused input tokens are greyed out. Generated tokens are shown with different colors to distinguish them.}
    \label{fig:forecasting_autoregressive}
\end{figure}

More importantly, we can now use the model in an autoregressive way with any sequence length up to the maximum length of $t$. This can be seen easily with an example. Suppose we have $t=4$, and we have observed two values $\mathbf{x}_1$ and $\mathbf{x}_2$. We call the model a first time by zero-padding the sequence to generate the third token:
$$
\begin{bmatrix} - \\  {\color{drawred}\widehat{\mathbf{x}}_3} \\ - \\ - \end{bmatrix} =f\left( \begin{bmatrix} \mathbf{x}_1 \\ \mathbf{x}_2 \\ \mathbf{0} \\ \mathbf{0} \end{bmatrix} \right)
$$
We are ignoring all output values except the second one (in fact, the third and fourth outputs are invalid due to the zero-padding). We add $\widehat{\mathbf{x}}_3$ to the sequence and continue calling the model autoregressively (we show in color the predicted values): \clearpage

{
\footnotesize
$$
\begin{bmatrix} - \\  - \\ {\color{drawgreen}\widehat{\mathbf{x}}_4} \\ - \end{bmatrix} =f\left( \begin{bmatrix} \mathbf{x}_1 \\ \mathbf{x}_2 \\ {\color{drawred}\widehat{\mathbf{x}}_3} \\ \mathbf{0} \end{bmatrix} \right) \;,\;\begin{bmatrix} - \\  - \\ - \\ {\color{drawblue}\widehat{\mathbf{x}}_5} \end{bmatrix} =f\left( \begin{bmatrix} \mathbf{x}_1 \\ \mathbf{x}_2 \\ {\color{drawred}\widehat{\mathbf{x}}_3} \\ {\color{drawgreen}\widehat{\mathbf{x}}_4} \end{bmatrix} \right) \;,\; \begin{bmatrix}  - \\ - \\ - \\{\color{orange}\widehat{\mathbf{x}}_6}  \end{bmatrix} =f\left( \begin{bmatrix} \mathbf{x}_2 \\ {\color{drawred}\widehat{\mathbf{x}}_3} \\ {\color{drawgreen}\widehat{\mathbf{x}}_4} \\ {\color{drawblue}\widehat{\mathbf{x}}_5} \end{bmatrix}  \right) \; \ldots
$$
}

In the last step we removed one of the original inputs to keep the constraint on the size of the input. This is also shown in Figure \ref{fig:forecasting_autoregressive}. Note that the model is trained only on real values, not on its own predictions: this is called \textbf{teacher forcing}. A variant of teacher forcing is to progressively replace some of the values in the mini-batches with values predicted by the model, as training proceeds and the model becomes more accurate. 

Causal autoregressive models are especially interesting in the case of text sequences (where we only have a single channel, the index of the tokens), since we can start from a single [BOS] token representing the beginning of the sequence and generate text sentences from scratch, or \textit{condition} the generation on a specific prompt by the user which is appended to the [BOS] token. A similar reasoning can be applied to audio models to generate speech or music \cite{oord2016wavenet}.

\section{Generative models}
\subsection{A probabilistic formulation}
\label{subsec:probabilistic_formulation_generative_models} \addteacup

An autoregressive model is a simple example of a \textbf{generative model}.\footnote{Remember from Chapter \ref{chap:supervised_learning} that we assume our supervised pairs $(x,y)$ come from some unknown probability distribution $p(x,y)$. By the product rule of probability we can decompose it equivalently as $p(y \mid x)p(x)$, or $p(x \mid y)p(y)$. Any model which approximates $p(x)$ or $p(x \mid y)$ is called \textbf{generative}, because you can use it to sample new input points. By contrast, a model that only approximates $p(y \mid x)$, like we did in the previous chapters, is called \textbf{discriminative}.} We will talk at length about other types of generative models in the next volume. For now, we provide some insights specific to autoregressive algorithms. We consider sequences with a single channel and discrete values, such as text. Autoregressive models over text tokens are the foundation of LLMs, and they can be used as the basis for multimodal architectures (Chapter \ref{chap:transformers_in_practice}).

Generative models are more naturally framed in the context of probabilities, so we begin by reframing our previous discussion with a probabilistic formalism. Denote by $\mathcal{X}$ the space of all possible sequences (e.g., all possible combinations of text tokens). In general, many of these sequences will be invalid, such as the sequence [“\textit{tt}”, “\textit{tt}”] in English. However, even very uncommon sequences may appear at least once or twice in very large corpora of text (imagine a character yelling “\textit{Scotttt!}”). 

We can generalize this by considering a probability distribution $p(x)$ over all possible sequences $x \in \mathcal{X}$. In the context of text, this is also called a \textbf{language model}. Generative modeling is the task of learning to sample efficiently from this distribution:\footnote{In this section $\sim$ is used to denote sampling from a probability distribution instead of the shape of a tensor.}
$$
x \sim p(x)
$$
To see how this connects to our previous discussion, note that by the product rule of probability we can always rewrite $p(x)$ as:
\begin{equation}
p(x)=\prod_ip(x_i \mid x_{:i})
\label{eq:prob_autoregressive}
\end{equation}
where we condition each value $x_i$ to all preceding values. If we assume that our model input length is large enough to accommodate all possible sequences, we can use a causal forecasting model to parameterize the probability distribution in \eqref{eq:prob_autoregressive}:
$$
p(x_i \mid x_{:i})=\text{Categorical}(x_i\mid f(x_{:i}))
$$
where we use a single, shared model for all time-steps. Maximum likelihood over this model is then equivalent to minimizing a cross-entropy loss over the predicted probabilities, as in Section \ref{subsec:logistic_regression}.

\subsection{Sampling in an autoregressive model}

In general, sampling from a probability distribution is non-trivial. However, for autoregressive models we can exploit the product decomposition in \eqref{eq:prob_autoregressive} to devise a simple iterative strategy:
\begin{enumerate}
\item Sample $x_1 \sim p(x_1)$. This is equivalent to conditioning on the empty set $p(x_1 \mid \{\})$. In practice, we always condition on an initial fixed token, such as the [BOS] token, so that our input is never empty.
\item Sample $x_2 \sim p(x_2 \mid x_1)$ by running again the network with the value we sampled at step (1), as in Figure \ref{fig:forecasting_autoregressive}.
\item Sample $x_3 \sim p(x_3 \mid x_1, x_2)$.
\item Continue until we reach a desired sequence length or until we get to an end-of-sentence token.
\end{enumerate}

We did this implicitly before by always sampling the element of highest probability:
$$
x_i=\underset{i}{\arg\max} \;f(x_{:i})
$$
However, we can also generalize this by sampling a value according to the probabilities predicted by $f$. Remember (Section \ref{sec:softmax}) that the softmax can be generalized by considering an additional temperature parameter. By varying this parameter during inference, we can vary smoothly between always taking the argmax value (very low temperature) to having an almost uniform distribution over tokens (very high temperature).

In the context of probabilistic modeling, sampling in this way from this class of models is called \textbf{ancestral sampling}, while in the context of language modeling we sometimes use the term \textbf{greedy decoding}. The use of the term “greedy” and this brief discussion is enough to highlight one potential drawback of this approach: while the product decomposition of $p(x)$ is exact, greedy decoding is not guaranteed to provide a sample corresponding to high values of $p(x)$. 

To see this, note that $f$ provides an estimate of the probability for a single token, but the probability of a sequence is given by a product of many such terms. Hence, sampling a token with high (local) probability at the beginning of a sequence may not correspond to a sequence having large (global) probability as a sentence. This is easy to visualize if you imagine the choice of the first token letting the decoding stage being “stuck” in a low-probability path.

A common mitigation to this problem is \textbf{beam search} (or \textbf{beam decoding}). In beam search, in the first step we sample $k$ different elements (called the beams, with $k$ being a user-defined parameter). In the second step, for each of our $k$ beams we sample $k$ possible continuations. Out of these $k^2$ pairs, we keep only the top-$k$ values in terms of their product probability $p(x_1)p(x_2\mid x_1)$ (or, equivalently, their log probability). We continue iteratively in this way until the end of the sequence.

Viewed under this lens, sampling the most probable sequence from our autoregressive model is a combinatorial search problem (think of a tree, where for each token we expand across all possible next tokens, and so on). From the point of view of computer programming, beam search is then an example of \textbf{breadth-first} search over this tree. In a sense, beam search is trading off a simple training procedure for a more expensive inference stage -- many other techniques exist to this end, including the possibility of guiding the decoding to satisfy an external reward function \cite{welleck2024decoding}.

\subsection{Conditional modeling}
\label{subsec:conditional_modeling}

As we mentioned earlier, in general we may not be interested so much in generating sequences from scratch, but in generating continuations of known sequences, such as a user’s question or interaction. This can be formalized by considering \textit{conditional} probability distributions in the form $p(x \mid c)$, where $c$ is the conditioning argument, such as a user’s prompt. Our previous discussion extends almost straightforwardly to this case. For example, the product decomposition is now written as:
$$
p(x \mid c)=\prod_ip(x_i \mid x_{:i},c)
$$
where we condition on the previous inputs \textit{and} the user’s context. Sampling and decoding are extended in a similar way. 

To perform conditional generation we parameterize $p(x_i \mid x_{:i},c)$ with a neural network $f(x,c)$ such that:
$$
p(x_i \mid x_{:i},c) \approx \text{Categorical}(x_i \mid f(x_{:i}, c))
$$ 
Hence, the major difference with the unconditional case is that we need a function $f(x,c)$ having two input arguments and which satisfies causality in the first argument. When working with autoregressive models, if both $x$ and $c$ are texts we can do this easily be considering $c$ as part of the input sequence and working with a single concatenated input $x^\prime = [c \Vert x]$. For example, with the user's prompt “\textit{The capital of France}”, taking for simplicity a word tokenizer we might have:\footnote{We ignore the presence of an end-of-sequence token (EOS) to stop the autoregressive generation.}
\begin{gather*}
f_{\text{causal}}([\text{The}, \text{capital}, \text{of}, \text{France}]) = {\color{drawred}\text{is}} \\
f_{\text{causal}}([\text{The}, \text{capital}, \text{of}, \text{France}, {\color{drawred}\text{is}}]) = {\color{drawgreen}\text{Paris}} \\
\end{gather*}
Hence, we can handle unconditional and conditional modeling simultaneously with a single model.\footnote{We will see in Chapter \ref{chap:transformers_in_practice} that almost any type of data can be converted into a sequence of tokens. Suppose we are generating a text sequence conditioned on an image prompt (e.g., \textbf{image captioning}). If both text and images are converted to tokens having the same embedding size, we can apply an autoregressive model by concatenating the tokens from the two input types (also called \textbf{modalities} in this context), where we view the image tokens as the conditioning set $c$.} In the next volume we will see other examples of conditional generative models in which more sophisticated strategies are needed. We will also extend upon this topic when we discuss decoder-only transformer models in Chapter \ref{chap:transformers_in_practice}. \clearpage

\section*{From theory to practice}

\begin{wrapfigure}{r}{3.0cm}
\vspace{-3em}\includegraphics[width=3.0cm]{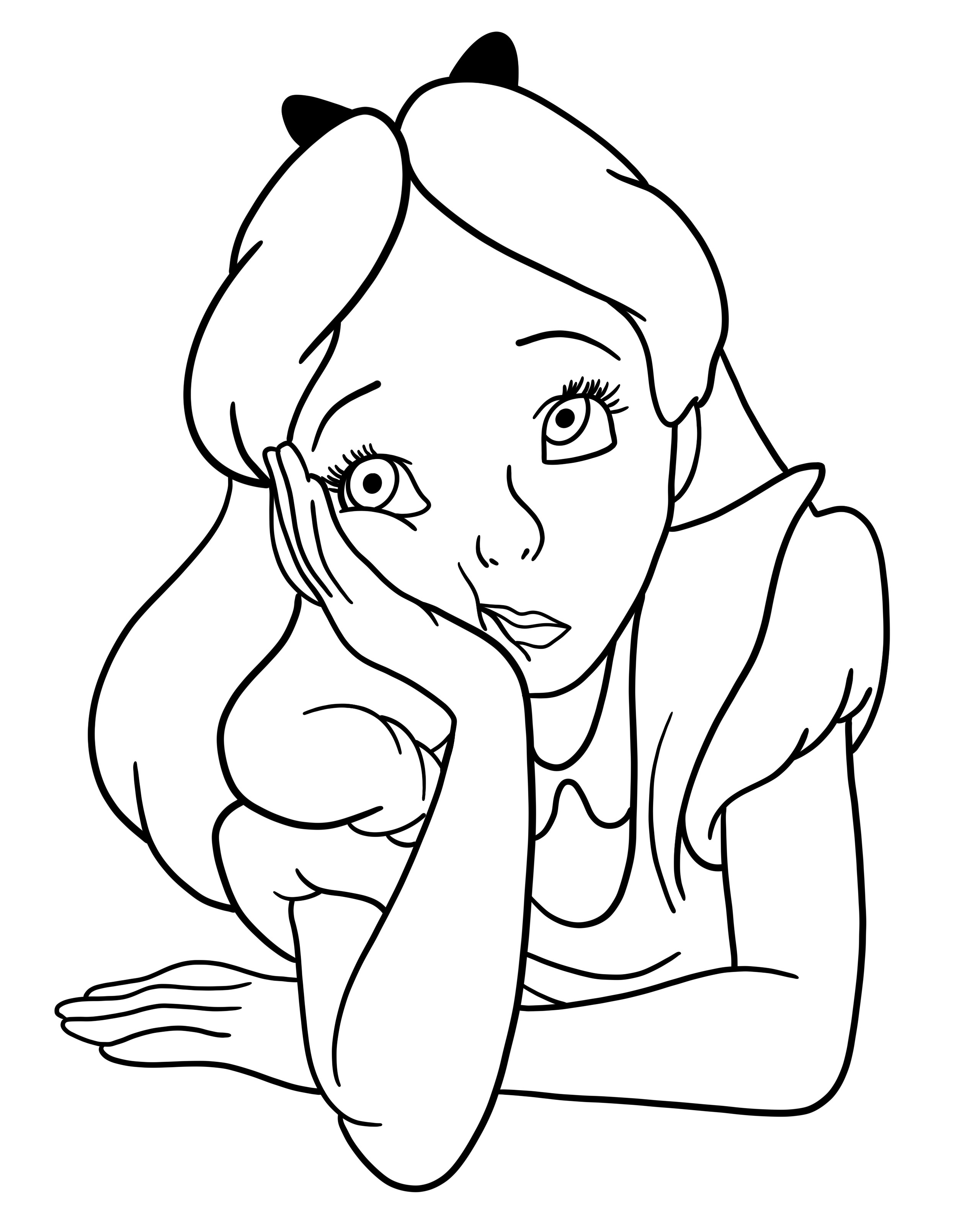}
\vspace{-2em}
\end{wrapfigure}

Working with text data is more complex than image classification, due to many subtleties involved with tokenization, data formatting, weird characters, and variable-length sequences. PyTorch has its own text library, {\footnotesize\texttt{torchtext}}, which at the time of writing is less documented than the main library and relies on another beta library ({\footnotesize\texttt{torchdata}}) to handle the data pipelines. Thus, we ignore it here, but we invite you to check it out on your own.

Hugging Face Datasets is probably the most versatile tool in this case, as it provides a vast array of datasets and pre-trained tokenizers, which can be exported immediately to PyTorch.\footnote{See this tutorial for a guide: \url{https://huggingface.co/docs/datasets/use_dataset}.} Familiarize yourself a bit with the library before proceeding with the exercise.

\begin{enumerate}
\item Choose a text classification dataset, such as the classic IMDB dataset.\footnote{\url{https://huggingface.co/datasets/stanfordnlp/imdb}} 
\item Tokenize it to obtain a dataset of the form $(x,y)$, where $x$ is a list of integers as in \eqref{eq:list_of_indices} and $y$ is the text label.
\item Build and train a 1D CNN model similar to Box \ref{code:custom_embeddings}. Experiment a bit with the model's design to see its impact on the final accuracy.
\end{enumerate}

PyTorch does not have a quick way to make a 1D convolution causal, so we will postpone our autoregressive experiments for when we introduce transformers.\footnote{If you want to try, you can emulate a causal convolution with proper padding; see Lecture 10.2 here: \url{https://fleuret.org/dlc/}. The entire course is really good if you are looking for streamed lectures.} Training your own tokenizer is a very good didactic exercise, although it is far beyond the scope of the book. For an introduction, you can check this minimalistic BPE implementation: \url{https://github.com/karpathy/minbpe}.

%% file: 9_building_deep_cnns.tex
\chapter[Scaling up the models]{Scaling up \\ the models}
\label{chap:deep_cnns}

\begin{supportbox}{About this chapter}
We now turn to the task of designing differentiable models having dozens (or hundreds) of layers. As we saw, the receptive field of convolutional models grows linearly with the number of layers, motivating architectures with such depth. This can be done by properly stabilizing training using a plethora of methods, ranging from data augmentation to normalization of the hidden states.
\end{supportbox}

\section{The ImageNet challenge}

Let us consider again the task of image classification, which holds a strong interest for neural networks, both practically and historically. In fact, interest in these models in the period 2012-2018 can be associated in large part to the \textbf{ImageNet Large Scale Visual Recognition Challenge}\footnote{\url{https://image-net.org/challenges/LSVRC/}} (later \textbf{ImageNet} for simplicity). ImageNet was a yearly challenge that run from 2010 to 2017 to evaluate state-of-the-art models for image classification. The challenge was run on a subset of the entire ImageNet dataset, consisting of approximately 1M images tagged across 1k classes. 

It is instructive to take a look at the early editions of the challenges. In 2010\footnote{\url{https://image-net.org/challenges/LSVRC/2010/}} and in 2011,\footnote{\url{https://image-net.org/challenges/LSVRC/2011/}} the winners were linear kernels methods built with a combination of specialized image descriptors and kernels, with a top-5\% error of 28\% (2010) and 26\% (2011). Despite a number of promising results,\footnote{\url{https://people.idsia.ch/~juergen/computer-vision-contests-won-by-gpu-cnns.html}} convolutional models trained by gradient descent remained a niche topic in computer vision. Then, In 2012 the winner model (AlexNet, \cite{krizhevsky2012imagenet}) achieved a top-5\% error of 15.3\%, 10\% lower than all (non-neural) competitors. 

\begin{SCfigure}
    \centering
    \includegraphics[width=0.6\textwidth]{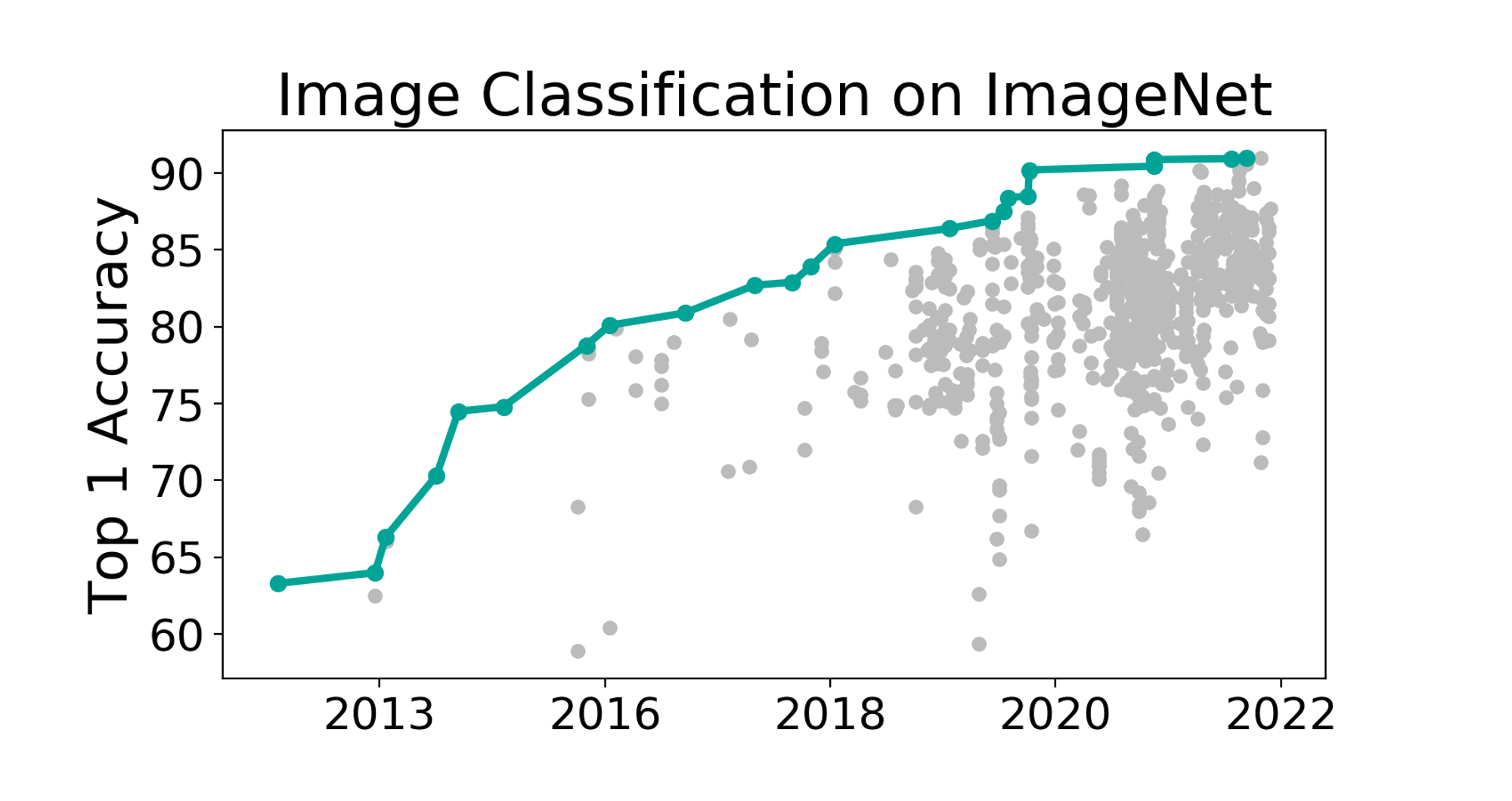}
    \caption[Reproduced from Papers With Code.]{Top-$1$ accuracy on the ImageNet dataset. Reproduced from Papers With Code.}
    \label{fig:papers_with_code}
\end{SCfigure}

This was followed by a veritable “Copernican revolution” (apologies to Copernicus) in the field, since in a matter of a few years almost all submissions turned to convolutional models, and the overall accuracy grew at an unprecedented speed, upward of 95\% (leading to the end of the challenge in 2017), as shown in Figure \ref{fig:papers_with_code}. In a span of 5 years, convolutional models trained with gradient descent became the leading paradigm in computer vision, including other subfields we are not mentioning here, from object detection to semantic segmentation and depth estimation.

AlexNet was a relatively simple model consisting of 5 convolutional layers and 3 fully-connected layers, totaling approximately 60M parameters, while the top-performing models in Figure \ref{fig:papers_with_code} require up to hundreds of layers. This is basic example of a scaling law (Chapter \ref{chap:introduction}): adding layers and compute power for training is proportionally linked to the accuracy of the model up to a saturation point given by the dataset. However, scaling up convolutional models beyond a few layers is non-trivial, as it runs into a number of problems ranging from slow optimization to gradient issues and numerical instabilities. As a consequence, a large array of techniques were developed in 2012-2017 to stabilize training of very large models.

In this chapter we provide an overview of some of these techniques. We focus on ideas and methods that are still fundamental nowadays, even for other architectures (e.g., transformers). We begin by three techniques to improve training that are well-known in machine learning: \textbf{weight regularization}, \textbf{data augmentation}, and \textbf{early stopping}. Then, we describe three of the most influential techniques popularized in 2012-2017: \textbf{dropout}, \textbf{batch normalization}, and \textbf{residual connections}, more or less in chronological order of introduction. For each method we describe the basic algorithm along with some variants that work well in practice (e.g., layer normalization).

\clearpage

\section{Data and training strategies}

\subsection{Weight regularization}

One possible way to improve training is to penalize solutions that may seem unplausible, such as having one or two extremely large weights. Denote by $\mathbf{w}$ the vector of all parameters of our model, and by $L(\mathbf{w}, \mathcal{S}_n)$ the loss function on our dataset (e.g., average cross-entropy). We can formalize the previous idea by defining a so-called \textbf{regularization term} $R(\mathbf{w})$ that scores solutions based on our preference, and penalize the loss by adding the regularization term to the original loss function:
$$
L_{\text{reg}}=L(\mathbf{w}, \mathcal{S}_n)+\lambda R(\mathbf{w})
$$
where we assume that a higher value of $R(\mathbf{w})$ corresponds to a worse solution, and $\lambda \ge 0$ is a scalar that weights the two terms. For $\lambda=0$ the regularization term has no effect, while for $\lambda \rightarrow \infty$ we simply select the best function based on our a priori knowledge.

This can also be justified as performing maximum a-priori (instead of maximum likelihood) inference based on the combination of a prior distribution on the weights $p(\mathbf{w})$ and a standard likelihood function on our data (Section \ref{sec:bayesian_learning}):

\begin{equation}
\mathbf{w}^*=\underset{\mathbf{w}}{\arg\max}\left\{ \log p(\mathcal{S}_n \;\vert\; \mathbf{w}) + \log p(\mathbf{w})\right\}
\label{eq:map_solution}
\end{equation}

where having a regularization term corresponds to a non-uniform prior distribution $p(\mathbf{w})$. We have already seen one example of regularization in Section \ref{subsec:regularizing_least_squares}, i.e., the $\ell_2$ norm of the weights:

$$
R(\mathbf{w})=\lVert \mathbf{w} \rVert^2 =\sum_i w_i^2
$$

For the same unregularized loss, penalizing the $\ell_2$ norm will favor solutions with a lower weight magnitude, corresponding to “less abrupt” changes in the output for a small deviation in the input.\footnote{With respect to \eqref{eq:map_solution},  $\ell_2$ regularization is equivalent to choosing a Gaussian prior on the weights with diagonal $\sigma^2 \mathbf{I}$ covariance.} Consider now the effect of the regularization term on the gradient term:

$$
\nabla L_{\text{reg}}=\nabla L(\mathbf{w},\mathcal{S}_n)+2\lambda\mathbf{w}
$$

Written in this form, this is sometimes called \textbf{weight decay}, because absent the first term, its net effect is to decay the weights by a small proportional factor $\lambda$ (sending them to $0$ exponentially fast in the number of iterations if $\nabla L(\mathbf{w}, \mathcal{S}_n)=0$). For (S)GD, $\ell_2$ regularization and weight decay coincide. However, for other types of optimization algorithms (e.g., momentum-based SGD, Adam), a post-processing is generally applied on the gradients. Denoting by $g(\nabla L(\mathbf{w}, \mathcal{S}_n))$ the post-processed gradients of the (unregularized) loss, we can write a generalized weight decay formulation (ignoring the constant term $2$) as:

$$
\mathbf{w}_t =\mathbf{w}_{t-1} \eqnmarkbox[drawred]{node}{-g(\nabla L(\mathbf{w}_{t-1}, \mathcal{S}_n))} \eqnmarkbox[drawgreen]{node2}{- \lambda \mathbf{w}_{t-1}}
$$
\annotate[yshift=1em]{above,right}{node}{Unregularized gradient}
\annotate[yshift=-1em]{below,left}{node2}{Weight decay term}

\vspace{1em}
This is different from pure $\ell_2$ regularization, in which case the gradients of the regularization term would be inside $g(\bullet)$. This is especially important for algorithms like Adam, for which the weight decay formulation (known as \textbf{AdamW} \cite{loshchilov2018fixing}) can work better.

We can also consider other types of regularization terms. For example, the $\ell_1$ norm:

$$
R(\mathbf{w})=\lVert \mathbf{w} \rVert_1=\sum_i \lvert x_i \rvert
$$

can favor \textit{sparse} solutions having a high percentage of zero values (and it corresponds to placing a Laplace prior on the weights). This can also be generalized to group sparse variants to enforce structured sparsity on the neurons \cite{scardapane2017group}.\footnote{Training sparse models is a huge topic with many connections also to efficient hardware execution. See \cite{bach2012optimization} for a review on sparse penalties in the context of convex models, and \cite{hoefler2021sparsity} for an overview of sparsity and pruning in general differentiable models.}  Sparse $\ell_1$ penalization is less common than for other machine learning models because it does not interact well with the strong non-convexity of the optimization problem and the use of gradient descent \cite{ziyin2023spred}. However, it is possible to re-parameterize the optimization problem to mitigate this issue at the cost of a larger memory footprint. In particular, \cite{ziyin2023spred}  showed that we can replace $\mathbf{w}$ with two equivalently shaped vectors $\mathbf{a}$ and $\mathbf{b}$, and:
\begin{equation}
\mathbf{w} = \mathbf{a} \odot \mathbf{b} \;,\; \lVert \mathbf{w} \rVert_1 \approx \lVert \mathbf{a} \rVert^2 + \lVert \mathbf{b} \rVert^2
\end{equation}
where $\approx$ means that the two problems can be shown to be almost equivalent under very general conditions \cite{ziyin2023spred}.

We can gain some geometric insights as to why (and how) regularization works by considering a convex loss function $L(\cdot, \cdot)$ (e.g., least-squares), in which case the regularized problem can be rewritten in an explicitly constrained form as:
\begin{equation}
\begin{matrix}\arg\min & L(\mathbf{w}, \mathcal{S}_n) \\ \text{subject to} & R(\mathbf{w})\le\mu \end{matrix}
\label{eq:constrained_problem}
\end{equation}

where $\mu$ depends proportionally on $\lambda$, with the unconstrained formulation arising by rewriting \eqref{eq:constrained_problem} with a Lagrange multiplier. In this case, $\ell_2$ regularization corresponds to constraining the solution to lie inside a circle centered in the origin, while $\ell_1$ regularization corresponds to having a solution inside (or on the vertices) of a regular polyhedron centered in the origin, with the sparse solutions lying at the vertices intersecting the axes.

\subsection{Early stopping}

From the point of view of optimization, minimizing a function $L(\mathbf{w})$ is the task of finding a stationary point as quickly as possible, i.e., a point $\mathbf{w}_t$ such that $\nabla L(\mathbf{w}_t) \approx 0$:

$$
\lVert L(\mathbf{w}_t)-L(\mathbf{w}_{t-1})\rVert^2\le\varepsilon
$$

for some tolerance $\varepsilon > 0$. However, this does not necessarily correspond to what we want when optimizing a model. In particular, in a low-data regime training for too long can result in overfitting and, in general, anything which improves generalization is good irrespective of its net effect on the value on $L(\bullet)$ or the descent direction (e.g., weight decay).

\textbf{Early stopping} is a simple example of the difference between pure optimization and learning. Suppose we have access to a small supervised dataset, separate from the training and test dataset, that we call \textbf{validation dataset}. At the end of every epoch, we track a metric of interest on the validation dataset, such as the accuracy or the F1-score. We denote the score at the $t$-th epoch as $a_t$. The idea of early stopping is to check this metric to see if it keeps improving: if not, we may be entering an overfitting regime and we should stop training. Because the accuracy can oscillate a bit due to random fluctuations, we do this robustly by considering a window of $k$ epochs (the \textbf{patience}):
$$
\text{If } a_t \le a_i ,\;\; \eqnmarkbox[drawred]{node}{\forall i =t-1,t-2, \ldots,t-k} \rightarrow \text{Stop training}
$$
\annotate[yshift=-1em]{below,left}{node}{Wait for $k$ epochs}

\vspace{1em}
For a high value of the patience hyper-parameter $k$, the algorithm will wait more, but we will be more robust to possible oscillations. If we have a mechanism to store the weights of the model (\textbf{checkpointing}) we can also rollback the weights to the last epoch that showed improvement, corresponding to the epoch number $t-k$.

Early stopping can be seen as a simple form of \textbf{model selection}, where we select the optimal number of epochs based on a given metric. Differently from the optimization of the model, we can optimize here for any metric of interest, such as the F1-score, even if not differentiable. 

Interestingly, for large over-parameterized models early stopping is not always beneficial, as the relation between epochs and validation error can be non-monotone with multiple phases of ascent and descent (a phenomenon called \textbf{multiple descents} \cite{rocks2022memorizing}) and sudden drops in the loss after long periods of stasis \cite{power2022grokking}. Hence, early stopping is useful mostly when optimizing on small datasets.

\subsection{Data augmentation} \addclock

Generally speaking, the most effective method to improve performance for a model is to increase the amount of available data. However, labelling data can be costly and time-consuming, and generating data artificially (e.g., with the help of large language models) requires customized pipelines to work effectively \cite{patel2024datadreamer}.

In many cases, it is possible to partially mitigate this issue by \textit{virtually} increasing the amount of available data by transforming them according to some pre-specified number of (semantic preserving) transformations. As a simple example, consider a vector input $\mathbf{x}$ and a transformation induced by adding Gaussian noise:
$$
\mathbf{x}^\prime=\mathbf{x}+\varepsilon,\;\varepsilon \sim \mathcal{N}(\mathbf{0},\sigma^2\mathbf{I})
$$
This creates a virtually infinite amount of data comprised in a small ball centered around $\mathbf{x}$. In addition, this data must not be stored in the disk, and the process can be simulated by applying the transformation at runtime every time a new mini-batch is selected. In fact, it is known that training in this way can make the model more robust and it is connected to $\ell_2$ regularization \cite{bishop1995training}. However, vectorial data is unstructured, and adding noise with too high variance can generate points that are invalid.

For images, we can do better by noting that there is in general a large number of transformations that can change an image while preserving its semantic: zooms, rotations, brightness modifications, contrast changes, etc. Denote by $T(x; c)$ one such transformation (e.g., rotation), parameterized by some parameter $c$ (e.g., the rotation angle). Most transformations include the base image as a special case (in this case, for example, with a rotation angle $c=0$). \textbf{Data augmentation} is the process of transforming images during training according to one or more of these transformations:
\begin{equation}
x^\prime=T(x;c),\;c\sim p(c)
\label{eq:data_augmentation}
\end{equation}
where $p(c)$ denotes the distribution of all valid parameters (e.g., rotation angles between $-20^\circ$ and $+20^\circ$). During training, each element of the dataset is sampled once per epoch, and each time a different transformation \eqref{eq:data_augmentation} can be applied, creating a (virtually) unlimited stream of unique data points.

Data augmentation is very common for images (or similar data, such as audio and video), but it requires a number of design choices: what transformations to include, which parameters to consider, and how to compose these transformations. A simple strategy called \textbf{RandAugment} \cite{cubuk2020randaugment} considers a wide set of transformations, and for every mini-batch samples a small number of them (e.g., 2 or 3), to be applied sequentially with the same magnitude. Still, the user must verify that the transformations are valid (e.g., if recognizing text, horizontal flipping can make the resulting image invalid). From a practical point of view, data augmentation can be included either as part of the data loading components (see Box \ref{code:data_augmentation}), or as part of the model.

\begin{mypy}{Data augmentation pipeline with two transformations applied in sequence, taken from the {\footnotesize\mintinline{python}{torchvision}} package. In PyTorch, augmentations can be passed to the data loaders or used independently. In other frameworks, such as TensorFlow and Keras, data augmentation can also be included natively as layers inside the model.}{code:data_augmentation}
# Image tensor (b, c, h, w)
img = torch.randint(0, 256, 
            size=(32, 3, 256, 256))

# Data augmentation pipeline
from torchvision.transforms import v2
transforms = v2.Compose([
    v2.RandomHorizontalFlip(p=0.5),
    v2.RandomRotation(10),
])

# Applying the data augmentation pipeline: 
# each function call returns a different 
# mini-batch starting from the same 
# input tensor.
img = transforms(img)
\end{mypy}

Data augmentation pipelines and methods can be more complex than simple intuitive transformations. Even for more sophisticated types, the intuition remains that, as long as the model is able to solve a task in a complex scenario (e.g., recognizing an object in all brightness conditions) it should perform even better in a realistic, mild scenario. Additionally, data augmentation can prevent overfitting by avoiding the repetition of the same input multiple times.

As an example of more sophisticated methods, we describe \textbf{mixup} \cite{zhang2017mixup} for vectors, and its extension \textbf{cutmix} \cite{yun2019cutmix} for images. For the former, suppose we sample two examples, $(\mathbf{x}_1, y_1)$ and $(\mathbf{x}_2, y_2)$. The idea of mixup is to create a new, virtual example which is given by their convex combination:
\begin{gather}
\mathbf{x}=\lambda\mathbf{x}_1+(1-\lambda)\mathbf{x}_2\\y=\lambda y_1+(1-\lambda)y_2
\label{eq:mixup}
\end{gather}
where $\lambda$ is chosen randomly in the interval $[0,1]$. This procedure should push the model to have a simple (linear) output in-between the two examples, avoiding abrupt changes in output. From a geometric viewpoint, for two points that are close, we can think of \eqref{eq:mixup} as slowly moving on the manifold of the data, by following the line that connects two points as $\lambda$ goes from $0$ to $1$.

\begin{figure}[t]
    \centering
    \includegraphics[width=0.8\textwidth]{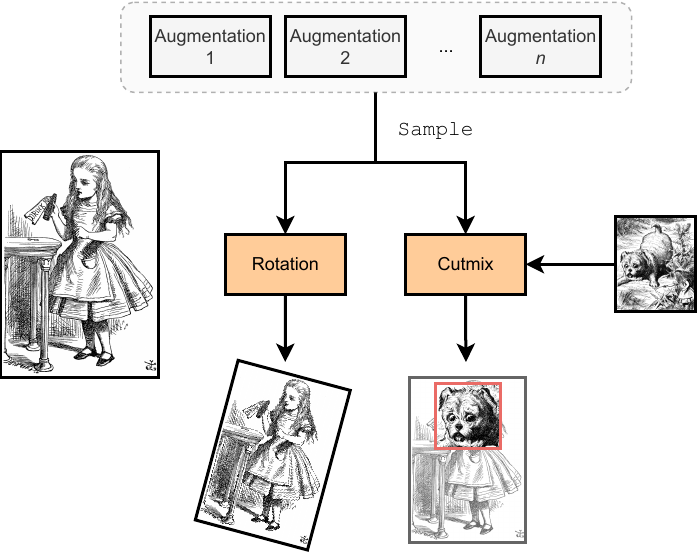}
    \caption{High-level overview of data augmentation. For every mini-batch, a set of data augmentations are randomly sampled from a base set, and they are applied to the images of the mini-batch. Here, we show an example of \textit{rotation} and an example of \textit{cutmix}. Illustrations by John Tenniel, reproduced from Wikimedia.}
    \label{fig:cutmix}
\end{figure}

Mixup may not work for images, because linearly interpolating two images pixel-by-pixel gives rise to blurred images. With \textbf{cutmix}, we sample instead a small patch of fixed shape (e.g., $32 \times 32$) on the first image. Denote by $\mathbf{M}$ a binary mask of the same shape as the images, with $1$ for pixels inside the patch, and $0$ for pixels outside the patch. In cutmix, we combine two images $x_1$ and $x_2$ by “stitching” a piece from the first one on top of the second one:
$$
x=\mathbf{M}\odot x_1+(1-\mathbf{M})\odot x_2
$$
while the labels are still linearly interpolated as before with a random coefficient $\lambda$. See Figure \ref{fig:cutmix} for an example of data augmentation using both rotation and cutmix.

\section{Dropout and normalization}

The strategies we have described in the previous section are very general, in the sense that they imply modifications to the optimization algorithm or to the dataset itself, and they can be applied to a wide range of algorithms. 

Instead, we now focus on three ideas that were popularized in the period between 2012 and 2016, mostly in the context of the ImageNet challenge. All three are specific to differentiable models, since they can be implemented as additional layers or connections in the model that simplify training of very deep models. We list the methods in roughly chronological order. As we will see in the following chapters, these methods remain fundamental also beyond convolutional models.

\subsection{Regularization via dropout}
\label{subsec:dropout}

When discussing data augmentation, we mentioned that one insight is that augmentation forces the network to learn in a more difficult setup, so that its performance in a simpler environment can improve in terms of accuracy and robustness. \textbf{Dropout} \cite{srivastava2014dropout} extends this idea to the internal embeddings of the model: by artificially introducing noise during training to the intermediate outputs of the model, the solution can improve.

There are many choices of possible noise types: for example, training with small amounts of Gaussian noise in the activation has always been a popular alternative in the literature of recurrent models. As the name suggests, dropout's idea is to randomly remove certain units (neurons) during the computation, reducing the dependence on any single internal feature and (hopefully) leading to training robust layers with a good amount of redundancy.

\begin{figure}
    \centering
    \includegraphics[width=\textwidth]{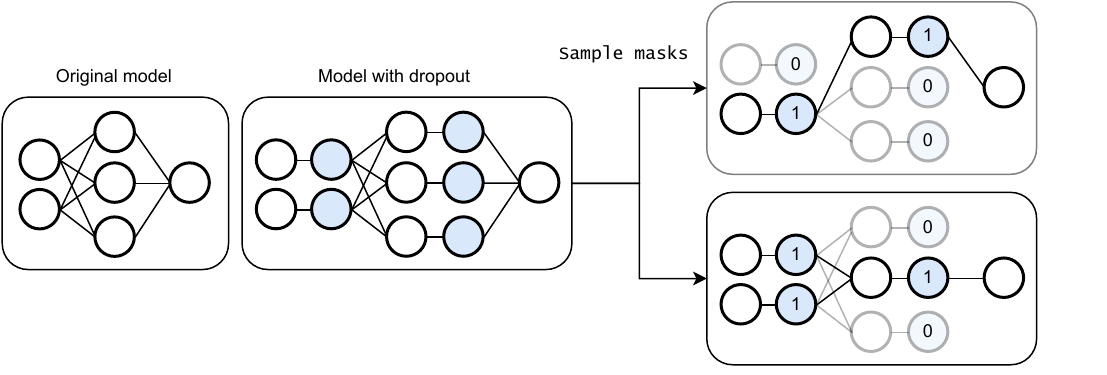}
    \caption{Schematic overview of dropout: starting from a base model, we add additional units after each layer of interest, shown in blue. At training time, each dropout unit is randomly assigned a binary value, masking part of the preceding layers. Hence, we select one out of exponentially many possible models having a subset of active hidden units every time a forward pass is made. Dropout can also be applied at the input level, by randomly removing some input features.}
    \label{fig:dropout}
\end{figure}

We define dropout in the case of a fully-connected layer, which is its most common use case. 

\begin{definition}[Dropout layer] \addbottle $\,$

Denote by $\mathbf{X} \sim (n, c)$ a mini-batch of internal activations of the model (e.g., the output of some intermediate fully-connected layer) with $n$ elements in the mini-batch and $c$ features. In a dropout layer, we first sample a binary matrix $\mathbf{M} \sim \text{Binary}(n,c)$ of the same size, whose elements are drawn from a Bernoulli distribution with probability $p$ (where $p \in [0,1]$ is a user’s hyper-parameter):\footnote{The samples from $\text{Bern}(p)$ are $1$ with probability $p$ and $0$ with probability $1-p$.}

\begin{equation}
M_{ij}\sim \text{Bern}(p)
\label{eq:sampling_m}
\end{equation}

The output of the layer is obtained by masking the input:

$$
\textnormal{Dropout}(\mathbf{X})=\mathbf{M} \odot \mathbf{X}
$$

The layer has a single hyper-parameter, $p$, and no trainable parameters.

\end{definition}

We call $1-p$ the \textbf{drop probability}. Hence, for any element in the mini-batch, a random number of units (approximately $(1-p) \%$) will be set to zero, effectively removing them. This is shown in Figure \ref{fig:dropout}, where the additional dropout units are shown in blue. Sampling the mask is part of the layer’s forward pass: for two different forward passes, the output will be different since different elements will be masked, as shown on the right in Figure \ref{fig:dropout}. 

As the figure shows, we can implement dropout as a layer, which is inserted after each layer that we want to regularize. For example, consider the fully-connected model with two layers shown in Figure \ref{fig:dropout}:

$$
y=(\text{FC}\circ\text{FC})(\mathbf{x})
$$

Adding dropout regularization over the input and over the output of the first layer returns a new model having \textit{four} layers:

$$
y=(\text{FC}\circ{\color{drawred}\text{Dropout}}\circ\text{FC} \circ {\color{drawred}\text{Dropout}})(\mathbf{x})
$$

See Box \ref{code:dropout_model} for an implementation in PyTorch.

\begin{mypy}{The model in Figure \ref{fig:dropout} implemented as a sequence of four layers in PyTorch. During training, the output of the model will be stochastic due to the presence of the two dropout layers.}{code:dropout_model}
model = nn.Sequential(
    nn.Dropout(0.3),
    nn.Linear(2, 3), nn.ReLU(),
    nn.Dropout(0.3),
    nn.Linear(3, 1)
)
\end{mypy}

While dropout can improve the performance, the output $y$ is now a random variable with respect to the sampling of the different masks inside the dropout layers, which is undesirable after training. For example, two forward passes of the network can return two different outputs, and some draws (e.g., with a very large number of zeroes) can be suboptimal. Hence, we require some strategy to replace the forward pass with a deterministic operation. 

Suppose we have $m$ dropout layers. Let us denote by $\mathbf{M}_i$ the mask in the $i$-th dropout layer, by $p(\mathbf{M}_1, \ldots, \mathbf{M}_m) = \prod_{i=1}^m p(\mathbf{M}_i)$ the probability distribution over the union of the masks, and by $f(\mathbf{x}; \mathbf{M})$ the deterministic output once a given set of masks $\mathbf{M} \sim p(\mathbf{M})$ are chosen. One choice is to replace the dropout effect with its expected value during inference:

$$
f(\mathbf{x})=\begin{cases} f(\mathbf{x}; \mathbf{M}), \;\; \mathbf{M}\sim p(\mathbf{M}) & \texttt{ [training]} \\ \mathbb{E}_{p(\mathbf{M})}\left[f(\mathbf{x}; \mathbf{M})\right]  & \texttt{ [inference]}\end{cases}
$$

We can approximate the expected value via Monte Carlo sampling (Appendix \ref{chap:probability_theory}) by repeatedly sampling masks values and averaging:

$$
\mathbf{E}_{p(\mathbf{M})}\left[f(\mathbf{x}; \mathbf{M})\right] \approx \frac{1}{k}\sum_{i=1}^k f(\mathbf{x}; \mathbf{Z}_i),\;\; \mathbf{Z}_i \sim p(\mathbf{M})
$$

which is simply the average of $k$ forward passes. This is called \textbf{Monte Carlo dropout} \cite{gal2016dropout}. The output is still stochastic, but with a proper choice of $k$, the variance can be contained. In addition, the outputs of the different forward passes can provide a measure of uncertainty over the prediction. 

However, performing multiple forward passes can be expensive. A simpler (and more common) option is to replace the random variables \textit{layer-by-layer}, which is a reasonable approximation. The expected value in this case can be written in closed form:

$$
\mathbb{E}_{p(\mathbf{M})}\left[\text{Dropout}(\mathbf{X})\right]=p\mathbf{X}
$$

which is the input rescaled by a constant factor $p$ (the probability of sampling a $1$ in the mask). This leads to an even simpler formulation, \textbf{inverted dropout}, where this correction is accounted for during training:

$$
\text{Dropout}(\mathbf{X})=\begin{cases} \displaystyle\frac{\mathbf{M}\odot\mathbf{X}}{p} & \texttt{ [training]} \\ \,\,\mathbf{X}  & \texttt{ [inference]}\end{cases}
$$

In this case, the dropout layer has no effect when applied during inference and can be directly removed. This is the preferred implementation in most frameworks. See Box \ref{code:dropout} for some comparisons.

\begin{mypy}{Applying the model from Box \ref{code:dropout_model} on a mini-batch of $16$ examples. For layers like dropout, a framework requires a way to differentiate between a  forward pass executed during training or during inference. In PyTorch, this is done by calling the {\footnotesize\mintinline{python}{train}} and {\footnotesize\mintinline{python}{eval}} methods of a model, which set an internal {\footnotesize\mintinline{python}{train}} flag on all layers. We also show a vectorized implementation of Monte Carlo dropout.}{code:dropout}
x = torch.randn((16, 2))

# Training with dropout
model.train()
y = model(x)

# Inference with dropout
model.eval()
y = model(x)

# Monte Carlo dropout for inference
k = 10
model.train()
y = model(x[:, None, :].repeat(1, k, 1))
                       .mean(1)
\end{mypy}

As we mentioned, dropout (possibly with a low drop probability, such as $p=0.8$ or $p=0.9$) is common for fully-connected layers. It is also common for attention maps (introduced in the next chapter). It is less common for convolutional layers, where dropping single elements of the input tensor results in sparsity patterns which are too unstructured. Variants of dropout have been devised which take into consideration the specific structure of images: for example, \textbf{spatial dropout} \cite{tompson2015efficient} drops entire channels of the tensor, while \textbf{cutout} \cite{devries2017improved} drops spatial patches of a single channel.

Other alternatives are also possible. For example, \textbf{DropConnect} \cite{wan2013regularization} drops single weights of a fully-connected layer:
$$
\text{DropConnect}(\mathbf{x})=(\mathbf{M}\odot\mathbf{W})\mathbf{x}+\mathbf{b}
$$
DropConnect in inference can also be approximated efficiently with moment matching \cite{wan2013regularization}. However, these are less common in practice, and the techniques described next are preferred.

\subsection{Batch (and layer) normalization} \addclock

When dealing with tabular data, a common pre-processing operation that we have not discussed yet is \textbf{normalization}, i.e., ensuring that all features (all columns of the input matrix) share similar ranges and statistics. For example, we can pre-process the data to squash all columns in a $[0,1]$ range (\textbf{min-max normalization}) or to ensure a zero mean and unitary variance for each column (called either \textbf{standard scaling} or \textbf{normal scaling} or \textbf{z-score scaling}).

\textbf{Batch normalization} (BN, \cite{ioffe2015batch}) replicates these ideas, but for the intermediate embeddings of the model. This is non trivial, since the statistics of a unit (e.g., its mean) will change from iteration to iteration after each gradient descent update. Hence, to compute the mean of a unit we should perform a forward pass on the entire training dataset at every iteration, which is unfeasible. As the name implies, BN’s core idea is to approximate these statistics using only the data \textit{in the mini-batch itself}. 

Consider again the output of any fully-connected layer $\mathbf{X} \sim (n,c)$, where $n$ is the mini-batch size. We will see shortly how to extend the ideas to images and other types of data. In BN, we normalize each feature (each column of $\mathbf{X}$) to have zero mean and unitary variance, based on the mini-batch alone. To this end, we start by computing the empirical column-wise mean $\mathbf{\mu} \sim (c)$ and variances $\sigma^2 \sim (c)$:
\begin{align}
\text{Mean of column } j \text{:}  &&& \mu_j = \frac{1}{n}\sum_iX_{ij} \label{eq:empirical_mean} \\ \text{Variance of column } j \text{:} &&& \sigma^2_j=\frac{1}{n}\sum_i(X_{ij}-\mu_j)^2  \label{eq:empirical_variance}
\end{align}
We then proceed to normalize the columns:

\vspace{1em}
$$
\mathbf{X}^\prime=\frac{\eqnmarkbox[drawred]{node}{\mathbf{X} - \mu}}{\eqnmarkbox[drawgreen]{node2}{\sqrt{\sigma^2 + \varepsilon}}}
$$
\annotate[yshift=1em]{above,right}{node}{Set the column mean to $0$}
\annotate[yshift=-1em]{below,right}{node2}{Set the column variance to $1$}

\vspace{1em}
where we consider the standard broadcasting rules ($\mu$ and $\sigma^2$ are broadcasted over the first dimension), and $\varepsilon > 0$ is a small positive term added to avoid division by zero. Differently from normalization for tabular data, where this operation is applied once to the entire dataset before training, in BN this operation must be recomputed for every mini-batch during each forward pass.

The choice of zero mean and unitary variance is just a convention, not necessarily the best one. To generalize it, we can let the optimization algorithm select the best choice, for a small overhead in term of parameters. Consider two trainable parameters $\mathbf{\alpha} \sim (c)$ and $\mathbf{\beta} \sim (c)$ (which we can initialize as $1$ and $0$ respectively), we perform:
$$
\mathbf{X}^{\prime\prime}=\alpha\mathbf{X}^\prime + \beta
$$
with similar broadcasting rules as above. The resulting matrix will have mean $\beta_i$ and variance $\alpha_i$ for the $i$-th column. The BN layer is defined as the combination of these two operations.

\begin{definition}[Batch normalization layer] \addbottle $\,$

Given an input matrix $\mathbf{X} \sim (n,c)$, a \textbf{batch normalization} (BN) layer applies the following normalization:

$$
\textnormal{BN}(\mathbf{X})=\alpha\left(\frac{\mathbf{X} - \mu}{\sqrt{\sigma^2 + \varepsilon}}\right) + \beta
$$

where $\mu$ and $\sigma^2$ are computed according to \eqref{eq:empirical_mean} and \eqref{eq:empirical_variance}, while $\alpha \sim (c)$ and $\beta \sim (c)$ are trainable parameters. The layer has no hyper-parameters. During inference, $\mu$ and $\sigma^2$ are fixed as described next.

\end{definition}

The layer has only $2c$ trainable parameters, and it can be shown to greatly simplify training of complex models when inserted across each block. In particular, it is common to consider BN placed in-between the linear and non-linear components of the model:
$$
\mathbf{H} = (\text{ReLU}\circ {\color{drawred}\text{BN}} \circ \text{Linear})(\mathbf{X})
$$
Centering the data before the ReLU can lead to better exploiting its negative (sparse) quadrant. In addition, this setup renders the bias in the linear layer redundant (as it conflates with the $\beta$ parameter), allowing to remove it. Finally, the double linear operation can be easily optimized by standard compilers in most frameworks.

BN is so effective that is has led to a vast literature on understanding why \cite{bjorck2018understanding}. The original derivation considered a problem known as \textbf{internal covariate shift}, i.e., the fact that, from the point of view of a single layer, the statistics of the inputs it receives will change during optimization due to the changes in weights of the preceding layers. However, current literature agrees that the effects of BN is more evident in the optimization itself, both in terms of stability and the possibility of using higher learning rates, due to a combination of scaling and centering effects on the gradients \cite{bjorck2018understanding}.\footnote{See also \url{https://iclr-blog-track.github.io/2022/03/25/unnormalized-resnets/} for a nice entry point into this literature (and the corresponding literature on developing \textbf{normalizer-free} models.}

Extending BN beyond tabular data is simple. For example, consider a mini-batch of image embeddings $X \sim (n,h,w,c)$. We can apply BN on each channel by considering the first three dimensions together, i.e., we compute a channel-wise mean as:

$$
\mu_z = \eqnmarkbox[drawred]{node}{\frac{1}{nhw}}\sum_{i, j,k} X_{ijkz}
$$
\annotate[yshift=-1em]{below,right
}{node}{Mean of channel $z$ (all pixels)} \clearpage

\subsubsection*{Batch normalization during inference}

BN introduces a dependency between the prediction over an input and the mini-batch it finds itself in, which is unwarranted during inference (stated differently, moving an image from one mini-batch to another will modify its prediction). However, we can exploit the fact that the model's parameters do not change after training, and we can freeze the mean and the variance to a preset value. There are two possibilities to this end:
\begin{enumerate}
\item After training, we perform another forward pass on the entire training set to compute the empirical mean and variance with respect to the dataset \cite{wu2021rethinking}.
\item More commonly, we can keep a rolling set of statistics that are updated after each forward pass of the model during training, and use these after training. Considering the mean only for simplicity, suppose we initialize another vector $\widehat{\mu} = \mathbf{0}$, corresponding to the “rolling mean of the mean”. After computing $\mu$ as in \eqref{eq:empirical_mean}, we update the rolling mean with an exponential moving average:
    $$
    \widehat{\mu} \gets	 \lambda\widehat{\mu} + (1-\lambda)\mu
    $$
    where $\lambda$ is set to a small value, e.g., $\lambda=0.01$. Assuming training converges, the rolling mean will also converge to an approximation of the average given by option (1). Hence, after training we can use BN by replacing $\mu$ with the (pre-computed) $\widehat{\mu}$, and similarly for the variance.\footnote{$\widehat{\mu}$ is the first example of a layer's tensor which is part of the layer's state, is adapted during training, but is not needed for gradient descent. In PyTorch, these are referred to as \textit{buffers}.}
\end{enumerate}    

\subsubsection*{Variants of batch normalization}

Despite its good empirical performance, BN has a few important drawbacks. We have already mentioned the dependence on the mini-batch, which has other implications: for example, the variance of $\mu$ during training will grow large for small mini-batches, and training can be unfeasible for very small mini-batch sizes. In addition, training can be difficult in distributed contexts (where each GPU holds a separate part of the mini-batch). Finally, replacing $\mu$ with a different value after training creates an undesirable mismatch between training and inference.

Variants of BN have been proposed to address these issues. A common idea is to keep the overall structure of the layer, but to modify the axes along which the normalization is performed. For example, \textbf{layer normalization} \cite{ba2016layer} computes the empirical mean and variance over \textit{the rows} of the matrix, i.e., for each input independently:

\begin{align}
\text{Mean of {\color{drawred}row} $i$:} &&& \mu_i = \frac{1}{\color{drawred}c}\sum_jX_{\color{drawred}ji}  \\ 
\text{Variance of {\color{drawred}row} $i$:}  &&& \sigma^2_i=\frac{1}{\color{drawred}c}\sum_j(X_{\color{drawred}ji}-\mu_i)^2  
\end{align}

\begin{SCfigure}
    \centering
    \hspace{0.5em}\includegraphics[width=0.6\textwidth]{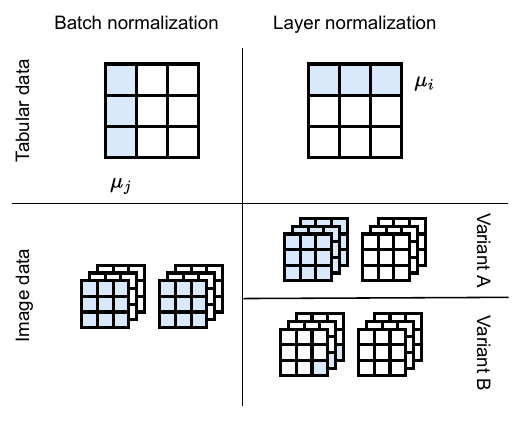}
    \caption{Comparison between BN and LN for tabular and image data. Blue regions show the sets over which we compute means and variances. For LN we have two variants, discussed better in the main text.}
    \label{fig:layer_normalization}
\end{SCfigure}

Consider Figure \ref{fig:layer_normalization}, where we show a comparison between BN and LN for tabular and image-like data. In particular, we show in blue all the samples used to compute a single mean and variance. For layer normalization, we can compute the statistics on $h,w,c$ simultaneously (variant A) or for each spatial location separately (variant B). The latter choice is common in transformer models, discussed in the next chapter. Other variants are also possible, e.g., \textbf{group normalization} restricts the operation to a subset of channels, with the case of a single channel known as \textbf{instance normalization}.\footnote{See \url{https://iclr-blog-track.github.io/2022/03/25/unnormalized-resnets/} for a nicer variant of Figure \ref{fig:layer_normalization}.}

In BN, the axes across which we compute the statistics in \eqref{eq:empirical_mean} and \eqref{eq:empirical_variance} are the same as the axes across which we apply the trainable parameters. In LN, the two are decoupled. For example, consider a PyTorch LN layer applied on mini-batches of dimension $(b, 3, 32, 32)$:

{\footnotesize
\noindent\mintinline{python}{nn.LayerNorm(normalized_shape=[3, 32, 32])}
}

This corresponds to variant A in Figure \ref{fig:layer_normalization}. In this case, $\alpha$ and $\beta$ will have the same shape as the axes over which we are computing the normalization, i.e., $\alpha, \beta \sim (3,32,32)$, for a total of $2\times3\times32\times32=6144$ trainable parameters. The specific implementation of LN and BN must be checked for each framework and model.

We close by mentioning another common variant of layer normalization, called \textbf{root mean square} normalization (RMSNorm) \cite{zhang2019root}. It simplifies LN by removing the mean centering and shifting, which for a single input vector $\mathbf{x} \sim (c)$ can be written as:
\begin{equation}
\text{RMSNorm}(\mathbf{x}) = \frac{\mathbf{x}}{\sqrt{\frac{1}{c}\sum_i x_i^2}} \odot \mathbf{\alpha} 
\end{equation}
When $\mathbf{\beta} = 0$ and the data is already zero-centered, LN and RMSNorm are identical.

\section{Residual connections}
\label{sec:residual_connections}
\subsection{Residual connections and residual networks} \addclock

\begin{SCfigure}
    \centering
    \hspace{1em}\includegraphics[width=0.6\textwidth]{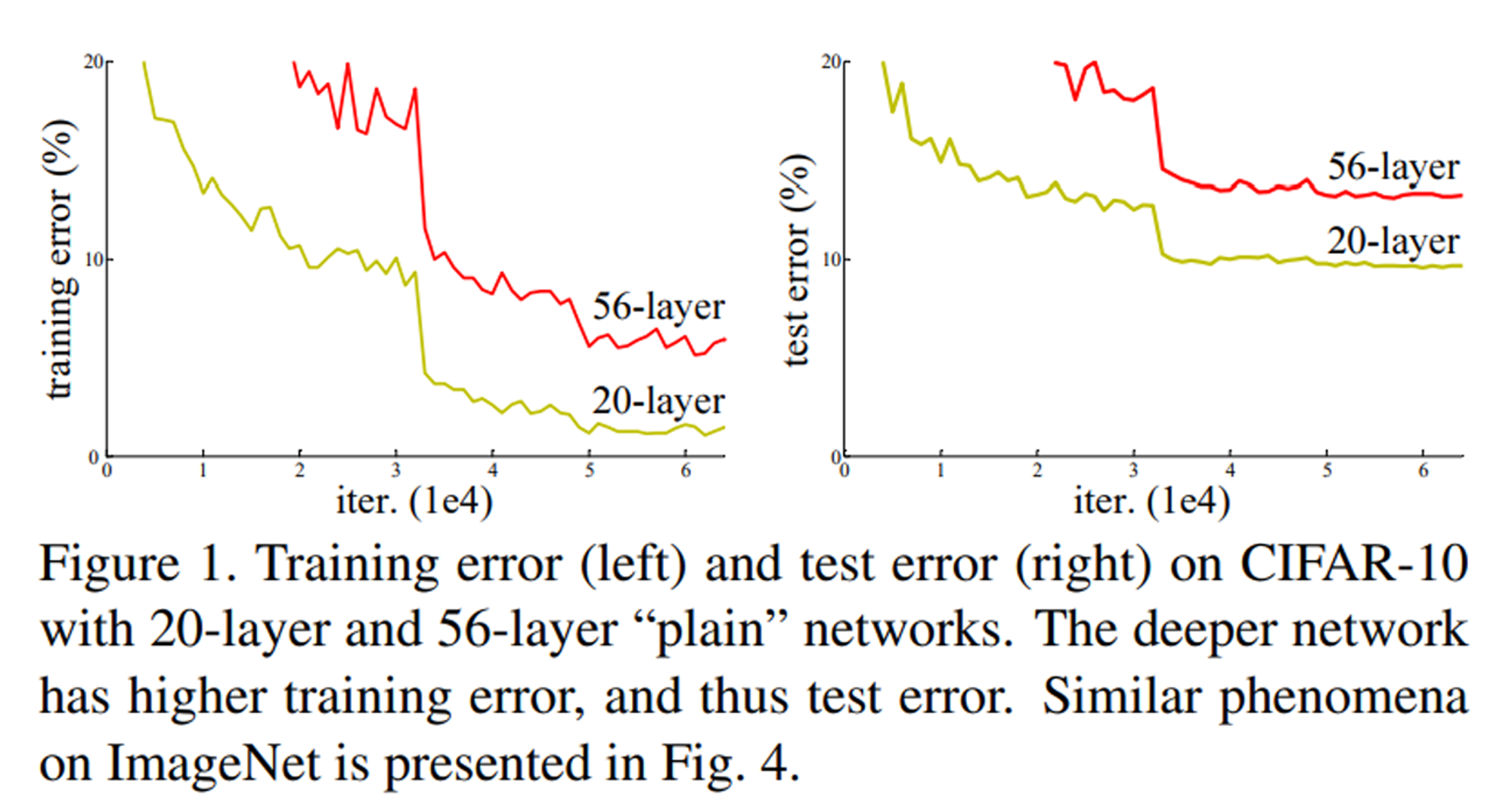}
    \caption{Bigger models do not always improve monotonically in training error, despite representing larger classes of functions. Reproduced from \cite{he2016deep}.}
    \label{fig:resnet}
\end{SCfigure}

The combination of all techniques seen in the previous section is enough to increase significantly the number of layers in our models, but only up to a certain upper bound. Consider three generic layers $f_1$, $f_2$, and $f_3$, and two models $g_1, g_2$ with $g_1$ being a subset of $g_2$:
\begin{gather*}
g_1(x) = (f_3 \circ f_1)(x) \\ g_2(x)=(f_3\circ {\color{drawred}f_2}\circ f_1)(x)
\end{gather*}
Intuitively, by the universal approximation theorem it should always be possible for the intermediate part, $f_2$, to approximate the identity function $f_2(x) \approx x$, in which case $g_2(x) \approx g_1(x)$. Hence, there is always a setting of the parameters in which the second (deeper) model should perform at least as well as the first (shallower) one. However, this was not observed in practice, as shown in Figure \ref{fig:resnet}.

We can solve this by biasing the blocks in the network towards the identity function. This can be done easily by rewriting a block $f(x)$ with what is called a \textbf{residual (skip) connection} \cite{he2016deep}:
$$
r(x)=f(x){\color{drawred} \; + \;x}
$$
Hence, we use the block to model deviations from the identity, $f(x) = r(x) - x$, instead of modeling deviations from the zero function. This small trick alone helps in training models up to hundreds of layers. We call $f(x)$ the \textbf{residual path}, $r(x)$ a \textbf{residual block}, and a convolutional model composed of residual blocks a \textbf{residual network} (abbreviated to ResNet).

Residual connections work well with batch normalization on the residual path, which can be shown to further bias the model towards the identity at the beginning of training \cite{de2020batch}. However, residual connections can be added only if the input and output dimensionality of $f(x)$ are identical. Otherwise, some rescaling can be added to the residual connection. For example, if $x$ is an image and $f(x)$ modifies the number of channels, we can add a $1 \times 1$ convolution:

$$
r(x)=f(x)+\text{Conv2D}_{1\times 1}(x)
$$

The benefit of a residual block can be understood also in terms of its backward pass. Consider the VJP of the residual block:

$$
\text{vjp}_r(\mathbf{v})=\text{vjp}_f(\mathbf{v}) + \mathbf{v}^\top\mathbf{I} = \eqnmarkbox[drawred]{node}{\text{vjp}_f(\mathbf{v})} + \eqnmarkbox[drawgreen]{node2}{\mathbf{v}^\top}
$$
\annotate[yshift=1em]{above,left}{node}{VJP of $f$}
\annotate[yshift=-1em]{below,left}{node2}{VJP of the skip connection}

\vspace{1em}
Hence, the forward pass lets the input $x$ pass through unmodified on the skip connection, while the backward pass adds the unmodified back-propagated gradient $\mathbf{v}$ to the original VJP, which can help mitigating gradient instabilities.

\subsection*{On the design of the residual block}

How to design the block $f(x)$? Consider the batch-normalized block introduced earlier:
$$
h = \underbrace{(\text{ReLU}\circ \text{BN} \circ \text{Conv2D})}_{=f(x)}(x) + x
$$
Because the output of ReLU is always positive, we have that $h \ge x$ (element-wise). Hence, a stack of residual blocks of this form can only increase the values of the input tensor, or set it to zero. For this reason, the original design proposed in \cite{he2016deep} considered a similar stack of blocks except for the last activation function. As an example, for two blocks we obtain the following design:
$$
h = (\text{BN} \circ \text{Conv2D} \circ\text{ReLU}\circ \text{BN} \circ \text{Conv2D})(x) + x
$$
A series of blocks of this form can be preceded by a small component with non-residual connections to reduce the image dimensionality, sometimes called the \textbf{stem}. The specific choice of hyper-parameters for this block has varied significantly over the years.

\begin{figure}[t]
    \centering
    \hspace{1em}\includegraphics[width=0.8\textwidth]{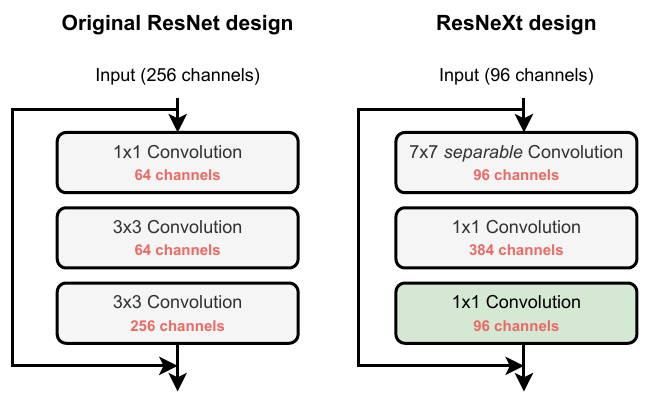}
    \caption{The original ResNet block \cite{he2016deep}, and the more recent ResNeXt \cite{liu2022convnet} block. As can be seen, the design has shifted from an early channel reduction to a later compression (\textbf{bottleneck}). Additional details (not shown) are the switch from BN to LN and the use of GELU activation functions. Adapted from \cite{liu2022convnet}.}
    \label{fig:resnext}
\end{figure}

The original ResNet block proposed a compression in the number of channels for the first operation, followed by a standard $3 \times 3$ convolution and a final upscaling in the number of channels. Recently, instead, \textbf{bottleneck} layers like the ResNeXt block \cite{liu2022convnet} (on the right in Figure \ref{fig:resnext}) have become popular. To increase the receptive field of the convolution, the initial layer is replaced by a depthwise convolution. To exploit the reduced number of parameters, the number of channels is \textit{increased} by a given factor (e.g., $3\times$, $4\times$), before being reduced by the last $1\times1$ convolution. \clearpage

\subsection{Additional perspectives on residual connections} \addteacup

We close the chapter by discussing two interesting perspectives on the use of residual connections, which have both been explored in-depth in current research. First, consider a network composed of two residual blocks:
\begin{gather}
h_1 = {\color{drawred}f_1}(x)+x\\h_2={\color{drawgreen}f_2}(h_1)+h_1
\end{gather}
If we unroll the computation:
$$
h_2 = {\color{drawgreen}f_2}({\color{drawred}f_1}(x) + x) + {\color{drawred}f_1}(x) + x
$$
This corresponds to the sum of several \textit{paths} in the network, where the input is either left unmodified, it goes through only a single transformation ($f_1$ or $f_2$), or through their combination.

\begin{SCfigure}
    \centering
    \includegraphics[width=0.6\textwidth]{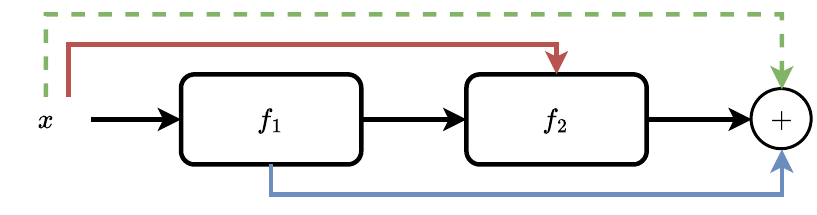}
    \caption{Residual paths: the black, red, and blue paths are implemented explicitly; the green path is only implicit.}
    \label{fig:residual_paths}
\end{SCfigure}

It should be clear that the number of such paths grows exponentially with the number of residual blocks. Hence, deep residual models can be seen as a combination (an ensemble) of a very large number of smaller models, implemented through weight-sharing. This view can be tested to show, for example, that ResNets tend to be robust to small deletions or modifications of their elements \cite{veit2016residual}. This is shown visually in Figure \ref{fig:residual_paths}.

Second, consider the following differential equation, expressed in terms of a continuous parameter $t$ representing time:
$$
\partial_tx_t=f(x,t)
$$
We are using a neural network with arguments $x$ and $t$ (a scalar) to parameterize the time derivative of some function. This is called an \textbf{ordinary differential equation} (ODE). A common problem with ODEs is integrating from a known starting value $x_0$ up to some specified time instant $T$:
$$
x_T=x_0+\int_{t=0}^Tf(x,t)dt
$$
Euler's method\footnote{\url{https://en.wikipedia.org/wiki/Euler_method}} for computing $x_T$ works by selecting a small step size $h$ and computing iteratively a first-order discretization:
$$
x_t=x_{t-1}+hf(x_{t-1}, t)
$$
Merging $h$ into $f$, this corresponds to a restricted form of residual model, where all residual blocks share the same weights, each layer corresponds to a discretized time-instant, and $x_T$ is the output of the network. Under this point of view, we can directly work with the original continuous-time equation, and compute the output by integrating it with modern ODE solvers. This is called a \textbf{neural ODE} \cite{chen2018neural}. Continuous-time variants of back-propagation can be derived that take the form of another ODE problem. We will see in the next volume an interesting connection between neural ODEs and a class of generative models known as \textbf{normalizing flows} \cite{papamakarios2021normalizing}.

\section*{From theory to practice}

\begin{wrapfigure}{r}{3.0cm}
\vspace{-3em}\includegraphics[width=3.0cm]{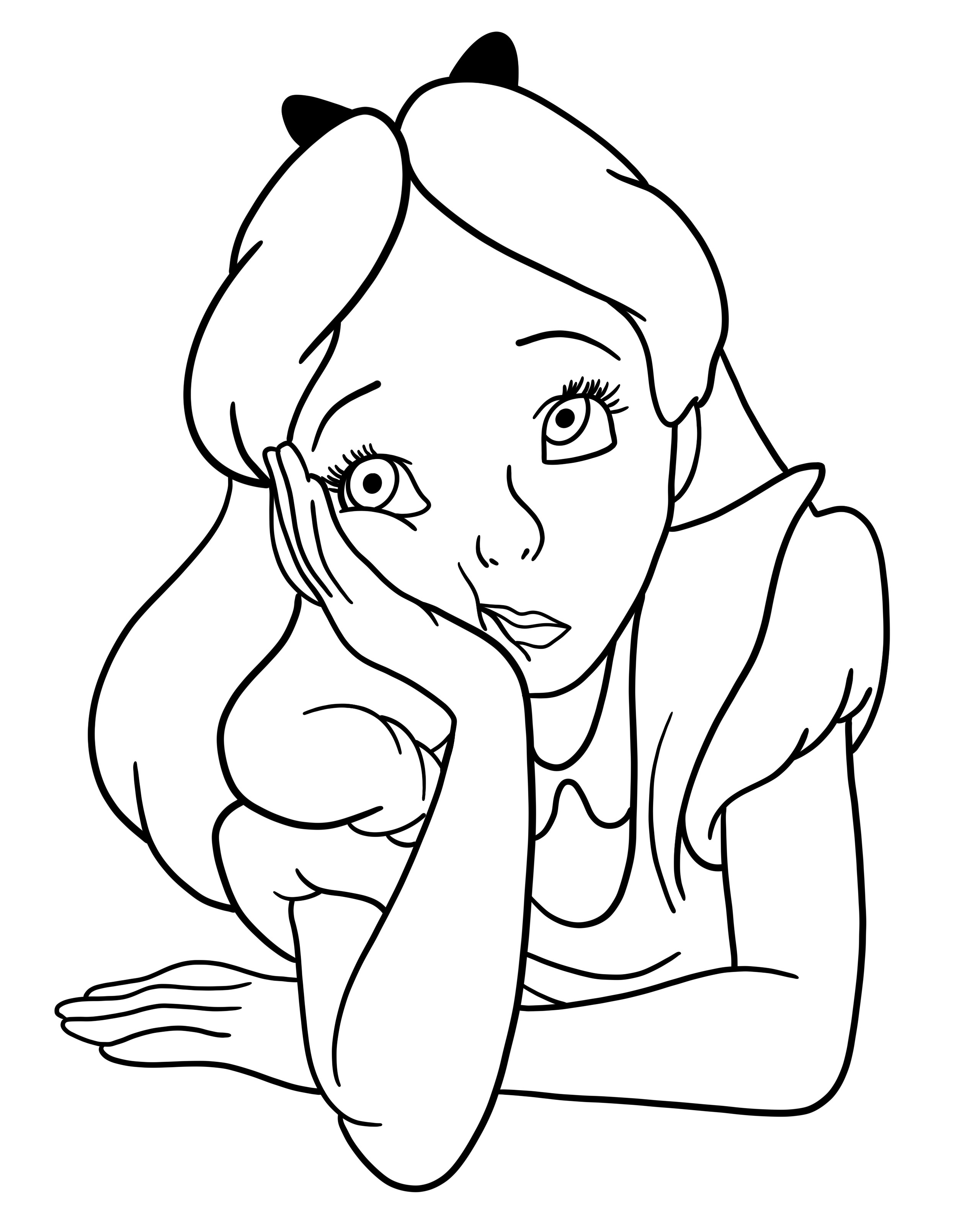}
\vspace{-2em}
\end{wrapfigure}

All the layers we discussed in this chapter (batch normalization, dropout, ...) are already implemented in PyTorch, Equinox, and practically every other framework. For what concerns the rest of the techniques we described, it depends on the framework: for example, weight decay in implemented natively in all PyTorch's optimizers, data augmentation can be found as transformations inside {\footnotesize\texttt{torchvision}} (and other corresponding libraries), while early stopping must be implemented manually.\footnote{In PyTorch, a common alternative is to use an external library such as PyTorch Lightning to handle the training process. Modifications to the training procedure, such as early stopping, are pre-implemented in the form of \textit{callback} functions.}

\begin{enumerate}
\item Before proceeding to the next chapter, I suggest you try implementing either dropout or batch normalization as a layer using only standard linear algebra routines, comparing the results with the built-in layers.
\item In Chapter \ref{chap:cnns} you should have implemented a simple convolutional model for image classification. Try progressively increasing its size, adding normalization, dropout, or residual connections as needed.
\item Take a standard architecture, such as a ResNet \cite{he2016deep}, or a  ResNeXt \cite{liu2022convnet}. Try implementing the entire model by following the suggestions from the original papers. Training on ImageNet-like datasets can be challenging on a consumer's GPU -- if you do not have access to good hardware or cloud GPU hours, you can keep focusing on simpler datasets, such as CIFAR-10.
\item At this point, you may have realized that training from scratch very large models (e.g., ResNet-50) on smaller datasets is practically impossible. One solution is to initialize the weights of the model from an online repository using, e.g., the weights of a model trained on ImageNet, and \textbf{fine-tuning} the model by modifying the last layer, corresponding to the classification head. By this point of the book, this should come as relatively easy -- I suggest using one of the many pre-trained models available on torchvision or on the Hugging Face Hub.\footnote{For an example tutorial: \url{https://pytorch.org/tutorials/beginner/transfer_learning_tutorial.html}.} We will cover fine-tuning more in-depth in the next volume.

\end{enumerate}

%% file: 10_transformers.tex
\chapter{Transformer models}
\label{chap:transformers}

\begin{supportbox}{About this chapter}
Convolutional models are strong baselines, especially for images and sequences where local relations prevail, but they are limited in handling very long sequences or non-local dependencies between elements of a sequence. In this chapter we introduce another class of models, called transformers, which are designed to overcome such challenges.
\end{supportbox}

\section{Long convolutions and non-local models}

After the key developments in the period 2012-2016, discussed in the previous chapter, the next important breakthrough in the design of differentiable models came in 2016-2017 with the popularization of the \textbf{transformer} \cite{vaswani2017attention}, an architecture designed to handle efficiently long-range dependencies in natural language processing. Due to its strong scaling laws, the architecture was then extended to other types of data, from images to time-series and graphs, and it is today a state-of-the-art model in many fields due to its very good scaling laws when trained on large amounts of data \cite{kaplan2020scaling,bordes2024introduction}. 

As we will see, an interesting aspect of the transformer is a decoupling between the data type (through the use of appropriate tokenizers) and the architecture, which for the most part remains data-agnostic. This opens up several interesting directions, such as simple multimodal architectures and transfer learning strategies. We begin by motivating the core component of the transformer, called the \textbf{multi-head attention} (MHA) layer. We will defer a discussion on the original transformer model from \cite{vaswani2017attention} to the next chapter.

\vspace{1em}

\begin{supportbox}{A bit of history}
Historically, this chapter is out of order: in 2015, the most common alternative to CNNs for text were \textbf{recurrent neural networks} (RNNs). As an isolated component, MHA was introduced for RNNs \cite{bahdanau2014neural}, before being used as the core component in the transformer model. We cover RNNs and their modern incarnation, linearized RNNs, in Chapter \ref{chap:rnns}. Recently, RNNs have become an attractive competitor to transformers for language modeling.
\end{supportbox}

\clearpage

\subsection{Handling long-range and sparse dependencies}

Consider these two sentences: 

\begin{quote}
“\textit{The {\color{drawred}cat} is on the {\color{drawgreen}table}}”
\end{quote}

and a longer one:

\begin{quote}
“\textit{The {\color{drawred}cat}, who belongs to my mother, is on the {\color{drawgreen}table}}”.
\end{quote}
In order to be processed by a differentiable model, the sentences must be tokenized and the tokens embedded as vectors (Chapter \ref{chap:convolutions_beyond_images}). From a semantic point of view, the tokens belonging to the red word ({\color{drawred}cat}) and to the green word ({\color{drawgreen}table}) share a similar dependency in both sentences. However, their relative offset varies in the two cases, and their distance can become arbitrarily large. Hence, dependencies in text can be both \textbf{long-range} and \textbf{input-dependent}.

Denote by $\mathbf{X} \sim (n, e)$ a sentence of $n$ tokens embedded in $e$-dimensional vectors and denote by $\mathbf{x}_i$ the $i$th token. We can rewrite a 1D convolution with kernel size $k$ on token $i$ as follows:
\begin{equation}
\mathbf{h}_i=\sum_{j=1}^{2k+1}\mathbf{W}_j \mathbf{x}_{i+k+1-j}
\label{eq:conv_1d}
\end{equation}
Each token inside the receptive field is processed with a fixed weight matrix $\mathbf{W}_i$ that only depends on the specific offset $i$. Modeling long-range dependencies inside the layer requires us to increase the receptive field of the layer, increasing the number of parameters linearly in the receptive field. 

\begin{figure}[t]
    \centering
    \includegraphics[width=\textwidth]{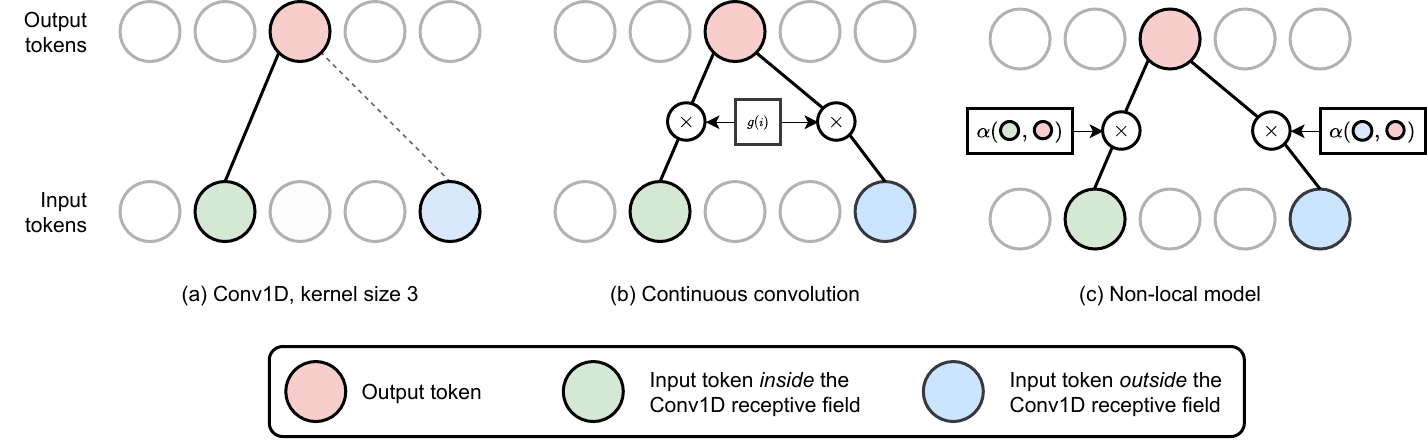}
    \caption{Comparison between different types of convolution for a 1D sequence. We show how one output token (in \colorbox{drawred!30}{red}) interacts with two tokens, one inside the receptive field of the convolution (in \colorbox{drawgreen!30}{green}), and one outside (in \colorbox{drawblue!30}{blue}). (a) In a standard convolution, the blue token is ignored because it is outside of the receptive field of the filter. (b) For a continuous convolution, both tokens are considered, and the resulting weight matrices are given by $g(-1)$ and $g(2)$ respectively. (c) In the non-local case, the weight matrices depend on a pairwise comparison between the tokens themselves.}
    \label{fig:biases}
\end{figure}

One possibility to solve this is the following: instead of explicitly learning the matrices $\mathbf{W}_1, \mathbf{W}_2, \ldots$, we can define them \textit{implicitly} by defining a separate neural block $g(i): \mathbb{R} \rightarrow \mathbb{R}^{e \times e}$ that outputs all weight matrices based on the relative offset $i$. Hence, we rewrite \eqref{eq:conv_1d} as:

$$
\mathbf{h}_i=\eqnmarkbox[drawred]{node}{\sum_{j=1}^n} g(i-j)\mathbf{x}_j
$$
\annotate[yshift=-1em]{below,right}{node}{The sum is now on \textit{all} tokens}

\vspace{1em}
This is called a \textbf{long convolution}, as the convolution spans the entire input matrix $\mathbf{X}$. It is also called a \textbf{continuous convolution} \cite{romero2022towards}, because we can use $g(\bullet)$ to parameterize intermediate positions or variable resolutions \cite{romero2022towards}. The number of parameters in this case only depends on the parameters of $g$, while it does not depend on $n$, the length of the sequence. Defining $g$ is non-trivial because it needs to output an entire weight matrix. We can recover a standard convolution easily:

\begin{equation}
    g(i, j) = \begin{cases} \mathbf{W}_{i-j} & \text{ if } \lvert i - j \rvert \le k \\ 0 & \text{ otherwise } \end{cases}
\end{equation}

This partially solves the problem of long-range dependencies, but it does not solve the problem of dependencies which are conditional on the input, since the weight given to a token depends only on the relative offset with respect to the index $i$. However, this formulation provides a simple way to tackle this problem by letting the trained function $g$ depend on the \textit{content} of the tokens instead of their positions:

\begin{equation}
\mathbf{h}_i=\sum_{j=1}^n {\color{drawred}g(\mathbf{x}_i, \mathbf{x}_j)}\mathbf{x}_j
\label{eq:nonlocal_convolution}
\end{equation}

In the context of computer vision, these models are also called \textbf{non-local} networks \cite{wang2018non}. We provide a comparison of standard convolutions, continuous convolutions, and non-local convolutions in Figure \ref{fig:biases}.

\subsection{The attention layer} \addclock

The MHA layer is a simplification of \eqref{eq:nonlocal_convolution}. First, working with functions having matrix outputs is difficult, so we restrict the layer to work with scalar weights. In particular, a simple measure of similarity between tokens is their inner (\textbf{dot}) product:

$$
g(\mathbf{x}_i, \mathbf{x}_j)=\mathbf{x}_i^\top\mathbf{x}_j
$$

As we will see, this results in an easily parallelizable algorithm for the entire sequence. For the following we consider a normalized version of the dot-product:

$$
g(\mathbf{x}_i, \mathbf{x}_j)=\frac{1}{\sqrt{e}}\mathbf{x}_i^\top\mathbf{x}_j
$$

This can be motivated as follows: if we assume $\mathbf{x}_i \sim \mathcal{N}(0, \sigma^2\mathbf{I})$, the variance of each element of $\mathbf{x}_i^\top\mathbf{x}_j$ is $\sigma^4$, hence the elements can easily grow very large in magnitude. The scaling factor ensures that the variance of the dot product remains bounded at $\sigma^2$.

Because we are summing over a potentially variable number of tokens $n$, it is also helpful to include a normalization operation, such as a softmax:\footnote{The notation $\text{softmax}_j$ in \eqref{eq:pre_self_attention} means we are applying the softmax normalization to the set $\left\{g(\mathbf{x}_i, \mathbf{x}_j)\right\}_{j=1}^n$, independently for each $i$. This is easier to see in the vectorized case, described below.}

\begin{equation}
\mathbf{h}_i=\sum_{j=1}^n {\color{drawgreen}\text{softmax}_j}(g(\mathbf{x}_i, \mathbf{x}_j))\mathbf{x}_j
\label{eq:pre_self_attention}
\end{equation}

In this context, we refer to $g(\bullet, \bullet)$ as the \textbf{attention scoring function}, and to the output of the softmax as the \textbf{attention scores}. Because of the normalization properties of the softmax, we can imagine that each token $i$ has a certain amount of “attention” it can allocate across the other tokens: by increasing the budget on a token, the attention over the other tokens will necessarily decrease due to the denominator in the softmax.

\vspace{-0.5em}
If we use a “dot-product attention”, our $g$ does not have trainable parameters. The idea of an attention layer is to recover them by adding trainable projections to the input before computing the previous equation. To this end, we define three trainable matrices ${\color{drawred}\mathbf{W}_k} \sim (e,k)$, ${\color{drawgreen}\mathbf{W}_v} \sim (e,v)$, ${\color{drawblue}\mathbf{W}_q} \sim (e,k)$, where $k$ and $v$ are hyper-parameters. Each token is projected using these three matrices, obtaining $3n$ tokens in total:

\begin{align}\text{Key tokens:} & \;\;\;{\color{drawred}\mathbf{k}_i}=\mathbf{W}_k^\top\mathbf{x}_i \label{eq:keys}\\
\text{Value tokens:} & \;\;\; {\color{drawgreen}\mathbf{v}_i}=\mathbf{W}_v^\top\mathbf{x}_i  \label{eq:values}\\ 
\text{Query tokens:} & \;\;\; {\color{drawblue}\mathbf{q}_i}=\mathbf{W}_q^\top\mathbf{x}_i\label{eq:queries}\end{align}

These processed tokens are called the \textbf{keys}, the \textbf{values}, and the \textbf{queries} (you can ignore the choice of terminology for now; we will return on this point at the end of the section). The \textbf{self-attention} (SA)  layer is obtained by combining the three projections \eqref{eq:keys}-\eqref{eq:values}-\eqref{eq:queries} with \eqref{eq:pre_self_attention}:

$$
\mathbf{h}_i=\sum_{j=1}^n \text{softmax}_j(g({\color{drawblue}\mathbf{q}_i}, {\color{drawred}\mathbf{k}_j}))\mathbf{\color{drawgreen}v_j}
$$

Hence, we compute the updated representation of token $i$ by comparing its query to all possible keys, and we use the normalized weights to combine the corresponding value tokens. Note that the dimensionality of keys and queries must be identical, while the dimensionality of the values can be different.

If we use the dot product, we can rewrite the operation of the SA layer compactly for all tokens. To this end, we define three matrices with the stack of all possible keys, queries, and values:
\begin{align}{\color{drawred}\mathbf{K}}=\mathbf{X}\mathbf{W}_k  \\ {\color{drawgreen}\mathbf{V}}=\mathbf{X}\mathbf{W}_v \\ {\color{drawblue}\mathbf{Q}}=\mathbf{X}\mathbf{W}_q \end{align}

The three derived matrices $\mathbf{K}, \mathbf{V}, \mathbf{Q}$ have shapes $(n, k)$, $(n, v)$, and $(n, k)$ respectively. As a side note, we can also implement them as a single matrix multiplication whose output is chunked in three parts:

$$
\left[ \mathbf{K} \mathbin\Vert \mathbf{V} \mathbin\Vert  \mathbf{Q} \right] = \mathbf{X} \left[ \mathbf{W}_k \mathbin\Vert \mathbf{W}_v \mathbin\Vert \mathbf{W}_q \right]
$$

where $\mathbin\Vert$ denotes concatenation. The SA layer is then written as:

$$
\text{SA}(\mathbf{X})=\text{softmax}\left(\frac{\mathbf{Q}\mathbf{K}^\top}{\sqrt{k}}\right)\mathbf{V}
$$

where we assume the softmax is applied row-wise. We can also make the projections explicit, as follows.

\begin{definition}[Self-attention layer] \addbottle $\,$

The \textbf{self-attention} (SA) layer is defined for an input $\mathbf{X} \sim (n,e)$ as:

\begin{equation}
\textnormal{SA}(\mathbf{X})=\textnormal{softmax}\left(\frac{\mathbf{X}\mathbf{W}_q\mathbf{W}_k^\top\mathbf{X}^\top}{\sqrt{k}}\right)\mathbf{X}\mathbf{W}_v
\end{equation}

The trainable parameters are $\mathbf{W}_q \sim (k,e)$, $\mathbf{W}_k \sim (k,e)$ and $\mathbf{W}_v \sim (v,e)$, where $k$ and $v$ are hyper-parameters. Hence, there are $2ke + ve$ trainable parameters, independent of $n$.

\end{definition}

\begin{figure}[t]
    \centering
    \includegraphics[width=\textwidth]{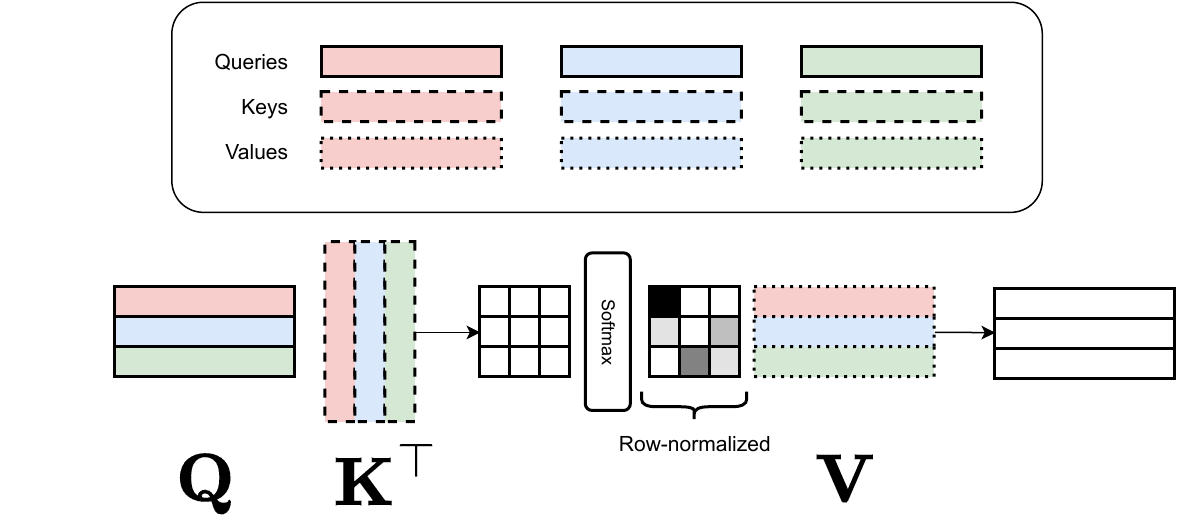}
    \caption{Visualization of the main operations of the SA layer (excluding projections).}
    \label{fig:self_attention}
\end{figure}

We show the operation of the layer visually in Figure \ref{fig:self_attention}.

\subsection{Multi-head attention}
\label{subsec:multi_head_attention}

The previous layer is also called a \textbf{single-head} attention operation. It allows to model pairwise dependencies across tokens with high flexibility. However, in some cases we may have multiple sets of dependencies to consider: taking again the example of “\textit{the cat, which belongs to my mother, is on the table}”, the dependencies between “\textit{cat}” and “\textit{table}” are different with respect to the dependencies between “\textit{cat}” and “\textit{mother}”, and we may want the layer to be able to model them separately.\footnote{And everything depends on the cat, of course.}

A multi-head layer achieves this by running multiple attention operations in parallel, each with its own set of trainable parameters, before aggregating the results with some pooling operation. To this end, we define a new hyper-parameter $h$, that we call the number of \textbf{heads} of the layer. We instantiate $h$ separate projections for the tokens, for a total of $3hn$ tokens ($3n$ for each “head”):
\begin{align}\mathbf{K}_e=\mathbf{X}\mathbf{W}_{k,e} \\ 
\mathbf{V}_e=\mathbf{X}\mathbf{W}_{v,e} \\ 
\mathbf{Q}_e=\mathbf{X} \mathbf{W}_{q,e}
\end{align}
$\mathbf{W}_{k,e}$ represents the key projection for the $e$-th head, and similarly for the other quantities. The \textbf{multi-head attention} (MHA) layer performs $h$ separate SA operations, stacks the resulting output embeddings, and projects them a final time to the desired dimensionality:

\vspace{1em}
\begin{equation}
\text{MHA}(\mathbf{X})=\begin{bmatrix}\eqnmarkbox[drawred]{node}{\text{SA}_1(\mathbf{X})} \;\mathbin\Vert\; \ldots \; \mathbin\Vert\;\text{SA}_h(\mathbf{X})\end{bmatrix}\eqnmarkbox[drawgreen]{node2}{\mathbf{W}_o}
\label{eq:mha_explicit}
\end{equation}
\annotate[yshift=1em]{above,right}{node}{Individual SA layer}
\annotate[yshift=-1em]{below,left}{node2}{Output projection}

\vspace{0.5em}
where:

$$
\text{SA}_i(\mathbf{X}) = \text{softmax}\left(\frac{\mathbf{Q}_i\mathbf{K}_i^\top}{\sqrt{k}}\right)\mathbf{V}_i
$$

Each SA operation returns a matrix of shape $(n,v)$. These $h$ matrices are concatenated across the second dimension to obtain a matrix $(n,hv)$, which is then projected with a matrix $\mathbf{W}_o \sim (hv, o)$, where $o$ is an additional hyper-parameter allowing flexibility in the choice of the output dimensionality.

\begin{supportbox}{Heads and circuits}

We will see shortly that the MHA layer is always combined with a residual connection (Section \ref{sec:residual_connections}). In this case we can write its output for the $i$-th token as:

\vspace{2em}
\begin{equation}
\mathbf{x}_i \leftarrow \mathbf{x}_i + \eqnmarkbox[drawred]{node}{\sum_e} \eqnmarkbox[drawgreen]{node2}{\sum_j} \alpha_e(\mathbf{x}_i, \mathbf{x}_j)\mathbf{W}_e^\top \mathbf{x}_j
\end{equation}
\annotate[yshift=1em]{above,right}{node}{Sum over heads}
\annotate[yshift=-1em]{below,right}{node2}{Sum over tokens}

\vspace{2em}
where $\alpha_e(\mathbf{x}_i, \mathbf{x}_j)$ is the attention score between tokens $i$ and $j$ in head $e$, and $\mathbf{W}_e$ combines the value projection of the $e$-th head with the $e$-th block of the output projection in \eqref{eq:mha_explicit}. The token embeddings are sometimes called the \textbf{residual stream} of the model.\footnote{This has been popularized in the context of \textbf{mechanistic interpretability}, which tries to retro-engineer the layers' behaviour to find interpretable components called \textit{circuits}: \url{https://transformer-circuits.pub}. The linearity of the stream is fundamental for the analisys.} Hence, the heads can be understood as “reading” from the residual stream (via the projection by $\mathbf{W}_e$ and the selection via the attention scores), and linearly “writing” back on the streams.
\end{supportbox}

\subsection*{An explanation of the terminology} \addteacup 

In order to understand why the three tokens are called queries, keys, and values, we consider the analogy of a SA layer with a standard Python dictionary, which is shown in Box \ref{code:dictionary}. 

\begin{mypy}{A dictionary in Python: a value is returned only if a perfect key-query match is found. Otherwise, we get an error.}{code:dictionary}
d = dict()
d["Alice"] = 2
d["Alice"]     # Returns 2
d["Alce"]      # Returns an error
\end{mypy}

Formally, a dictionary is a set of pairs of the form (key, value), where the key acts as an univocal ID to retrieve the corresponding value. For example, in the third and fourth line of Box \ref{code:dictionary} we query the dictionary with two different strings (“\textit{Alice}” and “\textit{Alce}”): the dictionary compares the query string to all keys which are stored inside, returning the corresponding value if a perfect match is found, an error otherwise.

\begin{figure}[t]
    \centering
    \includegraphics[width=1.0\textwidth]{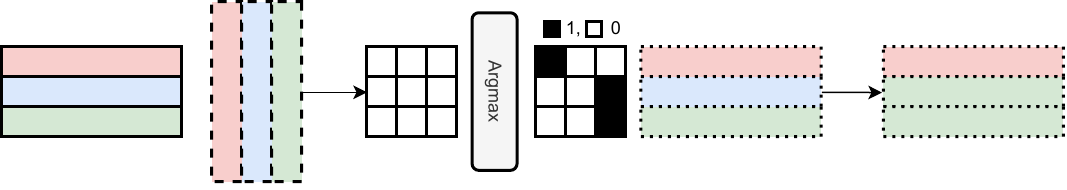}
    \caption{SA with a “hard” attention is equivalent to a vector-valued dictionary.}
    \label{fig:hard_attention}
\end{figure}

Given a measure of similarity over pair of keys, we can consider a variant of a standard dictionary which always returns the value corresponding to the closest key found in the dictionary. If the keys, queries, and values are vectors, this dictionary variant is equivalent to our SA layer if we replace the softmax operation with an argmax over the tokens, as shown in Figure \ref{fig:hard_attention}.

This “hard” variant of attention is difficult to implement because the gradients of the argmax operation are zero almost everywhere (we will cover discrete sampling and approximating the argmax operation with a discrete relaxation in the next volume). Hence, we can interpret the SA layer as a soft approximation in which each token is updated with a weighted combination of all values based on the corresponding key/query similarities. \clearpage

\section{Positional embeddings}
\label{sec:positional_embeddings}

With the MHA layer in hand, we consider the design of the complete transformer model, which requires another component, positional embeddings.

\subsection{Permutation equivariance}

It is interesting to consider what happens to the output of a MHA layer when the order of the tokens is re-arranged (\textit{permuted}). To formalize this, we introduce the concept of \textbf{permutation matrices}.

\begin{definition}[Permutation matrix] $\,$

A \textbf{permutation matrix} of size $n$ is a square binary matrix $\mathbf{P} \sim \text{Binary}(n,n)$ such that only a single 1 is present on each row or column:

$$
\mathbf{1}^\top\mathbf{P}=\mathbf{1},\;\; \mathbf{P}\mathbf{1}=\mathbf{1}
$$

If we remove the requirement for the matrix to have binary entries and we only constrain the entries to be non-negative, we obtain the set of \textbf{doubly stochastic} matrices (matrices whose rows and columns sum to one).

\end{definition}

The effect of applying a permutation matrix is to rearrange the corresponding rows / columns of a matrix. For example, consider the following permutation:

$$
\mathbf{P}=\begin{bmatrix} 1 & 0 & 0 \\ 0 & 0 & 1 \\ 0 & 1 & 0 \end{bmatrix}
$$

Looking at the rows, we see that the second and third elements are swapped by its application:

$$
\mathbf{P}\begin{bmatrix} {\color{drawred}\mathbf{x}_1} \\ {\color{drawgreen}\mathbf{x}_2} \\ {\color{drawblue}\mathbf{x}_3} \end{bmatrix} = \begin{bmatrix} {\color{drawred}\mathbf{x}_1} \\ {\color{drawblue}\mathbf{x}_3}  \\ {\color{drawgreen}\mathbf{x}_2}\end{bmatrix}
$$

Interestingly, the only effect of applying a permutation matrix to the inputs of a MHA layer is to rearrange the outputs of the layer in an equivalent way:

$$
\text{MHA}(\mathbf{P}\mathbf{X})=\mathbf{P}\,\cdot\,\text{MHA}(\mathbf{X})
$$

\begin{figure}[t]
    \centering
    \hspace*{2em}\includegraphics[width=\textwidth]{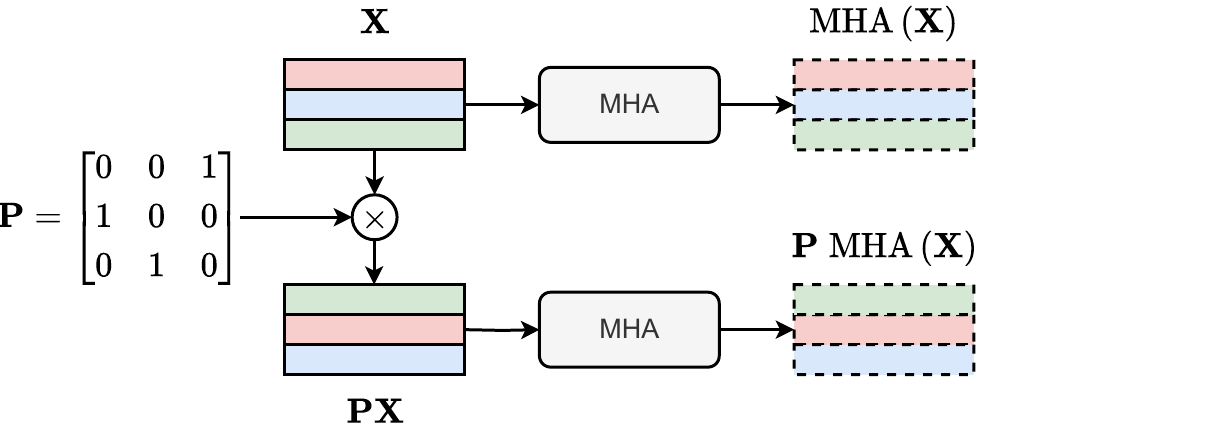}
    \caption{The output of a MHA layer after permuting the ordering of the tokens is trivially the permutation of the original outputs.}
    \label{fig:permutation_equivariance_mha}
\end{figure}

This is immediate to prove. We focus on the single headed variant as the multi-headed variant proceeds similarly. First, the softmax renormalizes the elements over the columns of a matrix, so it is trivially permutation equivariant across both rows and columns:

$$
\text{softmax}(\mathbf{P}\mathbf{X}\mathbf{P}^\top)=\mathbf{P}\left[\text{softmax}(\mathbf{X})\right]\mathbf{P}^\top
$$

From this we can immediately deduce the positional equivariance of SA:

\begin{gather}
\text{SA}(\mathbf{P}\mathbf{X}) = \text{softmax}\left(\mathbf{P}\frac{\mathbf{X}\mathbf{W}_q\mathbf{W}_k^\top\mathbf{X}^\top}{\sqrt{k}}\mathbf{P}^\top\right)\mathbf{P}\mathbf{X}\mathbf{W}_v \\ = \mathbf{P}\cdot \text{softmax}\left(\frac{\mathbf{X}\mathbf{W}_q\mathbf{W}_k^\top\mathbf{X}^\top}{\sqrt{k}}\right)\mathbf{X}\mathbf{W}_v = \mathbf{P}\cdot\text{SA}(\mathbf{X})
\end{gather}

where we make use of the fact that $\mathbf{P}^\top \mathbf{P} = \mathbf{I}$ for any permutation matrix. This can also be seen by reasoning on the SA layer for each token: the output is given by a sum of elements, each weighted by a pairwise comparison. Hence, for a given token the operation is \textbf{permutation invariant}. Instead, for the entire input matrix, the operation is \textbf{permutation equivariant}.

Translational equivariance was a desirable property for a convolutional layer, but permutation equivariance is \textit{undesirable} (at least here), because it discards the valuable ordering of the input sequence. As an example, the only effect of processing a text whose tokens have been reversed would be to reverse the output of the layer, despite the fact that the resulting reversed input is probably invalid. Formally, the SA and MHA layers are set functions, not sequence functions.\footnote{To be even more pedantic, they are \textit{multiset} functions since tokens can be repeated.}

Instead of modifying the layer or adding layers that are not permutation equivariant, the transformer operates by introducing a new concept of \textbf{positional embeddings}, which are auxiliary tokens that depend only on the position of a token in a sequence (\textbf{absolute positional embeddings}) or the offset of two tokens (\textbf{relative positional embeddings}). We describe the two in turn.

\subsection{Absolute positional embeddings}

\begin{figure}[t]
    \centering
    \hspace*{-3em}\includegraphics[width=0.7\textwidth]{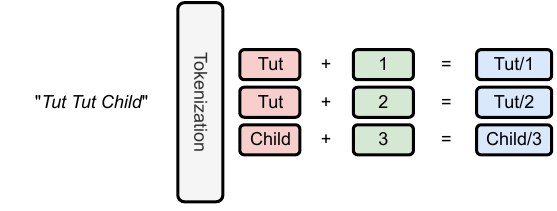}
    \caption{Positional embeddings (\colorbox{drawgreen!30}{green}) added to the  tokens' embeddings (\colorbox{drawred!30}{red}). The same token in different positions has different outputs (\colorbox{drawblue!30}{blue}).}
    \label{fig:positional_embeddings}
\end{figure}

Each token in the input matrix $\mathbf{X} \sim (n,e)$ represents the \textit{content} of the specific piece of text (e.g., a subword). Suppose we fix the maximum length of any sequence to $m$ tokens. To overcome positional equivariance, we introduce an additional set of \textbf{positional embeddings} $\mathbf{S} \sim (m,e)$, where the vector $\mathbf{S}_i$ uniquely encodes the concept of “being in position $i$”. Hence, the sum of the input matrix with the first rows of $\mathbf{S}$:
$$
\mathbf{X}^\prime =  \mathbf{X} + \mathbf{S}_{1:n}
$$
is such that $\idx{\mathbf{X}^\prime}{i}$ represents “token $\mathbf{X}_i$ in position $i$”.
Because it does not make sense to permute the positional embeddings (as they only depend on the position), the resulting layer is not permutation equivariant anymore:
$$
\text{MHA}(\mathbf{P}\mathbf{X} + \mathbf{S}) \neq \mathbf{P}\,\cdot\,\text{MHA}(\mathbf{X}+\mathbf{S})
$$
See Figure \ref{fig:positional_embeddings} for a visualization of this idea. 

How should we build positional embeddings? The easiest strategy is to consider $\mathbf{S}$ as part of the model's parameters, and train it together with the rest of the trainable parameters, similarly to the token embeddings. This strategy works well  when the number of tokens is relatively stable; we will see an example in the next chapter in the context of computer vision.

Alternatively, we can define some deterministic function from the set of tokens' positions to a given vector that uniquely identifies the position. Some strategies are clearly poor choices, for example:
\begin{enumerate}
\item We can associate to each position a scalar $p=i/m$ which is linearly increasing with the position. However, adding a single scalar to the token embeddings has a minor effect.
\item We can one-hot encode the position into a binary vector of size $m$, but the resulting vector would be extremely sparse and high-dimensional.
\end{enumerate}
A possibility, introduced in the original transformer paper \cite{vaswani2017attention}, is that of \textbf{sinusoidal embeddings}. To understand them, consider a sine function:
$$
y=\sin(x)
$$
The sine assigns a unique value to any input $x$ inside the range $[0, 2\pi]$. We can also vary the frequency of the sine:
$$
y=\sin(\omega x)
$$
This oscillates more or less rapidly based on the frequency $\omega$, and it assigns a unique value to any input in the range $[0, \frac{2\pi}{\omega}]$. 

There is an analogy with an (analogical) clock: the seconds' hand makes a full rotation with a frequency of $\frac{1}{60}$ Hz (once every minute). Hence, every “point in time” inside a minute can be distinguished by looking at the hand, but two time instants in general can only be identified modulo 60 seconds. We overcome this in a clock by adding a separate hand (the minute hand) that rotates with a much slower frequency of $\frac{1}{3600}$ Hz. Hence, by looking at the pair of coordinates (second, minute) (the “embedding” of time) we can distinguish any point inside an hour. Adding yet another hand with an even slower frequency (the hour hand) we can distinguish any point inside a day. This can be generalized: we could design clocks with lower or higher frequencies to distinguish months, years, or milliseconds.

\begin{SCfigure}
    \centering
    \hspace{1em}\includegraphics[width=0.6\textwidth]{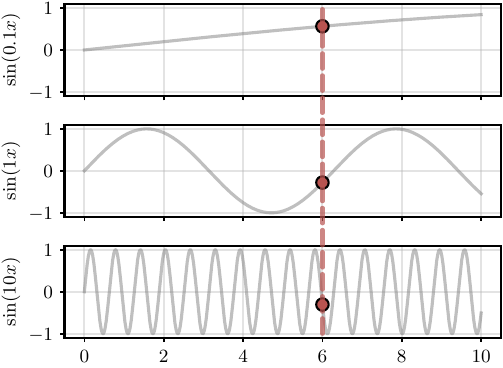}
    \caption{We show three $\sin$ functions with $\omega=0.1$, $\omega=1$, and $\omega=10$. The embedding for position $x=6$ is given by the corresponding values (red circles).}
    \label{fig:positional_embeddings_plot}
\end{SCfigure}

A similar strategy can be applied here: we can distinguish each position $i$ by encoding it through a set of $e$ sines (with $e$ an hyper-parameter) of increasing frequencies:
$$
\mathbf{S}_i=\left[ \sin(\omega_1i), \sin(\omega_2i),\ldots,\sin(\omega_ei) \right]
$$
In practice, the original proposal from \cite{vaswani2017attention} uses only $e/2$ possible frequencies, but adds both sines and cosines:
$$
\mathbf{S}_i=\left[ \sin(\omega_1i), \cos(\omega_1i),\ldots,\sin(\omega_{e/2}i), \cos(\omega_{e/2}i) \right]
$$
This can be justified by noting that in this embedding, two positions are related via a simple linear transformation, a rotation, that depends only on the relative offset of the two positions.\footnote{See \url{https://kazemnejad.com/blog/transformer_architecture_positional_encoding/} for a worked-out computation.} Any choice of frequency is valid provided they are sufficiently large and increasing at a super-linear rate. The choice from \cite{vaswani2017attention} was a geometric progression:
$$
\omega_i=\frac{1}{10000^{i/e}}
$$
that varies from $\omega_0=1$ to $\omega_e=\frac{1}{10000}$. See Figure \ref{fig:positional_embeddings_plot} for a visualization.

\subsection{Relative positional embeddings}

Trainable positional embeddings and sinuisodal positional embeddings are examples of \textbf{absolute} embeddings, because they encode a specific position in the sequence. An alternative that has become common with very long sequences are \textbf{relative positional embeddings}. In this case, instead of adding a positional encoding to a token, we modify the attention function to make it dependent on the offset between any two tokens:
$$
g(\mathbf{x}_i, \mathbf{x}_j)\rightarrow g(\mathbf{x}_i, \mathbf{x}_j, i-j)
$$
This is a combination of the two ideas we introduced at the beginning of this chapter (Figure \ref{fig:biases}). Note that while absolute embeddings are added only once (at the input), relative embeddings must be added every time an MHA layer is used. As an example, we can add a trainable bias matrix $\mathbf{B} \sim (m,m)$ and rewrite the dot product with an offset-dependent bias:
$$
g(\mathbf{x}_i,\mathbf{x}_j)=\mathbf{x}_i^\top\mathbf{x}_j+B_{ij}
$$
A simpler variant, \textbf{attention with linear biases} (ALiBi) \cite{press2021train}, considers a single trainable scalar in each head which is multiplied by a matrix of offsets. More advanced strategies, such as \textbf{rotary positional embeddings} (RoPE), are also possible \cite{su2024roformer}.

\section{Building the transformer model}
\subsection{The transformer block and model}
\label{subsec:transformer_block}

A model could be built, in principle, from a stack of multiple MHA layers (with the softmax providing the non-linearity necessary to avoid the collapse of multiple linear projections). Empirically, however, it is found that the MHA works best when interleaved with a separate fully-connected block that operates on each token independently. These two operations can be understood as mixing the tokens (MHA), and mixing the channels (MLP), similarly to the depthwise-separable convolution  model.

In particular, for the MLP block it is common to choose a bottleneck architecture composed of two fully-connected layers of the form:
$$
\text{MLP}(\mathbf{x})=\mathbf{W}_2\phi\left(\mathbf{W}_1\mathbf{x}\right)
$$
where $\mathbf{x} \sim (e)$ is a token, $\mathbf{W}_1 \sim (p, e)$, with $p$ selected as an integer multiple of $e$ (e.g., $p=3e$ or $p=4e$), and $\mathbf{W}_2 \sim (e,p)$ reprojecting back to the original embedding dimension. Biases are generally removed as the increased hidden dimension provides sufficient degrees of freedom.

\begin{SCfigure}
    \centering
    \hspace{2em}\includegraphics[width=0.4\textwidth]{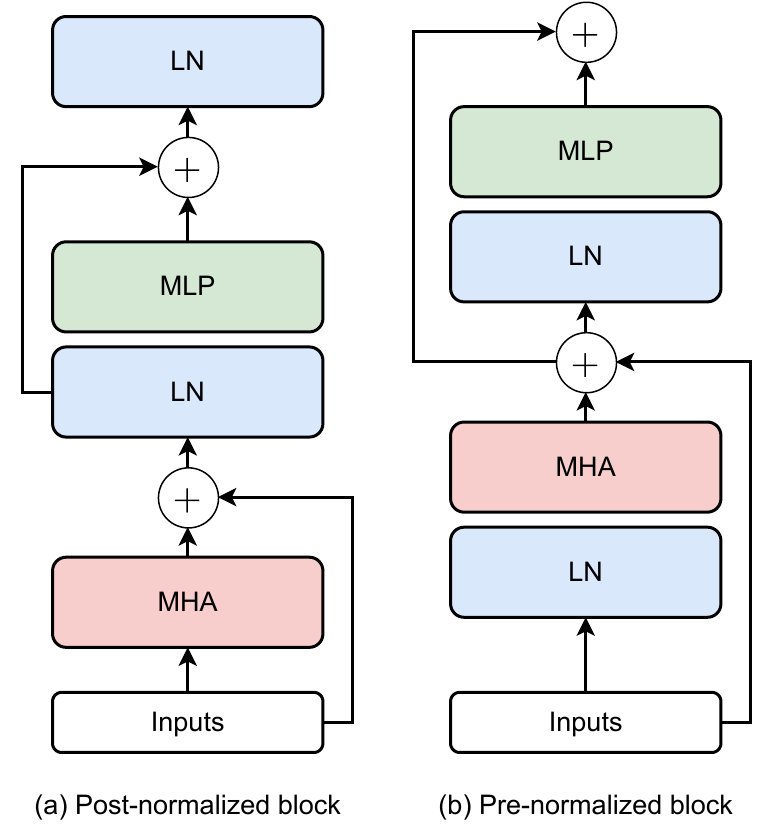}
    \caption{Schematic view of pre-normalized and post-normalized transformer blocks. In the post-normalized variant the LN block is applied after the MHA or MLP operation, while in the pre-normalized one before each layer.}
    \label{fig:pre_post_normalization}
\end{SCfigure}

To ensure efficient training of deep models we also need a few additional regularization strategies. In particular, it is common to include two layer normalization steps and two residual connections, respectively for the MHA and MLP blocks. Depending on where the layer normalization is applied, we obtain two variants of the basic transformer block, sometimes denoted as \textbf{pre-normalized} and \textbf{post-normalized}. These are shown in Figure \ref{fig:pre_post_normalization}.

While the post-normalized version corresponds to the original transformer block, the pre-normalized variant is generally found to be more stable and faster to train \cite{xiong2020layer}. The design of the block in Figure \ref{fig:pre_post_normalization} is, fundamentally, an empirical choice, and many variants have been proposed and tested in the literature. We review some of these later on in Section \ref{subsec:mha_variants}.

We can now complete the description of a basic transformer model:
\begin{enumerate}
\item Tokenize and embed the original input sequence in a matrix $\mathbf{X} \sim (n,e)$.\
\item If using absolute positional embeddings, add them to the input matrix.
\item Apply 1 or more blocks of the form discussed above.
\item Include a final head depending on the task.
\end{enumerate}
The output of step (3) is a set of processed tokens $\mathbf{H} \sim (n,e)$, where neither $n$ nor $e$ are changed by the transformer model (the former because we do not have local pooling operations on sets, the latter because of the residual connections in the block). Considering for example a classification task, we can apply a standard classification head by pooling over the tokens and proceeding with a fully-connected block:
$$
y=\text{softmax}\left(\text{MLP}\left(\frac{1}{n}\sum_i\mathbf{H}_i\right)\right)
$$
This part is identical to its corresponding CNN design. However, the transformer has a number of interesting properties, mostly stemming by the fact that it manipulates its input as a (multi)set, without modifying its dimensionality throughout the architecture. We investigate one simple example next. \clearpage

\subsection{Class tokens and register tokens}
\label{subsec:class_register_tokens}

While up to now we have assumed that each token corresponds to one part of our input sequence, nothing prevents us from adding \textit{additional} tokens to the input of the transformer. This is strictly dependent on its specific architecture: a CNN, for example, requires its input to be precisely ordered, and it is not clear how we could add additional tokens to an image or to a sequence. This is a very powerful idea, and we only consider two specific implementations here.

First, we consider the use of a \textbf{class token} \cite{dosovitskiy2020image}, an additional token which is added explicitly for classification in order to replace the global pooling operation above. Suppose we initialize a single trainable token $\mathbf{c} \sim (e)$, which is added to the input matrix:
$$
\mathbf{X} \leftarrow\begin{bmatrix}\mathbf{X} \\ \mathbf{c}^\top \end{bmatrix}
$$
The new matrix has shape $(n+1, e)$. The class token is identical for all sequences in a mini-batch. After step (3) above, the transformer outputs a matrix $\mathbf{H} \sim (n+1,e)$ of updated representations for all tokens, including the class one. The idea is that, instead of pooling over the tokens, the model should be able to “compress” all information related to the classification task inside the class token, and we can rewrite the classification head by simply discarding all other tokens:\footnote{In the language of circuits and heads from Section \ref{subsec:multi_head_attention}, we could say equivalently that the model must learn to move all information related to the task in the residual stream of the class token.}
$$
y=\text{softmax}\left(\text{MLP}\left(\mathbf{H}_{n+1}\right)\right)
$$
Additional trainable tokens can be useful even if not explicitly used. For example, \cite{darcet2023vision} has shown that adding a few additional tokens (called \textbf{registers} in this case) can improve the quality of the attention maps by providing the model with the possibility of using the registers to “store” auxiliary information that does not depend explicitly on a given position.

\section*{From theory to practice}

\begin{wrapfigure}{r}{3.0cm}
\vspace{-6em}\includegraphics[width=3.0cm]{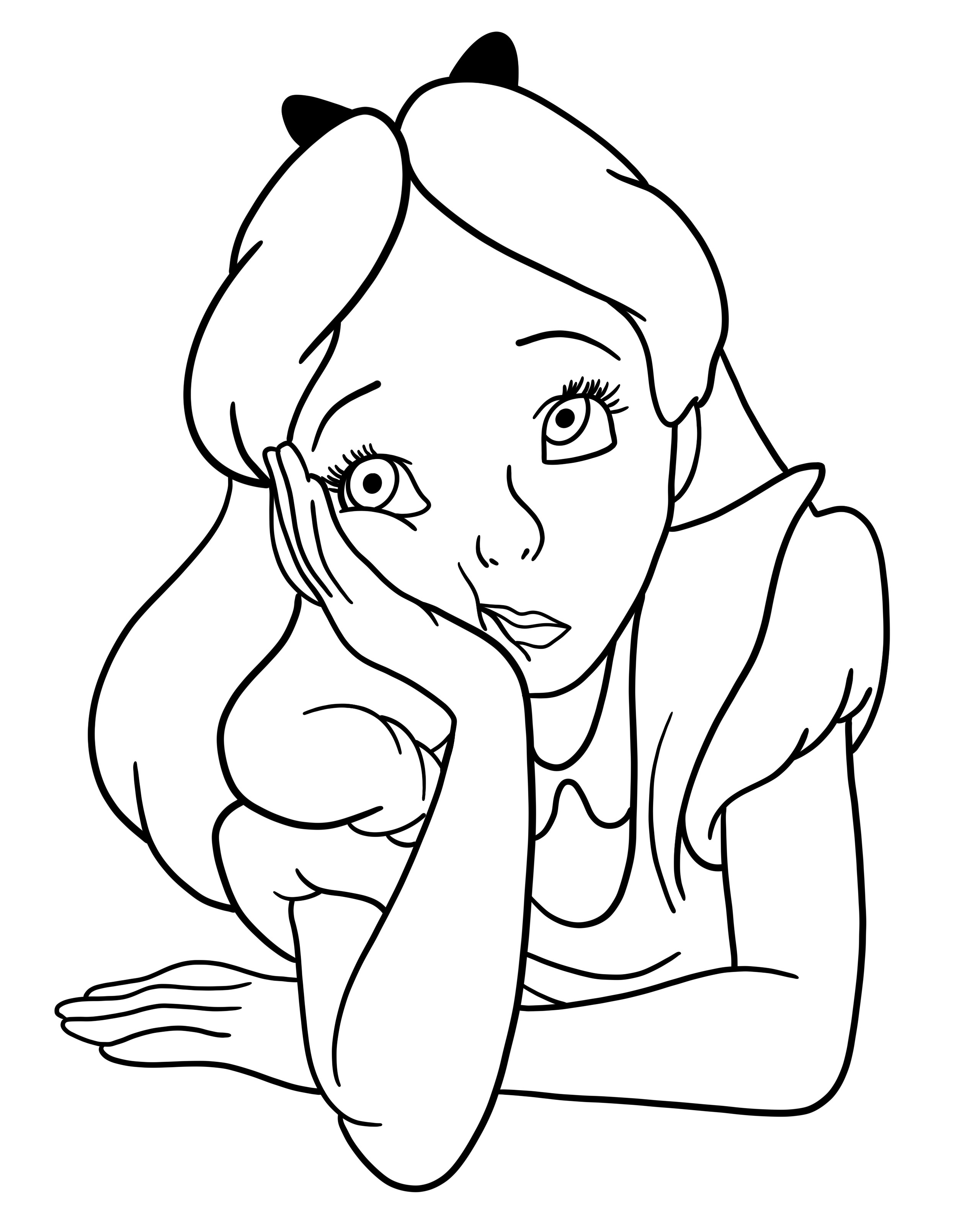}
\vspace{-3em}
\end{wrapfigure}

\begin{figure}[t]
    \centering
    \hspace{1em}\includegraphics[width=\textwidth]{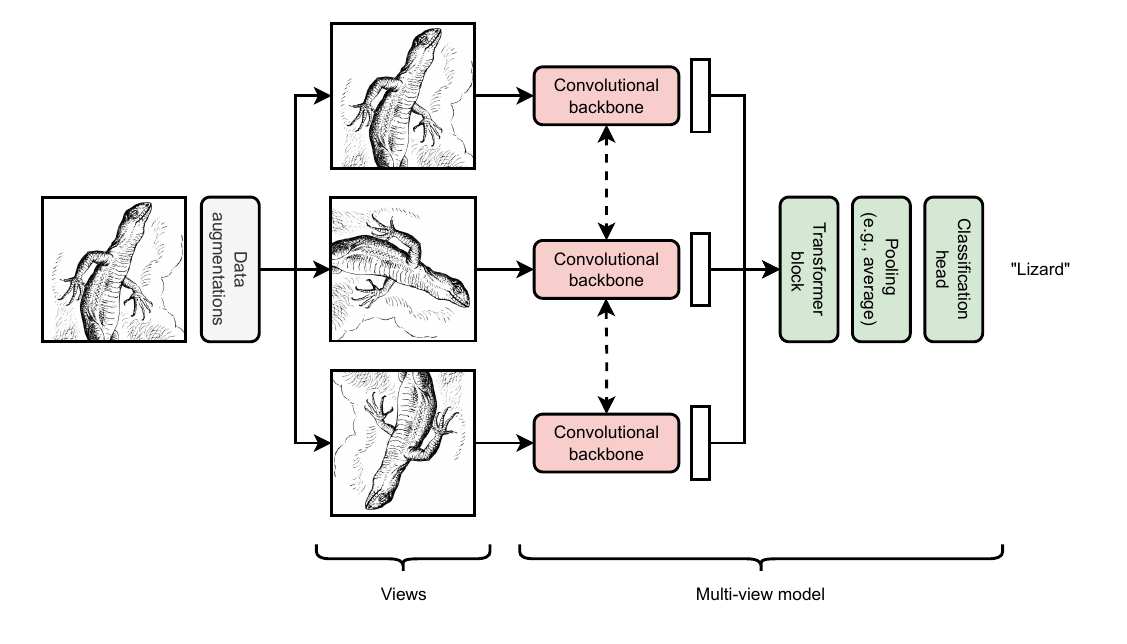}
    \caption{Multi-view model to be implemented in this chapter. The image is augmented through a set of random data augmentation strategies to obtain a set of \textbf{views} of the input (\colorbox{gray!30}{gray}). Each \textit{view} is processed by the same convolutional backbone to obtain a fixed-sized dimensional embedding (\colorbox{drawred!30}{red}). The set of embeddings are processed by a transformer block before the final classification (\colorbox{drawgreen!30}{green}). Illustration by John Tenniel.}
    \label{fig:multiview_model}
\end{figure}

We will introduce many important concepts related to transformers in the next chapter. Thus, for this chapter I am suggesting a slightly unconventional exercise which combines a convolutional backbone with a transformer-like head -- as depicted in Figure \ref{fig:multiview_model}.

The convolutional models you developed in Chapters \ref{chap:cnns} and \ref{chap:deep_cnns} were applied to a \textit{single} image. However, sometimes we have available a \textit{set} of images of the same object to be recognized -- for example, in a monitoring system, we may have multiple screenshots of a suspicious person. This is called a \textbf{multi-view} system in the literature, and each image is called a \textbf{view} of the object. A multi-view model should provide a single prediction for the entire set of views, while being invariant to the order of the views in input. For this exercise we will implement a simple multi-view model -- see Figure \ref{fig:multiview_model}.

\begin{enumerate}
\item Using any image classification dataset, you can simulate a multi-view model by applying a fixed number of data transformations to the input (gray block in Figure \ref{fig:multiview_model}). Ignoring the batch dimension, for each input image of shape $x \sim (h,w,c)$ (height, width, channels), you obtain a \textit{multi-view} input of shape $x^\prime \sim (v,h,w,c)$, where $v$ is the number of views. A single label $y$ is associated to this tensor -- the label of the original image. The number of views can also be different from mini-batch to mini-batch, as no part of the model is constrained to a pre-specified number of view.
\item The multi-view model is composed of three components. Denote by $g(x)$ a model that processes a single view to a fixed-dimensional embedding -- for example, this can be any convolutional backbone you trained for the previous exercises. The first part of the full model (red part in Figure \ref{fig:multiview_model}) applies $g$ in parallel to all views, $\mathbf{h}_i = g(x_i) \sim (e)$, where $e$ is a hyper-parameter (the output size of the backbone).
\item After concatenating the embeddings of the views we obtain a matrix $\mathbf{H} \sim (v, e)$. In order for the full model to be permutation invariant, any component applied on $\mathbf{H}$ must be permutation equivariant.\footnote{An average operation over the views is the simplest example of permutation invariant layer. Hence, removing the MHA block from Figure \ref{fig:multiview_model} is also a valid baseline. Alternatively, deep sets \cite{zaheer2017deep} characterize the full spectrum of linear, permutation invariant layers.} For the purposes of this exercise, implement and apply a single transformer block as per Section \ref{subsec:transformer_block}. You can implement MHA using basic PyTorch, or you can try a more advanced implementation using {\footnotesize\texttt{einops}}.\footnote{See \url{https://einops.rocks/pytorch-examples.html}.} You can also compare with the pre-implemented version in \mintinline{python}{torch.nn}.
\item The transformer block does not modify the input shape. To complete the model, perform an average over the views (which represent the tokens in this scenario), and apply a final classification head. You can also experiment with adding a class token (Section \ref{subsec:class_register_tokens}). It is easy to show that a model built in this way is permutation invariant with respect to the views.
\end{enumerate}

%% file: 11_advanced_transformers.tex
\chapter[Transformers in practice]{Transformers \\ in practice}
\label{chap:transformers_in_practice}

\begin{supportbox}{About this chapter}
We now consider a few variations of the basic transformer model, including encoder-decoder architectures, causal MHA layers, and applications to the image and audio domains.
\end{supportbox}

\section{Encoder-decoder transformers}

The model we described in Chapter \ref{chap:transformers} can be used to perform regression or classification of a given sequence. However, the original transformer \cite{vaswani2017attention} was a more complex model, designed for what are called \textbf{sequence-to-sequence} (seq2seq) tasks. In a seq2seq task, both input and output are sequences, and there is no trivial correspondence between their tokens. A notable example is \textbf{machine translation}, in which the output is the translation of the input sequence in a different language.

One possibility to build a differentiable model for seq2seq tasks is an \textbf{encoder-decoder} (ED) design \cite{sutskever2014sequence}. An ED model is composed of two blocks: an encoder that processes the input sequence to a transformed representation (possibly of a fixed dimensionality), and a decoder that autoregressively generates the output sequence conditioned on the output of the encoder. The transformer model we described before can be used to build the encoder: transformers of this type for classification are called \textbf{encoder-only} transformers. In order to build the decoder we need two additional components: a way to make the model causal (to perform autoregression), and a way to condition its computation to a separate input (the output of the encoder).

\subsection{Causal multi-head attention} \addclock

Let us consider first the problem of making the transformer block causal. The only component in which tokens interact is the MHA block. Hence, having a causal variant of MHA is enough to make the entire model causal. Remember that, for convolutions, we designed a causal variant by appropriately masking the weights in the convolutional filter. For MHA, we can mask instead all interactions between tokens that do not satisfy the causality property:
$$
\text{Masked-SA}(\mathbf{X})=\text{softmax}\left(\frac{\mathbf{Q}\mathbf{K}^\top {\color{drawred_l}\odot \,\mathbf{M}}}{\sqrt{k}}\right)\mathbf{V}
$$
It is essential to perform the masking inside the softmax. Consider the following (wrong) variant:
$$
\text{\color{red}\textbf{Wrong}:} \;\;\left(\text{softmax}\left(\frac{\mathbf{Q}\mathbf{K}^\top }{\sqrt{k}}\right){\color{drawred_l}\odot \,\mathbf{M}}\right)\mathbf{V}
$$
Because of the denominator in the softmax, all tokens participate in the computation of each token, irrespective of the later masking. Also note that setting $M_{ij}=0$ for non-causal links does not work, because $\exp(0)=1$. Hence, the correct implementation of a masked variant of MHA is to select an upper triangular matrix with $-\infty$ on the upper part, since $\exp(-\infty)=0$ as desired:
$$
M_{ij} =\begin{cases} -\infty & \text{ if } i > j \\ 1 & \text{ otherwise } \end{cases}
$$
Practically, the values can be set to a very large, negative number instead (e.g., $-10^9$).

\begin{figure}[t]
    \centering
    \hspace{1em}\includegraphics[width=0.8\textwidth]{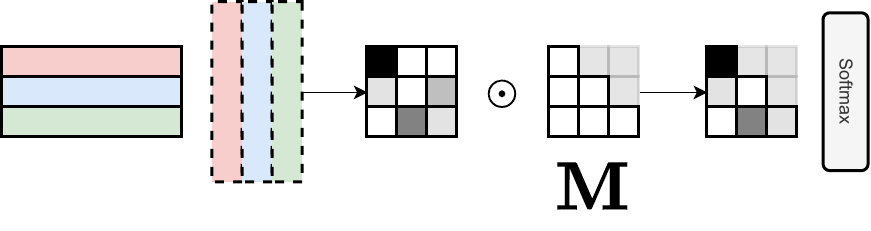}
    \caption{Visual depiction of causal attention implemented with attention masking.}
    \label{fig:masked_attention}
\end{figure}

\subsection{Cross-attention}

Second, let us consider the problem of making the output of the MHA layer conditional on a separate block of inputs. To this end, let us rewrite the MHA operation by explicitly separating the three appearances of the input matrix:

$$
\text{SA}({\color{drawred}\mathbf{X}_1}, {\color{drawgreen}\mathbf{X}_2}, {\color{drawblue}\mathbf{X}_3})=\text{softmax}\left(\frac{{\color{drawred}\mathbf{X}_1}\mathbf{W}_q\mathbf{W}_k^\top{\color{drawgreen}\mathbf{X}_2^\top}}{\sqrt{k}}\right){\color{drawblue}\mathbf{X}_3}\mathbf{W}_v
$$

The SA layer corresponds to $\mathbf{X}_1 = \mathbf{X}_2 = \mathbf{X}_3 = \mathbf{X}$ (which, coincidentally, explains the name we gave to it). However, the formulation also works if we consider keys, values, and queries belonging to separate sets. One important case is \textbf{cross-attention} (CA), in which we assume that the keys and values are computed from a second matrix $\mathbf{Z} \sim (m,e)$:

\begin{equation}
\text{CA}(\mathbf{X}, \mathbf{Z}) = \text{softmax}\left(\frac{\eqnmarkbox[drawred]{node}{\mathbf{X}\mathbf{W}_q\mathbf{W}_k^\top\mathbf{Z}^\top}}{\sqrt{k}}\right)\mathbf{Z}\mathbf{W}_v
\label{eq:ca}
\end{equation}
\annotate[yshift=1em]{above,left}{node}{Cross-attention between $\mathbf{X}$ and $\mathbf{Z}$}

such that $\text{CA}(\mathbf{X}, \mathbf{Z}) = \text{SA}(\mathbf{X}, \mathbf{Z}, \mathbf{Z})$. The interpretation is that the embeddings of $\mathbf{X}$ are updated based on their similarity with a set of external (key, values) pairs provided by $\mathbf{Z}$: we say that $\mathbf{X}$ is \textit{cross-attending} on $\mathbf{Z}$. Note that this formulation is very similar to a concatenation of the two sets of input tokens followed by an appropriate masking of the attention matrix.

\subsection*{Comparison with feedforward layers} \addteacup
\label{subsec:comparison_ca_mlp}

Consider a simplified variant of the cross-attention operation in \eqref{eq:ca}, in which we parameterize explicitly the keys and values matrices:\footnote{See also the discussion on the perceiver network in Section \ref{subsec:time_complexity_mha}.}

\begin{equation}
\text{NeuralMemory}(\mathbf{X}) = \text{softmax}\left(\frac{\mathbf{X}\mathbf{W}_q {\color{drawred}\mathbf{K}}}{\sqrt{k}}\right){\color{drawgreen}\mathbf{V}}
\label{eq:memory_layer}
\end{equation}

The layer is now parameterized by a query projection matrix $\mathbf{W}_q$ and by the two matrices ${\color{drawred}\mathbf{K}}$ and ${\color{drawgreen}\mathbf{V}}$. \eqref{eq:memory_layer} is called a \textbf{memory layer} \cite{sukhbaatar2015end}, in the sense that rows of the key and value matrices are used by the model to store interesting patterns to be retrieved dynamically by an attention-like operation. If we further simplify the layer by setting $\mathbf{W}_q = \mathbf{I}$, ignoring the normalization by $\sqrt{k}$, and replacing the softmax with a generic activation function $\phi$, we obtain a two-layer MLP:

\begin{equation}
\text{MLP}(\mathbf{X}) = \phi\left(\mathbf{X}\mathbf{K}\right)\mathbf{V}
\end{equation}

Hence, MLPs in transformer networks can be seen as approximating an attention operation over trainable keys and values. Visualizing the closest tokens in the training data shows human-understandable patterns \cite{geva2020transformer}.

\subsection{The encoder-decoder transformer}

\begin{figure}[t]
    \centering
    \includegraphics[width=0.9\textwidth]{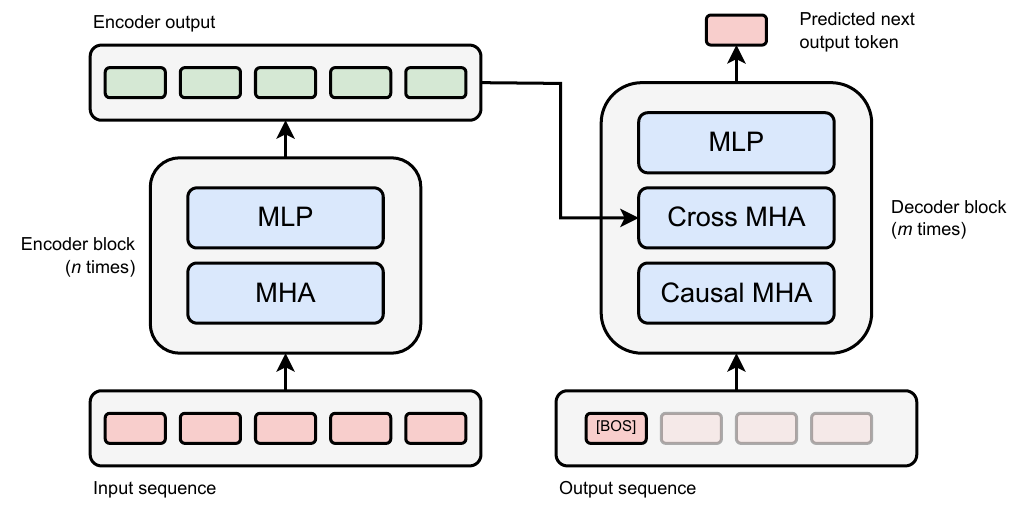}
    \caption{Encoder-decoder architecture, adapted from \cite{vaswani2017attention}. Padded tokens in the decoder are greyed out.}
    \label{fig:transformer_model}
\end{figure}

With these two components in hand, we are ready to discuss the original transformer model, shown in Figure \ref{fig:transformer_model}.\footnote{A pedantic note: technically, Transformer (upper-cased) is a proper noun in \cite{vaswani2017attention}. In the book, I use transformer (lower-cased) to refer to any model composed primarily of attention layers.} First, the input sequence $\mathbf{X}$ is processed by a standard transformer model (called the \textbf{encoder}), providing an updated embedding sequence $\mathbf{H}$. Next, the output sequence is predicted autoregressively by another transformer model (called the \textbf{decoder}). Differently from the encoder, the decoder has three components for each block:

\begin{enumerate}
\item A masked variant of the MHA layer (to ensure autoregression is possible).
\item A cross-attention layer where the queries are given by the input sequence embedding $\mathbf{H}$.
\item A standard token-wise MLP.
\end{enumerate}

\textbf{Decoder-only} models are also possible, in which case the second block of the decoder is removed and only masked MHA and MLPs are used. Most modern LLMs are built by decoder-only models trained to autoregressively generate text tokens \cite{radford2019language}, as discussed below. In fact, encoder-decoder models have become less common with the realization that many seq2seq tasks can be solved directly with decoder-only models by concatenating the input sequence to the generated output sequence, as described in Section \ref{subsec:conditional_modeling}.

\section{Computational considerations}

\subsection{Time complexity and linear-time transformers}
\label{subsec:time_complexity_mha}

The MHA performance does not come without a cost: since every token must attend to all other tokens, its complexity is higher than a simpler convolutional operation. To understand this, we look at its complexity from two points of view: memory and time. We use a naive implementation of the SA layer for reference, shown in Box \ref{code:sa}.

\begin{mypy}{Simple implementation of the SA layer, explicitly parameterized in terms of the query, key, and value matrices.}{code:sa}
def self_attention(Q: Float[Array, "n k"], 
                   K: Float[Array, "n k"], 
                   V: Float[Array, "n v"]
                 ) -> Float[Array, "n v"]:
	return nn.softmax(Q @ K.T) @ V
\end{mypy}

Let us look first at the time complexity. The operation inside the softmax scales as $\mathcal{O}(n^2k)$ because it needs to compute $n^2$ dot products (one for each pair of tokens). Compare this to a 1D convolutional layer, which scales only linearly in the sequence length. \textit{Theoretically}, this quadratic growth in complexity can be problematic for very large sequences, which are common in, e.g., LLMs. 

This has led to the development of several strategies for speeding up autoregressive generation (e.g., speculative decoding \cite{leviathan2023fast}), as well as linear or sub-quadratic variants of transformers. As an example, we can replace the SA layer with a cross-attention layer having a \textit{trainable} set of tokens $\mathbf{Z}$, where the number of tokens can be chosen as hyper-parameter and controlled by the user. This strategy was popularized by the Perceiver architecture \cite{jaegle2021perceiver} to distill the original set of tokens into smaller latent bottlenecks. There are many alternative strategies for designing linearized transformers: we discuss a few  variants in Section \ref{subsec:mha_variants} and Chapter \ref{chap:rnns}.

Importantly, an implementation such as the one in Box \ref{code:sa} can be shown to be heavily memory-bound on modern hardware \cite{dao2022flashattention}, meaning that its compute cost is dominated by memory and I/O operations. Hence, the theoretical gains of linear-time attention variants are not correlated with actual speedup on hardware. Combined with a possible reduction in performance, this makes them less attractive than a strongly-optimized implementation of MHA, such as the one described next.

\subsection{Online softmax}

In terms of memory, the implementation in Box \ref{code:sa} has also a quadratic $n^2$ complexity factor because the attention matrix $\mathbf{Q}\mathbf{K}^\top$ is fully materialized during computation. However, this is unnecessary and this complexity can be drastically reduced to a linear factor by chunking the computation in blocks and only performing the softmax normalization at the end \cite{rabe2021self}. 

To understand this, consider a single query vector $\mathbf{q}$, and suppose we split our keys and values into two blocks, which are loaded in turn in memory:

\begin{equation}
\mathbf{K} = \begin{bmatrix}\mathbf{K}_1 \\ \mathbf{K}_2 \end{bmatrix} \;,\;\;\; \mathbf{V} = \begin{bmatrix}\mathbf{V}_1 \\ \mathbf{V}_2 \end{bmatrix}
\end{equation}

If we ignore the denominator in the softmax, we can decompose the SA operation, computing the output for each chunk in turn:

\begin{equation}
\text{SA}(\mathbf{q}, \mathbf{K}, \mathbf{V}) = \frac{1}{L_1 + L_2} \left[ \mathbf{h}_1 + \mathbf{h}_2 \right]
\label{eq:attention_two_chunks}
\end{equation}

where for the two chunks $i=1,2$ we have defined two auxiliary quantities:
\begin{gather}
\mathbf{h}_i = \exp\left( \mathbf{K}_i\mathbf{q} \right)\mathbf{V}_i  \\
L_i = \sum_j \idx{\exp\left( \mathbf{K}_i\mathbf{q} \right)}{j}
\end{gather}
Remember we are loading the chunks in memory separately, hence for chunk $1$ we compute $\mathbf{h}_1$ and $L_1$; then we offload the previous chunk and we compute $\mathbf{h}_2$ and $L_2$ for chunk 2. Note that the operation is not fully-decomposable unless we keep track of the additional statistics $L_i$ (which is needed to compute the normalization coefficients of the softmax operation). More in general, for multiple chunks $i=1, \ldots, m$ we will have:

\begin{equation}
\text{SA}(\mathbf{q}, \mathbf{K}, \mathbf{V}) = \frac{1}{\sum_{i=1}^m L_i} \left[ \sum_{i=1}^m \mathbf{h}_i\right]
\label{eq:chunked_sa}
\end{equation}

\begin{figure}[t]
    \centering
    \includegraphics[width=0.8\textwidth]{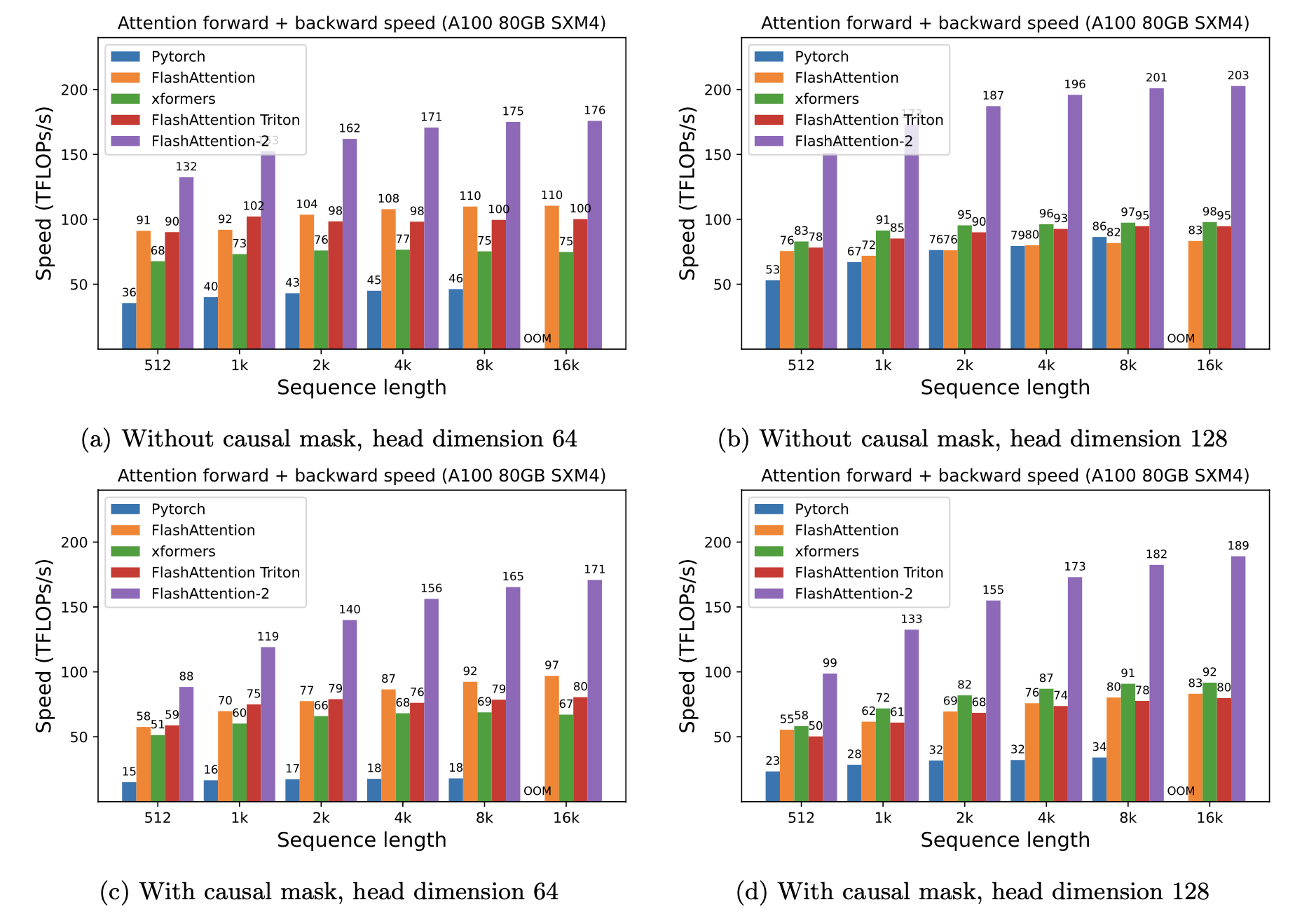}
    \caption[Official benchmark of FlashAttention and FlashAttention-2 on an NVIDIA A100 GPU card.]{Official benchmark of FlashAttention and FlashAttention-2 on an NVIDIA A100 GPU card, reproduced from \url{https://github.com/Dao-AILab/flash-attention}.}
    \label{fig:flash_attention}
\end{figure}

Hence, we can design a simple iterative algorithm where for every block of keys and values loaded in memory, we update and store the cumulative sum of the numerator and denominator in \eqref{eq:chunked_sa}, only performing the normalization at the end. This trick (sometimes called \textit{online} softmax), combined with an IO-aware implementation and kernel fusion has led to highly memory- and compute- efficient implementations of attention such as FlashAttention-2.\footnote{\url{https://github.com/Dao-AILab/flash-attention}} Distributed implementations of attention (e.g., RingAttention \cite{liu2023ring}) can also be devised by assigning groups of queries to different devices and rotating chunks of keys and queries among the devices. Optimizing the operation for specific hardware can lead to some counter-intuitive behaviours, such as \textit{increased} speed for larger sequence lengths - see Figure \ref{fig:flash_attention}.

\subsection{The KV cache}

\begin{SCfigure}
    \centering
    \hspace{-0.5em}\includegraphics[width=0.6\textwidth]{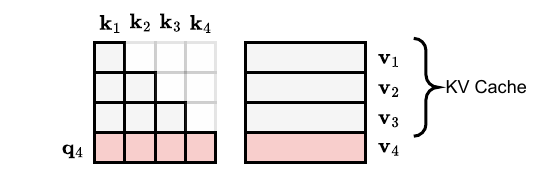}
    \caption{To compute masked self-attention on a new token, most of the previous computation can be reused (in gray). This is called the \textbf{KV cache}.}
    \label{fig:kv_cache}
\end{SCfigure}

An important implementative aspect of MHA occurs when dealing with autoregressive generation in decoder-only models. For each new token to be generated, only a new row of the attention matrix and one value token must be computed, meaning that the previous keys and values can be stored in memory, as shown in Figure \ref{fig:kv_cache}. This is called the \textbf{KV cache} and it is a standard in most optimized implementations of MHA.

The size of the KV cache is linearly increasing in the sequence length. Once again, you can compare this to an equivalent implementation of a causal convolutional layer, where memory is upper-bounded by the size of the receptive field. Designing expressive layers with a fixed memory cost in autoregressive generation is a motivating factor for Chapter \ref{chap:rnns}.

\subsection{Transformers for images and audio} \addteacup
\label{subsec:transformers_image_audio}

Transformers were originally developed for text, and they soon became the default choice for language modeling. In particular, the popular GPT-2 model \cite{radford2019language} (and later variants) is a decoder-only architecture which is pre-trained by forecasting tokens in text sequences. Most open-source LLMs, such as LLaMa \cite{touvron2023llama}, follow a similar architecture. By constrast, BERT \cite{devlin2018bert} is another popular family of pre-trained word embeddings based on an encoder-only architecture trained to predict masked tokens (\textbf{masked language modeling}). Differently from GPT-like models, BERT-like models cannot be used to generate text but only to perform text embedding or as the first part of a fine-tuned architecture. Encoder-decoder models for language modeling also exist (e.g., the T5 family \cite{raffel2020exploring}), but they have become less popular.\footnote{\textbf{Diffusion language models} \cite{yang2025mmada} (DLMs) are a recent alternative to autoregressive LLMs, and they are trained with a denoising objective vaguely reminiscent of BERT models. All types of LLMs undergo several \textit{post-training} steps beyond token prediction (e.g., instruction tuning), which we do not have space to cover here.}

From a high-level point of view, a transformer is composed of three components: a tokenization / embedding step, which converts the original input into a sequence of vectors; positional embeddings to encode information about the ordering of the original sequence; and the transformer blocks themselves. Hence, transformers for other types of data can be designed by defining the appropriate tokenization procedure and positional embeddings.

\begin{figure}[t]
    \centering
    \includegraphics[width=1.0\textwidth]{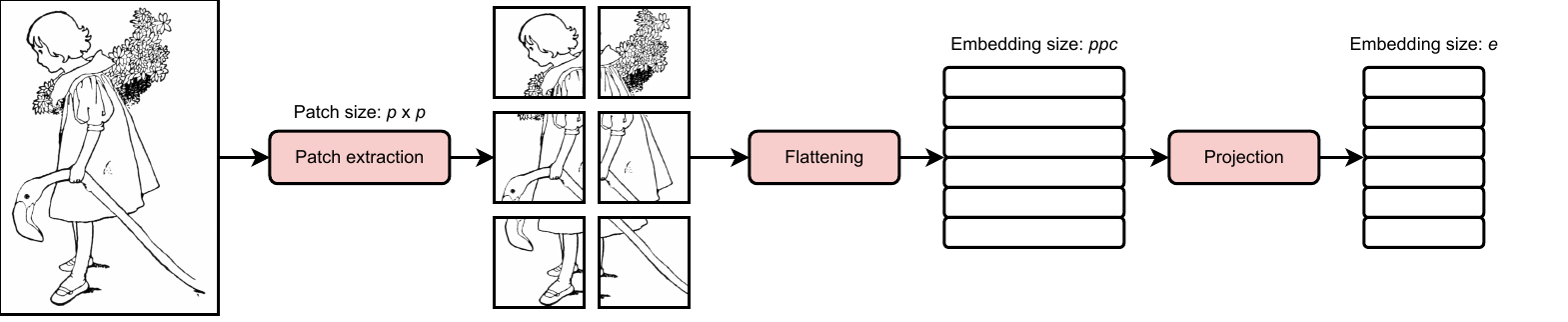}
    \caption{Image tokenization: the image is split into non-overlapping patches of shape $p \times p$ (with $p$ an hyper-parameter). Then, each patch is flattened and undergoes a further linear projection to a user-defined embedding size $e$. $c$ is the number of channels of the input image.}
    \label{fig:image_tokenization}
\end{figure}

Let us consider first computer vision. Tokenizing an image at the pixel level is too expensive, because of the quadratic growth in complexity with respect to the sequence length. The core idea of Vision Transformers (ViTs, \cite{dosovitskiy2020image}) is to split the original input into non-overlapping \textbf{patches} of fixed length, which are then flattened and projected to an embedding of pre-defined size, as shown in Figure \ref{fig:image_tokenization}.

The embedding step in Figure \ref{fig:image_tokenization} can be achieved with a convolutional layer, having stride equal to the kernel size. Alternatively, libraries like einops\footnote{\url{http://einops.rocks}} extend the einsum operation (Section \ref{sec:linear_algebra}) to allow for grouping of elements into blocks of pre-determined shape. An example is shown in Box \ref{code:einops}.

The original ViT used trainable positional embeddings along with an additional class token to perform image classification. ViTs can also be used for image generation by predicting the patches in a row-major or column-major order. In this case, we can train a separate module that converts each patch into a discrete set of tokens using, e.g., a \textbf{vector-quantized variational autoencoder} \cite{chang2022maskgit}, or we can work directly with continuous outputs \cite{tschannen2023givt}. For image generation, however, other non-autoregressive approaches such as diffusion models and flow matching tend to be preferred; we cover them in the next volume.

\begin{mypy}{einops can be used to decompose an image into patches with a simple extension of the einsum syntax.}{code:einops}
from einops import rearrange
# A batch of images
xb = torch.randn((32, 3, 64, 64)) 

# Define the operation: differently from 
# standard einsum, we can split the output 
# in blocks using brackets
op = 'b c (h ph) (w pw) \
                -> b (h w) (ph pw c)'

# Run the operation with a given patch size
patches = rearrange(xb, op, ph=8, pw=8)
print(patches.shape) # [Out]: (32, 64, 192)
\end{mypy}

By developing proper tokenization mechanisms and positional embeddings, transformers have also been developed for audio, in particular for speech recognition. In this case, it is common to have a small 1D convolutional model (with pooling) as the tokenization block \cite{baevski2020wav2vec,radford2023robust}. For example, Wav2Vec \cite{baevski2020wav2vec} is an encoder-only model whose output is trained with an extension of the cross-entropy loss, called \textbf{connectionist temporal classification} loss \cite{graves2012connectionist}, to align the output embeddings to the transcription. Because labeled data with precise alignments is scarce, Wav2Vec models are pre-trained on large amounts of unlabeled audio with a variant of a masked language modeling loss. By contrast, Whisper \cite{radford2023robust} is an encoder-decoder model where the decoder is trained to autoregressively generate the transcription. This provides more flexibility to the model and reduces the need for strongly labeled data, but at the cost of possible \textit{hallucinations} in the transcription phase. \textbf{Neural audio codecs} can also be trained to compress audio into a sequence of discrete tokens \cite{defossez2022high}, which in turn form the basis for generative applications such as text-to-speech generation \cite{wang2023neural}.

\begin{figure}[t]
    \centering
    \includegraphics[width=0.8\textwidth]{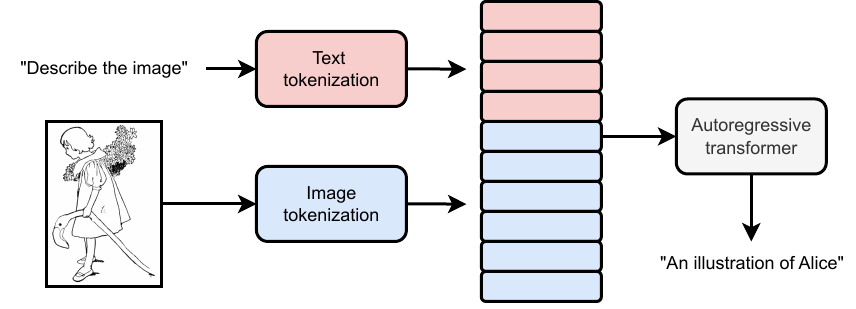}
    \caption{An example of a \textbf{bimodal} transformer that operates on both images and text: the outputs of the two tokenizers are concatenated and sent to the model.}
    \label{fig:multimodal_transformer}
\end{figure}

Transformers can also be defined for time-series \cite{ansari2024chronos}, graphs (covered in the next chapter), and other types of data. The decoupling between data and architecture is also the basis for \textbf{multimodal} variants, which can take as input (or provide as output) different modalities. This is achieved by tokenizing each modality (image, audio, ...) with its tokenizer, and concatenating the different tokens together into a single sequence \cite{bordes2024introduction}. We show an example for an \textbf{image-text} model in Figure \ref{fig:multimodal_transformer}.

\section{Transformer variants}
\label{subsec:mha_variants}

We close the chapter by discussing a few interesting variation on the basic transformer block. First, several variants have been devised for very large transformers to slightly reduce the computational time or parameter’s count. As an example, \textbf{parallel blocks} \cite{dehghani2023scaling} perform the MLP and MHA operation in parallel:
$$
\mathbf{H} = \mathbf{H}+\text{MLP}(\mathbf{H})+\text{MHA}(\mathbf{H})
$$
In this way, the initial and final linear projections in the MLP and MHA layers can be fused for a more efficient implementation. As another example, \textbf{multi-query MHA} \cite{shazeer2019fast} shares the same key and value projection matrix for each head, varying only the queries.

\begin{SCfigure}
    \centering
    \hspace{1em}\includegraphics[width=0.55\textwidth]{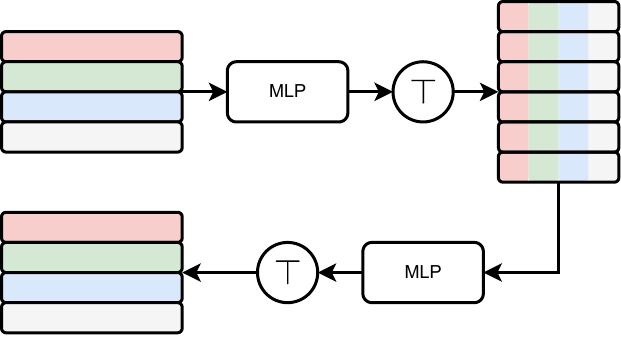}
    \caption{Mixer block, composed of alternating MLPs on the rows and columns of the input matrix.}
    \label{fig:mlp_mixer}
\end{SCfigure}

More in general, we can replace the MHA layer with a simpler (linear complexity in the sequence length) operation, while keeping the overall structure of the transformer block, i.e., alternating token and channel mixing with layer normalization and residual connections. As an example, suppose the sequence length is fixed (e.g., for computer vision, the number of patches can be fixed a priori). In this case, the MHA layer can be replaced by an MLP operating on a single input channel, corresponding to one dimension of the embedding. This type of model is called a \textbf{mixer} model \cite{tolstikhin2021mlp} - see Figure \ref{fig:mlp_mixer}.
Ignoring the normalization operations, this can be written as alternating MLPs on transpositions of the input matrix:
\begin{align}
\mathbf{H}& =\text{MLP}(\mathbf{H})+\mathbf{H} \\ \mathbf{H}& = \left[\text{MLP}(\mathbf{H}^\top)+\mathbf{H}^\top\right]^\top
\end{align}
Other variants of the mixer model are also possible using, e.g., 1D convolutions, Fourier transforms, or pooling. In particular, in the S2-MLP \cite{yu2022s2}  model the token mixing operation is replaced by an even simpler MLP applied on a shifted version of its input. The general class of such models has been called \textbf{MetaFormers} by \cite{yu2022metaformer}.

Gated (multiplicative) interactions can also be used in the composition of the block. In this case, several blocks are executed in parallel but their output is combined via Hadamard multiplication. We can write a generic gated unit as:
\begin{equation}
f(\mathbf{X}) =\phi_1(\mathbf{X})\odot \phi_2(\mathbf{X})
\label{eq:glu_again}
\end{equation}
where $\phi_1$ and $\phi_2$ are trainable blocks. For example, with $\phi_1(\mathbf{X}) = \sigma(\mathbf{X}\mathbf{A})$ and $\phi_2(\mathbf{X})=\mathbf{X}\mathbf{B}$ we obtain the \textbf{gated linear unit} (GLU) described in Section \ref{sec:activation_functions}.

As a few representative examples, the \textbf{gMLP} model \cite{liu2021pay} uses
gated units instead of a channel mixing block in a mixer model; the \textbf{LLaMa} family of models \cite{touvron2023llama} uses GLU-like units instead of the standard MLP block; while the \textbf{gated attention unit} (GAU) \cite{hua2022transformer} uses a simpler attention-like model having a single head for $\phi_1$ and a linear projection for $\phi_2$. These designs are especially popular in some recent variants of recurrent models, discussed later on in Chapter \ref{chap:rnns}.

To simplify the design even further, the multilinear operator network (MONet) removes all activation functions to define a block which is composed only of linear projections and element-wise multiplications \cite{cheng2024multilinear}:
$$
\mathbf{H}=\mathbf{E}(\mathbf{A}\mathbf{X}\odot\mathbf{B}\mathbf{X}+\mathbf{D}\mathbf{X})
$$
where $\mathbf{E}$ is similar to the output projection in the transformer block, $\mathbf{D}\mathbf{X}$ acts as a residual connection, and $\mathbf{B}$ is implemented via a low-rank decomposition to reduce the number of parameters \cite{cheng2024multilinear}. In order to introduce token mixing, a token-shift operation is implemented in all odd-numbered blocks in the model.

\section*{From theory to practice}

\begin{wrapfigure}{r}{3.0cm}
\vspace{-6em}\includegraphics[width=3.0cm]{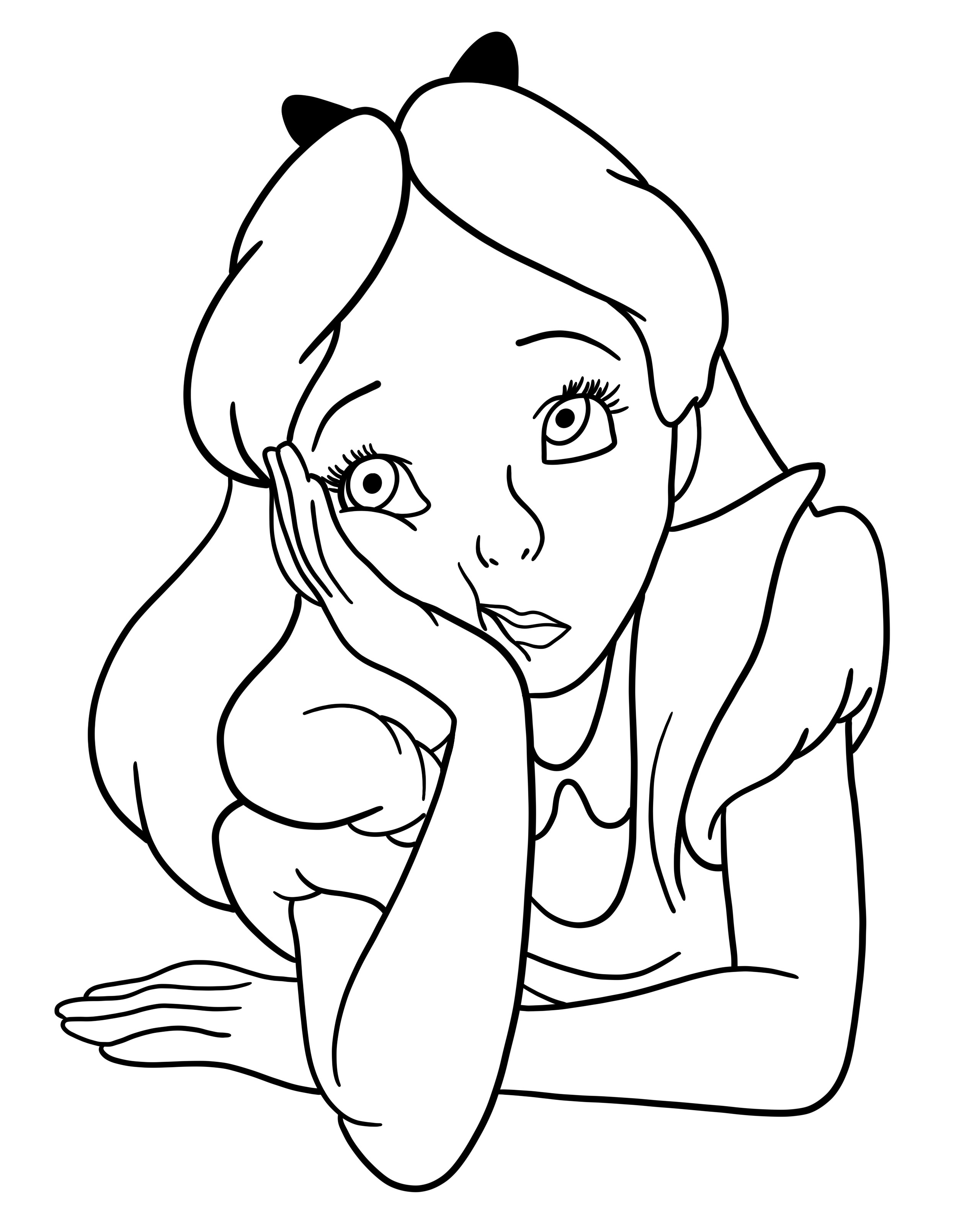}
\vspace{-2em}
\end{wrapfigure}

There are many interesting exercises that can be done at this point -- you are almost a master in designing differentiable models! To begin, using any image classification dataset, you can try implementing from scratch a Vision Transformer as described in Section \ref{subsec:transformers_image_audio}, following \cite{dosovitskiy2020image} for choosing the hyper-parameters. Training a ViT from scratch on a small dataset is quite challenging \cite{lee2021vision,steiner2021train}, so be ready for some disappointment unless you have sufficient computational power to consider million-size datasets. You can also try a simpler variant, such as the Mixer model described in Section \ref{subsec:mha_variants}. All these exercises should be relatively simple. 

\begin{enumerate}
\item For tokenizing the image you can use Einops as in Box \ref{code:einops} or other strategies (e.g., a convolution with large stride). For small images you can also try using each pixel as token.
\item For positional embeddings, all strategies described in Section \ref{sec:positional_embeddings} are valid. The simplest one for a ViT is to initialize a matrix of \textit{trainable} embeddings, but I suggest you experiment with sinusoidal and relative embeddings as practice.
\end{enumerate}

You can also try implementing a small GPT-like model. There are many sophistications in the tokenization of text data that we do  not cover. However, the minGPT repository\footnote{\url{https://github.com/karpathy/minGPT}} is a fantastic didactic implementation that you can use as starting point.

%% file: 12_graph_models.tex
\chapter{Graph models}
\label{chap:gnns}

\begin{supportbox}{About this chapter}
In this chapter we consider graph-structured data, i.e., nodes connected by a set of (known) relations. Graph are pervasive in the real world, ranging from proteins to traffic networks, social networks, and recommender systems. We introduce specialized layers to work on graphs, broadly categorized as either message-passing layers or graph transformers architectures.
\end{supportbox}

\section{Learning on graph-based data}
\subsection{Graphs and features on graphs}

Up to now we have considered data which is either completely unstructured  (tabular data represented as a vector) or structured in simple ways, including sets, sequences, and grids such as images. However, many types of data are defined by more sophisticated dependencies between its constituents. For example, molecules are composed by atoms which are only sparsely connected via chemical bonds. Networks of many kinds (social networks, transportation networks, energy networks) are composed of millions of units (people, products, users) which interact only through a small set of connections, e.g., roads, feedbacks, or friendships. These are more naturally defined in the language of \textbf{graph theory}. The aim of this chapter is to introduce differentiable models to work with data defined in such a way.

\begin{figure}[t]
    \centering
    \includegraphics[width=0.8\textwidth]{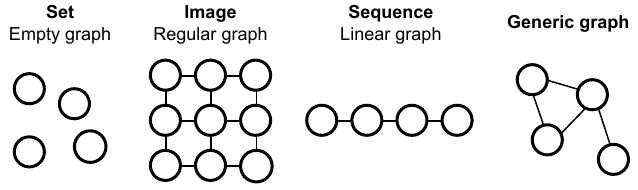}
    \caption{Graphs generalize many types of data: sets can be seen as empty graphs (or graphs having only self-loops), images as regular graphs, and sequences as linear graphs. In this chapter we look at more general graph structures.}
    \label{fig:graphs}
\end{figure}

In its simplest form, a graph can be described by a pair of sets $\mathcal{G} = (\mathcal{V}, \mathcal{E})$, where $\mathcal{V} = \left\{1, \ldots, n\right\}$ is the set of \textbf{nodes} (\textbf{vertices}), while:
$$
\mathcal{E} = \left\{(i,j) \mid \eqnmarkbox[drawred]{node}{i,j \in \mathcal{N}}\right\}
$$
\annotate[yshift=-1em]{below,right}{node}{Two nodes of the graph}

is the set of \textbf{edges} present in the graph. In most datasets, the number of nodes $n$ and the number of edges $m = \lvert \mathcal{E}\rvert$ can vary from graph to graph. 

Graph generalize many concepts we have already seen: for example, graphs containing only \textbf{self-loops} of the form $(i,i)$ represent sets of objects, while graphs containing all possible edges (\textbf{fully-connected graphs}) are  connected to attention layers, as we show next. Images can be represented as a graph by associating each pixel to a node of the graph and connecting close pixels based on a regular grid-like structure - see Figure \ref{fig:graphs}.\footnote{There are many variants of this basic setup, including heterogenous graphs (graphs with different types of nodes), directed graphs, signed graphs, etc. Most of them can be handled by variations of the techniques we describe next.}

\begin{figure}[t]
    \centering
    \includegraphics[width=1.0\textwidth]{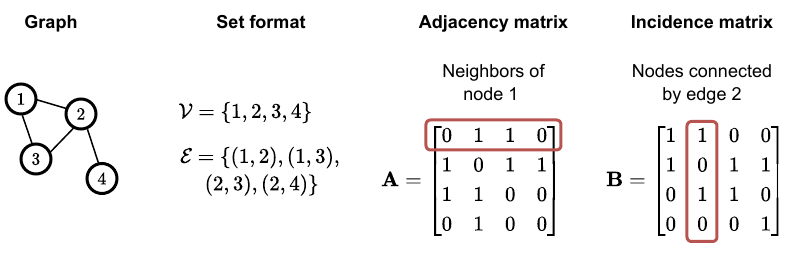}
    \caption{We can represent the graph connectivity in three ways: as a set $\mathcal{E}$ of pairs (second column); as an $(n, n)$ adjacency matrix (third column); or as an $(n, \lvert \mathcal{E} \rvert)$ incidence matrix (fourth column).}
    \label{fig:adjacency_matrix}
\end{figure}

Connections in a graph can be equivalently represented by a matrix representation called the \textbf{adjacency matrix}. This is a binary square matrix $\mathbf{A} \sim \text{Binary}(n,n)$ such that:
$$
A_{ij} = \begin{cases} 1 & \text{ if } (i,j) \in \mathcal{E} \\ 0 & \text{ otherwise} \end{cases}
$$
In this format, a set is represented by the identity matrix $\mathbf{A} = \mathbf{I}$, a fully-connected graph by a matrix of all ones, and an image by a Toeplitz matrix. A graph where connections are always bidirectional (i.e., $(i,j)$ and $(j,i)$ are always present as pairs among the edges) is called \textbf{undirected}, and we have $\mathbf{A}^\top = \mathbf{A}$. We will deal with undirected graphs for simplicity, but the methods can be easily extended to the directed case. We note that there are also alternative matrix representations, e.g., the incidence matrix $\mathbf{B} \sim \text{Binary}(n, \lvert \mathcal{E}\rvert)$ is such that $B_{ij} = 1$ if node $i$ participates in edge $j$, and we have $\mathbf{B}\mathbf{1}^\top = 2$ because each edge connects exactly two nodes. See Figure \ref{fig:adjacency_matrix} for an example.

We will assume our graphs to have self-loops, i.e., $A_{ii} = 1$. If the adjacency matrix does not have self-loops, we can add them by re-assigning it as:
$$
\mathbf{A} \gets\mathbf{A} + \mathbf{I}
$$

\subsection{Graph features}

Graphs come with a variety of possible features describing them. For example, atoms and bonds in a molecule can be described by categorical features denoting their types; roads in a transportation network can have a capacity and a traffic flow; and two friends in a social networks can be described by how many years they have known each other.

In general, these features can be of three types: \textbf{node features} associated to each node, \textbf{edge features} associated to each edge, and \textbf{graph features} associated to the entire graph. We will begin with the simplest case of having access to only unstructured node features, i.e., each node $i$ has associated a vector $\mathbf{x}_i \sim (c)$. The complete graph can then be described by two matrices $\mathbf{X} \sim (n,c)$, that we call the \textbf{feature matrix}, and the adjacency matrix $\mathbf{A} \sim (n,n)$.

In most cases, the ordering of the nodes is irrelevant, i.e., if we consider a permutation matrix $\mathbf{P} \sim \text{Binary}(n,n)$ (see Section \ref{sec:positional_embeddings}), a graph and its permuted version are fundamentally identical, in other words:

$$
(\mathbf{X}, \mathbf{A}) \;\; \text{ is the same graph as } \;\;(\mathbf{P}\mathbf{X},\mathbf{P}\mathbf{A}\mathbf{P}^\top)
$$

Note that the permutation matrix acts by swapping the rows in $\mathbf{X}$, while it swaps both rows and columns in the adjacency matrix.

Some features can also be extracted directly from the topology of the graph. For example, we can associate to each node a scalar value $d_i$, called the \textbf{degree}, which describes how many nodes it is connected to:

$$
d_i=\sum_j A_{ij}
$$

\begin{figure}[t]
    \centering
    \begin{subfigure}[b]{0.24\textwidth}
    \includegraphics[width=\textwidth]{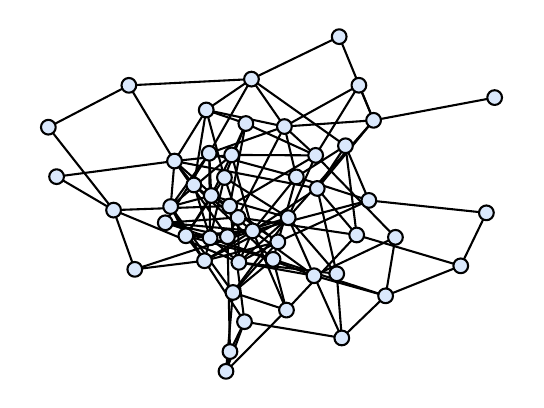}
    \caption{\scriptsize Erdős–Rényi}
    \end{subfigure}
    \begin{subfigure}[b]{0.24\textwidth}
    \includegraphics[width=\textwidth]{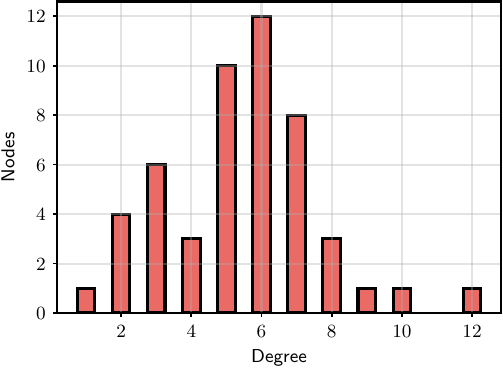}
    \caption{\scriptsize Degree}
    \end{subfigure}
    \begin{subfigure}[b]{0.24\textwidth}
    \includegraphics[width=\textwidth]{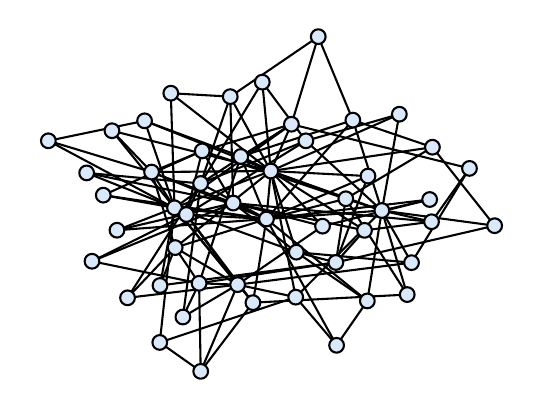}
    \caption{\scriptsize Barabasi-Albert}
    \end{subfigure}
    \begin{subfigure}[b]{0.24\textwidth}
    \includegraphics[width=\textwidth]{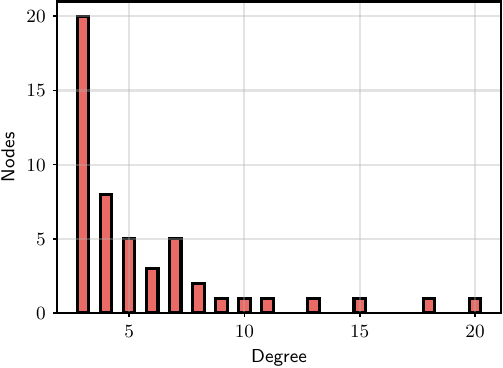}
    \caption{\scriptsize Degree}
    \end{subfigure}
    \caption{(a) Random graph generated by drawing each edge independently from a Bernoulli distribution (\textbf{Erdős–Rényi model}). (b) These graphs show a Gaussian-like degree distribution. (c) Random graph generated by adding nodes sequentially, and for each of them drawing $3$ connections towards existing nodes with a probability proportional to their degree (\textbf{preferential attachment process} or \textbf{Barabasi-Albert model}). (d) These graphs have a few nodes with many connections acting as hubs for the graph.}
    \label{fig:random_graphs}
\end{figure}

The distribution of the degrees across the graph is an important characteristic of the graph itself, as shown in Figure \ref{fig:random_graphs}. We can collect the degrees into a single diagonal matrix called the \textbf{degree} matrix:

$$
\mathbf{D} =\begin{bmatrix} d_1 & \ldots & 0 \\ \vdots &\ddots & \vdots \\0 & \ldots & d_n \end{bmatrix}
$$

We can use the degree matrix to define several types of \textit{weighted} adjacency matrices. For example, the row-normalized adjacency matrix is defined as:

$$
\mathbf{A}^\prime \leftarrow \mathbf{D}^{-1}\mathbf{A}_{ij} \;\; \rightarrow\;\; A^\prime_{ij} = \frac{1}{d_i}A_{ij}
$$

This is normalized in the sense that $\sum_i A^\prime_{ij} = \mathbf{1}$. We can also define a column-normalized adjacency matrix as $\mathbf{A}^\prime = \mathbf{A}\mathbf{D}^{-1}$. Both 
matrices can be interpreted as “random walks” over the graph, in the sense that, given a node $i$, the corresponding row or column of the normalized adjacency matrix represents a probability distribution of moving at random towards any of its neighbours. A more general symmetrically normalized adjacency matrix is given by:

$$
\mathbf{A}^\prime=\mathbf{D}^{-1/2}\mathbf{A}\mathbf{D}^{-1/2}
$$

This is defined by $A^\prime_{ij} = \frac{A_{ij}}{\sqrt{d_i d_j}}$, giving a weight to each connection based on the degree of both nodes it connects to. Both the adjacency matrix and its weighted variants have the property that $A_{ij} = 0$ whenever $(i,j) \notin \mathcal{E}$. In signal processing terms, these are called \textbf{graph-shift} matrices. \clearpage

\begin{supportbox}{Sparsity in matrices}
Consider a generic adjacency matrix for a $6$-nodes graph (try drawing the graph as an exercise):
$$
\mathbf{A} = \begin{bmatrix} 0& 1& 1& 1& 1& 1\\1& 0& 0& 0& 1& 0\\1& 0& 0& 0& 1& 1\\1& 0& 0& 0& 0& 1\\1& 1& 1& 0& 0& 0\\1& 0& 1& 1& 0& 0\end{bmatrix}
$$

The adjacency is very sparse (many zeros). This is an important property, because sparse matrices have customized implementations and techniques for manipulating them, with better computational complexity than their dense counterparts.\footnote{As an example, in JAX: \url{https://jax.readthedocs.io/en/latest/jax.experimental.sparse.html}.}

\end{supportbox}

\subsection{Diffusion operations over graphs}

The fundamental graph operation we are interested into is something called \textbf{diffusion}, which corresponds to a smoothing of the node features with respect to the graph topology. To understand it, consider a scalar feature on each node, that we collect in a vector $\mathbf{x} \sim (n)$, and the following operation over the features:

$$
\mathbf{x}^\prime=\mathbf{A}\mathbf{x}
$$

where $\mathbf{A}$ can be the adjacency matrix, a normalized variant, or any weighted adjacency matrix. We can re-write this operation node-wise as:

$$
x^\prime_i = \sum_{j \in \mathcal{N}(i)} A_{ij}x_j
$$

where we have defined the $1$-hop neighborhood:

\vspace{1em}
$$
\mathcal{N}(i) = \left\{ j\mid \eqnmarkbox[drawred]{node}{(i,j)\in\mathcal{E}} \right\}
$$
\annotate[yshift=1em]{above,left}{node}{All edges with node $i$ as a vertex}

\vspace{-1em}
If we interpret the node feature as a physical quantity, projection by the adjacency matrix can be seen as a “diffusion” process which replaces the quantity at each node by a weighted average of the quantity in its neighborhood.

Another fundamental matrix in the context of graph analysis is the \textbf{Laplacian matrix}:

$$
\mathbf{L}=\mathbf{D}-\mathbf{A}
$$

where the degree matrix is computed as $D_{ii} = \sum_j A_{ij}$ irrespective of whether the adjacency matrix is normalized or not. One step of diffusion by the Laplacian can be written as:

\begin{equation}
\idx{\mathbf{L}\mathbf{x}}{i} = \sum_{(i,j) \in \mathcal{E}} A_{ij}(x_i-x_j)
\label{eq:laplacian}
\end{equation}

\begin{figure}[t]
    \centering
    \begin{subfigure}[b]{0.24\textwidth}
    \includegraphics[width=\textwidth]{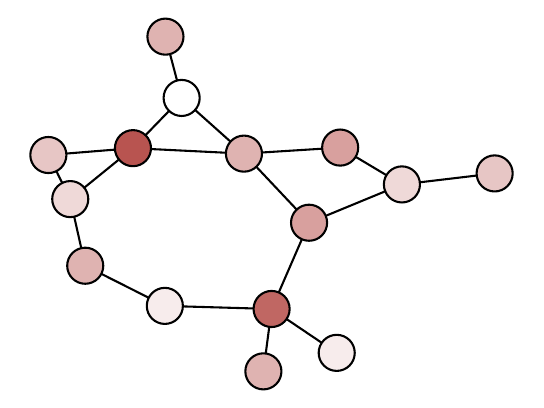}
    \caption{Initial graph}
    \end{subfigure}
    \begin{subfigure}[b]{0.24\textwidth}
    \includegraphics[width=\textwidth]{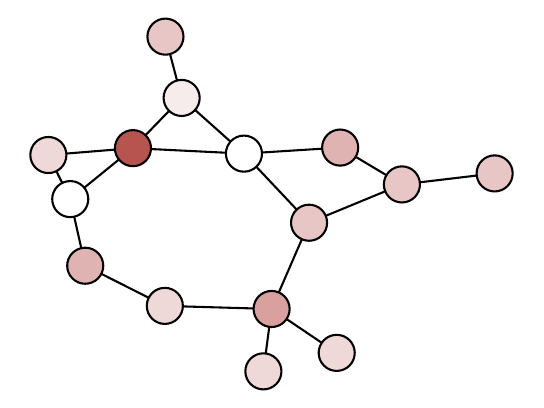}
    \caption{10 steps}
    \end{subfigure}
    \begin{subfigure}[b]{0.24\textwidth}
    \includegraphics[width=\textwidth]{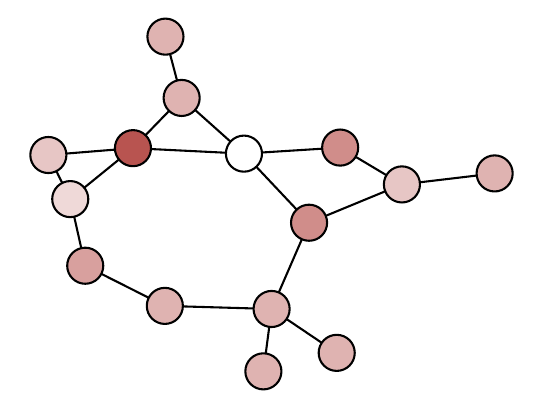}
    \caption{20 steps}
    \end{subfigure}
    \begin{subfigure}[b]{0.24\textwidth}
    \includegraphics[width=\textwidth]{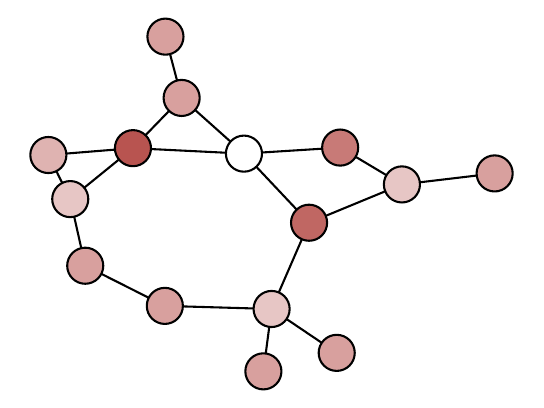}
    \caption{30 steps}
    \end{subfigure}
    \caption{(a) A random graph with 15 nodes and a scalar feature on each node (denoted with variable colors). (b)-(d) The result after $10$, $20$, and $30$ steps of diffusion with the Laplacian matrix. The features converge to a stable state.}
    \label{fig:diffusion}
\end{figure}

We can see from here that the Laplacian is intimately linked to the idea of a gradient over a graph, and its analysis is at the core of the field of \textbf{spectral graph theory}. As an example, in \eqref{eq:laplacian} $\mathbf{1}$ is always an eigenvector of the Laplacian associated to a zero eigenvalue (in particular, the smallest one). We show an example of diffusion with the Laplacian matrix in Figure \ref{fig:diffusion}.

\subsection{Manifold regularization}

From \eqref{eq:laplacian} we can also derive a quadratic form built on the Laplacian:

\begin{equation}
\mathbf{x}^\top\mathbf{L}\mathbf{x}=\sum_{(i,j)\in \mathcal{E}}A_{ij}(x_i-x_j)^2
\label{eq:laplacian_regularizer}
\end{equation}

Informally, this is a scalar value that measures how “smooth” the signal over the graph is, i.e., how quickly it changes for pairs of nodes that are connected in the graph. To see a simple application of this concept, consider a tabular classification dataset $\mathcal{S}_n = \left\{(\mathbf{x}_i, y_i)\right\}$. Suppose we build a graph over this dataset, where each node is an element of the dataset, and the adjacency matrix is built based on the distance between features:

\begin{equation}
A_{ij}=\begin{cases} \exp(-\lVert \mathbf{x}_i - \mathbf{x}_j \rVert^2) & \text{ if } \lVert \mathbf{x}_i - \mathbf{x}_j \rVert^2 < \tau \\ 0 & \text{otherwise} \end{cases}
\label{eq:dataset_graph}
\end{equation}

where $\tau$ is a user-defined hyper-parameter. Given a classification model $f(\mathbf{x})$, we may want to constrain its output to be similar for similar inputs, where similarity is defined proportionally to \eqref{eq:dataset_graph}. To this end, we can define the features of the graph as the outputs of our model:

$$
\mathbf{f} = \begin{bmatrix} f(\mathbf{x}_1) \\ \vdots \\ f(\mathbf{x}_n) \end{bmatrix} \sim (n)
$$

The quadratic form \eqref{eq:laplacian_regularizer} tells us exactly how much similar inputs vary in terms of their predictions:

\begin{equation}
\mathbf{f}^\top\mathbf{L}\mathbf{f} = \sum_{i,j} A_{ij}(f(\mathbf{x}_i)-f(\mathbf{x}_j))^2
\label{eq:manifold_regularizer}
\end{equation}

The optimal model can be found by a regularized optimization problem, where the regularizer is given by \eqref{eq:manifold_regularizer} :
$$
f^*(\mathbf{x})=\arg\min \left\{\sum_{i=1}^nL(y_i, f(\mathbf{x}))  + \lambda \;\;\mathbf{f}^\top \mathbf{L}\mathbf{f}\right\}
$$
where $L$ is a generic loss function and $\lambda$ is a scalar hyper-parameter:

This is called \textbf{manifold regularization} \cite{belkin2006manifold} and it can be used as a generic regularization tool to force the model to be smooth over a graph, where the adjacency is either given or is built by the user as in \eqref{eq:dataset_graph}. This is especially helpful in a \textbf{semi-supervised} scenario where we have a small labeled dataset and a large unlabeled one from the same distribution, since the regularizer in \eqref{eq:manifold_regularizer} does not require labels \cite{belkin2006manifold}. However, the prediction of the model depends only on a single element $\mathbf{x}_i$, and the graph is thrown away after training. In the next section, we will introduce more natural ways of embedding the connectivity inside the model itself.

\section{Graph convolutional layers}
\subsection{Properties of a graph layer}

In order to design models whose predictions are conditional on the connectivity, we can augment standard layers $f(\mathbf{X})$ with knowledge of the adjacency matrix, i.e., we consider layers of the form:
$$
\mathbf{H} =f(\mathbf{X}, \mathbf{A})
$$
where as before $\mathbf{X} \sim (n, c)$ (with $n$ the number of nodes and $c$ the features at each node) and $\mathbf{H} \sim (n, c^\prime)$, i.e., the operation does not change the connectivity of the graph, and it returns an updated embedding $\mathbf{H}_i \sim (c^\prime)$ for each node $i$ in the graph. For what follows, $\mathbf{A}$ can be the adjacency or any matrix with the same sparsity pattern (a graph-shift matrix), including a weighted adjacency matrix, the Laplacian matrix, and so on.

Since permuting the nodes in a graph should have no impact on the final predictions, the layer should not depend on the specific ordering of the nodes, i.e., for any permutation matrix $\mathbf{P}$ the output of the layer should be \textbf{permutation equivariant}:
$$
f(\mathbf{P}\mathbf{X}, \mathbf{P}\mathbf{A}\mathbf{P}^\top)=\mathbf{P}\,\cdot\,f(\mathbf{X}, \mathbf{A})
$$
We can define a notion of “locality” for a graph layer, similar to the image case. To this end, we first introduce the concept of a subgraph. Given a subset of nodes $\mathcal{T} \in \mathcal{V}$ from the full graph, we define the \textbf{subgraph} induced by $\mathcal{T}$ as:
$$
\mathcal{G}_{\mathcal{T}}= (\mathbf{X}_{\mathcal{T}}, \mathbf{A}_{\mathcal{T}})
$$
where $\mathbf{X}_{\mathcal{T}}$ is a $(\lvert\mathcal{T}\rvert,c)$ matrix collecting the features of the nodes in $\mathcal{T}$, and $\mathbf{A} \sim (\lvert\mathcal{T}\rvert,\lvert\mathcal{T}\rvert)$ is the corresponding block of the full adjacency matrix.

\begin{definition}[Graph locality] $\,$

A graph layer $\mathbf{H} =f(\mathbf{X}, \mathbf{A})$ is \textbf{local} if for every node, $\mathbf{H}_i = f(\mathbf{X}_{\mathcal{N}(i)}, \mathbf{A}_{\mathcal{N}(i)})$, where $\mathcal{N}(i)$ is the $1$-hop neighborhood of node $i$.

\end{definition}

This is similar to considering all pixels at distance $1$ in the image case, except that (a) nodes in $\mathcal{N}(i)$ have no specific ordering in this case, and (b) the size of $\mathcal{N}(i)$ can vary a lot depending on $i$. Hence, we cannot define a convolution like we did in the image case, as its definition requires these two properties (think of the weight tensor in a convolutional layer).

For what follows, note that we can extend our definition of locality beyond 1-hop neighbors. For example, the 2-hop neighborhood $\mathcal{N}^2(i)$ is defined as all nodes at distance at most $2$:
$$
\mathcal{N}^2(i) = \bigcup_{j \in \mathcal{N}(i)} \mathcal{N}(j) 
$$
where $\cup$ is the set union operator. We can extend the definition of locality to take higher-order neighborhoods into consideration and design the equivalent of $3 \times 3$ filters, $5 \times 5$ filters, and so on.

\subsection{The graph convolutional layer} \addclock

In order to define a graph layer that mimicks the convolutional layer, we need it to be permutation equivariant (instead of translation equivariant) and local. The MHA layer is naturally permutation equivariant, but it is not local and it does not depend explicitly on the adjacency matrix $\mathbf{A}$. We will see possible extensions to this end in the next section. For now, let us focus on a simpler fully-connected layer:
$$
f(\mathbf{X}, \_)= \phi(\mathbf{X}\mathbf{W}+\mathbf{b})
$$
where $\mathbf{W} \sim (c, c^\prime)$ and $\mathbf{b} \sim (c^\prime)$. This is also naturally permutation equivariant, but it does not depend on the connectivity of the graph, which is ignored. To build an appropriate differentiable layer, we can alternate the layer's operation with a diffusion step.

\begin{definition}[Graph convolution] \addbottle $\,$

Given a graph represented by a node feature matrix $\mathbf{X} \sim (n,c)$ and a generic graph-shift matrix $\mathbf{A} \sim (n,n)$ (the adjacency, the Laplacian, ...), a \textbf{graph convolutional} (GC) layer is given by \cite{kipf2016semi}:

$$
f(\mathbf{X}, {\color{drawred}\mathbf{A}})= \phi({\color{drawred}\mathbf{A}}(\mathbf{X}\mathbf{W}+\mathbf{b}))
$$

where the trainable parameters are $\mathbf{W} \sim (c, c^\prime)$ and $\mathbf{b} \sim (c^\prime)$, with $c^\prime$ an hyper-parameter. $\phi$ is a standard activation function, such as a ReLU.

\end{definition}

\begin{figure}
    \centering
    \hspace*{0.5em}\includegraphics[width=0.7\textwidth]{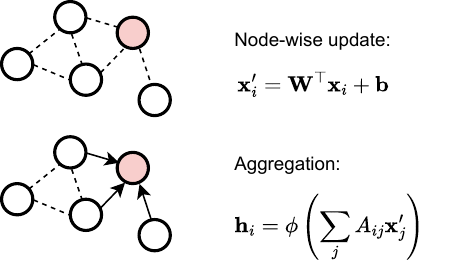}
    \caption{Two stages of a GC layer: each node updates its embedding in parallel to all other nodes; the output is given by a weighted average of all updated embeddings in the node's neighbourhood.}
    \label{fig:graph_convolution_layer}
\end{figure}

Note the similarity with a standard convolutional layer: we are performing a “channel mixing” operation via the matrix $\mathbf{W}$, and a “node mixing” operation via the matrix $\mathbf{A}$, the difference being that the former is untrainable in this case (due to, once again, variable degrees between nodes and the need to make the layer permutation equivariant). See Figure \ref{fig:graph_convolution_layer} for a visualization. The analogy can also be justified more formally by leveraging concepts from graph signal processing, which is beyond the scope of this book \cite{bronstein2017geometric}. 

Ignoring the bias, we can rewrite this for a single node $i$ as:

$$
\mathbf{H}_i=\phi\left(\sum_{j \in \mathcal{N}(i)}A_{ij}\mathbf{X}_j\mathbf{W}\right)
$$

Hence, we first perform a simultaneous update of all node embeddings (given by the right multiplication by $\mathbf{W}$). Then, each node computes a weighted average of the updated node embeddings from itself and its neighbors. Since the number of neighbors can vary from node to node, working with the normalized variants of the adjacency matrix can help significantly in training. It is trivial to show permutation equivariance for the layer:

$$
f({\color{drawred}\mathbf{P}}\mathbf{X},{\color{drawred}\mathbf{P}}\mathbf{A}{\color{drawred}\mathbf{P}^\top})=\phi\left({\color{drawred}\mathbf{P}}\mathbf{A}{\color{drawred}\mathbf{P}^\top}{\color{drawred}\mathbf{P}}\mathbf{X}\mathbf{W}\right) = {\color{drawred}\mathbf{P}}\,\cdot\,f(\mathbf{X},\mathbf{A})
$$

\subsection{Building a graph convolutional network}

A single GC layer is local, but the stack of multiple layers is not. For example, consider a two-layered GC model:

\begin{equation}
f(\mathbf{X},\mathbf{A})=\phi(\mathbf{A}\eqnmarkbox[drawred]{node}{\phi\left(\mathbf{A}\mathbf{X}\mathbf{W}_1\right)}\mathbf{W}_2)
\label{eq:two_layer_gcn}
\end{equation}
\annotate[yshift=-1em]{below,right}{node}{First GC layer}

\vspace{1em}
with two trainable weight matrices $\mathbf{W}_1$ and $\mathbf{W}_2$. Similarly to the image case, we can define a notion of receptive field.

\begin{definition}[Graph receptive field] $\,$

Given a generic graph neural network $\mathbf{H} = f(\mathbf{X}, \mathbf{A})$, the \textbf{receptive field} of node $i$ is the smallest set of nodes $\mathcal{V}(i) \in \mathcal{V}$ such that $\mathbf{H}_i = f(\mathbf{X}_{\mathcal{V}(i)}, \mathbf{A}_{\mathcal{V}(i)})$.

\end{definition}

For a single GC layer, the receptive field is $\mathcal{V}(i) = \mathcal{N}(i)$. For a two-layer network as in  \eqref{eq:two_layer_gcn}, we need to consider neighbors of neighbors, and the receptive field becomes $\mathcal{V}(i) = \mathcal{N}^2(i)$. In general, for a stack of $k$ layers we will have a receptive field of $\mathcal{V}(i) = \mathcal{N}^k(i)$. The smallest number of steps which is needed to move from any two nodes in the graph is called the \textbf{diameter} of the graph. The diameter defines the smallest number of layers which is required to achieve a global receptive field for all the nodes. 

\begin{supportbox}{Polynomial GC layers}

Alternatively, we can increase the receptive field of \textit{a single} GC layer. For example, if we remove the self-loops from the adjacency matrix, we can make the layer local with respect to $\mathcal{N}^2(i)$ instead of $\mathcal{N}(i)$ by also considering the square of the adjacency matrix:

$$
\mathbf{H} = \phi\left(\mathbf{X}\mathbf{W}_0 + \mathbf{A}\mathbf{X}\mathbf{W}_1 + \mathbf{A}^2\mathbf{X}\mathbf{W}_2\right)
$$

where we have three sets of parameters $\mathbf{W}_0$, $\mathbf{W}_1$, and $\mathbf{W}_2$ to handle self-loops, neighbors, and neighbors of neighbors respectively. This is called a \textbf{polynomial} GC layer. Larger receptive fields can be obtained with higher powers. More complex layers can be designed by considering ratios of polynomials \cite{bianchi2021graph}.

\end{supportbox}

We can combine GC layers with standard normalization layers, residual connections, dropout, or any other operation that is permutation equivariant. Differently from the image case, pooling is harder because there is no immediate way to subsample a graph connectivity. Pooling layers can still be defined by leveraging tools from graph theory or adding additional trainable components, but they are less common \cite{grattarola2022understanding}.

\begin{figure}[t]
    \centering
    \includegraphics[width=0.9\textwidth]{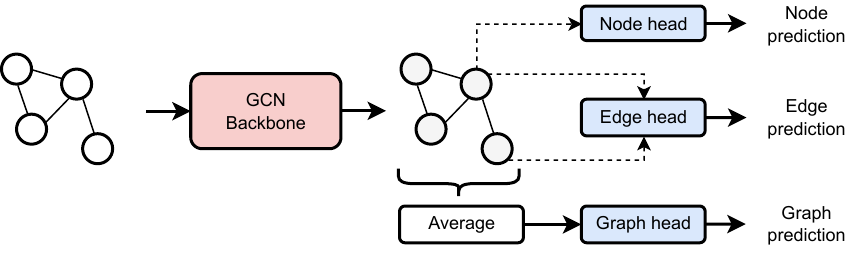}
    \caption{Different types of graph heads: (a) node tasks need to process the features of a single node; (b) edge tasks require heads that are conditioned on two nodes simultaneously; (c) graph tasks can be achieved by pooling all node representations into a fixed-dimensional vector.}
    \label{fig:graph_heads}
\end{figure}

Denote by $\mathbf{H} = f(\mathbf{X}, \mathbf{A})$ a generic combination of layers providing an updated embedding for each node (without modifying the connectivity). In analogy with CNNs, we call it the \textbf{backbone} network. We can complete the design of a generic \textbf{graph convolutional network} (GCN) by adding a small head of top of these representations:
$$
y=(g\circ f)(\mathbf{X},\mathbf{A})
$$
The design of the head depends on the task we are trying to solve. The most common tasks fall into one of three basic categories: node-level tasks (e.g., node classification), edge-level task (e.g., edge classification), or graph-level tasks (e.g., graph classification). We briefly consider an example for each of them in turn, see Figure \ref{fig:graph_heads}.

\textbf{Node classification}

First, suppose the input graph describes some kind of social network, where each user is associated to a node. For a given subset of users, $\mathcal{T} \subseteq \mathcal{V}$, we know a label $y_i, i \in \mathcal{T}$ (e.g., whether the user if a real user, a bot, or another kind of automated profile). We are interested in predicting the label for all other nodes. In this case, we can obtain a node-wise prediction by processing each updated node embedding, e.g.:

$$
\hat{y}_i=g(\mathbf{H}_i) =\text{softmax}(\text{MLP}(\mathbf{H}_i))
$$

Running this operation over the entire matrix $\mathbf{H}$ gives us a prediction for all nodes, but we only know the true labels for a small subset. We can train the GCN by discarding the nodes outside of the training set:

$$
\arg\min \frac{1}{\lvert \mathcal{T} \rvert}\sum_{i \in \mathcal{T}} \text{CE}(\hat{y}_i,y_i)
$$

where $\text{CE}$ is the cross-entropy loss. Importantly, even if we are discarding the output predictions for nodes outside our training set, their input features are still involved in the training process due to the diffusion steps inside the GCN. The rest of the nodes can then be classified by running the GCN a final time after training. This scenario, where only a subset of the training data is labeled, is called a \textbf{semi-supervised} problem.

\textbf{Edge classification}

As a second example, suppose we have a label for a subset of \textit{edges}, i.e., $\mathcal{T}_E \subseteq \mathcal{E}$. As an example, our graph could be a traffic network, of which we know the traffic flow only on a subset of roads. In this case, we can obtain an edge-wise prediction by adding an head that depends on the features of the two connected nodes, e.g., by concatenating them:

$$
\hat{y}_{ij} = g(\mathbf{H}_i, \mathbf{H}_j)= \text{MLP}\left(\left[ \mathbf{H}_i \;\Vert \; \mathbf{H}_j\right]\right)
$$

For binary classification (e.g., predicting the affinity of two users with a scalar value between $0$ and $1$) we can simplify this by considering the dot product between the two features:

$$
\hat{y}_{ij} = \sigma( \mathbf{H}_i^\top \mathbf{H}_j)
$$

Like before, we can train the network by minimizing a loss over the known edges.

\textbf{Graph classification}

Finally, suppose we are interested in classifying (or regressing) the entire graph. As an example, the graph could be a molecule of which we want to predict some chemical property, such as reactivity against a given compound. We can achieve this by pooling the node representations (e.g., via a sum), and processing the resulting fixed-dimensional embedding:

$$
y=\text{MLP}\left(\frac{1}{n}\sum_{i=1}^n \mathbf{H}_i\right)
$$

The final pooling layer makes the network \textit{invariant} to the permutation of the nodes. In this case, our dataset will be composed of multiple graphs (e.g., several molecules), making it similar to a standard image classification task. For node and edge tasks, instead, some datasets may be composed of a single graph (e.g., a large social network), while other datasets can have more than a single graph (e.g., several unconnected road networks from different towns). This opens up the question of how to efficiently build mini-batches of graphs. \clearpage

\subsection{On the implementation of graph neural networks} \addteacup

As mentioned, the peculiarity of working with graphs is that several matrices can be very sparse. For example, consider the following adjacency matrix:

$$
\mathbf{A} = \begin{bmatrix} 0 & 0 & 1 \\ 0 & 0 & 0 \\ 1 & 0 & 0 \end{bmatrix}
$$

This corresponds to a three-node graph with a single bidirectional edge between nodes $1$ and $3$. We can store this more efficiently by only storing the indices of the non-zero values, e.g., in code:

{\begin{center}\footnotesize
\noindent\mintinline{python}{A = [[0,2], [2,0]]}
\end{center}
}

This is called a \textbf{coordinate list} format. For very sparse matrices, specialized formats like this one can reduce storage but also significantly improve the runtime of operating on sparse matrices or on combinations of sparse and dense matrices. As an example, {\footnotesize \texttt{pytorch-sparse}}\footnote{\url{https://github.com/rusty1s/pytorch_sparse}} supports highly-efficient implementations of transposition and several types of matrix multiplications in PyTorch. This is also reflected on the layers’ implementation. The forward pass of the layers in PyTorch Geometric\footnote{\url{https://pytorch-geometric.readthedocs.io/en/latest/get_started/introduction.html\#learning-methods-on-graphs}} (one of the most common libraries for working with graph neural networks in PyTorch) is parameterized by providing as inputs the features of the graph and the connectivity as a list of edge coordinates. \clearpage

\begin{SCfigure}
    \centering
    \hspace{2em}\includegraphics[width=0.4\textwidth]{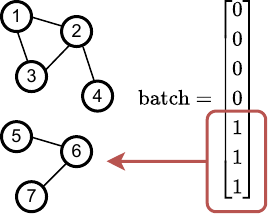}
    \caption{Two graphs in a mini-batch can be seen as a single graph with two disconnected components. In order to distinguish them, we need to introduce an additional vector containing the mapping between nodes and graph IDs.}
    \label{fig:graph_batch}
\end{SCfigure}

Working with sparse matrices has another interesting consequence in terms of mini-batches. Suppose we have $b$ graphs $(\mathbf{X}_i, \mathbf{A}_i)_{i=1}^b$. For each graph we have the same number of node features $c$ but a different number of nodes $n_i$, so that $\mathbf{X}_i \sim (n_i, c)$ and $\mathbf{A}_i \sim \text{Binary}(n_i, n_i)$. In order to build a mini-batch, we can create two rank-$3$ tensors:
\begin{gather}
X \sim (b,n,c) \\A\sim \text{Binary}(b,n,n)
\end{gather}
where $n = \max(n_1, \ldots, n_b)$, and both matrices are padded with zeros to fill up the two tensors. However, a more elegant alternative can be obtained by noting that in a GC layer, two nodes that are not connected by any path (a sequence of edges) will never communicate. Hence, we can build a \textit{single} graph describing the entire mini-batch by simply merging all the nodes:
\begin{gather}
\mathbf{X} = \begin{bmatrix}\mathbf{X}_1 \\ \vdots \\ \mathbf{X}_b \end{bmatrix} \\ \mathbf{A} = \begin{bmatrix} \mathbf{A}_1 & \ldots & \mathbf{0} \\ \vdots& \ddots & \vdots \\ \mathbf{0} & \ldots & \mathbf{A}_b \end{bmatrix}
\end{gather}
where $\mathbf{X} \sim (\sum_i n_i, c)$ and $\mathbf{A} \sim \text{Binary}(\sum_i n_i, \sum_i n_i)$. The adjacency matrix of the mini-batch has a block-diagonal structure, where all elements outside the diagonal blocks are zero (nodes from different graphs are not connected). While seemingly wasteful, this actually increases the sparsity ratio of the graph, making better use of the sparse matrix operations. Hence, for graph datasets in many cases there is no real difference between working with a single graph or a mini-batch of graphs.

\begin{figure}[t]
    \centering
    \hspace*{-2em}\includegraphics[width=\textwidth]{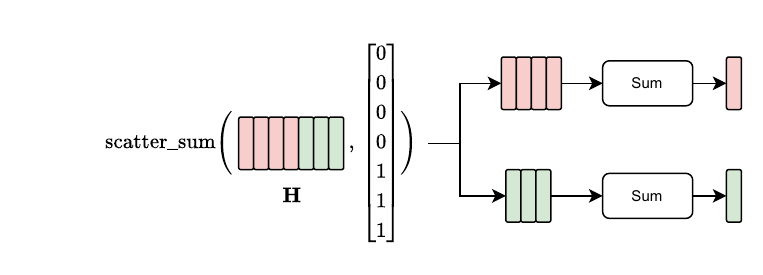}\hspace*{5em}
    \caption{Example of scattered sum on the graph of Figure \ref{fig:graph_batch}. In this example nodes (1,2,3,4) belong to graph 1, and nodes (5,6,7) to graph 2. After pooling, we obtain a pooled representation for each of the two graphs.}
    \label{fig:scatter_sum}
\end{figure}

To keep track of the correspondence between nodes and graphs, we can augment the representation with an additional vector $\mathbf{b} \sim (\sum_i n_i)$ such that $b_i$ is an index in $[0, \ldots, b-1]$ identifying one of the $b$ input graphs - see Figure \ref{fig:graph_batch}. For graph classification, we can exploit $\mathbf{b}$ to perform pooling separately on groups of nodes corresponding to different graphs. Suppose $\mathbf{H} \sim (n,c^\prime)$ is the output of the GCN backbone, then:
\begin{equation}
\text{scatter\_sum}\left(\mathbf{H}, \mathbf{b}\right) = \mathbf{Y} \sim (b, c^\prime)
\label{eq:scatter_sum}
\end{equation}
is called a \textbf{scattered sum} operation, and it is defined such that $\mathbf{Y}_i$ is the sum of all rows of $\mathbf{H}$ where $b_j = i$, as shown in Figure \ref{fig:scatter_sum}. Similar operations can be defined for other types of pooling operations, including averages and maximums.

As a separate problem, sometimes we may have a single graph that does not fit into memory: in this case, mini-batches should be formed by \textit{sampling} subgraphs from the original graph \cite{hamilton2017inductive}. This is a relatively complex task that goes beyond the scope of this chapter. \clearpage

\section{Beyond graph convolutional layers}

With the GC layer as a template, we now overview a few extensions, either in terms of adaptivity or graph features that can be handled. We close by discussing \textbf{graph transformers}, a different family of layers in which the graph is embedded into a structural embedding which is summed to the node features.

\subsection{Graph attention layers}

One issue with GC layers is that the weights that are used to sum up contributions from the neighborhoods are fixed and are given by the adjacency matrix (or a proper normalized variant). This is equivalent to the assumption that, apart from the relative number of connections, all neighbors are similarly important. A graph where nodes are connected mostly with similar nodes is called \textbf{homophilic}: empirically, homophily is a good predictor of the performance of graph convolutional layers \cite{li2022graph}. Not all graphs are homophilic: for example, in a dating network, most people will be connected with people from the opposite sex. Hence, in these scenarios we need techniques that can adapt the weights across nodes.

For sufficiently small graphs, we can let the non-zero elements of the weight matrix $\mathbf{A}$ adapt from their starting value through gradient descent. However, the number of trainable parameters in this case increases quadratically with the number of nodes, and this solution does not apply to a scenario with more than a single graph. If we assume that an edge depends only on the features of the two nodes it connects, we can generalize the GC layer with an attention-like operation:
$$
\mathbf{h}_i=\phi\left(\sum_{j \in \mathcal{N}(i)}{\color{drawred}\text{softmax}(\alpha(\mathbf{x}_i, \mathbf{x}_j))}\mathbf{W}^\top\mathbf{x}_j\right)
$$
where $\alpha$ is some generic MLP block having two inputs and a scalar output, and the softmax is applied, for each node, to the set of outputs of $\alpha$ with respect to $\mathcal{N}(i)$, to normalize the weights irrespective of the size of the neighborhood. Due to the similarity to the attention layer, these are called \textbf{graph attention} (GAT) layers \cite{velivckovic2017graph}. Seen from the perspective of the entire graph, this is very similar to a MHA layer, where the attention operation is restricted only on nodes having an edge that connects them.

The choice of $\alpha$ is relatively free. Instead of a dot product, the original GAT formulation considered an MLP applied on a concatenation of features:

$$
\alpha(\mathbf{x}_i, \mathbf{x}_j)=\text{LeakyReLU}(\mathbf{a}^\top \left[ \mathbf{V}\mathbf{x}_i \;\lVert\; \mathbf{V}\mathbf{x}_j \right])
$$

where $\mathbf{V}$ and $\mathbf{a}$ are trainable. This was later found to be restrictive, in the sense that the ordering between elements does not depend on the central node \cite{brody2021attentive}. A less restrictive variant, called \textbf{GATv2} \cite{brody2021attentive} is obtained as:

$$
\alpha(\mathbf{x}_i, \mathbf{x}_j)= \mathbf{a}^\top\text{LeakyReLU}( \mathbf{V}\left[ \mathbf{x}_i \;\lVert\; \mathbf{x}_j \right])
$$

Both GAT and GATv2 are very popular baselines nowadays.

\subsection{Message-passing neural networks} \addteacup

Suppose we have available additional \textbf{edge features} $\mathbf{e}_{ij}$, e.g., in a molecular dataset we may know a one-hot encoded representation of the type of each molecular bond. We can generalize the GAT layer to include these features by properly modifying the attention function:

$$
\alpha(\mathbf{x}_i, \mathbf{x}_j)= \mathbf{a}^\top\text{LeakyReLU}( \mathbf{V}\left[ \mathbf{x}_i \;\lVert\; \mathbf{x}_j \;\Vert\; {\color{drawred}\mathbf{e}_{ij}} \right])
$$

We can further generalize all the layers seen up to now (GC, GAT, GATv2, GAT with edge features) by abstracting away their basic components. Consider a very general layer formulation:

\begin{equation}
\mathbf{h}_i =\psi\left(\mathbf{x}_i, \text{Aggr}\left(\left\{M(\mathbf{x}_i, \mathbf{x}_j, \mathbf{e}_{ij})\right\}_{\mathcal{N}(i)}\right) \right)
\label{eq:message_passing_layer}
\end{equation}

where:

\begin{enumerate}
\item $M$ builds a feature vector (which we call a \textbf{message}) relative to the edge between node $i$ and node $j$. Contrary to GC and GAT layers, we are not restricting the message to be scalar-valued.
\item $\text{Aggr}$ is a generic permutation invariant function (e.g., a sum) to aggregate the messages from all nodes connected to node $i$.
\item $\psi$ is a final block that combines the aggregated message with the node features $\mathbf{x}_i$. In this way, two nodes with the same neighborhood can still be distinguished.
\end{enumerate}

As an example, in a GC layer the message is built as $M(\_, \mathbf{x}_j, \_)=A_{ij}\mathbf{W}^\top\mathbf{x}_j$, the aggregation is a simple sum, and $\psi(\_, \mathbf{x})=\phi(\mathbf{x})$. The general layer \eqref{eq:message_passing_layer} was introduced in \cite{gilmer2017neural} with the name of \textbf{message-passing} layer, and it has become a very popular way to categorize (and generalize) layers operating on graphs \cite{velivckovic2022message}.

Let us consider a few examples of using this message-passing framework. First, we may want to give more importance to the central node in the message-passing phase. We can do this by modifying the $\psi$ function:

$$
\psi(\mathbf{x}, \mathbf{m})=\phi(\mathbf{V}\mathbf{x}+\mathbf{m})
$$

where $\mathbf{V}$ is a generically trainable matrix (this was introduced in \cite{morris2019weisfeiler} and popularized in PyTorch Geometric as the GraphConv\footnote{\url{https://pytorch-geometric.readthedocs.io/en/latest/generated/torch_geometric.nn.conv.GraphConv.html}} layer). Second, suppose nodes have available more complex features such as a time series per node (e.g., a distributed set of sensors). Note that in the message-passing framework, node-wise operations are decoupled from the way messages are aggregated and processed. Denoting by $x_i$ the time-series at node $i$, we can generalize the GC layer by simply modifying the message function with a layer working on time series, e.g., a Conv1d layer:

$$
h_i=\sum_{j \in \mathcal{N}(i)} A_{ij} \text{Conv1d}(x_i)
$$

This is an example of a \textbf{spatio-temporal} GC layer \cite{yu2017spatio}. Furthermore, up to now we have assumed that only node features should be updated. However, it is easy to also update edge features by an additional edge update layer:

$$
\mathbf{e}_{ij} \leftarrow\text{MLP}(\mathbf{e}_{ij},\mathbf{h}_i, \mathbf{h}_j)
$$

This can also be seen as a message-passing iteration, in which the edge aggregates messages from its neighbors (the two connected nodes). This line of reasoning allows to further generalize these layers to consider more extended neighborhoods and graph features \cite{battaglia2018relational}.

This is a very brief overview that provides a gist of many possible message-passing variants. There are many topics we are not able to cover in detail due to space: among these, we single out building MP layers for higher-order graphs (in which edges connect more that a pair of nodes) \cite{chien2021you} and MP layers for point cloud data, in which we are interested in satisfying additional symmetries (rotational and translational symmetries) \cite{satorras2021n,eijkelboom2023n}. \clearpage

\subsection{Graph transformers}

We have seen two techniques to employ the graph structure: the first one is to add a regularization term that forces the network’s outputs to be smooth relative to the graph; the second one is to constrain the operations of the graph to follow the graph connectivity. In particular, in the GAT layer we have used a standard attention operation by properly masking the pairwise comparisons. However, we have also seen in the previous chapter that transformers have become popular because they provide an architecture that is completely agnostic from the type of data. Can we design the equivalent of a \textbf{graph transformer} \cite{muller2023attending}?

Recall that the two basic steps for building a transformer are tokenization of the input data and definition of the positional embedding. Tokenization for a graph is simple: for example, we can consider each node as a token, or (if edge features are given) each node and each edge as separate tokens after embedding them in a shared space. Let us ignore for now edge features. Consider the generic architecture taking as input the node features:

$$
\mathbf{H} = \text{Transformer}(\mathbf{X})
$$

This is permutation equivariant but completely agnostic to the connectivity. We can partially solve this by augmenting the node features with some graph-based features, such as the degree of the node, or the shortest path distance to some pre-selected nodes (anchors) \cite{rampavsek2022recipe,muller2023attending}. More in general, however, we can consider an embedding of the graph connectivity  into what we call a \textbf{structural embedding}: \clearpage

$$
\mathbf{H} = \text{Transformer}(\mathbf{X} + \text{Embedding}(\mathbf{A}))
$$

\begin{figure}[t]
    \centering
    \hspace{1em}\includegraphics[width=0.8\textwidth]{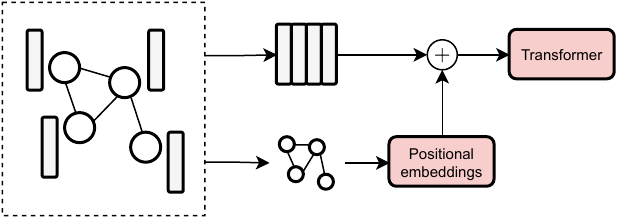}
    \caption{General idea of a graph transformer: the connectivity is embedded into a set of positional embeddings, which are added to the collected features. The result is then processed by a standard transformer network.}
    \label{fig:graph_transformer}
\end{figure}

Each row of $\text{Embedding}(\mathbf{A})$ provides a vectorial embedding of the connectivity of the graph relative to a single node, ignoring all features (see Figure \ref{fig:graph_transformer}). Luckily, embedding the structure of a graph into a vector space is a broad field. As an example, we describe here a common embedding procedure based on random walks \cite{dwivedi2021graph}. Recall that the following matrix:

$$
\mathbf{R} =\mathbf{A}\mathbf{D}^{-1}
$$

can be interpreted as a “random walk”, in which $R_{ij}$ is the probability of moving from node $i$ to node $j$. We can iterate the random walk multiple times, for a fixed $k$ set a priori by the user:
$$
\mathbf{R}, \mathbf{R}^2, \ldots,\mathbf{R}^k
$$

Random walk embeddings are built by collecting all the walk probabilities of a node returning on itself, and projecting them to a fixed-dimensional embedding:

$$
\text{Embedding}(\mathbf{A})=\begin{bmatrix} \text{diag}(\mathbf{R}) \\ \text{diag}(\mathbf{R}^2) \\ \vdots \\ \text{diag}(\mathbf{R}^k)\end{bmatrix}\mathbf{W}
$$

Under specific conditions on the graph structure, this can be shown to provide a unique representation for each node \cite{dwivedi2021graph}. Alternative types of embeddings can be obtained by considering eigen-decompositions of the Laplacian matrix \cite{lim2022sign}. For a fuller exposition of graph transformers, we refer to \cite{muller2023attending}. Building graph transformers opens up the possibility of GPT-like foundation models for the graph domain, and also of adding graph-based data as an additional modality to existing language models \cite{maoposition}.

\section*{From theory to practice}

\begin{wrapfigure}{r}{3.0cm}
\vspace{-5em}\includegraphics[width=3.0cm]{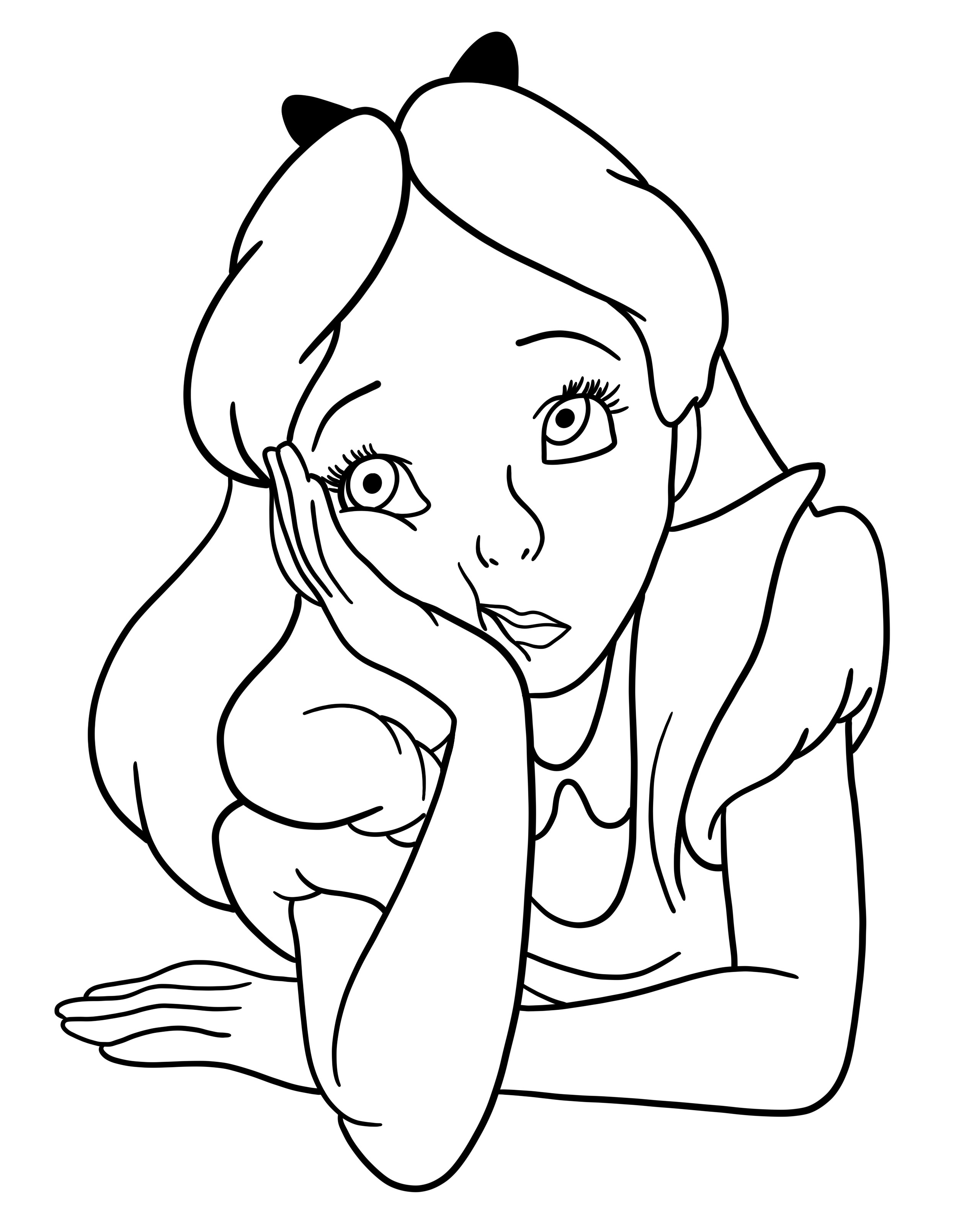}
\vspace{-2em}
\end{wrapfigure}

Handling efficiently graph data requires extensions of the basic frameworks, due to the problems described in this chapter (e.g., sparsity). Common libraries include PyTorch Geometric for PyTorch, and Jraph for JAX. Both have ample sets of tutorials, for example for node classification in small citation networks.\footnote{Recommended example in PyTorch Geometric: \url{https://pytorch-geometric.readthedocs.io/en/latest/get_started/introduction.html}.}

\begin{figure}[t]
    \centering
    \hspace{1em}\includegraphics[width=0.8\textwidth]{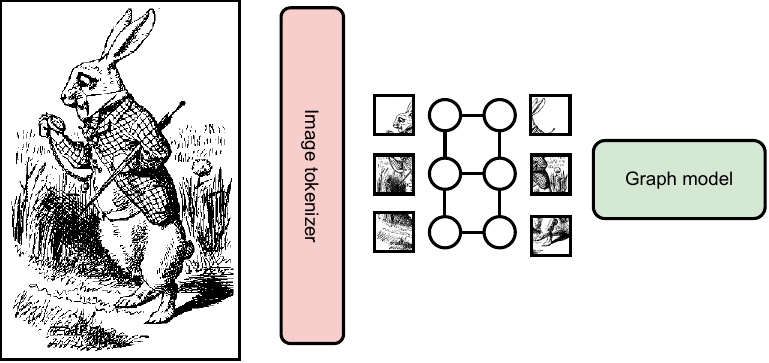}
    \caption{A GNN for computer vision: the image is tokenized into patches, an adjacency matrix is built over the patches, and the two are propagated through a graph model.}
    \label{fig:image_gnn}
\end{figure}

If you implemented a Vision Transformer in Chapter \ref{chap:transformers_in_practice}, I suggest a funny exercise which has (mostly) didactic value, as shown in Figure \ref{fig:image_gnn}. Suppose we tokenize the image into patches, but instead of adding positional embeddings, we construct an adjacency matrix $\mathbf{A} \sim (p, p)$ (where $p$ is the number of patches) as:
\begin{equation}
A_{ij} = \begin{cases} 1 & \text{ if the two patches share a border in the image} \\ 0 & \text{ otherwise} \end{cases}
\label{eq:image_adjacency_matrix}
\end{equation}

We now have a graph classification dataset, where the node features are given by the patch embedding, and the adjacency matrix by \eqref{eq:image_adjacency_matrix}. Thus, we can perform image classification by adapting the GNN from the previously-mentioned tutorials.

%% file: 13_recurrent_models.tex
\chapter{Recurrent models}
\label{chap:rnns}

\begin{supportbox}{About this chapter}
Transformer models are very effective at processing sequences, but they are hindered by their quadratic complexity in the sequence length. One possibility is to replace them with recurrent layers, having only constant-time for processing each element of a sequence, irrespective of its length. In this final chapter we provide an overview of several recurrent models and their characteristics. The field has been moving very rapidly in the last two years, and we provide a wide overview at the expense of precision -- see \cite{tiezzi2024state} for a recent survey.
\end{supportbox}

\section{Linearized attention models}
\subsection{Replacing the dot product}
\label{sec:linearized_transformer_model}

To provide some intuition on why recurrent neural networks (RNNs) can be useful, we begin with a generalization of the attention layer (called the \textbf{linearized attention layer} \cite{katharopoulos2020transformers}) that can be written in a recurrent form. We start by rewriting the SA layer in an abstract form with a generic scalar-valued attention function $\alpha(\bullet, \bullet)$ instead of the dot product:
\begin{equation}
\mathbf{h}_i=\frac{\sum_{j=1}^n\alpha\left(\mathbf{q}_i, \mathbf{k}_j\right)\mathbf{v}_j}{\sum_{j=1}^n \alpha\left(\mathbf{q}_i, \mathbf{k}_j\right)}
\label{eq:generalized_attention}
\end{equation}
where for the standard SA, $\alpha(\mathbf{x}, \mathbf{y})=\exp(\mathbf{x}^\top\mathbf{y})$. If the elements of the sequence must be processed in order (as in autoregressive generation), \eqref{eq:generalized_attention} is inconvenient because its cost grows quadratically in the sequence length. Even if a KV cache is used, memory still grows linearly. By comparison, a convolutional layer has fixed time and memory cost for each element to be processed, but information is lost if a token is outside the receptive field. What we would like, then, is a mechanism to compress all the information of the sequence into a fixed-size input (which we will call a \textbf{memory} or \textbf{state} tensor), so that the cost of running the model on our current input token plus the memory is constant. We call models of this form \textbf{recurrent}.

To begin, note that any non-negative $\alpha$ is a valid similarity function. In machine learning, this requirement is equivalent to $\alpha$ being what is called a \textbf{kernel function} \cite{hofmann2008kernel}. Many such kernel functions can be written as a generalized dot product:
\begin{equation}
\alpha(\mathbf{x}, \mathbf{y})=\phi(\mathbf{x})^\top \phi(\mathbf{y})
\label{eq:kernel}
\end{equation}
for some function $\phi: \mathbb{R}^c \rightarrow \mathbb{R}^e$ performing a feature expansion (this is the cornerstone of methods such as support vector machines, but discussing it at length would go beyond the scope of the book - also, it will not be required beyond this section). 

\clearpage

\begin{supportbox}{Kernel functions}
As an example of kernel function, the \textbf{polynomial kernel function} $\alpha(\mathbf{x}, \mathbf{y})=(1 + \mathbf{x}^\top\mathbf{y})^d$ can be rewritten as \eqref{eq:kernel} if $\phi(\bullet)$ explicitly computes all polynomials of its input up to order $d$ \cite{hofmann2008kernel}. Some kernel functions correspond to infinite-dimensional expansions (e.g., the Gaussian kernel), in which case \eqref{eq:kernel} can still be recovered in terms of an approximated kernel expansion, such as working with random Fourier features \cite{scardapane2017randomness}.
\end{supportbox}

Based on \eqref{eq:kernel} we can rewrite \eqref{eq:generalized_attention} as:

$$
\mathbf{h}_i=\frac{\sum_{j=1}^n \phi(\mathbf{q}_i)^\top\phi(\mathbf{k}_j)\mathbf{v}_j^\top}{\sum_{j=1}^n \phi(\mathbf{q}_i)^\top\phi(\mathbf{k}_j)}
$$

where we have added a transpose operation on $\mathbf{v}_j$ to be consistent with the dimensions. Because $\phi(\mathbf{q}_i)$ does not depend on $j$ we can bring it outside the sum to obtain:

\begin{equation}
\mathbf{h}_i=\frac{{\color{drawred}\phi(\mathbf{q}_i)^\top}\sum_{j=1}^n \phi(\mathbf{k}_j)\mathbf{v}_j^\top}{{\color{drawred}\phi(\mathbf{q}_i)^\top}\sum_{j=1}^n \phi(\mathbf{k}_j)}
\label{eq:linearized_attention}
\end{equation}

This is called a \textbf{linearized attention} model \cite{katharopoulos2020transformers}. Computing \eqref{eq:linearized_attention} for all tokens has complexity $\mathcal{O}(n(e^2 + ev))$, which is linear in the sequence length and advantageous whenever $n <e^2$. $\phi$ can be chosen freely, e.g., in \cite{katharopoulos2020transformers} they consider a quadratic feature expansion or even a simpler $\phi(\mathbf{x})=\text{ELU}(\mathbf{x})+1$ for short sequences.

\clearpage

\subsection{A recurrent formulation}

We now rewrite the linearized attention model in a recurrent form, by considering what happens for a causal variant of the layer. First, we modify \eqref{eq:linearized_attention} by constraining the sum only on past input elements to make it causal:

\begin{equation}
\mathbf{h}_i=\frac{\phi(\mathbf{q}_i)^\top\eqnmarkbox[drawred]{node}{\sum_{j=1}^{\color{drawred}i} \phi(\mathbf{k}_j)\mathbf{v}_j^\top}}{\phi(\mathbf{q}_i)^\top\eqnmarkbox[drawgreen]{node2}{\sum_{j=1}^{\color{drawred}i} \phi(\mathbf{k}_j)}}
\label{eq:atf}
\end{equation}
\annotate[yshift=1em]{above,right}{node}{Attention memory $\mathbf{S}_i$}
\annotate[yshift=-1em]{below,right}{node2}{Normalizer memory $\mathbf{z}_i$}

\vspace{1em}
This is our first example of a \textbf{recurrent layer}. To understand the name, we note that the attention and normalizer memories can be written recursively as:
\begin{gather}
\mathbf{S}_i =\mathbf{S}_{i-1} + \phi(\mathbf{k}_i)\mathbf{v}_i^\top \label{eq:atf_recurrent_1} \\ 
\mathbf{z}_i = \mathbf{z}_{i-1}+\phi(\mathbf{k}_i) \label{eq:atf_recurrent_2}
\end{gather}
where the base case of the recurrence is given by their initialization:
\begin{gather}
\mathbf{S}_0=\mathbf{0} \\ \mathbf{z}_0=\mathbf{0}
\label{eq:atf_recurrent_init}
\end{gather}
The output is then given by:
\begin{gather}
\mathbf{h}_i=\frac{\phi(\mathbf{q}_i)^\top\mathbf{S}_i}{\phi(\mathbf{q}_i)^\top\mathbf{z}_i} \label{eq:atf_recurrent_3}
\end{gather}
Equations \eqref{eq:atf_recurrent_1}-\eqref{eq:atf_recurrent_3} are particularly interesting for an autoregressive scenario: for any new token to be generated, we update the two memory states (equations \eqref{eq:atf_recurrent_1} and \eqref{eq:atf_recurrent_2}), and we use these updated states to compute the output for the $i$-th element. Importantly, the total computation for generating a new token is constant, and the cost in memory is also fixed since the previous memories $\mathbf{S}_{i-1}$ and $\mathbf{z}_{i-1}$ can be discarded. We can alternate between the two formulations of the layer: we can use a vectorized variant for training (for efficient implementation on GPUs) and the recurrent formulation for inference. 

\section{Classical recurrent layers}
\subsection{General formulation}

\begin{SCfigure}
    \centering
    \hspace{2em}\includegraphics[width=0.5\textwidth]{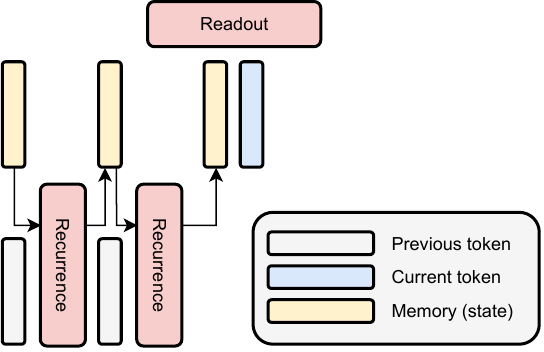}
    \caption{Overview of a recurrent layer: past tokens are shown in gray, current input token in blue, the memory state in yellow.}
    \label{fig:recurrence}
\end{SCfigure}

Let us now abstract away the key components of a recurrent layer, using the previous section as reference. First, we need a \textbf{state} of fixed size, which is used to compress all useful information up to the $i$-th element of the sequence. We denote it generically as $\mathbf{s}_i$, and without lack of generality we assume it is a single vector from now on. Second, we need a \textbf{transition function} (recurrence) that updates the state vector based on the previous value and the value of the current token, which we denote as $f(\mathbf{s}_{i-1}, \mathbf{x}_i)$. Third, we need what we call a \textbf{readout function} that provides an output for the $i$-th element of the sequence. We denote it as $g(\mathbf{s}_i, \mathbf{x}_i)$. See also Figure \ref{fig:recurrence} for a visualization. 

\begin{definition}[Recurrent layer] \addbottle $\,$

Given a sequence of tokens $\mathbf{x}_1, \mathbf{x}_2, \ldots$, a generic recurrent layer can be written as: 
\begin{gather}
\mathbf{s}_i = f(\mathbf{s}_{i-1},\mathbf{x}_i) \label{eq:recurrent_layer_state_update} \\ 
\mathbf{h}_i = g(\mathbf{s}_i, \mathbf{x}_i) \label{eq:recurrent_layer_readout}
\end{gather}

where the \textbf{state vector} $\mathbf{s}_i \sim (e)$ is initialized as zero by convention, $\mathbf{s}_0 = \mathbf{0}$. The size of the state vector, $e$, and the size of the output vector $\mathbf{h}_i \sim (o)$ are hyper-parameters. We call $f$ the \textbf{state transition} function and $g$ the \textbf{readout} function.

\end{definition}

In this format, a recurrent layer represents a discrete-time, input-driven dynamical system, and it is a causal layer by definition. In control engineering, this is also known as a \textbf{state-space model}. For tasks in which causality is unnecessary, \textbf{bidirectional layers} \cite{schuster1997bidirectional} can also be defined. In a bidirectional layer we initialize two recurrent layers (with separate parameters), one of which processes the sequence left-to-right, and the second one right-to-left. Their output states are then concatenated to provide the final output. 

Recurrent neural networks (RNNs) can be built by stacking multiple recurrent layers on the updated sequence $\mathbf{h}_1, \mathbf{h}_2, \ldots, \mathbf{h}_n$ \cite{pascanu2013construct}. Interestingly, a recurrent layer has no requirement on the length of the sequence, which can (in principle) be unbounded. For this reason, RNNs with unbounded precision or growing architectures can be shown to be Turing-complete \cite{chung2021turing}. \clearpage

\begin{supportbox}{Implicit layers}
What happens if we apply a recurrent layers to a \textit{single} token $\mathbf{x}$?
\begin{equation}
\mathbf{s}_i = f(\mathbf{s}_{i-1}, \mathbf{x})
\label{eq:recurrent_layer_single_token}
\end{equation}
If we run the state transition several time starting from a known initialization $\mathbf{s}_0$, this is similar to a model with several layers (one per transition) sharing the same parameters. Suppose we run \eqref{eq:recurrent_layer_single_token} an \textit{infinite} number of times. If the dynamic system has a stable attractor, the output will be defined by the fixed-point equation:
\begin{equation}
\mathbf{s} = f(\mathbf{s}, \mathbf{x})
\label{eq:fixed_point_layer}
\end{equation}
If we take \eqref{eq:fixed_point_layer} as the definition of a layer, we obtain what is called an \textbf{implicit layer} \cite{bai2019deep}. The implementation of implicit layers can be made feasible by using fast solvers for the fixed-point equation and computing the backward pass with the use of the \textbf{implicit function theorem} \cite{bai2019deep}. Implicit graph layers can also be defined by running each diffusion operation to a stable state \cite{gori2005new,scarselli2008graph}. 

\end{supportbox}

\vspace{-0.5em}
\subsection{“Vanilla” recurrent layers} \addclock

Historically, recurrent layers were instantiated by considering two fully-connected layers as transition and readout functions:
\begin{gather}
f(\mathbf{s}_{i-1},\mathbf{x}_i)= \phi(\mathbf{A}\mathbf{s}_{i-1}+\mathbf{B}\mathbf{x}_i) \\
g(\mathbf{s}_i,\mathbf{x}_i)=\mathbf{C}\mathbf{s}_i+\mathbf{D}\mathbf{x}_i
\end{gather}
where as always we ignore biases for simplicity, and we have four trainable matrices $\mathbf{A} \sim (e,e)$, $\mathbf{B} \sim (e,c)$, $\mathbf{C} \sim (o,e)$, and $\mathbf{D} \sim (o,c)$, where $c$ is the input dimensionality (the size of each token). A layer in this form is sometimes referred to generically as a “\textit{recurrent layer}”, a “\textit{vanilla recurrent layer}”, or an \textbf{Elman} recurrent layer. When the two matrices $\mathbf{A}$ and $\mathbf{B}$ are left untrained and we only have a single layer, these models are called \textbf{echo state networks} (ESNs) or \textbf{reservoir computers} \cite{lukovsevivcius2009reservoir}. ESNs can be a powerful baseline for time series forecasting, especially when the untrained matrices (the reservoir) are initialized in a proper way \cite{gauthier2021next}.

Despite their historical significance, layers of this form are extremely inefficient (and hard) to train. To see this, note that by its design the computation across elements of the sequence cannot be parallelized efficiently, as shown in Box \ref{code:recurrence}. Hence, we need to resort to iterative (for-loops) implementations, and even highly customized CUDA implementations\footnote{\url{https://docs.nvidia.com/deeplearning/performance/dl-performance-recurrent/index.html}} are slower than most alternative sequence layers. 

\begin{mypy}{Vanilla recurrence in PyTorch. It is impossible to parallelize the for-loop because of the dependencies in the recurrence. In PyTorch, the state update is called a \textbf{recurrent cell}, while the recurrent layers, such as {\footnotesize\mintinline{python}{torch.nn.RNN}}, wrap a cell and perform the complete for-loop.}{code:recurrence}
# Input tensor
x = torch.randn(batch_size, 
                sequence_length, 
                features)

# State tensor
s = torch.zeros(batch_size, 
                state_size)

# State update
state_update = nn.RNNCell(features, 
                          state_size)
for i in range(x.shape[1]):
  s = state_update(x[:, i, :], s)
\end{mypy}

Another issue stems from the gradients involved in the layer’s computations. Consider a simplified case having only the transition function. We can unroll the full computation as:

\begin{gather*}\mathbf{s}_1=f(\mathbf{s}_0,\mathbf{x}_1) \\ \mathbf{s}_2=f(\mathbf{s}_1,\mathbf{x}_2) \\ \vdots \\ \mathbf{s}_n=f(\mathbf{s}_{n-1},\mathbf{x}_n)\end{gather*}

\clearpage

This is similar to a model with $n$ layers, except that the parameters are shared (the same) across the layers. Below we focus on the quantity $\partial_{\mathbf{A}} \mathbf{s}_n$ (the weight Jacobian with respect to $\mathbf{A}$), but similar considerations apply to all gradients. Let us define the following cumulative product:

\begin{equation}\widetilde{\mathbf{s}}_i = \prod_{j={i+1}}^{n}\partial_{\mathbf{s}_{j-1} }f(\mathbf{s}_{j-1},\mathbf{x}_j)\end{equation}

\begin{figure}[t]
    \centering
    \includegraphics[width=0.65\textwidth]{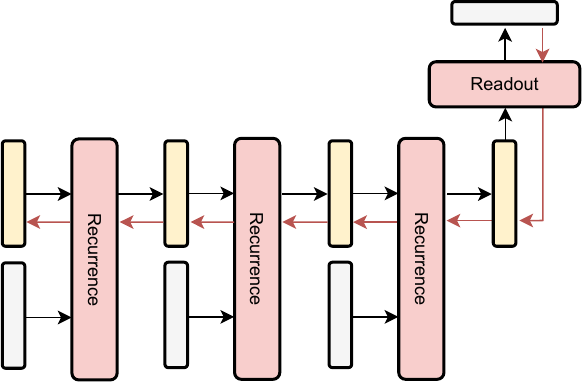}
    \caption{Backward pass for a recurrent layer: the adjoint values have to be propagated through all the transition steps. Each state then contributes a single term to the full gradient of the parameters.}
    \label{fig:recurrence_backward}
\end{figure}

This represents the gradient of the transition function from the end of the sequence backwards to element $i$, as shown in Figure \ref{fig:recurrence_backward}. Because of weight sharing, the gradient we are looking for has a separate term for each element in the sequence which involves these cumulative products:

\vspace{0.5em}
\begin{equation} 
\partial_{\mathbf{A}} \mathbf{s}_n = \eqnmarkbox[drawred]{node}{\partial_{\mathbf{A}} f(\mathbf{s}_{n-1}, \mathbf{x}_n)} + \sum_{i=1}^{n-1}  \eqnmarkbox[drawgreen]{node2}{\widetilde{\mathbf{s}}_i \bigl[ \partial_{\mathbf{A}} f(\mathbf{s}_{i-1}, \mathbf{x}_i)\bigr]}
\label{eq:backpropagation_through_time}
\end{equation}
\annotate[yshift=1em]{above,right}{node}{Gradient from element $n$}
\annotate[yshift=-1em]{below,left}{node2}{Gradient from element $i$}

The first term corresponds to a “standard” weight Jacobian, describing the influence of $\mathbf{A}$ on the last element of the sequence. The terms in the summation are the additional contributions, one for each element of the sequence, which are weighted by the chained input Jacobian computed over the sequence itself. 

Written in this form, reverse mode automatic differentiation is also called \textbf{backpropagation through time} (BPTT), and it can be a strong source of instability or gradient problems during gradient descent. To see this, note that each input Jacobian in the inner product in \eqref{eq:backpropagation_through_time} involves a multiplication by the derivative of the activation function $\phi$. Some of the earliest analyses of vanishing and exploding gradients were done in this context \cite{hochreiter1998recurrent}. For long sequences, stability of the layer is guaranteed only when the eigenvalues of the transition matrix are properly constrained \cite{gallicchio2017echo}. Layer normalization was also originally developed to stabilize training in RNNs, by computing statistics over the states’ sequence \cite{ba2016layer}.

Several techniques have been developed to partially solve these instabilities in the context of recurrent layers. For example, the sum in \eqref{eq:backpropagation_through_time} can be truncated to a given interval (\textbf{truncated BPTT}), or the gradients can be thresholded if they exceed a pre-defined upper bound (\textbf{clipped gradients}).

\subsection{Gated recurrent networks}

Over the years, several variants of the vanilla layer were proposed to improve its performance. In this section we focus on a popular class of such models, called \textbf{gated} RNNs. One issue of RNNs is that the entire state gets overwritten at each transition, which is reflected in the partial products in \eqref{eq:backpropagation_through_time}. However, we can assume that, for many sequences, only a few elements of these transitions are important: as an example, in an audio signal, empty regions or regions with no information are typical. In these cases, we may be interested in \textit{sparsifying} the transition (similarly to how most attention weights tend to be close to zero) and, consequently, setting most elements in $\widetilde{\mathbf{s}}_i$ to $1$. This can be achieved with the addition of specialized gating layers.

We consider the simplest form of gated RNN, called \textbf{light gated recurrent unit} (Li-GRU, \cite{ravanelli2018light}), having a single gate. For our purposes, a gating function is simply a layer that outputs values in the range $[0,1]$ that can be used to “mask” the input. As an example, a gate over the state can be obtained by a fully-connected layer with a sigmoid activation function:
$$
\gamma(\mathbf{s}_{i-1}, \mathbf{x}_i)=\sigma\left(\mathbf{V}\mathbf{s}_{i-1}+\mathbf{U}\mathbf{x}_i\right)
$$

where $\mathbf{V}$ and $\mathbf{U}$ have similar shapes to $\mathbf{A}$ and $\mathbf{B}$. We can interpret this as follows: if $\gamma_i \approx 0$, the $i$-th feature of the state should be kept untouched, while if $\gamma_1 \approx 1$, we should propagate its updated value as output. Hence, we can rewrite the transition function by properly masking the new and old values as:
\begin{align*}
f(\mathbf{s}_{i-1}, \mathbf{x}_i) = & \overbrace{\gamma(\mathbf{s}_{i-1}, \mathbf{x}_i)\odot\phi\left(\mathbf{A}\mathbf{s}_{i-1}+\mathbf{B}\mathbf{x}_i\right)}^{\text{New values}} \\ &  + \underbrace{(1-\gamma(\mathbf{s}_{i-1}, \mathbf{x}_i))\odot\mathbf{s}_{i-1}}_{\text{Old values}}
\end{align*}

\clearpage

This can be seen as a soft (differentiable) approximation to a “real” gate having only binary values, or as a convex combination of the original layer and a skip connection. We can theoretically control the goodness of this approximation by adding an additional regularizer to the loss that constrains the outputs of the gate to lie as close as possible to $0$ or $1$. 

Other gated recurrent layers can be obtained by adding additional gates to this design: the original  \textbf{gated recurrent unit} (GRU) adds a so-called “reset gate” to the layer \cite{cho2014learning}, while \textbf{long-short term memory} units (LSTMs) have a third “forget gate” \cite{hochreiter1997long}. LSTMs were the first gated variant to be introduced in the literature, and for a long time they have been the most successful deep architecture for processing sequences \cite{schmidhuber2015deep}. Because of this, research on LSTM models is still very active \cite{beck2024xlstm}.

\section{Structured state space models}
\subsection{Linear recurrent layers}

We now consider a simplified class of recurrent layers, in which we remove the intermediate nonlinearity in the transition function:
\begin{align}
f(\mathbf{s}_{i-1},\mathbf{x}_i)= \mathbf{A}\mathbf{s}_{i-1}+\mathbf{B}\mathbf{x}_i \label{eq:s4_1} \\
g(\mathbf{s}_i,\mathbf{x}_i)=\mathbf{C}\mathbf{s}_i+\mathbf{D}\mathbf{x}_i \label{eq:s4_2}
\end{align}
Written in this form, \eqref{eq:s4_1}-\eqref{eq:s4_2} are called \textbf{state space models} (SSM).\footnote{Confusingly, any recurrent layer in the form \eqref{eq:recurrent_layer_state_update}-\eqref{eq:recurrent_layer_readout} is an SSM, but in the neural network's literature the term SSM has come to be associated only with the linear variant. Sometimes we refer to them as \textbf{structured} SSMs because, as we will see, we need to properly constrain the transition matrix to make them effective.}  Intuitively, an SSM layer is “less expressive” than a standard recurrent layer (because of the lack of nonlinearities). However, this can be recovered by adding activation functions after the output, or by interleaving these layers with token-wise MLPs \cite{orvieto2023universality}.

Interest in this class of models (re)-started in 2020, when \cite{gu2020hippo} analyzed a theoretical construction for the matrix $\mathbf{A}$ in \eqref{eq:s4_1} that could efficiently compress one-dimensional input sequences. The result was called the HiPPO (\textbf{High-Order Polynomial Projection Operator}) matrix. A family of neural networks built by a stack of SSM layers based on the HiPPO theory soon followed, leading to the \textbf{Structured State Space for Sequence Modeling} (S4) layer in 2021 \cite{gu2021efficiently} and the simplified S4 model (S5) in 2022 \cite{smith2022simplified}. 

Because of their roots in HiPPO theory, the proposed SSM layers up to S4 considered a stack of 1D models, one for each channel of the input, with transition matrices initialized as HiPPO matrices. By contrast, S5 introduced a standard multi-input, multi-output model of the form in \eqref{eq:s4_1}-\eqref{eq:s4_2}, which is the one we describe here. In particular, we focus our analysis on a simplified variant known as the \textbf{linear recurrent unit} (LRU) \cite{orvieto2023resurrecting}.

This formulation has a number of interesting properties, mostly stemming from the associativity of the linear transition function. To see this, we start by noting that the recurrence has a closed form solution:

\begin{equation}
\mathbf{s}_i =\sum_{j=1}^i\mathbf{A}^{i-j}\mathbf{B}\mathbf{x}_j
\end{equation}

We can view this summation from two different points of view. First, we can aggregate all coefficients with respect to the input sequence into a rank-3 tensor:

$$
K=\text{stack}\left( \mathbf{A}^{n-1}\mathbf{B}, \mathbf{A}^{n-2}\mathbf{B},\ldots,\mathbf{A}\mathbf{B},\mathbf{B}\right)
$$

We can compute all outputs via a single 1D convolution of filter size equal to the length of the sequence (a \textit{long} convolution) between the input sequence stacked into a single matrix $\mathbf{X} \sim (n,c)$ and the pre-computed kernel $K$:

$$
\mathbf{S}=\text{Conv1D}(\mathbf{X},K)
$$

Hence, the SSM layer can be interpreted as a convolution \cite{gu2021combining}. If the transition matrix is applied on a single channel, this can be exploited to speed-up computations by operating in the frequency domain, e.g., in the FlashConv implementation.\footnote{\url{https://www.together.ai/blog/h3}} However, a more efficient solution can be found by exploiting a family of algorithms known as \textbf{associative (parallel) scans} (or \textbf{all-prefix-sums}).

\subsection{An interlude: associative scans}

We introduce parallel scans in their general formulation before seeing their application to linear SSMs. Consider a sequence of elements $(x_1, x_2, \ldots, x_n)$, and an operation $\star$ which is assumed binary (it acts on any two elements of the sequence) and associative. We want to compute all partial applications of this operator to the sequence (using separate colors for readability):

$$
{\color{drawred_l}x_1}, \;\; {\color{drawgreen}x_1\star x_2}, \;\; {\color{drawblue}x_1\star x_2 \star x_3}, \;\; \ldots, \;\; x_1\star x_2 \star \cdots\star x_n
$$

This can be done trivially by an iterative algorithm which computes the elements one-by-one, adding one element at every iteration (this corresponds to how a standard recurrent layer would be computed). However, we can devise an efficient \textit{parallel} algorithm by exploiting the associativity of the operator $\star$ \cite{blelloch1990prefix}. The key intuition is that multiple pairs of elements can be computed in parallel and then aggregated recursively.

\begin{SCfigure}
    \centering
    \hspace{1em}\includegraphics[width=0.45\textwidth]{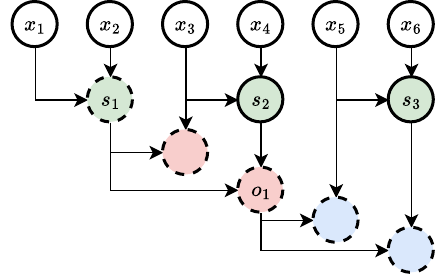}
    \caption{Parallel scan on a sequence of six elements: circles of the same color can be computed in parallel; dashed circles are the outputs of the parallel scan.}
    \label{fig:associative_scan}
\end{SCfigure}

As a simple example, consider a sequence of 6 elements $x_1,x_2,x_3, x_4, x_5, x_6$ (an in-depth example applied to SSMs can be found in \cite{smith2022simplified}). We will denote by $\hat{x}_i$ the $i$-th prefix we want to compute. The overall procedure is shown schematically in Figure \ref{fig:associative_scan}. We first aggregate pairs of adjacent values as:
\begin{align*}
s_1 = x_1 \star x_2  & \rightarrow\hat{x}_2 \\s_2 = x_3 \star x_4 &\\ s_3 = x_5 \star x_6 &
\end{align*}

\clearpage

where we use arrows to denote outputs of the algorithm. We now perform a second level of aggregations:
\begin{align*}
s_1 \star x_3 & \rightarrow \hat{x}_3 \\o_1 = s_1 \star s_2 & \rightarrow \hat{x}_4
\end{align*}
And finally:
\begin{align*} 
o_1 \star x_5 & \rightarrow \hat{x}_5 \\ o_1 \star s_3 & \rightarrow \hat{x}_6
\end{align*}
While this looks strange (we made 7 steps instead of 5), the three blocks of computations can be trivially parallelized if we have access to 3 separate threads. In general, by organizing the set of computations in a balanced fashion, we are able to compute the parallel scan in $\mathcal{O}(T \log n)$, where $T$ is the cost of the binary operator $\star$. An example of implementation is the associative scan function in JAX.\footnote{\url{https://jax.readthedocs.io/en/latest/_autosummary/jax.lax.associative_scan.html}}

It is easy to show that the transition function in a linear SSM is an example of an all-prefix-sums problem. We define the elements of our sequence as pairs $x_i = (\mathbf{A}, \mathbf{B}\mathbf{x}_i)$, and the binary operator as:
$$
(\mathbf{Z}, \mathbf{z})\star (\mathbf{V}, \mathbf{v})=(\mathbf{V}\mathbf{Z}, \mathbf{V}\mathbf{z}+\mathbf{v})
$$
The prefixes of $\star$ are then given by \cite{smith2022simplified}:

$$
x_1 \star x_2 \star \ldots \star x_i=(\mathbf{A}^i, \mathbf{s}_i)
$$

Hence, running a parallel scan gives us the powers of $\mathbf{A}$ as the first elements of the output, and all the states of the layer as the second element of the output. The complexity of this operation is upper bounded by the complexity of $\mathbf{A}^{i-1}\mathbf{A}$, which scales as $\mathcal{O}(n^3)$. To make the entire procedure viable, we can constrain $\mathbf{A}$ so that its powers can be computed more efficiently. This is the topic of the next section.

\subsection{Diagonal SSMs}

A common strategy to make the previous ideas feasible is to work with diagonal transition matrices (or diagonal matrices plus a low-rank term \cite{gu2021efficiently}). In this case, powers of $\mathbf{A}$ can be computed easily by taking powers of the diagonal entries in linear time. In addition, as we will see, working with diagonal matrices allows us to control the dynamics of the transition function to avoid numerical instabilities.

In particular, a square matrix $\mathbf{A}$ is said to be \textbf{diagonalizable} if we can find another square (invertible) matrix $\mathbf{P}$ and a diagonal matrix $\mathbf{\Lambda}$ such that:

\begin{equation}
\mathbf{A} = \mathbf{P}\mathbf{\Lambda}\mathbf{P}^{-1}
\label{eq:diagonalizable_A}
\end{equation}

Diagonalizable matrices are (in a sense) “simpler” that generic matrices, For example, if such a decomposition exists, it is easy to show that powers can also be computed efficiently as:

$$
\mathbf{A}^i =\mathbf{P}\mathbf{\Lambda}^i\mathbf{P}^{-1}
$$

Suppose that the transition matrix is diagonalizable. Then, we can re-write the SSM in an equivalent form having a diagonal transition matrix. We begin by substituting \eqref{eq:diagonalizable_A} into the definition of the SSM and multiplying on both sides by $\mathbf{P}^{-1}$:

$$
\eqnmarkbox[drawred]{node}{\mathbf{P}^{-1}\mathbf{s}_i} =\sum_{j=1}^i\mathbf{\Lambda}^{i-j}\eqnmarkbox[drawgreen]{node2}{\mathbf{P}\mathbf{B}}\mathbf{x}_j
$$
\annotate[yshift=-1em]{below,left}{node}{New state vector $\bar{\mathbf{s}}_i$}
\annotate[yshift=-1em]{below,right}{node2}{New input-state matrix $\bar{\mathbf{B}}$}

\vspace{1em}
We now rewrite the readout function in terms of the new variable $\bar{\mathbf{s}}$:

$$
\mathbf{y}_i = \eqnmarkbox[drawred]{node}{\mathbf{C}\mathbf{P}}\bar{\mathbf{s}}_i + \mathbf{D}\mathbf{x}_i
$$
\annotate[yshift=1em]{above,right}{node}{New readout matrix $\bar{\mathbf{C}}$}

Putting everything together:
\begin{gather}
\bar{\mathbf{s}}_i=\mathbf{\Lambda}\bar{\mathbf{s}}_{i-1}+\bar{\mathbf{B}}\mathbf{x}_i \\ \mathbf{y}_i=\bar{\mathbf{C}}\bar{\mathbf{s}}_i+ \mathbf{D}\mathbf{x}_i
\end{gather}

Hence, whenever a diagonalization of $\mathbf{A}$ exists, we can always rewrite the SSM into an equivalent form having a diagonal transition matrix. In this case, we can directly train the four matrices $\mathbf{\Lambda} = \text{diag}(\lambda), \lambda \sim (e)$, $\bar{\mathbf{B}} \sim (e,c)$, $\bar{\mathbf{C}} \sim (o,e)$ and $\mathbf{D} \sim (o,c)$, with the diagonal matrix being parameterized by a single vector of dimension $e$.

Not all matrices can be diagonalized. However, an approximate diagonalization can always be found if one allows for matrices $\mathbf{P}$ and $\mathbf{\Lambda}$ to have complex-valued entries \cite{orvieto2023resurrecting}. Care must be taken to parameterize the values over the diagonal so that the eigenvalues of the transition matrix stay $< 1$ in absolute value, to avoid diverging dynamics. We refer to \cite{orvieto2023resurrecting} for a description of both points and for a complete analysis of the resulting LRU layer.

\vspace{-1em}
\section{Additional variants}
Balancing the different strengths of convolutions, recurrence, and attention is an active research topic. To close the book, we list some recurrent layers (or layers that can be interpreted as recurrent) that have been introduced very recently in the literature.
\subsection{Attention-free transformers}
One issue of the linearized transformer model (Section \ref{sec:linearized_transformer_model}) is the quadratic complexity in the feature dimension $e$. The attention-free transformer (ATF) was introduced as a variant of the basic attention layer that is instead linear in both sequence length and in the number of features \cite{zhai2021attention}.

The core idea is to replace the dot product interactions between keys, query, and values with a simpler \textit{multiplicative interaction} (element-wise):
\begin{equation}
\mathbf{h}_i=\sigma(\mathbf{q}_i) \odot\frac{\sum_j \exp\left(\mathbf{k}_j \right) \odot \mathbf{v}_j}{\sum_j\exp\left(\mathbf{k}_j \right)}
\label{eq:atf_new}
\end{equation}
This is similar to the self-attention layer, except that we replace all dot products with element-wise (Hadamard) multiplications. It is also inspired by the linearized attention layer in that the query is only used as a global modulation factor, in this case after normalizing it with a sigmoid operation. In fact, we can recover a standard attention formulation by rewriting \eqref{eq:atf_new} for a single dimension $z$ (exploiting the fact that we only perform element-wise operations):
$$
h_{iz}=\frac{\sigma(q_{iz})\sum_{j}\exp(k_{jz})}{\sum_j \exp(k_{jz})}v_{jz}
$$
Hence, the ATF layer can be re-interpreted as a channel-wise variant of attention, in the sense that for every channel we can rewrite it as an attention operation over the elements of the sequence. To increase flexibility, \cite{zhai2021attention} also considered adding relative embeddings $\mathbf{W} \sim (m, m)$ (where $m$ is the maximum allowed length of the sequences):
\begin{equation}\mathbf{h}_i=\sigma(\mathbf{q}_i) \odot\frac{\sum_j \exp\left(\mathbf{k}_j +W_{ij}\right) \odot \mathbf{v}_j}{\sum_j\exp\left(\mathbf{k}_j +W_{ij}\right)}\end{equation}

The relative embeddings can also be trained via a low-rank factorization to reduce the number of parameters. See \cite{zhai2021attention} for this and for additional variants of the basic ATF layer (e.g., hybridizing it with convolutional operations). We can also convert \eqref{eq:atf_new} to a causal (recurrent) variant by properly restricting the summation.

\subsection{The Receptance Weighted Key Value (RWKV) model}

The RWKV model \cite{peng2023rwkv} extends the ATF layer by incorporating a few additional architectural modifications. At the time of writing, this is one of the only pre-trained RNNs matching transformers at the largest scale, so we describe it in more detail. First, the relative embeddings are simplified by considering a single vector $\mathbf{w} \sim (e)$ which is scaled for each offset:
$$
w_{ij}=-(i-j)\mathbf{w}
$$
In addition, experiments showed that having a separate offset $\mathbf{u}$ (in place of $\mathbf{w})$ for the current element is beneficial. Written in causal form, this gives:

\begin{equation*}\footnotesize
\mathbf{h}_i={\color{drawred}\mathbf{W}_o\Biggl(\Bigr.}\sigma(\mathbf{q}_i) \odot\frac{{\color{drawred}\sum_{j=1}^{i-1}}\exp\left(\mathbf{k}_j +w_{ij}\right) \odot \mathbf{v}_j{\color{drawred}+\exp\left(\mathbf{k}_i+\mathbf{u}\right) \odot \mathbf{v}_i}}{{\color{drawred}\sum_{j=1}^{i-1}}\exp\left(\mathbf{k}_j +w_{ij}\right){\color{drawred}+\exp\left(\mathbf{k}_i+\mathbf{u}\right)}}{\color{drawred}\Bigl.\Biggr)}
\label{eq:rwkv}
\end{equation*}

where we highlight the differences from the basic ATF layer in red. The query is called the \textbf{receptance} in \cite{peng2023rwkv}, and an additional output projection $\mathbf{W}_o$ is added at the end. Second, the RWKV model modifies the standard MLP in the transformer block with a differently \textit{gated} token-wise block. For a given input token $\mathbf{x}$ this can be written as:
\begin{equation}\mathbf{y}=\sigma(\mathbf{W}_1\mathbf{x})\odot \mathbf{W}_2\max(0, \mathbf{W}_3\mathbf{x})^2\end{equation}

where $\mathbf{W}_1$, $\mathbf{W}_2$, and $\mathbf{W}_3$ are trainable parameters. This is a standard MLP except for the left-most gate and the use of the squared ReLU. As a final modification, all three projections in the first block (and also the two appearances of $\mathbf{x}$ in \eqref{eq:rwkv}) are replaced with convex combinations of $\mathbf{x}_i$ and $\mathbf{x}_{i-1}$ to improve performance, which is called \textit{token shift}.

\subsection{Selective state space models}

We have seen three classes of recurrent models: standard recurrent layers (and their gated versions), linearized attention layers, and structured state space models. Although they look different, it is relatively easy to move from one class of models to the other. To see this, let us consider a linearized attention layer where we ignore the denominator:
\begin{align}
\mathbf{S}_i=\mathbf{S}_{i-1}+\phi(\mathbf{k}_i)\mathbf{v}_i^\top \label{eq:lin_att_simplified_1} \\ \mathbf{h}_i=\phi(\mathbf{q}_i)^\top\mathbf{S}_i
\label{eq:lin_att_simplified_2}\end{align}
Apart from the matrix-valued state, we see this has the form of a SSM layer, except that some matrices (e.g., $\mathbf{C} = \phi(\mathbf{q}_i)^\top$) are not fixed but they depend on the specific input token. From the point of view of dynamic systems, we say that standard SSMs describe \textit{time-invariant} systems, while \eqref{eq:lin_att_simplified_1}-\eqref{eq:lin_att_simplified_2} describe a \textit{time-varying} system. This has inspired another class of SSM layers whose matrices are not constrained to be time-invariant, which have been called \textbf{selective} SSMs. Most of these models leverage the idea of attention layers of projecting the input multiple times before the layer’s computations.

\begin{SCfigure}
    \centering
    \hspace{2em}\includegraphics[width=0.4\textwidth]{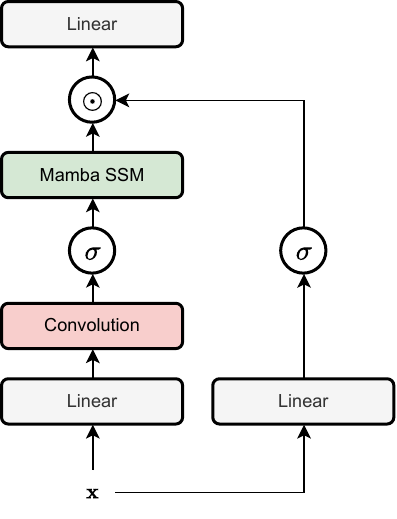}
    \caption{Mamba block (residual connections around the block and normalization are not shown). $\sigma$ is the sigmoid function. Adapted from \cite{gu2023mamba}.}
    \label{fig:mamba}
\end{SCfigure}

As an example, we focus here on the so-called Mamba layer \cite{gu2023mamba} which, at the time of writing, is one of the few SSM layers that was scaled to match the performance of transformer models at very large contexts and parameters’ counts. First, in order to make the SSM layer time-varying, a subset of its matrices are made input-dependent:\footnote{The matrix $\mathbf{D}$ can be seen as a simple residual connection and it is left untouched. The original layer has a slightly different parameterization where $\mathbf{A} = \exp(\Delta \bar{\mathbf{A}})$, for some trainable $\bar{\mathbf{A}}$ and input-dependent scalar value $\Delta$. This does not change our discussion.}
\begin{gather}
\mathbf{s}_i = A(\mathbf{x}_i)\mathbf{s}_{i-1} + B(\mathbf{x}_i)\mathbf{x}_i \\ \mathbf{h}_i = C(\mathbf{x}_i)\mathbf{s}_i + \mathbf{D}\mathbf{x}_i 
\end{gather}
where $A(\bullet)$, $B(\bullet)$, and $C(\bullet)$ are linear projections of their input tokens. To make this feasible, the layer is applied to each channel of the input independently, and the transition  matrix is selected as diagonal, so that all matrices of the SSM can be represented with a single vector of values. This layer looses a simple parallel scan implementation and requires a customized hardware-aware implementation \cite{gu2023mamba}. It can be shown that the Mamba SSM variant and several other SSM layers are degenerate case of a gated recurrent layer \cite{gu2021combining,gu2023mamba}.

To make the overall architecture simpler, Mamba avoids alternating MLPs and SSMs, in favour of a gated architecture (similar to the gated attention unit from Section \ref{subsec:mha_variants}) where an MLP is used to weight the outputs from the SSM. An additional depthwise convolution is added for improved flexibility - see Figure \ref{fig:mamba}.

%% file: appendix_a.tex
\chapter{Probability theory}
\label{chap:probability_theory}

\begin{supportbox}{About this chapter}
Machine learning deals with a wide array of uncertainties (such as in the data collection phase), making the use of probability fundamental. We review here - informally - basic concepts associated with probability distributions and probability densities that are helpful in the main text. This appendix introduces many concepts, but many of them should be familiar. For a more in-depth exposition of probability in the context of machine learning and neural networks, see \cite{bishop2006pattern,bishop2024deep}.
\end{supportbox}

\section{Basic laws of probability}

Consider a simple lottery, where you can buy tickets with 3 possible outcomes: “no win”, “small win”, and “large win”. For any 10 tickets, 1 of them will have a large win, 3 will have a small win, and 6 will have no win. We can represent this with a probability distribution describing the relative frequency of the three events (we assume an unlimited supply of tickets):
\begin{gather*}
p(w=\text{`no win'})=6/10 \\
p(w=\text{`small win'})=3/10\\
p(w=\text{`large win'})=1/10
\end{gather*}
Equivalently, we can associate an integer value $w=\left\{1,2,3\right\}$ to the three events, and write $p(w=1)=6/10$, $p(w=2)=3/10$, and $p(w=3)=1/10$. We call $w$ a \textbf{random variable}. In the following we always write $p(w)$ in place of $p(w=i)$ for readability when possible. The elements of the probability distribution must be positive and they must sum to one:
$$
p(w)\ge0,\;\sum_wp(w)=1
$$
The space of all such vectors is called the \textbf{probability simplex}. 

\begin{tcolorbox}
Remember that we use $\mathbf{p} \sim \Delta(n)$ to denote a vector of size $n$ belonging to the probability simplex. 
\end{tcolorbox}

Suppose we introduce a second random variable $r$, a binary variable describing whether the ticket is real (1) or fake (2). The fake tickets are more profitable but less probable overall, as summarized in Table \ref{tab:lottery_tickets}.
\begin{table}[h]
\centering
\caption{Relative frequency of winning at an hypothetical lottery, in which tickets can be either real or fake, shown for a set of $100$ tickets.}
\label{tab:lottery_tickets}
\begin{tabular}{@{}lcc@{}}
\toprule
 & $r=1$ (real ticket) & $r=2$ (fake ticket) \\ \midrule
$w=1$ (no win) & 58 & 2 \\
$w=2$ (small win) & 27 & 3 \\
$w=3$ (large win) & 2 & 8 \\ \midrule
Sum & 87 & 13 \\ \bottomrule
\end{tabular}
\end{table}

We can use the numbers in the table to describe a \textbf{joint probability distribution}, describing the probability of two random variables taking a certain value jointly:
$$
p(r=2,w=3)=8/100
$$
Alternatively, we can define a \textbf{conditional probability distribution}, e.g., answering the question “\textit{what is the probability of a certain event given that another event has occurred?}”:
$$
p(r=1 \mid w=3) = \frac{p(r=1, w=3)}{p(w=3)} = 0.2
$$
This is called the \textbf{product rule} of probability. As before, we can make the notation less verbose by using the random variable in-place of its value:
\begin{equation}
p(r,w)=p(r\mid w)p(w)
\label{eq:product_rule}
\end{equation}
If $p(r \mid w)=p(r)$ we have $p(r,w)=p(r)p(w)$, and we say that the two variables are \textbf{independent}. We can use conditional probabilities to \textbf{marginalize} over one random variable:
\begin{equation}
p(w)=\sum_{r} p(w,r) = \sum_r p(w \mid r)p(r)
\label{eq:sum_rule}
\end{equation}

This is called the \textbf{sum rule} of probability. The product and sum rules are the basic axioms that define the algebra of probabilities. By combining them we obtain the fundamental \textbf{Bayes’s rule}: \clearpage

\begin{equation}
    p(r\mid w)=\frac{p(w \mid r)p(r)}{p(w)} =\frac{p(w \mid r)p(r)}{\sum_{r^\prime}p(w \mid r^\prime)p(r^\prime)}
    \label{eq:bayes_rule}
\end{equation}

Bayes’s rule allows us to “reverse” conditional distributions, e.g., computing the probability that a winning ticket is real or fake, by knowing the relative proportions of winning tickets in both categories (try it).

\section{Real-valued distributions}

In the real-valued case, defining $p(x)$ is more tricky, because $x$ can take infinitely possible values, each of which has probability $0$ by definition. However, we can work around this by defining a probability \textbf{cumulative density function} (CDF):
$$
P(x)=\int_{0}^xp(t)dt
$$
and defining the probability density function $p(x)$ as its derivative. We ignore most of the subtleties associated with working with probability densities, which are best tackled in the context of measure theory \cite{bogachev2007measure}. We only note that the product and sum rules continue to be valid in this case by suitably replacing sums with integrals:
\begin{gather}
p(x,y)=p(x\mid y)p(y) \\ 
p(x)=\int_y p(x \mid y)p(y)dy
\end{gather}
Note that probability densities are not constrained to be less than one.

\section{Common distributions}

The previous random variables are example of \textbf{categorical probability distributions}, describing the situation in which a variable can take one out of $k$ possible values. We can write this down compactly by defining as $\mathbf{p} \sim \Delta(k)$ the vector of probabilities, and by $\mathbf{x} \sim \text{Binary}(k)$ a one-hot encoding of the observed class:

$$
p(\mathbf{x})=\text{Cat}(\mathbf{x}; \mathbf{p})=\prod_ip_i^{x_i}
$$

We use a semicolon to differentiate the input of the distribution from its parameters. If $k=2$, we can equivalently rewrite the distribution with a single scalar value $p$. The resulting distribution is called a \textbf{Bernoulli distribution}:

$$
p(x)=\text{Bern}(x; p)= p^x(1-p)^{(1-x)}
$$

In the continuous case, we will deal repeatedly with the \textbf{Gaussian} distribution, denoted by $\mathcal{N}(x; \mu, \sigma^2)$, describing a bell-shaped probability centered in $\mu$ (the mean) and with a spread of $\sigma^2$ (the variance):

$$
p(x)=\mathcal{N}(x;\mu,\sigma^2)= \frac{1}{\sqrt{2\pi \sigma^2}}\exp\left(-\frac{1}{2}\left(\frac{x-\mu}{\sigma}\right)^2\right)
$$

\clearpage

In the simplest case of mean zero and unitary variance, $\mu=0$, $\sigma^2=1$, this is also called the \textbf{normal distribution}. For a vector $\mathbf{x} \sim (k)$, a multivariate variant of the Gaussian distribution is obtained by considering a mean vector $\mathbf{\mu} \sim (k)$ and a covariance matrix $\Sigma \sim (k,k)$:
\begin{align*}
p(\mathbf{x})& =\mathcal{N}(\mathbf{x};\mu, \Sigma) =\\& \left(2\pi\right)^{-k/2}\det(\Sigma)^{-1/2}\exp\left((\mathbf{x}-\mu)^\top\Sigma^{-1}(\mathbf{x}-\mu)\right)
\end{align*}

Two interesting cases are Gaussian distributions with a diagonal covariance matrix, and the even simpler \textbf{isotropic} Gaussian having a diagonal covariance with all entries identical:
$$
\Sigma=\sigma^2\mathbf{I}
$$

The first can be visualized as an axis-aligned ellipsoid, the isotropic one as an axis-aligned sphere.

\section{Moments and expected values}

In many cases we need to summarize a probability distribution with one or more values. Sometimes a finite number of values are enough: for example, having access to $\mathbf{p}$ for a categorical distribution or to $\mu$ and $\sigma^2$ for a Gaussian distribution completely describe the distribution itself. These are called \textbf{sufficient statistics}. 

More in general, for any given function $f(x)$ we can define its \textbf{expected value} as:

\begin{equation}
\mathbb{E}_{p(x)}\left[f(x)\right]=\sum_{x}f(x)p(x)
\label{eq:expected_value}
\end{equation}

In the real-valued case, we obtain the same definition by replacing the sum with an integral. Of particular interest, when $f(x)=x^p$ we have the \textbf{moments} (of order $p$) of the distribution, with $p=1$ called the \textbf{mean} of the distribution:

$$
\mathbb{E}_{p(x)}\left[x\right]=\sum_{x}xp(x)
$$

We may want to estimate some expected values despite not having access to the underlying probability distribution. If we have access to a way of sampling elements from $p(x)$, we can apply the so-called \textbf{Monte Carlo estimator}:

\begin{equation}
\mathbb{E}_{p(x)}\left[f(x)\right]\approx \frac{1}{n}\sum_{x_i \sim p(x)}f(x_i)
\label{eq:montecarlo_estimation}
\end{equation}

where $n$ controls the quality of the estimation and we use $x_i \sim p(x)$ to denote the operation of sampling from the probability distribution $p(x)$. For the first-order moment, this reverts to the very familiar notation for computing the mean of a quantity from several measurements:

$$
\mathbb{E}_{p(x)}\left[x\right]=\frac{1}{n}\sum_{x_i \sim p(x)}x_i
$$

\section{Distance between distributions}

At times we may also require some form of distance between probability distributions, in order to evaluate how close two distributions are. The \textbf{Kullback-Leibler} (KL) divergence between $p(x)$ and $q(x)$ is a common choice:

$$
\text{KL}(p \;\lVert\; q) = \int p(x)\log\frac{p(x)}{q(x)}dx
$$

The KL divergence is not a proper metric (it is asymmetric and does not respect the triangle inequality). It is lower bounded at 0, but it is not upper bounded. The divergence can only be defined if for any $x$ such that $q(x)=0$, then $p(x)=0$ (i.e., the support of $p$ is a subset of the support of $q$). The minimum of $0$ is achieved whenever the two distributions are identical. The KL divergence can be written as an expected value, hence it can be estimated via Monte Carlo sampling as in \eqref{eq:montecarlo_estimation}.

\section{Maximum likelihood estimation} \addclock
\label{sec:maximum_likelihood_estimation}

Monte Carlo sampling shows that we can estimate quantities of interest concerning a probability distribution if we have access to samples from it. However, we may be interested in estimating the probability distribution itself. Suppose we have a guess about its functional form $f(x; s)$, where $s$ are the sufficient statistics (e.g., mean and variance of a Gaussian distribution), and a set of $n$ samples $x_i \sim p(x)$. We call these samples identical (because they come from the same probability distribution) and independently distributed, in short, i.i.d. Because of independence, their joint distribution factorizes for any choice of $s$:

$$
p(x_1, \ldots, x_n)=\prod_{i=1}^n f(x_i; s)
$$

Large products are inconvenient computationally, but we can equivalently rewrite this as a sum through a logarithmic transformation:

$$
L(s)= \sum_{i=1}^n\log(f(x_i;s))
$$

Finding the parameters $s$ that maximize the previous quantity is called the \textbf{maximum likelihood} (ML) approach. Because of its importance, we reframe it briefly below.

\begin{definition}[Maximum likelihood] \addbottle $\,$

Given a parametric family of probability distributions $f(x; s)$, and a set of $n$ values $\left\{x_i\right\}_{i=1}^n$ which are i.i.d. samples from an unknown distribution $p(x)$, the best approximation to $p(x)$ according to the \textbf{maximum likelihood} (ML) principle is:

$$
s^*=\underset{s}{\arg\max} \sum_{i=1}^n \log(f(x_i;s))
$$
\end{definition}

If $f$ is differentiable, we can maximize the objective through gradient descent. This is the core approach we follow for training differentiable models. For now, we close the appendix by describing simple examples of ML estimation in the case of standard probability distributions. We do not provide worked out calculations, for which we refer to \cite{bishop2006pattern,bishop2024deep}. 

\subsubsection*{Maximum likelihood for the Bernoulli distribution}

Consider first the case of a Bernoulli distribution with unknown parameter $p$. In this case, the ML estimator is:

$$
p^*=\frac{\sum_i x_i}{n}
$$

which is the ratio of positive samples over the entire dataset.

\subsubsection*{Maximum likelihood for the Gaussian distribution}

For the Gaussian distribution, we can rewrite its log likelihood as:

$$
L(\mu, \sigma^2) = - \frac{n}{2}\log(2\pi \sigma^2) - \frac{1}{2\sigma^2}\sum_{i=1}^n(x_i-\mu)^2
$$

Maximizing for $\mu$ and $\sigma^2$ separately returns the known rules for computing the empirical mean and variance of a Gaussian distribution:
\begin{gather}
\mu^*=\frac{1}{n}\sum_i x_i \\\sigma^{2*}=\frac{1}{n}\sum_i(x_i-\mu^*)^2
\end{gather}

The two can be computed sequentially. Because we are using an estimate for the mean inside the variance’s formula, the resulting estimation is shown to be slightly biased. This can be corrected by modifying the normalization term to $\frac{1}{n-1}$; this is known as Bessel’s correction.\footnote{\url{https://en.wikipedia.org/wiki/Bessel\%27s\_correction}} For large $n$, the difference between the two variants is minimal.

%% file: appendix_b.tex
\chapter[1D universal approximation]{1D universal \\ approximation}
\label{sec:universal_approximation}

\begin{supportbox}{About this chapter}
While formally proving the universal approximation theorem is beyond the scope of this book, it is helpful to get an intuitive feeling for how such proofs can be constructed. In this appendix we follow and extend the visual intuitions from a 2019 online book chapter by M. Nielsen,\footnote{\url{http://neuralnetworksanddeeplearning.com/chap4.html}} to which we refer for an extended discussion (and some interactive visualizations), especially for the case of multi-dimensional inputs.
\end{supportbox}

We focus on the original approximation theorem by Cybenko \cite{cybenko1989approximation} which considers models having one hidden layer with sigmoid activation functions. We also restrict the analysis to functions with a single input and a single output, that can be visualized easily. The reasoning can be extended to other activation functions and to higher dimensions.

The outline of this visual proof is relatively simple: 

\begin{enumerate}
\item As a first step, we show how to manually set the weights of a model with a single neuron in the hidden layer to approximate a step function.
\item Then, we proceed to show how adding another unit in the hidden layer allows to approximate any function which is constant over a small interval, and zero everywhere else (we call these interval functions “bin” functions). 
\item Finally, we describe a simple procedure to approximate a generic function by first binning it to the desired accuracy, and then adding as many neurons as needed to approximate all bins in turn. For $m$ bins we obtain a network with $2m$ neurons. For a generic function with multiple inputs, this number would grow exponentially in the number of dimensions, making the proof non constructive in a practical case.
\end{enumerate}

\section{Approximating a step function}

To begin, let us consider a single neuron in the hidden layer, in which case we can write the network’s equation as (ignoring the output bias term, as it is not helpful in our derivation):

$$
f(x)=a\sigma(wx+s)
$$

For the purposes of visualization, we rewrite this by adding a minus sign on the bias, and we factor the multiplication term on the entire input of $\sigma$ (the two variants are clearly equivalent):

\begin{equation}
f(x) = \eqnmarkbox[drawred]{node}{a}\sigma(\eqnmarkbox[drawgreen]{node2}{w}(x-\eqnmarkbox[drawblue]{node3}{s}))
\label{eq:proof_1}
\end{equation}
\annotate[yshift=-1em]{below,left}{node}{Amplitude}
\annotate[yshift=-1em]{below,right}{node2}{Slope}
\annotate[yshift=-1em]{below,right}{node3}{Shift}

\vspace{1em}
This is similar to the “tunable” variant of sigmoid we introduce in Section \ref{sec:activation_functions}. In particular, in this formulation $\color{drawred}a$ controls the amplitude of the sigmoid, $\color{drawgreen}w$ controls the slope, and $\color{drawblue}s$ shifts the function by a fixed amount.

\begin{SCfigure}
    \centering
    \hspace{1em}\includegraphics[width=0.5\textwidth]{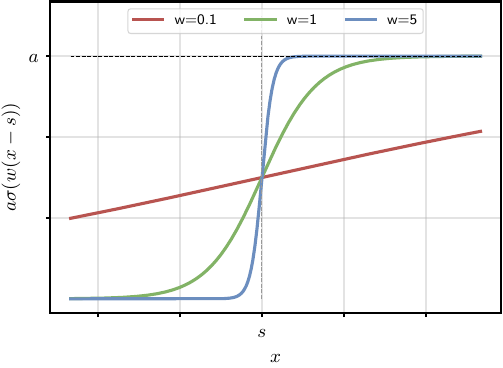}
    \caption{A network with a single neuron in the hidden layer can be visualized as a sigmoid with controllable slope, center, and amplitude. We show here an example where we fix the amplitude and the center, but we vary the slope.}
    \label{fig:tunable_sigmoid}
\end{SCfigure}

\begin{figure}
    \centering
    \begin{subfigure}[b]{0.32\textwidth}
    \includegraphics[width=0.95\textwidth]{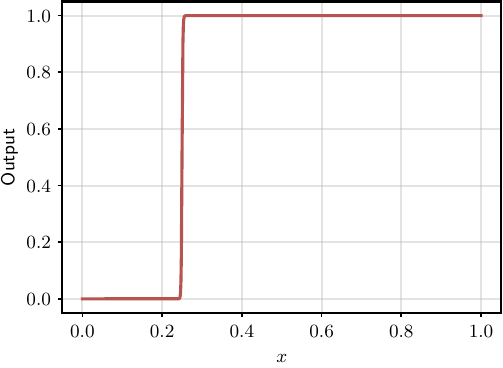}
    \caption{1 neuron}
    \label{fig:step_function}
    \end{subfigure}
    \hfill
    \begin{subfigure}[b]{0.32\textwidth}
    \includegraphics[width=0.95\textwidth]{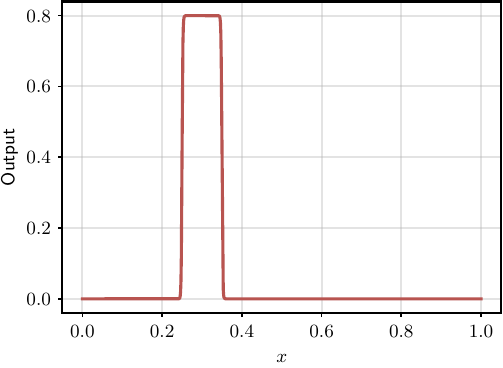}
    \caption{2 neurons}
    \label{fig:bin_function}
    \end{subfigure}
    \begin{subfigure}[b]{0.32\textwidth}
    \includegraphics[width=0.95\textwidth]{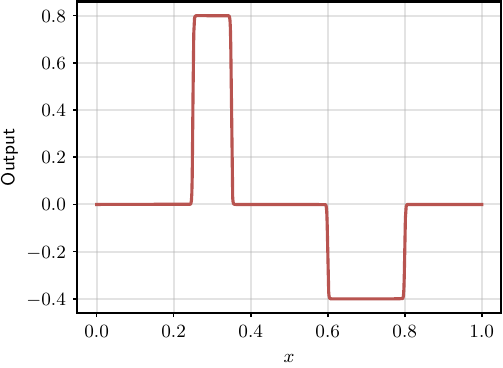}
    \caption{4 neurons}
    \label{fig:bin_function_2}
    \end{subfigure}
    \caption{(a) A neural network with one input, one hidden neuron, and one output can approximate any step function (here shown with $a=1$ and $s=0.3$). (b) With two hidden neurons and one output we can approximate any function which is constant over a small interval. (c) With four neurons, we can approximate any function which is piecewise constant over two non-zero intervals. Note that bins can be negative by defining a negative amplitude.}
\end{figure}

We show in Figure \ref{fig:tunable_sigmoid} several plots of \eqref{eq:proof_1}, where we fix $a$ and $s$ while varying $w$. As can be seen, by increasing $w$ the slope gets steeper. Fixing it to a very large constant (say, $w=10^4$), we are left with a very good approximation to a step function, of which we can control the location of the step (the $s$ parameter) and the amplitude (the $a$ parameter), as shown in Figure \ref{fig:step_function}.

\section{Approximating a constant function}

If we add a second neuron with opposite amplitude (and slightly shifted position), we can approximate a function which is constant over a small interval (we call it a “bin” function). Defining a width $\Delta$ we can write: 
\begin{equation}\footnotesize
f(x) = a\sigma\left(w\left(x-\eqnmarkbox[drawred]{node}{s -\frac{\Delta}{2}}\right)\right)  -a\sigma\left(w\left(x-\eqnmarkbox[drawgreen]{node2}{s+\frac{\Delta}{2}}\right)\right)
\label{eq:proof_2}
\end{equation}
\annotate[yshift=-1em]{below,left}{node}{Go up [down] at $s-\frac{\Delta}{2}$}
\annotate[yshift=-1em]{below,left}{node2}{Go down [up] at $s+\frac{\Delta}{2}$}

\vspace{1em}
where we recall that $w$ is now a large constant, e.g., $10^4$. \eqref{eq:proof_2} describes a function (equivalent to a model with one hidden layer having two neurons) which increases by $a$ at $s-\frac{\Delta}{2}$, is constant with value $f(x)=a$ over the interval $\left[s-\frac{\Delta}{2}, s+\frac{\Delta}{2}\right]$, and then decreases to $0$ afterwards. An example is  shown in Figure \ref{fig:bin_function}.

For the following, we can rewrite the previous function as $f(x; a, s, \Delta)$ to highlight the dependence on the three parameters $a$, $s$, and $\Delta$. \clearpage

\section{Approximating a generic function}
Because $f_{a, s, \Delta}(x)$ is effectively $0$ outside the corresponding interval, two functions defined over non-intersecting intervals will not influence each other, i.e., the “bin” function we just defined is highly localized. Hence, by adding two additional neurons in the hidden layer we can define a function which is constant over two separate intervals (an example of which is shown in Figure \ref{fig:bin_function_2}):

$$
f(x)=f(x;a_1,s_2,\Delta_1)+f(x;a_2,s_2,\Delta_2)
$$

The rest of the proof is now trivial and proceeds by binning the function we want to approximate in many small intervals. Given any (continuous) function $g(x)$ over an interval (which we assume $[0,1]$ for simplicity), we first bin the input domain into $m$ equispaced intervals, where $m$ controls the accuracy of the approximation (the higher $m$, the better the approximation). Hence, the $i$-th bin spans the interval:

$$
B_i = \left[\frac{i}{m}-\frac{\Delta}{2}, \frac{i}{m}+\frac{\Delta}{2}\right]
$$

where $\Delta$ is the size of each bin. For each bin, we compute the average value of $g(x)$ inside the interval itself:

$$
g_i = \frac{1}{\Delta}\int_{x \in B_i} g(x)dx
$$

Finally, we define a network with $2m$ neurons in the hidden layer, two for each bin. Each bin function is centered in the bin and takes value $g_i$:

\begin{equation}
f(x)=\sum_{i=1}^m f\left(x; \eqnmarkbox[drawgreen]{node2}{g_i}, \eqnmarkbox[drawred]{node}{\frac{i}{m}},\Delta\right)
\label{eq:proof_3}
\end{equation}
\annotate[yshift=-2em]{below,left}{node}{The $i$-th bin is centered in $\frac{i}{m}$}
\annotate[yshift=-1em]{below,left}{node2}{(Approximated) constant value}

\begin{figure}[t]
    \centering
    \begin{subfigure}[b]{0.32\textwidth}
    \includegraphics[width=0.95\textwidth]{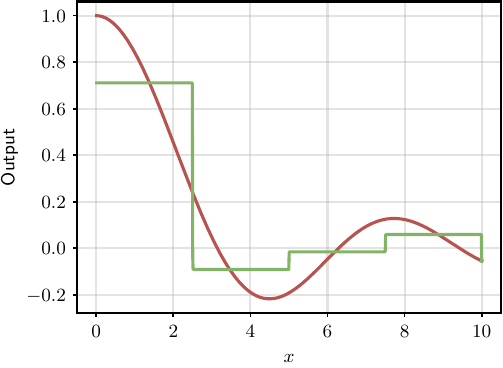}
    \caption{5 bins}
    \end{subfigure}
    \hfill
    \begin{subfigure}[b]{0.32\textwidth}
    \includegraphics[width=0.95\textwidth]{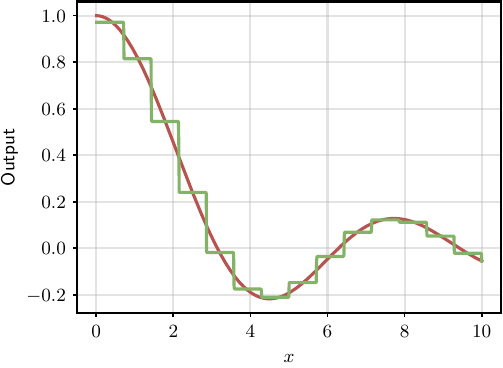}
    \caption{15 bins}
    \end{subfigure}
    \hfill
    \begin{subfigure}[b]{0.32\textwidth}
    \includegraphics[width=0.95\textwidth]{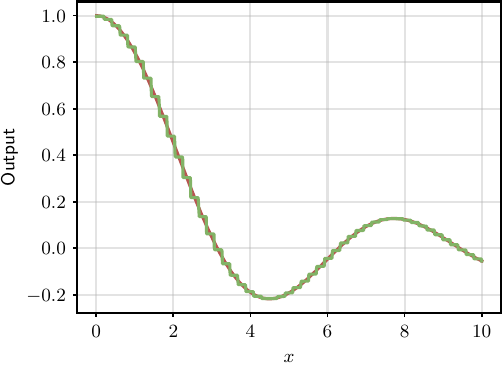}
    \caption{50 bins}
    \end{subfigure}
    \hfill
    \caption{Approximating $g(x) = \frac{\sin(x)}{x}$ in $[0,10]$ with (a) $m=5$, (b) $m=15$, and (c) $m=50$ bins. The original function is in red, the approximation \eqref{eq:proof_3} in green. The average squared error in the three cases decreases exponentially (approximately $0.02$, $0.002$, and $0.00016$).}
    \label{fig:sin_approximation}
\end{figure}

\vspace{2.5em}
We show in Figure \ref{fig:sin_approximation} an example of such approximation in the case of $g(x)=\frac{\sin(x)}{x}$ for increasing number of bins ($m=5$, $m=15$, $m=50$). It should be clear that the MSE is inversely proportional to $m$, and we can decrease the error as much as desired by simply increasing the resolution of the approximation.

Similar reasonings can be applied to multi-dimensional inputs and different activation functions.\footnote{\url{http://neuralnetworksanddeeplearning.com/chap4.html}}

%% file: main_small.bbl
\newcommand{\etalchar}[1]{$^{#1}$}
\begin{thebibliography}{GWFM{\etalchar{+}}13}

\bibitem[AB21]{angelopoulos2021gentle}
A.~N. Angelopoulos and S.~Bates.
\newblock A gentle introduction to conformal prediction and distribution-free uncertainty quantification.
\newblock {\em arXiv preprint arXiv:2107.07511}, 2021.

\bibitem[ADIP21]{apicella2021survey}
A.~Apicella, F.~Donnarumma, F.~Isgr{\`o}, and R.~Prevete.
\newblock A survey on modern trainable activation functions.
\newblock {\em Neural Networks}, 138:14--32, 2021.

\bibitem[AHS23]{ainsworth2022git}
S.~K. Ainsworth, J.~Hayase, and S.~Srinivasa.
\newblock Git re-basin: Merging models modulo permutation symmetries.
\newblock In {\em ICLR}, 2023.

\bibitem[AHSB14]{agostinelli2014learning}
F.~Agostinelli, M.~Hoffman, P.~Sadowski, and P.~Baldi.
\newblock Learning activation functions to improve deep neural networks.
\newblock {\em arXiv preprint arXiv:1412.6830}, 2014.

\bibitem[AJB{\etalchar{+}}17]{arpit2017closer}
D.~Arpit, S.~Jastrz{\k{e}}bski, N.~Ballas, D.~Krueger, E.~Bengio, M.~S. Kanwal, T.~Maharaj, A.~Fischer, A.~Courville, Y.~Bengio, et~al.
\newblock A closer look at memorization in deep networks.
\newblock In {\em ICML}, 2017.

\bibitem[AR00]{anderson2000talking}
J.~A. Anderson and E.~Rosenfeld.
\newblock {\em Talking nets: An oral history of neural networks}.
\newblock MIT Press, 2000.

\bibitem[ARAA{\etalchar{+}}16]{al2016theano}
R.~Al-Rfou, G.~Alain, A.~Almahairi, C.~Angermueller, D.~Bahdanau, N.~Ballas, F.~Bastien, J.~Bayer, A.~Belikov, A.~Belopolsky, et~al.
\newblock Theano: A {Python} framework for fast computation of mathematical expressions.
\newblock {\em arXiv preprint arXiv:1605.02688}, pages 1--19, 2016.

\bibitem[ASA{\etalchar{+}}23]{akyurek2022learning}
E.~Aky{\"u}rek, D.~Schuurmans, J.~Andreas, T.~Ma, and D.~Zhou.
\newblock What learning algorithm is in-context learning? investigations with linear models.
\newblock In {\em ICLR}, 2023.

\bibitem[AST{\etalchar{+}}24]{ansari2024chronos}
A.~F. Ansari, L.~Stella, C.~Turkmen, X.~Zhang, P.~Mercado, H.~Shen, O.~Shchur, S.~S. Rangapuram, S.~P. Arango, S.~Kapoor, et~al.
\newblock Chronos: Learning the language of time series.
\newblock {\em arXiv preprint arXiv:2403.07815}, 2024.

\bibitem[AZLL19]{allen2019learning}
Z.~Allen-Zhu, Y.~Li, and Y.~Liang.
\newblock Learning and generalization in overparameterized neural networks, going beyond two layers.
\newblock In {\em NeurIPS}, 2019.

\bibitem[BAY22]{brody2021attentive}
S.~Brody, U.~Alon, and E.~Yahav.
\newblock How attentive are graph attention networks?
\newblock In {\em ICLR}, 2022.

\bibitem[BB23]{bishop2024deep}
C.~M. Bishop and H.~Bishop.
\newblock {\em Deep learning: Foundations and concepts}.
\newblock Springer Nature, 2023.

\bibitem[BBCJ20]{biesialska2020continual}
M.~Biesialska, K.~Biesialska, and M.~R. Costa-Jussa.
\newblock Continual lifelong learning in natural language processing: A survey.
\newblock In {\em COLING}, 2020.

\bibitem[BBL{\etalchar{+}}17]{bronstein2017geometric}
M.~M. Bronstein, J.~Bruna, Y.~LeCun, A.~Szlam, and P.~Vandergheynst.
\newblock Geometric deep learning: going beyond {Euclidean} data.
\newblock {\em IEEE Signal Processing Magazine}, 34(4):18--42, 2017.

\bibitem[BCB15]{bahdanau2014neural}
D.~Bahdanau, K.~Cho, and Y.~Bengio.
\newblock Neural machine translation by jointly learning to align and translate.
\newblock In {\em ICLR}, 2015.

\bibitem[BCZ{\etalchar{+}}16]{bolukbasi2016man}
T.~Bolukbasi, K.-W. Chang, J.~Y. Zou, V.~Saligrama, and A.~T. Kalai.
\newblock Man is to computer programmer as woman is to homemaker? debiasing word embeddings.
\newblock In {\em NeurIPS}, 2016.

\bibitem[Ben09]{bengio2009learning}
Y.~Bengio.
\newblock Learning deep architectures for {AI}.
\newblock {\em Foundations and Trends{\textregistered} in Machine Learning}, 2(1):1--127, 2009.

\bibitem[BGHN24]{blasiok2024does}
J.~Blasiok, P.~Gopalan, L.~Hu, and P.~Nakkiran.
\newblock When does optimizing a proper loss yield calibration?
\newblock In {\em NeurIPS}, 2024.

\bibitem[BGLA21]{bianchi2021graph}
F.~M. Bianchi, D.~Grattarola, L.~Livi, and C.~Alippi.
\newblock Graph neural networks with convolutional {ARMA} filters.
\newblock {\em IEEE Transactions on Pattern Analysis and Machine Intelligence}, 44(7):3496--3507, 2021.

\bibitem[BGMMS21]{bender2021dangers}
E.~M. Bender, T.~Gebru, A.~McMillan-Major, and S.~Shmitchell.
\newblock On the dangers of stochastic parrots: Can language models be too big?
\newblock In {\em ACM FAccT}. ACM, 2021.

\bibitem[BGSW18]{bjorck2018understanding}
N.~Bjorck, C.~P. Gomes, B.~Selman, and K.~Q. Weinberger.
\newblock Understanding batch normalization.
\newblock In {\em NeurIPS}, 2018.

\bibitem[BHB{\etalchar{+}}18]{battaglia2018relational}
P.~W. Battaglia, J.~B. Hamrick, V.~Bapst, A.~Sanchez-Gonzalez, V.~Zambaldi, M.~Malinowski, A.~Tacchetti, D.~Raposo, A.~Santoro, R.~Faulkner, et~al.
\newblock Relational inductive biases, deep learning, and graph networks.
\newblock {\em arXiv preprint arXiv:1806.01261}, 2018.

\bibitem[Bis95]{bishop1995training}
C.~M. Bishop.
\newblock Training with noise is equivalent to {Tikhonov} regularization.
\newblock {\em Neural Computation}, 7(1):108--116, 1995.

\bibitem[Bis06]{bishop2006pattern}
C.~Bishop.
\newblock {\em Pattern recognition and machine learning}.
\newblock Springer, 2006.

\bibitem[BJMO12]{bach2012optimization}
F.~Bach, R.~Jenatton, J.~Mairal, and G.~Obozinski.
\newblock Optimization with sparsity-inducing penalties.
\newblock {\em Foundations and Trends{\textregistered} in Machine Learning}, 4(1):1--106, 2012.

\bibitem[BKH16]{ba2016layer}
J.~L. Ba, J.~R. Kiros, and G.~E. Hinton.
\newblock Layer normalization.
\newblock {\em arXiv preprint arXiv:1607.06450}, 2016.

\bibitem[BKK19]{bai2019deep}
S.~Bai, J.~Z. Kolter, and V.~Koltun.
\newblock Deep equilibrium models.
\newblock In {\em NeurIPS}, 2019.

\bibitem[Ble90]{blelloch1990prefix}
G.~E. Blelloch.
\newblock {\em Prefix sums and their applications}.
\newblock School of Computer Science, Carnegie Mellon University Pittsburgh, PA, USA, 1990.

\bibitem[BN24]{bernstein2024old}
J.~Bernstein and L.~Newhouse.
\newblock Old optimizer, new norm: An anthology.
\newblock {\em arXiv preprint arXiv:2409.20325}, 2024.

\bibitem[BNS06]{belkin2006manifold}
M.~Belkin, P.~Niyogi, and V.~Sindhwani.
\newblock Manifold regularization: A geometric framework for learning from labeled and unlabeled examples.
\newblock {\em Journal of Machine Learning Research}, 7(11), 2006.

\bibitem[BP20]{bolte2020mathematical}
J.~Bolte and E.~Pauwels.
\newblock A mathematical model for automatic differentiation in machine learning.
\newblock In {\em NeurIPS}, 2020.

\bibitem[BPA{\etalchar{+}}24]{bordes2024introduction}
F.~Bordes, R.~Y. Pang, A.~Ajay, A.~C. Li, A.~Bardes, S.~Petryk, O.~Ma{\~n}as, Z.~Lin, A.~Mahmoud, B.~Jayaraman, et~al.
\newblock An introduction to vision-language modeling.
\newblock {\em arXiv preprint arXiv:2405.17247}, 2024.

\bibitem[BPRS18]{baydin2018automatic}
A.~G. Baydin, B.~A. Pearlmutter, A.~A. Radul, and J.~M. Siskind.
\newblock Automatic differentiation in machine learning: a survey.
\newblock {\em Journal of Marchine Learning Research}, 18:1--43, 2018.

\bibitem[BPS{\etalchar{+}}24]{beck2024xlstm}
M.~Beck, K.~P{\"o}ppel, M.~Spanring, A.~Auer, O.~Prudnikova, M.~Kopp, G.~Klambauer, J.~Brandstetter, and S.~Hochreiter.
\newblock {xLSTM}: Extended long short-term memory.
\newblock {\em arXiv preprint arXiv:2405.04517}, 2024.

\bibitem[BR07]{bogachev2007measure}
V.~I. Bogachev and M.~A.~S. Ruas.
\newblock {\em Measure theory}.
\newblock Springer, 2007.

\bibitem[BR24]{blondel2024elements}
M.~Blondel and V.~Roulet.
\newblock The elements of differentiable programming.
\newblock {\em arXiv preprint arXiv:2403.14606}, 2024.

\bibitem[BZMA20]{baevski2020wav2vec}
A.~Baevski, Y.~Zhou, A.~Mohamed, and M.~Auli.
\newblock wav2vec 2.0: A framework for self-supervised learning of speech representations.
\newblock In {\em NeurIPS}, 2020.

\bibitem[CCGC24]{cheng2024multilinear}
Y.~Cheng, G.~G. Chrysos, M.~Georgopoulos, and V.~Cevher.
\newblock Multilinear operator networks.
\newblock In {\em ICLR}, 2024.

\bibitem[CMMB22]{cinus2022effect}
F.~Cinus, M.~Minici, C.~Monti, and F.~Bonchi.
\newblock The effect of people recommenders on echo chambers and polarization.
\newblock In {\em AAAI ICWSM}, 2022.

\bibitem[CPPM22]{chien2021you}
E.~Chien, C.~Pan, J.~Peng, and O.~Milenkovic.
\newblock You are {AllSet}: A multiset function framework for hypergraph neural networks.
\newblock In {\em ICLR}, 2022.

\bibitem[CRBD18]{chen2018neural}
R.~T. Chen, Y.~Rubanova, J.~Bettencourt, and D.~K. Duvenaud.
\newblock Neural ordinary differential equations.
\newblock In {\em NeurIPS}, 2018.

\bibitem[CS21]{chung2021turing}
S.~Chung and H.~Siegelmann.
\newblock Turing completeness of bounded-precision recurrent neural networks.
\newblock In {\em NeurIPS}, 2021.

\bibitem[CVMG{\etalchar{+}}14]{cho2014learning}
K.~Cho, B.~Van~Merri{\"e}nboer, C.~Gulcehre, D.~Bahdanau, F.~Bougares, H.~Schwenk, and Y.~Bengio.
\newblock Learning phrase representations using rnn encoder-decoder for statistical machine translation.
\newblock In {\em EMNLP}. ACL, 2014.

\bibitem[CW82]{coppersmith1982asymptotic}
D.~Coppersmith and S.~Winograd.
\newblock On the asymptotic complexity of matrix multiplication.
\newblock {\em SIAM Journal on Computing}, 11(3):472--492, 1982.

\bibitem[Cyb89]{cybenko1989approximation}
G.~Cybenko.
\newblock Approximation by superpositions of a sigmoidal function.
\newblock {\em Mathematics of Control, Signals and Systems}, 2(4):303--314, 1989.

\bibitem[CZJ{\etalchar{+}}22]{chang2022maskgit}
H.~Chang, H.~Zhang, L.~Jiang, C.~Liu, and W.~T. Freeman.
\newblock {MaskGIT}: Masked generative image transformer.
\newblock In {\em IEEE/CVF CVPR}, 2022.

\bibitem[CZSL20]{cubuk2020randaugment}
E.~D. Cubuk, B.~Zoph, J.~Shlens, and Q.~V. Le.
\newblock {RandAugment}: Practical automated data augmentation with a reduced search space.
\newblock In {\em IEEE/CVF CVPR Workshops}, 2020.

\bibitem[DBK{\etalchar{+}}21]{dosovitskiy2020image}
A.~Dosovitskiy, L.~Beyer, A.~Kolesnikov, D.~Weissenborn, X.~Zhai, T.~Unterthiner, M.~Dehghani, M.~Minderer, G.~Heigold, S.~Gelly, et~al.
\newblock An image is worth 16x16 words: Transformers for image recognition at scale.
\newblock In {\em ICLR}, 2021.

\bibitem[DCLT18]{devlin2018bert}
J.~Devlin, M.-W. Chang, K.~Lee, and K.~Toutanova.
\newblock {BERT}: Pre-training of deep bidirectional transformers for language understanding.
\newblock In {\em NAACL}. ACL, 2018.

\bibitem[DCSA23]{defossez2022high}
A.~D{\'e}fossez, J.~Copet, G.~Synnaeve, and Y.~Adi.
\newblock High fidelity neural audio compression.
\newblock {\em Transactions on Machine Learning Research}, 2023.

\bibitem[DDM{\etalchar{+}}23]{dehghani2023scaling}
M.~Dehghani, J.~Djolonga, B.~Mustafa, P.~Padlewski, J.~Heek, J.~Gilmer, A.~P. Steiner, M.~Caron, R.~Geirhos, I.~Alabdulmohsin, et~al.
\newblock Scaling vision transformers to 22 billion parameters.
\newblock In {\em ICML}, 2023.

\bibitem[DFAG17]{dauphin2017language}
Y.~N. Dauphin, A.~Fan, M.~Auli, and D.~Grangier.
\newblock Language modeling with gated convolutional networks.
\newblock In {\em ICML}, 2017.

\bibitem[DFE{\etalchar{+}}22]{dao2022flashattention}
T.~Dao, D.~Fu, S.~Ermon, A.~Rudra, and C.~R{\'e}.
\newblock Flash{A}ttention: Fast and memory-efficient exact attention with {IO}-awareness.
\newblock In {\em NeurIPS}, 2022.

\bibitem[DLL{\etalchar{+}}22]{dwivedi2021graph}
V.~P. Dwivedi, A.~T. Luu, T.~Laurent, Y.~Bengio, and X.~Bresson.
\newblock Graph neural networks with learnable structural and positional representations.
\newblock In {\em ICLR}, 2022.

\bibitem[DOMB24]{darcet2023vision}
T.~Darcet, M.~Oquab, J.~Mairal, and P.~Bojanowski.
\newblock Vision transformers need registers.
\newblock In {\em ICLR}, 2024.

\bibitem[DS20]{de2020batch}
S.~De and S.~Smith.
\newblock Batch normalization biases residual blocks towards the identity function in deep networks.
\newblock In {\em NeurIPS}, 2020.

\bibitem[DT17]{devries2017improved}
T.~DeVries and G.~W. Taylor.
\newblock Improved regularization of convolutional neural networks with cutout.
\newblock {\em arXiv preprint arXiv:1708.04552}, 2017.

\bibitem[DZPS19]{du2018gradient}
S.~S. Du, X.~Zhai, B.~Poczos, and A.~Singh.
\newblock Gradient descent provably optimizes over-parameterized neural networks.
\newblock In {\em ICLR}, 2019.

\bibitem[EHB23]{eijkelboom2023n}
F.~Eijkelboom, R.~Hesselink, and E.~J. Bekkers.
\newblock E $(n) $ equivariant message passing simplicial networks.
\newblock In {\em ICML}, 2023.

\bibitem[FAL17]{finn2017model}
C.~Finn, P.~Abbeel, and S.~Levine.
\newblock Model-agnostic meta-learning for fast adaptation of deep networks.
\newblock In {\em ICML}, 2017.

\bibitem[Fle23]{fleuret2023little}
F.~Fleuret.
\newblock {\em The Little Book of Deep Learning}.
\newblock Lulu Press, Inc., 2023.

\bibitem[GBGB21]{gauthier2021next}
D.~J. Gauthier, E.~Bollt, A.~Griffith, and W.~A. Barbosa.
\newblock Next generation reservoir computing.
\newblock {\em Nature Communications}, 12(1):5564, 2021.

\bibitem[GD23]{gu2023mamba}
A.~Gu and T.~Dao.
\newblock Mamba: Linear-time sequence modeling with selective state spaces.
\newblock {\em arXiv preprint arXiv:2312.00752}, 2023.

\bibitem[GDE{\etalchar{+}}20]{gu2020hippo}
A.~Gu, T.~Dao, S.~Ermon, A.~Rudra, and C.~R{\'e}.
\newblock Hippo: Recurrent memory with optimal polynomial projections.
\newblock In {\em NeurIPS}, 2020.

\bibitem[GFGS06]{graves2012connectionist}
A.~Graves, S.~Fern{\'a}ndez, F.~Gomez, and J.~Schmidhuber.
\newblock Connectionist temporal classification: labelling unsegmented sequence data with recurrent neural networks.
\newblock In {\em ICML}, 2006.

\bibitem[GG16]{gal2016dropout}
Y.~Gal and Z.~Ghahramani.
\newblock Dropout as a bayesian approximation: Representing model uncertainty in deep learning.
\newblock In {\em ICML}, 2016.

\bibitem[GGR22]{gu2021efficiently}
A.~Gu, K.~Goel, and C.~R{\'e}.
\newblock Efficiently modeling long sequences with structured state spaces.
\newblock In {\em ICLR}, 2022.

\bibitem[GJG{\etalchar{+}}21]{gu2021combining}
A.~Gu, I.~Johnson, K.~Goel, K.~Saab, T.~Dao, A.~Rudra, and C.~R{\'e}.
\newblock Combining recurrent, convolutional, and continuous-time models with linear state space layers.
\newblock In {\em NeurIPS}, 2021.

\bibitem[GM17]{gallicchio2017echo}
C.~Gallicchio and A.~Micheli.
\newblock Echo state property of deep reservoir computing networks.
\newblock {\em Cognitive Computation}, 9:337--350, 2017.

\bibitem[GMS05]{gori2005new}
M.~Gori, G.~Monfardini, and F.~Scarselli.
\newblock A new model for learning in graph domains.
\newblock In {\em IEEE IJCNN}. IEEE, 2005.

\bibitem[GOV22]{grinsztajn2022tree}
L.~Grinsztajn, E.~Oyallon, and G.~Varoquaux.
\newblock Why do tree-based models still outperform deep learning on typical tabular data?
\newblock In {\em NeurIPS}, 2022.

\bibitem[GPE{\etalchar{+}}23]{golkar2023xval}
S.~Golkar, M.~Pettee, M.~Eickenberg, A.~Bietti, M.~Cranmer, G.~Krawezik, F.~Lanusse, M.~McCabe, R.~Ohana, L.~Parker, et~al.
\newblock {xVal}: A continuous number encoding for large language models.
\newblock {\em arXiv preprint arXiv:2310.02989}, 2023.

\bibitem[GPSW17]{guo2017calibration}
C.~Guo, G.~Pleiss, Y.~Sun, and K.~Q. Weinberger.
\newblock On calibration of modern neural networks.
\newblock In {\em ICML}, 2017.

\bibitem[Gri12]{griewank2012invented}
A.~Griewank.
\newblock Who invented the reverse mode of differentiation?
\newblock {\em Documenta Mathematica, Extra Volume ISMP}, 389400, 2012.

\bibitem[GSBL20]{geva2020transformer}
M.~Geva, R.~Schuster, J.~Berant, and O.~Levy.
\newblock Transformer feed-forward layers are key-value memories.
\newblock In {\em EMNLP}. ACL, 2020.

\bibitem[GSR{\etalchar{+}}17]{gilmer2017neural}
J.~Gilmer, S.~S. Schoenholz, P.~F. Riley, O.~Vinyals, and G.~E. Dahl.
\newblock Neural message passing for quantum chemistry.
\newblock In {\em ICML}, 2017.

\bibitem[GW08]{griewank2008evaluating}
A.~Griewank and A.~Walther.
\newblock {\em Evaluating derivatives: principles and techniques of algorithmic differentiation}.
\newblock SIAM, 2008.

\bibitem[GWFM{\etalchar{+}}13]{goodfellow2013maxout}
I.~Goodfellow, D.~Warde-Farley, M.~Mirza, A.~Courville, and Y.~Bengio.
\newblock Maxout networks.
\newblock In {\em ICML}, 2013.

\bibitem[GZBA22]{grattarola2022understanding}
D.~Grattarola, D.~Zambon, F.~M. Bianchi, and C.~Alippi.
\newblock Understanding pooling in graph neural networks.
\newblock {\em IEEE Transactions on Neural Networks and Learning Systems}, 2022.

\bibitem[HABN{\etalchar{+}}21]{hoefler2021sparsity}
T.~Hoefler, D.~Alistarh, T.~Ben-Nun, N.~Dryden, and A.~Peste.
\newblock Sparsity in deep learning: Pruning and growth for efficient inference and training in neural networks.
\newblock {\em Journal of Machine Learning Research}, 22(241):1--124, 2021.

\bibitem[HBE{\etalchar{+}}24]{ho2024algorithmic}
A.~Ho, T.~Besiroglu, E.~Erdil, D.~Owen, R.~Rahman, Z.~C. Guo, D.~Atkinson, N.~Thompson, and J.~Sevilla.
\newblock Algorithmic progress in language models.
\newblock {\em arXiv preprint arXiv:1710.05941}, 2024.

\bibitem[HDLL22]{hua2022transformer}
W.~Hua, Z.~Dai, H.~Liu, and Q.~Le.
\newblock Transformer quality in linear time.
\newblock In {\em ICML}, 2022.

\bibitem[HG16]{hendrycks2016gaussian}
D.~Hendrycks and K.~Gimpel.
\newblock Gaussian error linear units ({GELUs}).
\newblock {\em arXiv preprint arXiv:1606.08415}, 2016.

\bibitem[HHWW14]{huang2014deep}
P.~Huang, Y.~Huang, W.~Wang, and L.~Wang.
\newblock Deep embedding network for clustering.
\newblock In {\em ICPR}. IEEE, 2014.

\bibitem[Hoc98]{hochreiter1998recurrent}
S.~Hochreiter.
\newblock Recurrent neural net learning and vanishing gradient.
\newblock {\em International Journal Of Uncertainity, Fuzziness and Knowledge-Based Systems}, 6(2):107--116, 1998.

\bibitem[Hor91]{hornik1991approximation}
K.~Hornik.
\newblock Approximation capabilities of multilayer feedforward networks.
\newblock {\em Neural Networks}, 4(2):251--257, 1991.

\bibitem[HR22]{hardt2022patterns}
M.~Hardt and B.~Recht.
\newblock {\em Patterns, predictions, and actions: Foundations of machine learning}.
\newblock Princeton University Press, 2022.

\bibitem[HS97]{hochreiter1997long}
S.~Hochreiter and J.~Schmidhuber.
\newblock Long short-term memory.
\newblock {\em Neural Computation}, 9(8):1735--1780, 1997.

\bibitem[HSS08]{hofmann2008kernel}
T.~Hofmann, B.~Sch{\"o}lkopf, and A.~J. Smola.
\newblock Kernel methods in machine learning.
\newblock {\em The Annals of Statistics}, 36(3):1171--1220, 2008.

\bibitem[HTF09]{hastie2009elements}
T.~Hastie, R.~Tibshirani, and J.~H. Friedman.
\newblock {\em The elements of statistical learning: data mining, inference, and prediction}.
\newblock Springer, 2009.

\bibitem[HWG25]{hwang2025dynamic}
S.~Hwang, B.~Wang, and A.~Gu.
\newblock Dynamic chunking for end-to-end hierarchical sequence modeling.
\newblock {\em arXiv preprint arXiv:2507.07955}, 2025.

\bibitem[HYL17]{hamilton2017inductive}
W.~Hamilton, Z.~Ying, and J.~Leskovec.
\newblock Inductive representation learning on large graphs.
\newblock In {\em NeurIPS}, 2017.

\bibitem[HZC{\etalchar{+}}17]{howard2017mobilenets}
A.~G. Howard, M.~Zhu, B.~Chen, D.~Kalenichenko, W.~Wang, T.~Weyand, M.~Andreetto, and H.~Adam.
\newblock {MobileNets}: Efficient convolutional neural networks for mobile vision applications.
\newblock {\em arXiv preprint arXiv:1704.04861}, 2017.

\bibitem[HZRS15]{he2015delving}
K.~He, X.~Zhang, S.~Ren, and J.~Sun.
\newblock Delving deep into rectifiers: Surpassing human-level performance on {ImageNet} classification.
\newblock In {\em IEEE ICCV}, 2015.

\bibitem[HZRS16]{he2016deep}
K.~He, X.~Zhang, S.~Ren, and J.~Sun.
\newblock Deep residual learning for image recognition.
\newblock In {\em IEEE/CVF CVPR}, 2016.

\bibitem[ICS22]{irie2022dual}
K.~Irie, R.~Csord{\'a}s, and J.~Schmidhuber.
\newblock The dual form of neural networks revisited: Connecting test time predictions to training patterns via spotlights of attention.
\newblock In {\em ICML}, 2022.

\bibitem[IS15]{ioffe2015batch}
S.~Ioffe and C.~Szegedy.
\newblock Batch normalization: Accelerating deep network training by reducing internal covariate shift.
\newblock In {\em ICML}, 2015.

\bibitem[JGB{\etalchar{+}}21]{jaegle2021perceiver}
A.~Jaegle, F.~Gimeno, A.~Brock, O.~Vinyals, A.~Zisserman, and J.~Carreira.
\newblock Perceiver: General perception with iterative attention.
\newblock In {\em ICML}, 2021.

\bibitem[JK{\etalchar{+}}17]{jain2017non}
P.~Jain, P.~Kar, et~al.
\newblock Non-convex optimization for machine learning.
\newblock {\em Foundations and Trends{\textregistered} in Machine Learning}, 10(3-4):142--363, 2017.

\bibitem[JLB{\etalchar{+}}22]{jospin2022hands}
L.~V. Jospin, H.~Laga, F.~Boussaid, W.~Buntine, and M.~Bennamoun.
\newblock Hands-on {Bayesian} neural networks—a tutorial for deep learning users.
\newblock {\em IEEE Computational Intelligence Magazine}, 17(2):29--48, 2022.

\bibitem[KB15]{kingma2014adam}
D.~P. Kingma and J.~Ba.
\newblock Adam: A method for stochastic optimization.
\newblock In {\em ICLR}, 2015.

\bibitem[KG21]{kidger2021equinox}
P.~Kidger and C.~Garcia.
\newblock {E}quinox: neural networks in {JAX} via callable {P}y{T}rees and filtered transformations.
\newblock {\em Differentiable Programming Workshop, NeurIPS}, 2021.

\bibitem[KL18]{kakade2018provably}
S.~M. Kakade and J.~D. Lee.
\newblock Provably correct automatic sub-differentiation for qualified programs.
\newblock In {\em NeurIPS}, 2018.

\bibitem[KMH{\etalchar{+}}20]{kaplan2020scaling}
J.~Kaplan, S.~McCandlish, T.~Henighan, T.~B. Brown, B.~Chess, R.~Child, S.~Gray, A.~Radford, J.~Wu, and D.~Amodei.
\newblock Scaling laws for neural language models.
\newblock {\em arXiv preprint arXiv:2001.08361}, 2020.

\bibitem[KPR{\etalchar{+}}17]{kirkpatrick2017overcoming}
J.~Kirkpatrick, R.~Pascanu, N.~Rabinowitz, J.~Veness, G.~Desjardins, A.~A. Rusu, K.~Milan, J.~Quan, T.~Ramalho, A.~Grabska-Barwinska, et~al.
\newblock Overcoming catastrophic forgetting in neural networks.
\newblock {\em Proceedings of the National Academy of Sciences}, 114(13):3521--3526, 2017.

\bibitem[KSH12]{krizhevsky2012imagenet}
A.~Krizhevsky, I.~Sutskever, and G.~E. Hinton.
\newblock {ImageNet} classification with deep convolutional neural networks.
\newblock In {\em NeurIPS}, 2012.

\bibitem[KVPF20]{katharopoulos2020transformers}
A.~Katharopoulos, A.~Vyas, N.~Pappas, and F.~Fleuret.
\newblock Transformers are {RNNs}: Fast autoregressive transformers with linear attention.
\newblock In {\em ICML}, 2020.

\bibitem[KW17]{kipf2016semi}
T.~N. Kipf and M.~Welling.
\newblock Semi-supervised classification with graph convolutional networks.
\newblock In {\em ICLR}, 2017.

\bibitem[Lau19]{laue2019equivalence}
S.~Laue.
\newblock On the equivalence of automatic and symbolic differentiation.
\newblock {\em arXiv preprint arXiv:1904.02990}, 2019.

\bibitem[LBBH98]{lecun1998gradient}
Y.~LeCun, L.~Bottou, Y.~Bengio, and P.~Haffner.
\newblock Gradient-based learning applied to document recognition.
\newblock {\em Proceedings of the IEEE}, 86(11):2278--2324, 1998.

\bibitem[LBH15]{lecun2015deep}
Y.~LeCun, Y.~Bengio, and G.~Hinton.
\newblock Deep learning.
\newblock {\em Nature}, 521(7553):436--444, 2015.

\bibitem[LCX{\etalchar{+}}23]{li2023generalized}
J.~Li, Y.~Cheng, Z.~Xia, Y.~Mo, and G.~Huang.
\newblock Generalized activation via multivariate projection.
\newblock {\em arXiv preprint arXiv:2309.17194}, 2023.

\bibitem[LDR23]{lialin2023scaling}
V.~Lialin, V.~Deshpande, and A.~Rumshisky.
\newblock Scaling down to scale up: A guide to parameter-efficient fine-tuning.
\newblock {\em arXiv preprint arXiv:2303.15647}, 2023.

\bibitem[LDSL21]{liu2021pay}
H.~Liu, Z.~Dai, D.~So, and Q.~V. Le.
\newblock Pay attention to {MLP}s.
\newblock In {\em NeurIPS}, 2021.

\bibitem[LH19]{loshchilov2018fixing}
I.~Loshchilov and F.~Hutter.
\newblock Decoupled weight decay regularization.
\newblock In {\em ICLR}, 2019.

\bibitem[Lim21]{lim2021tensors}
L.-H. Lim.
\newblock Tensors in computations.
\newblock {\em Acta Numerica}, 30:555--764, 2021.

\bibitem[LJ09]{lukovsevivcius2009reservoir}
M.~Luko{\v{s}}evi{\v{c}}ius and H.~Jaeger.
\newblock Reservoir computing approaches to recurrent neural network training.
\newblock {\em Computer Science Review}, 3(3):127--149, 2009.

\bibitem[LKM23]{leviathan2023fast}
Y.~Leviathan, M.~Kalman, and Y.~Matias.
\newblock Fast inference from transformers via speculative decoding.
\newblock In {\em ICML}, 2023.

\bibitem[LLLG22]{li2022graph}
Y.~Li, B.~Lin, B.~Luo, and N.~Gui.
\newblock Graph representation learning beyond node and homophily.
\newblock {\em IEEE Transactions on Knowledge and Data Engineering}, 35(5):4880--4893, 2022.

\bibitem[LLS21]{lee2021vision}
S.~H. Lee, S.~Lee, and B.~C. Song.
\newblock Vision transformer for small-size datasets.
\newblock {\em arXiv preprint arXiv:2112.13492}, 2021.

\bibitem[LMW{\etalchar{+}}22]{liu2022convnet}
Z.~Liu, H.~Mao, C.-Y. Wu, C.~Feichtenhofer, T.~Darrell, and S.~Xie.
\newblock A {ConvNet} for the 2020s.
\newblock In {\em IEEE/CVF CVPR}, 2022.

\bibitem[LPW{\etalchar{+}}17]{lu2017expressive}
Z.~Lu, H.~Pu, F.~Wang, Z.~Hu, and L.~Wang.
\newblock The expressive power of neural networks: A view from the width.
\newblock In {\em NeurIPS}, 2017.

\bibitem[LRZ{\etalchar{+}}23]{lim2022sign}
D.~Lim, J.~Robinson, L.~Zhao, T.~Smidt, S.~Sra, H.~Maron, and S.~Jegelka.
\newblock Sign and basis invariant networks for spectral graph representation learning.
\newblock In {\em ICLR}, 2023.

\bibitem[LTM{\etalchar{+}}22]{liu2022few}
H.~Liu, D.~Tam, M.~Muqeeth, J.~Mohta, T.~Huang, M.~Bansal, and C.~A. Raffel.
\newblock Few-shot parameter-efficient fine-tuning is better and cheaper than in-context learning.
\newblock In {\em NeurIPS}, 2022.

\bibitem[LWV{\etalchar{+}}24]{liu2024kan}
Z.~Liu, Y.~Wang, S.~Vaidya, F.~Ruehle, J.~Halverson, M.~Solja{\v{c}}i{\'c}, T.~Y. Hou, and M.~Tegmark.
\newblock {KAN}: {Kolmogorov-Arnold} networks.
\newblock {\em arXiv preprint arXiv:2404.19756}, 2024.

\bibitem[LZA23]{liu2023ring}
H.~Liu, M.~Zaharia, and P.~Abbeel.
\newblock Ring attention with blockwise transformers for near-infinite context.
\newblock In {\em Foundation Models for Decision Making Workshop, NeurIPS}, 2023.

\bibitem[MCT{\etalchar{+}}]{maoposition}
H.~Mao, Z.~Chen, W.~Tang, J.~Zhao, Y.~Ma, T.~Zhao, N.~Shah, M.~Galkin, and J.~Tang.
\newblock Position: Graph foundation models are already here.
\newblock In {\em ICML}.

\bibitem[Met22]{metz2022genius}
C.~Metz.
\newblock {\em Genius makers: the mavericks who brought AI to Google, Facebook, and the world}.
\newblock Penguin, 2022.

\bibitem[MGMR24]{muller2023attending}
L.~M{\"u}ller, M.~Galkin, C.~Morris, and L.~Ramp{\'a}{\v{s}}ek.
\newblock Attending to graph transformers.
\newblock {\em Transactions on Machine Learning Research}, 2024.

\bibitem[MKS{\etalchar{+}}20]{mukhoti2020calibrating}
J.~Mukhoti, V.~Kulharia, A.~Sanyal, S.~Golodetz, P.~Torr, and P.~Dokania.
\newblock Calibrating deep neural networks using focal loss.
\newblock In {\em NeurIPS}, 2020.

\bibitem[MRF{\etalchar{+}}19]{morris2019weisfeiler}
C.~Morris, M.~Ritzert, M.~Fey, W.~L. Hamilton, J.~E. Lenssen, G.~Rattan, and M.~Grohe.
\newblock Weisfeiler and {Leman} go neural: Higher-order graph neural networks.
\newblock In {\em AAAI Conference on Artificial Intelligence}, volume~33, pages 4602--4609, 2019.

\bibitem[MRT18]{mohri2018foundations}
M.~Mohri, A.~Rostamizadeh, and A.~Talwalkar.
\newblock {\em Foundations of machine learning}.
\newblock MIT Press, 2018.

\bibitem[MSC{\etalchar{+}}13]{mikolov2013distributed}
T.~Mikolov, I.~Sutskever, K.~Chen, G.~S. Corrado, and J.~Dean.
\newblock Distributed representations of words and phrases and their compositionality.
\newblock In {\em NeurIPS}, 2013.

\bibitem[MZBG18]{marra2018learning}
G.~Marra, D.~Zanca, A.~Betti, and M.~Gori.
\newblock Learning neuron non-linearities with kernel-based deep neural networks.
\newblock {\em arXiv preprint arXiv:1807.06302}, 2018.

\bibitem[NCN{\etalchar{+}}23]{niculae2023discrete}
V.~Niculae, C.~F. Corro, N.~Nangia, T.~Mihaylova, and A.~F. Martins.
\newblock Discrete latent structure in neural networks.
\newblock {\em arXiv preprint arXiv:2301.07473}, 2023.

\bibitem[ODG{\etalchar{+}}23]{orvieto2023universality}
A.~Orvieto, S.~De, C.~Gulcehre, R.~Pascanu, and S.~L. Smith.
\newblock On the universality of linear recurrences followed by nonlinear projections.
\newblock In {\em HLD 2023 Workshop, ICML}, 2023.

\bibitem[ODZ{\etalchar{+}}16]{oord2016wavenet}
A.~v.~d. Oord, S.~Dieleman, H.~Zen, K.~Simonyan, O.~Vinyals, A.~Graves, N.~Kalchbrenner, A.~Senior, and K.~Kavukcuoglu.
\newblock {WaveNet}: A generative model for raw audio.
\newblock In {\em ISCA SSW Workshop}, 2016.

\bibitem[OSG{\etalchar{+}}23]{orvieto2023resurrecting}
A.~Orvieto, S.~L. Smith, A.~Gu, A.~Fernando, C.~Gulcehre, R.~Pascanu, and S.~De.
\newblock Resurrecting recurrent neural networks for long sequences.
\newblock In {\em ICML}, 2023.

\bibitem[PAA{\etalchar{+}}23]{peng2023rwkv}
B.~Peng, E.~Alcaide, Q.~Anthony, A.~Albalak, S.~Arcadinho, H.~Cao, X.~Cheng, M.~Chung, M.~Grella, K.~K. GV, et~al.
\newblock {RWKV}: Reinventing {RNNs} for the transformer era.
\newblock In {\em EMNLP}. ACL, 2023.

\bibitem[PABH{\etalchar{+}}21]{puny2021frame}
O.~Puny, M.~Atzmon, H.~Ben-Hamu, I.~Misra, A.~Grover, E.~J. Smith, and Y.~Lipman.
\newblock Frame averaging for invariant and equivariant network design.
\newblock {\em arXiv preprint arXiv:2110.03336}, 2021.

\bibitem[PBE{\etalchar{+}}22]{power2022grokking}
A.~Power, Y.~Burda, H.~Edwards, I.~Babuschkin, and V.~Misra.
\newblock Grokking: Generalization beyond overfitting on small algorithmic datasets.
\newblock In {\em 1st Mathematical Reasoning in General Artificial Intelligence Workshop, ICLR}, 2022.

\bibitem[PBL20]{poggio2020theoretical}
T.~Poggio, A.~Banburski, and Q.~Liao.
\newblock Theoretical issues in deep networks.
\newblock {\em Proceedings of the National Academy of Sciences}, 117(48):30039--30045, 2020.

\bibitem[PGCB14]{pascanu2013construct}
R.~Pascanu, C.~Gulcehre, K.~Cho, and Y.~Bengio.
\newblock How to construct deep recurrent neural networks.
\newblock In {\em ICLR}, 2014.

\bibitem[PKP{\etalchar{+}}19]{parisi2019continual}
G.~I. Parisi, R.~Kemker, J.~L. Part, C.~Kanan, and S.~Wermter.
\newblock Continual lifelong learning with neural networks: A review.
\newblock {\em Neural Networks}, 113:54--71, 2019.

\bibitem[PNR{\etalchar{+}}21]{papamakarios2021normalizing}
G.~Papamakarios, E.~Nalisnick, D.~J. Rezende, S.~Mohamed, and B.~Lakshminarayanan.
\newblock Normalizing flows for probabilistic modeling and inference.
\newblock {\em Journal of Machine Learning Research}, 22(57):1--64, 2021.

\bibitem[PP08]{petersen2008matrix}
K.~B. Petersen and M.~S. Pedersen.
\newblock {\em The matrix cookbook}.
\newblock Technical University of Denmark, 2008.

\bibitem[PPR{\etalchar{+}}24]{pagnoni2024byte}
A.~Pagnoni, R.~Pasunuru, P.~Rodriguez, J.~Nguyen, B.~Muller, M.~Li, C.~Zhou, L.~Yu, J.~Weston, L.~Zettlemoyer, et~al.
\newblock Byte latent transformer: Patches scale better than tokens.
\newblock {\em arXiv preprint arXiv:2412.09871}, 2024.

\bibitem[PPVF21]{pesme2021implicit}
S.~Pesme, L.~Pillaud-Vivien, and N.~Flammarion.
\newblock Implicit bias of {SGD} for diagonal linear networks: a provable benefit of stochasticity.
\newblock In {\em NeurIPS}, 2021.

\bibitem[PRCB24]{patel2024datadreamer}
A.~Patel, C.~Raffel, and C.~Callison-Burch.
\newblock Datadreamer: A tool for synthetic data generation and reproducible {LLM} workflows.
\newblock In {\em ACL}. ACL, 2024.

\bibitem[Pri23]{prince2023understanding}
S.~J. Prince.
\newblock {\em Understanding Deep Learning}.
\newblock MIT Press, 2023.

\bibitem[PS{\etalchar{+}}03]{poggio2003mathematics}
T.~Poggio, S.~Smale, et~al.
\newblock The mathematics of learning: Dealing with data.
\newblock {\em Notices of the AMS}, 50(5):537--544, 2003.

\bibitem[PSL22]{press2021train}
O.~Press, N.~A. Smith, and M.~Lewis.
\newblock Train short, test long: Attention with linear biases enables input length extrapolation.
\newblock In {\em ICLR}, 2022.

\bibitem[QPF{\etalchar{+}}24]{qiu2024compute}
S.~Qiu, A.~Potapczynski, M.~Finzi, M.~Goldblum, and A.~G. Wilson.
\newblock Compute better spent: Replacing dense layers with structured matrices.
\newblock {\em arXiv preprint arXiv:2406.06248}, 2024.

\bibitem[RBOB18]{ravanelli2018light}
M.~Ravanelli, P.~Brakel, M.~Omologo, and Y.~Bengio.
\newblock Light gated recurrent units for speech recognition.
\newblock {\em IEEE Transactions on Emerging Topics in Computational Intelligence}, 2(2):92--102, 2018.

\bibitem[RGD{\etalchar{+}}22]{rampavsek2022recipe}
L.~Ramp{\'a}{\v{s}}ek, M.~Galkin, V.~P. Dwivedi, A.~T. Luu, G.~Wolf, and D.~Beaini.
\newblock Recipe for a general, powerful, scalable graph transformer.
\newblock In {\em NeurIPS}, 2022.

\bibitem[RHM86]{rumelhart1986general}
D.~E. Rumelhart, G.~E. Hinton, and J.~L. McClelland.
\newblock A general framework for parallel distributed processing.
\newblock In {\em Parallel Distributed Processing Volume 1}. MIT Press, 1986.

\bibitem[RKG{\etalchar{+}}22]{romero2022towards}
D.~W. Romero, D.~M. Knigge, A.~Gu, E.~J. Bekkers, E.~Gavves, J.~M. Tomczak, and M.~Hoogendoorn.
\newblock Towards a general purpose {CNN} for long range dependencies in $n${D}.
\newblock {\em arXiv preprint arXiv:2206.03398}, 2022.

\bibitem[RKX{\etalchar{+}}23]{radford2023robust}
A.~Radford, J.~W. Kim, T.~Xu, G.~Brockman, C.~McLeavey, and I.~Sutskever.
\newblock Robust speech recognition via large-scale weak supervision.
\newblock In {\em ICML}, 2023.

\bibitem[RM22]{rocks2022memorizing}
J.~W. Rocks and P.~Mehta.
\newblock Memorizing without overfitting: Bias, variance, and interpolation in overparameterized models.
\newblock {\em Physical Review Research}, 4(1):013201, 2022.

\bibitem[RS21]{rabe2021self}
M.~N. Rabe and C.~Staats.
\newblock Self-attention does not need $\mathcal{O}(n^{2})$ memory.
\newblock {\em arXiv preprint arXiv:2112.05682}, 2021.

\bibitem[RSR{\etalchar{+}}20]{raffel2020exploring}
C.~Raffel, N.~Shazeer, A.~Roberts, K.~Lee, S.~Narang, M.~Matena, Y.~Zhou, W.~Li, and P.~J. Liu.
\newblock Exploring the limits of transfer learning with a unified text-to-text transformer.
\newblock {\em The Journal of Machine Learning Research}, 21(1):5485--5551, 2020.

\bibitem[RWC{\etalchar{+}}19]{radford2019language}
A.~Radford, J.~Wu, R.~Child, D.~Luan, D.~Amodei, I.~Sutskever, et~al.
\newblock Language models are unsupervised multitask learners.
\newblock {\em OpenAI blog}, 2019.

\bibitem[RZL17]{ramachandran2017searching}
P.~Ramachandran, B.~Zoph, and Q.~V. Le.
\newblock Searching for activation functions.
\newblock {\em arXiv preprint arXiv:1710.05941}, 2017.

\bibitem[SAL{\etalchar{+}}24]{su2024roformer}
J.~Su, M.~Ahmed, Y.~Lu, S.~Pan, W.~Bo, and Y.~Liu.
\newblock Roformer: Enhanced transformer with rotary position embedding.
\newblock {\em Neurocomputing}, 568:127063, 2024.

\bibitem[Sch15]{schmidhuber2015deep}
J.~Schmidhuber.
\newblock Deep learning in neural networks: An overview.
\newblock {\em Neural Networks}, 61:85--117, 2015.

\bibitem[SCHU17]{scardapane2017group}
S.~Scardapane, D.~Comminiello, A.~Hussain, and A.~Uncini.
\newblock Group sparse regularization for deep neural networks.
\newblock {\em Neurocomputing}, 241:81--89, 2017.

\bibitem[SGT{\etalchar{+}}08]{scarselli2008graph}
F.~Scarselli, M.~Gori, A.~C. Tsoi, M.~Hagenbuchner, and G.~Monfardini.
\newblock The graph neural network model.
\newblock {\em IEEE Transactions on Neural Networks}, 20(1):61--80, 2008.

\bibitem[Sha19]{shazeer2019fast}
N.~Shazeer.
\newblock Fast transformer decoding: One write-head is all you need.
\newblock {\em arXiv preprint arXiv:1911.02150}, 2019.

\bibitem[Sha20]{shazeer2020glu}
N.~Shazeer.
\newblock {GLU} variants improve transformer.
\newblock {\em arXiv preprint arXiv:2002.05202}, 2020.

\bibitem[SHK{\etalchar{+}}14]{srivastava2014dropout}
N.~Srivastava, G.~Hinton, A.~Krizhevsky, I.~Sutskever, and R.~Salakhutdinov.
\newblock Dropout: a simple way to prevent neural networks from overfitting.
\newblock {\em The Journal of Machine Learning Research}, 15(1):1929--1958, 2014.

\bibitem[SHW21]{satorras2021n}
V.~G. Satorras, E.~Hoogeboom, and M.~Welling.
\newblock E(n) equivariant graph neural networks.
\newblock In {\em ICML}, 2021.

\bibitem[SKF{\etalchar{+}}99]{shibata1999byte}
Y.~Shibata, T.~Kida, S.~Fukamachi, M.~Takeda, A.~Shinohara, T.~Shinohara, and S.~Arikawa.
\newblock Byte pair encoding: A text compression scheme that accelerates pattern matching.
\newblock 1999.

\bibitem[SKZ{\etalchar{+}}21]{steiner2021train}
A.~Steiner, A.~Kolesnikov, X.~Zhai, R.~Wightman, J.~Uszkoreit, and L.~Beyer.
\newblock How to train your {ViT}? data, augmentation, and regularization in vision transformers.
\newblock {\em Transactions on Machine Learning Researc}, 2021.

\bibitem[SLJ{\etalchar{+}}15]{szegedy2015going}
C.~Szegedy, W.~Liu, Y.~Jia, P.~Sermanet, S.~Reed, D.~Anguelov, D.~Erhan, V.~Vanhoucke, and A.~Rabinovich.
\newblock Going deeper with convolutions.
\newblock In {\em IEEE CVPR}, 2015.

\bibitem[SMDH13]{sutskever2013importance}
I.~Sutskever, J.~Martens, G.~Dahl, and G.~Hinton.
\newblock On the importance of initialization and momentum in deep learning.
\newblock In {\em ICML}, 2013.

\bibitem[SP97]{schuster1997bidirectional}
M.~Schuster and K.~K. Paliwal.
\newblock Bidirectional recurrent neural networks.
\newblock {\em IEEE Transactions on Signal Processing}, 45(11):2673--2681, 1997.

\bibitem[SSBD14]{shalev2014understanding}
S.~Shalev-Shwartz and S.~Ben-David.
\newblock {\em Understanding machine learning: From theory to algorithms}.
\newblock Cambridge University Press, 2014.

\bibitem[Sti81]{stigler1981gauss}
S.~M. Stigler.
\newblock Gauss and the invention of least squares.
\newblock {\em The Annals of Statistics}, pages 465--474, 1981.

\bibitem[SVL14]{sutskever2014sequence}
I.~Sutskever, O.~Vinyals, and Q.~V. Le.
\newblock Sequence to sequence learning with neural networks.
\newblock In {\em NeurIPS}, 2014.

\bibitem[SVVTU19]{scardapane2019kafnets}
S.~Scardapane, S.~Van~Vaerenbergh, S.~Totaro, and A.~Uncini.
\newblock Kafnets: Kernel-based non-parametric activation functions for neural networks.
\newblock {\em Neural Networks}, 110:19--32, 2019.

\bibitem[SW17]{scardapane2017randomness}
S.~Scardapane and D.~Wang.
\newblock Randomness in neural networks: an overview.
\newblock {\em Wiley Interdisciplinary Reviews: Data Mining and Knowledge Discovery}, 7(2):e1200, 2017.

\bibitem[SWF{\etalchar{+}}15]{sukhbaatar2015end}
S.~Sukhbaatar, J.~Weston, R.~Fergus, et~al.
\newblock End-to-end memory networks.
\newblock In {\em NeurIPS}, 2015.

\bibitem[SWL23]{smith2022simplified}
J.~T. Smith, A.~Warrington, and S.~W. Linderman.
\newblock Simplified state space layers for sequence modeling.
\newblock In {\em ICLR}, 2023.

\bibitem[TCB{\etalchar{+}}24]{tiezzi2024state}
M.~Tiezzi, M.~Casoni, A.~Betti, M.~Gori, and S.~Melacci.
\newblock State-space modeling in long sequence processing: A survey on recurrence in the transformer era.
\newblock {\em arXiv preprint arXiv:2406.09062}, 2024.

\bibitem[TEM23]{tschannen2023givt}
M.~Tschannen, C.~Eastwood, and F.~Mentzer.
\newblock {GIVT}: Generative infinite-vocabulary transformers.
\newblock {\em arXiv preprint arXiv:2312.02116}, 2023.

\bibitem[TGJ{\etalchar{+}}15]{tompson2015efficient}
J.~Tompson, R.~Goroshin, A.~Jain, Y.~LeCun, and C.~Bregler.
\newblock Efficient object localization using convolutional networks.
\newblock In {\em IEEE/CVF CVPR}, 2015.

\bibitem[THK{\etalchar{+}}21]{tolstikhin2021mlp}
I.~O. Tolstikhin, N.~Houlsby, A.~Kolesnikov, L.~Beyer, X.~Zhai, T.~Unterthiner, J.~Yung, A.~Steiner, D.~Keysers, J.~Uszkoreit, et~al.
\newblock {MLP-Mixer}: An all-{MLP} architecture for vision.
\newblock In {\em NeurIPS}, 2021.

\bibitem[TLI{\etalchar{+}}23]{touvron2023llama}
H.~Touvron, T.~Lavril, G.~Izacard, X.~Martinet, M.-A. Lachaux, T.~Lacroix, B.~Rozi{\`e}re, N.~Goyal, E.~Hambro, F.~Azhar, et~al.
\newblock Llama: Open and efficient foundation language models.
\newblock {\em arXiv preprint arXiv:2302.13971}, 2023.

\bibitem[TNHA24]{teney2024neural}
D.~Teney, A.~M. Nicolicioiu, V.~Hartmann, and E.~Abbasnejad.
\newblock Neural redshift: Random networks are not random functions.
\newblock In {\em IEEE/CVF CVPR}, 2024.

\bibitem[Unc15]{uncini2015fundamentals}
A.~Uncini.
\newblock {\em Fundamentals of adaptive signal processing}.
\newblock Springer, 2015.

\bibitem[Vap13]{vapnik2013nature}
V.~Vapnik.
\newblock {\em The nature of statistical learning theory}.
\newblock Springer Science \& Business Media, 2013.

\bibitem[VCC{\etalchar{+}}18]{velivckovic2017graph}
P.~Veli{\v{c}}kovi{\'c}, G.~Cucurull, A.~Casanova, A.~Romero, P.~Lio, and Y.~Bengio.
\newblock Graph attention networks.
\newblock In {\em ICLR}, 2018.

\bibitem[Vel22]{velivckovic2022message}
P.~Veli{\v{c}}kovi{\'c}.
\newblock Message passing all the way up.
\newblock {\em arXiv preprint arXiv:2202.11097}, 2022.

\bibitem[VSP{\etalchar{+}}17]{vaswani2017attention}
A.~Vaswani, N.~Shazeer, N.~Parmar, J.~Uszkoreit, L.~Jones, A.~N. Gomez, {\L}.~Kaiser, and I.~Polosukhin.
\newblock Attention is all you need.
\newblock In {\em NeurIPS}, 2017.

\bibitem[VWB16]{veit2016residual}
A.~Veit, M.~J. Wilber, and S.~Belongie.
\newblock Residual networks behave like ensembles of relatively shallow networks.
\newblock In {\em NeurIPS}, 2016.

\bibitem[WBF{\etalchar{+}}24]{welleck2024decoding}
S.~Welleck, A.~Bertsch, M.~Finlayson, H.~Schoelkopf, A.~Xie, G.~Neubig, I.~Kulikov, and Z.~Harchaoui.
\newblock From decoding to meta-generation: Inference-time algorithms for large language models.
\newblock {\em arXiv preprint arXiv:2406.16838}, 2024.

\bibitem[WCW{\etalchar{+}}23]{wang2023neural}
C.~Wang, S.~Chen, Y.~Wu, Z.~Zhang, L.~Zhou, S.~Liu, Z.~Chen, Y.~Liu, H.~Wang, J.~Li, et~al.
\newblock Neural codec language models are zero-shot text to speech synthesizers.
\newblock {\em arXiv preprint arXiv:2301.02111}, 2023.

\bibitem[WFD{\etalchar{+}}23]{wang2023scientific}
H.~Wang, T.~Fu, Y.~Du, W.~Gao, K.~Huang, Z.~Liu, P.~Chandak, S.~Liu, P.~Van~Katwyk, A.~Deac, et~al.
\newblock Scientific discovery in the age of artificial intelligence.
\newblock {\em Nature}, 620(7972):47--60, 2023.

\bibitem[WGGH18]{wang2018non}
X.~Wang, R.~Girshick, A.~Gupta, and K.~He.
\newblock Non-local neural networks.
\newblock In {\em IEEE/CVF CVPR}, 2018.

\bibitem[WJ21]{wu2021rethinking}
Y.~Wu and J.~Johnson.
\newblock Rethinking "{B}atch" in {BatchNorm}.
\newblock {\em arXiv preprint arXiv:2105.07576}, 2021.

\bibitem[WZZ{\etalchar{+}}13]{wan2013regularization}
L.~Wan, M.~Zeiler, S.~Zhang, Y.~Le~Cun, and R.~Fergus.
\newblock Regularization of neural networks using {DropConnect}.
\newblock In {\em ICML}, 2013.

\bibitem[XYH{\etalchar{+}}20]{xiong2020layer}
R.~Xiong, Y.~Yang, D.~He, K.~Zheng, S.~Zheng, C.~Xing, H.~Zhang, Y.~Lan, L.~Wang, and T.~Liu.
\newblock On layer normalization in the transformer architecture.
\newblock In {\em ICML}, 2020.

\bibitem[YCC{\etalchar{+}}24]{yuan2024revisiting}
L.~Yuan, Y.~Chen, G.~Cui, H.~Gao, F.~Zou, X.~Cheng, H.~Ji, Z.~Liu, and M.~Sun.
\newblock Revisiting out-of-distribution robustness in {NLP}: Benchmarks, analysis, and llms evaluations.
\newblock In {\em NeurIPS}, 2024.

\bibitem[YHO{\etalchar{+}}19]{yun2019cutmix}
S.~Yun, D.~Han, S.~J. Oh, S.~Chun, J.~Choe, and Y.~Yoo.
\newblock {CutMix}: Regularization strategy to train strong classifiers with localizable features.
\newblock In {\em IEEE/CVF ICCV}, 2019.

\bibitem[YLC{\etalchar{+}}22]{yu2022s2}
T.~Yu, X.~Li, Y.~Cai, M.~Sun, and P.~Li.
\newblock S2-{MLP}: Spatial-shift {MLP} architecture for vision.
\newblock In {\em WACV}, 2022.

\bibitem[YLZ{\etalchar{+}}22]{yu2022metaformer}
W.~Yu, M.~Luo, P.~Zhou, C.~Si, Y.~Zhou, X.~Wang, J.~Feng, and S.~Yan.
\newblock Metaformer is actually what you need for vision.
\newblock In {\em IEEE/CVF CVPR}, 2022.

\bibitem[YTL{\etalchar{+}}25]{yang2025mmada}
L.~Yang, Y.~Tian, B.~Li, X.~Zhang, K.~Shen, Y.~Tong, and M.~Wang.
\newblock Mmada: Multimodal large diffusion language models.
\newblock {\em arXiv preprint arXiv:2505.15809}, 2025.

\bibitem[YYZ17]{yu2017spatio}
B.~Yu, H.~Yin, and Z.~Zhu.
\newblock Spatio-temporal graph convolutional networks: A deep learning framework for traffic forecasting.
\newblock In {\em IJCAI}, 2017.

\bibitem[ZBH{\etalchar{+}}21]{zhang2021understanding}
C.~Zhang, S.~Bengio, M.~Hardt, B.~Recht, and O.~Vinyals.
\newblock Understanding deep learning (still) requires rethinking generalization.
\newblock {\em Communications of the ACM}, 64(3):107--115, 2021.

\bibitem[ZCDLP17]{zhang2017mixup}
H.~Zhang, M.~Cisse, Y.~N. Dauphin, and D.~Lopez-Paz.
\newblock mixup: Beyond empirical risk minimization.
\newblock In {\em ICLR}, 2017.

\bibitem[ZER{\etalchar{+}}23]{zador2023catalyzing}
A.~Zador, S.~Escola, B.~Richards, B.~{\"O}lveczky, Y.~Bengio, K.~Boahen, M.~Botvinick, D.~Chklovskii, A.~Churchland, C.~Clopath, et~al.
\newblock Catalyzing next-generation artificial intelligence through {neuroAI}.
\newblock {\em Nature Communications}, 14(1):1597, 2023.

\bibitem[ZJM{\etalchar{+}}21]{zbontar2021barlow}
J.~Zbontar, L.~Jing, I.~Misra, Y.~LeCun, and S.~Deny.
\newblock Barlow twins: Self-supervised learning via redundancy reduction.
\newblock In {\em ICML}, 2021.

\bibitem[ZKR{\etalchar{+}}17]{zaheer2017deep}
M.~Zaheer, S.~Kottur, S.~Ravanbakhsh, B.~Poczos, R.~R. Salakhutdinov, and A.~J. Smola.
\newblock Deep sets.
\newblock In {\em NeurIPS}, 2017.

\bibitem[ZLLS23]{zhang2023dive}
A.~Zhang, Z.~C. Lipton, M.~Li, and A.~J. Smola.
\newblock {\em Dive into deep learning}.
\newblock Cambridge University Press, 2023.

\bibitem[ZS19]{zhang2019root}
B.~Zhang and R.~Sennrich.
\newblock Root mean square layer normalization.
\newblock In {\em NeurIPS}, 2019.

\bibitem[ZTS{\etalchar{+}}21]{zhai2021attention}
S.~Zhai, W.~Talbott, N.~Srivastava, C.~Huang, H.~Goh, R.~Zhang, and J.~Susskind.
\newblock An attention free transformer.
\newblock {\em arXiv preprint arXiv:2105.14103}, 2021.

\bibitem[ZW23]{ziyin2023spred}
L.~Ziyin and Z.~Wang.
\newblock spred: Solving $l_1$ penalty with {SGD}.
\newblock In {\em ICML}, 2023.

\end{thebibliography}
